\documentclass[12pt,openany]{cmuthesis}

\usepackage{amsmath,amsfonts,bm}

\def\1{\bm{1}}

\def\rvc{{\mathbf{c}}}

\def\rve{{\mathbf{e}}}

\def\rvu{{\mathbf{i}}}

\def\rvk{{\mathbf{k}}}

\def\rvm{{\mathbf{m}}}

\def\rvp{{\mathbf{p}}}

\def\rvs{{\mathbf{s}}}
\def\rvt{{\mathbf{t}}}
\def\rvu{{\mathbf{u}}}
\def\rvv{{\mathbf{v}}}

\def\rvx{{\mathbf{x}}}

\def\rvz{{\mathbf{z}}}

\def\ervb{{\textnormal{b}}}
\def\ervc{{\textnormal{c}}}

\def\erve{{\textnormal{e}}}

\def\ervm{{\textnormal{m}}}

\def\ervp{{\textnormal{p}}}

\def\ervs{{\textnormal{s}}}
\def\ervt{{\textnormal{t}}}

\def\ervv{{\textnormal{v}}}

\def\ervx{{\textnormal{x}}}

\def\ervz{{\textnormal{z}}}

\def\rmC{{\mathbf{C}}}
\def\rmD{{\mathbf{D}}}

\def\rmI{{\mathbf{I}}}

\def\rmP{{\mathbf{P}}}

\def\rmV{{\mathbf{V}}}

\def\rmX{{\mathbf{X}}}

\def\rmZ{{\mathbf{Z}}}

\def\vzero{{\bm{0}}}
\def\vone{{\bm{1}}}

\def\vs{{\bm{s}}}

\DeclareMathAlphabet{\mathsfit}{\encodingdefault}{\sfdefault}{m}{sl}
\SetMathAlphabet{\mathsfit}{bold}{\encodingdefault}{\sfdefault}{bx}{n}

\def\gI{{\mathcal{I}}}

\def\gL{{\mathcal{L}}}
\def\gM{{\mathcal{M}}}

\def\gO{{\mathcal{O}}}

\def\gT{{\mathcal{T}}}
\def\gU{{\mathcal{U}}}

\def\sC{{\mathbb{C}}}

\def\sR{{\mathbb{R}}}

\def\sT{{\mathbb{T}}}

\def\sZ{{\mathbb{Z}}}

\newcommand{\pllm}{\mathds{P}_{\rm{LLM}}}

\newcommand{\E}{\mathbb{E}}
\newcommand{\Ls}{\mathcal{L}}

\DeclareMathOperator*{\argmin}{arg\,min}

\DeclareMathOperator{\sign}{sign}

\DeclareMathOperator{\allpadded}{ispad}
\DeclareMathOperator{\gumbel}{Gumbel}

\usepackage{pifont}% http://ctan.org/pkg/pifont
\newcommand{\cmark}{\ding{51}}%
\usepackage{libertine}
\usepackage{libertinust1math}
\usepackage[T1]{fontenc}
\usepackage{fullpage}
\usepackage{graphicx}
\usepackage{amsmath}
\usepackage[numbers,sort]{natbib}
\usepackage[pagebackref,pageanchor=true,plainpages=false, pdfpagelabels, bookmarks,bookmarksnumbered,hidelinks,breaklinks,colorlinks,
linkcolor=teal,citecolor=teal,urlcolor=teal
]{hyperref}
\usepackage{subcaption}
\usepackage{amssymb}
\usepackage{booktabs}
\usepackage{makecell}
\usepackage[shortlabels]{enumitem}
\usepackage{centernot}
\usepackage{amsfonts}
\usepackage{algorithm}
\usepackage{algpseudocode}
\usepackage[normalem]{ulem}
\usepackage{dsfont}
\usepackage{multirow}
\usepackage{multicol}
\usepackage{cuted}
\usepackage{capt-of}
\usepackage{bbm}
\usepackage{balance}
\usepackage[accsupp]{axessibility}
\usepackage{xspace}
\usepackage{arydshln}
\usepackage{adjustbox}
\usepackage{rotating}
\usepackage{wrapfig,lipsum,booktabs}
\usepackage{fancyvrb}
\usepackage[acronym]{glossaries}
\usepackage{mathtools}
\usepackage{bookmark}
\usepackage{CJKutf8}

\usepackage[letterpaper,twoside,vscale=.8,hscale=.75,nomarginpar,hmarginratio=1:1]{geometry}
\usepackage[capitalize]{cleveref}
\crefname{section}{Section}{Secs.}
\Crefname{section}{Section}{Sections}
\Crefname{table}{Table}{Tables}
\crefname{table}{Tab.}{Tabs.}
\definecolor{mygray}{gray}{0.4}
\newenvironment{mytiny}{\color{mygray}\scriptsize}{}
\makeatletter
\DeclareRobustCommand\onedot{\futurelet\@let@token\@onedot}
\def\@onedot{\ifx\@let@token.\else.\null\fi\xspace}

\def\eg{\emph{e.g}\onedot} \def\Eg{\emph{E.g}\onedot}
\def\ie{\emph{i.e}\onedot} 
\def\cf{\emph{c.f}\onedot} 
\def\etc{\emph{etc}\onedot} \def\vs{\emph{vs}\onedot}

\makeatother

\newcommand{\dataname}{DocumentNet}
\newcommand{\taskname}{VDER}
\newcommand{\modelname}{MAGVIT}
\newcommand{\methodname}{COMMIT}
\newcommand{\cls}{\texttt{[CLS]}\xspace}
\newcommand{\mask}{\texttt{[MASK]}\xspace}

\makenoidxglossaries
\newacronym{AI}{AI}{Artificial Intelligence}
\newacronym{LLM}{LLM}{Large Language Model}
\newacronym{NLP}{NLP}{Natural Language Processing}
\newacronym{AGI}{AGI}{Artificial General Intelligence}
\newacronym{VDER}{VDER}{Visually-rich Document Entity Retrieval}
\newacronym{MAGVIT}{MAGVIT}{MAsked Generative VIdeo Transformer}
\newacronym{FVD}{FVD}{Fr\'echet Video Distance}
\newacronym{SPAE}{SPAE}{Semantic Pyramid AutoEncoder}
\newacronym{HEVC}{HEVC}{High Efficiency Video Coding}
\newacronym{VVC}{VVC}{Versatile Video Coding}
\newacronym{PaLM}{PaLM}{Pathways Language Model}
\newacronym{GPT}{GPT}{Generative Pre-trained Transformer}

\begin {document} 
\frontmatter

%initialize page style, so contents come out right (see bot) -mjz
\pagestyle{empty}

\title{  %{\it \huge Thesis Proposal}\\
{\bf Towards Multi-Task Multi-Modal Models}:\\
A Video Generative Perspective
}
\author{Yu, Lijun \begin{CJK*}{UTF8}{gkai}于力军\end{CJK*}}
\date{CMU-LTI-24-003\\April 2024}
\Year{2024}
\trnumber{}

\committee{
Alexander G. Hauptmann, Chair \\
Yonatan Bisk \\
Lu Jiang \\
Ming-Hsuan Yang (Google, UC Merced) \\
}

\support{}
\disclaimer{}

% copyright notice generated automatically from Year and author.
% permission added if \permission{} given.


\maketitle

\begin{dedication}
To generate anything.
\end{dedication}

\pagestyle{plain} % for toc, was empty

%% Obviously, it's probably a good idea to break the various sections of your thesis
%% into different files and input them into this file...

\begin{abstract}
Advancements in language foundation models have primarily fueled the recent surge in artificial intelligence. In contrast, generative learning of non-textual modalities, especially videos, significantly trails behind language modeling.
This thesis chronicles our endeavor to build multi-task models for generating videos and other modalities under diverse conditions, as well as for understanding and compression applications.

We start with two pixel-space prototypes for separate multi-task and multi-modal setups. Despite their effectiveness, these models are constrained by task-specific modules and predefined label spaces, underscoring the need for more universally applicable designs.

Given the high dimensionality of visual data, we pursue concise and accurate latent representations. Our video-native spatial-temporal tokenizers preserve high fidelity. We unveil a novel approach to mapping bidirectionally between visual observation and interpretable lexical terms. Furthermore, our scalable visual token representation proves beneficial across generation, compression, and understanding tasks. This achievement marks the first instances of language models surpassing diffusion models in visual synthesis and a video tokenizer outperforming industry-standard codecs.

Within these multi-modal latent spaces, we study the design of multi-task generative models. Our masked multi-task transformer excels at the quality, efficiency, and flexibility of video generation. We enable a frozen language model, trained solely on text, to generate visual content. Finally, we build a scalable generative multi-modal transformer trained from scratch, enabling the generation of videos containing high-fidelity motion with the corresponding audio given diverse conditions.

Throughout the course, we have shown the effectiveness of integrating multiple tasks, crafting high-fidelity latent representation, and generating multiple modalities. This work suggests intriguing potential for future exploration in generating non-textual data and enabling real-time, interactive experiences across various media forms.

\end{abstract}

\begin{acknowledgments}
We live in an exciting era for artificial intelligence, especially video generation. I am fortunate to have started my video research in 2018 when I visited Dr. Alexander G. Hauptmann, who later became my Ph.D. advisor.
Subsequently, I have studied video generation techniques and explored more modalities with Dr. Lu Jiang, my internship mentor, Dr. Ming-Hsuan Yang, and Dr. Yonatan Bisk.
The last chapter details the valuable opportunities I have had to collaborate with many brilliant minds at Carnegie Mellon, Google, and other institutions.
Despite ``My Heart is in the Work'' being a valuable factor for success, it turns out that identifying and seizing the correct opportunities play a more significant role in history.
I would like to express my sincere gratitude to everyone who has facilitated my unique Ph.D. journey, which has exceeded my expectations.

While this thesis chronicles my notable technical contributions, the side effects of doing a Ph.D. include a so-called ``permanent head damage'', where I understand myself more profoundly than before.
Our curiosity about science and truth drives us to explore the unknown realm, where our desire for fortune and fame accompanies it.
My curiosity would not have been satisfied to the current extent without the generous support from the Language Technologies Institute, Google Research, Siebel Scholar Foundation, and Baidu Scholarship, as well as the projects funded by IARPA, NIST, DSTA, and Traffic 21, along with the valuable assistance from the LTI staff team.
For further curiosity, ``\begin{CJK*}{UTF8}{gkai}买岛建国当国家元首\end{CJK*}'' might not be an infeasible path.

Ten thousand scrolls are no better than ten thousand miles. The production of this thesis took place around the globe, from the Dalton Highway in the Arctic Circle to the tropical Hainan and Honolulu islands.
I appreciate everyone who has shared joyful memories, especially those who ``\begin{CJK*}{UTF8}{gkai}如胡适先生一般爱好打牌\end{CJK*}'' and who ``\begin{CJK*}{UTF8}{gkai}如老大爷一般爱好遛弯\end{CJK*}''.
Notable mentions are dedicated to Dr. Wenhe Liu, Kevin, Haoyang, Xiaoyu, Zora, Xinyu, Liying, and Yufei.
The unconditional love from my family and my partner has always been my source of strength in pursuing my curiosity.

Happy graduation!

\end{acknowledgments}
\cleardoublepage

\setcounter{tocdepth}{1}
\tableofcontents
\printnoidxglossaries
\cleardoublepage
\listoffigures
\listoftables

\mainmatter

%% Double space document for easy review:
%\renewcommand{\baselinestretch}{1.66}\normalsize

% The other requirements Catherine has:
%
%  - avoid large margins.  She wants the thesis to use fewer pages, 
%    especially if it requires colour printing.
%
%  - The thesis should be formatted for double-sided printing.  This
%    means that all chapters, acknowledgements, table of contents, etc.
%    should start on odd numbered (right facing) pages.
%
%  - You need to use the department standard tech report title page.  I
%    have tried to ensure that the title page here conforms to this
%    standard.
%
%  - Use a nice serif font, such as Times Roman.  Sans serif looks bad.
%
% Other than that, just make it look good...

\glsresetall
\cleardoublepage
\part*{Introduction}
\addcontentsline{toc}{part}{Introduction}
\cleardoublepage
\chapter*{Introduction}

\section*{Motivation}
\addcontentsline{toc}{section}{Motivation}

% How can we build an \Gls{AI} that understands the cosmos in ways better than all human beings and spearheads unparalleled breakthroughs in the field of natural science?
% While it appears to be formidable, we endeavor to progress toward this ambitious objective through a series of strategic considerations.
% and acts as a ``world model'', hopefully in ways greatly better than human beings?
% We will reach a key milestone when AI can understand 
% A key milestone would be 
% One key property is the creativity, flexibility 

\paragraph{}
Since its inception nearly seven decades ago, the field of \Gls{AI}~\cite{mccarthy2006proposal} has undergone significant evolutionary strides, marked by a succession of pivotal milestones. This journey witnessed the transition from rule-based expert systems~\cite{campbell2002deep} to the data-driven paradigms ushered in by machine learning~\cite{samuel1959some}, subsequently transcending to the realms of deep learning where the focus shifted from feature engineering~\cite{lowe1999object} to the acquisition of representations directly from raw data~\cite{lecun1989backpropagation}. The advent of foundation models~\cite{bommasani2021opportunities} further epitomizes this evolutionary trajectory, promoting the sharing of knowledge across tasks, thereby obviating the need for task-specific models. Within this continuum, BERT~\cite{devlin2019bert} emerges as a quintessential exemplar of foundation models, epitomized by its training on extensive data via self-supervision and its proficiency in adapting to a plethora of downstream tasks. This dissertation delves into the \emph{multi-task} versatility at the heart of methodological innovations, tracing the evolution from hierarchically structured supervised modules to cohesive, universally applicable self-supervised frameworks.

% Artificial intelligence, since its debut nearly 70 years ago~\cite{mccarthy2006proposal}, has seen significant advancements with a series of milestones.
% Among them, machine learning~\cite{samuel1959some} enables data-driven problem solving instead of rule-based expert systems~\cite{campbell2002deep}.
% Later, deep learning~\cite{lecun1989backpropagation} acquires representation from raw data rather than feature engineering~\cite{lowe1999object}.
% More recently, \emph{foundation models}~\cite{bommasani2021opportunities} promotes knowledge sharing between tasks in place of separate models.
% While the characteristics of foundation models remain evolving, BERT~\cite{devlin2019bert} has been a representative example, which is trained on broad data with self-supervision and adaptable to a wide range of downstream tasks.
% In this thesis, such \emph{multi-task} versatility lies in the core of methodology designs, evolving from cascaded supervised modules to universal self-supervised frameworks.

\paragraph{}
\Glspl{LLM}~\cite{brown2020language,googlepalm2,touvron2023llama}, emblematic of foundation models, are architectured with \emph{generative} goals, crafting text outputs from diverse inputs. Notably, certain adaptations of \Glspl{LLM}~\cite{openai2023gpt,liu2024visual} have expanded their input capacity to encompass images, though their outputs are exclusively textual. This text-centric output is a manifestation of a human-conceived low-bandwidth abstraction, leading to projections of an impending scarcity of high-quality textual data~\cite{villalobos2022will}. In stark contrast, there exists a prodigious generation of raw signal data, particularly \emph{videos}, which often surpasses the computational resources available for their effective utilization in training paradigms. Moreover, the progression of self-supervised generative learning for these non-textual data types significantly lags behind that of language models, thereby curtailing the potential of associated tasks. The crux of this dissertation is anchored in the exploration of generative learning aimed at producing outputs beyond text, including videos, images, and audio, thus embracing a more holistic \emph{multi-modal} approach.

% \Glspl{LLM}~\cite{brown2020language,googlepalm2,touvron2023llama}, as a typical type of foundation models, are trained with \emph{generative} objectives to produce text.
% Although some variants~\cite{openai2023gpt,liu2024visual} of \Glspl{LLM} accept additional information such as images as inputs, the output remains text-only.
% As a low-bandwidth abstraction induced by human, we are likely running out of high-quality text data soon~\cite{villalobos2022will}.
% In contrast, significant amounts of data in raw signals, especially \emph{videos}, are regularly produced, where compute is generally insufficient to fully leverage them for training.
% However, the development of self-supervised generative learning of these modalities significantly trails behind language modeling~\cite{chen2020generative}, limiting the capabilities of relevant tasks.
% This thesis primarily focuses on generative learning to produce non-textual modalities, from videos alone to \emph{multi-modal} outputs spanning video, image, and audio.

\paragraph{}
The transformer architecture~\cite{vaswani2017attention}, initially conceived to interpret text tokens, stands as the cornerstone for scalable models across various domains. Yet, when it comes to handling raw signals, such as videos, we encounter a paradigm marked by considerably greater complexity due to their inherently higher dimensional nature, encapsulating high spatial-temporal resolutions alongside multiple channels.
% Transformers~\cite{vaswani2017attention}, as the commonly adopted scalable architecture, are originally designed to process text tokens.
% Raw signals like videos are characterized by significantly higher dimensions, often involving high spatial-temporal resolutions and multiple channels.
While straightforward downscaling techniques~\cite{dosovitskiy2020image} may suffice for discriminative models when predicting labels, they present formidable hurdles for generative models tasked with producing content in these high-dimensional spaces, particularly in the context of high-resolution image or extended video generation.
To address this challenge, we embark on a journey to construct learned \emph{latent representations} within highly-compressed spaces and subsequently formulate generative models tailored to operate within these constrained dimensions.

\section*{Thesis Organization}
\addcontentsline{toc}{section}{Thesis Organization}

\begin{table}[tp]
\centering
\begin{tabular}{@{}llll@{}}
\toprule
 Modality & Evaluation & Representation & Model \\ \midrule
% \multicolumn{4}{l}{\textbf{\cref{part:pre}: Discriminative Tasks}} \\
\\
 Video & Understanding & & \makecell[l]{\textit{\small \cref{part:pre} \cref{chap:argus} (A)}\\\makecell[l]{Multi-Task Cascaded Modules}} \\
\\
\makecell[l]{Image\\\ + Text} & Understanding & & \makecell[l]{\textit{\small \cref{part:pre} \cref{chap:doc} (C)}\\Masked Transformer} \\
% \midrule
% \multicolumn{4}{l}{\textbf{\cref{part:latent}: Multi-Modal Discrete Latent Spaces}} \\
% \multicolumn{4}{r}{\textbf{\cref{part:model}: Multi-Task Generative Models}} \\
% \\ \hdashline
\\
Video & Generation & \makecell[l]{\textit{\small \cref{part:latent} \cref{chap:3dvq} (B)}\\Spatial-Temporal\\Vector-Quantized} & \makecell[l]{\textit{\small \cref{part:model} \cref{chap:magvit} (A)}\\Masked Generative\\Video Transformer} \\
\\
\makecell[l]{Video\\\ + Image\\\ + Text} & \makecell[l]{Generation\\\ + Understanding} & \makecell[l]{\textit{\small \cref{part:latent} \cref{chap:spae} (BC)}\\Visual Lexical\\\ } & \makecell[l]{\textit{\small \cref{part:model} \cref{chap:llm} (AC)}\\Frozen Large\\Language Model} \\
\\
\makecell[l]{Video\\\ + Image\\\ + Audio\\\ + Text} & \makecell[l]{Generation\\\ + Compression\\\ + Understanding} & \makecell[l]{\textit{\small \cref{part:latent} \cref{chap:magvitv2} (BC)}\\Scalable Visual\\Token} & \makecell[l]{\textit{\small \cref{part:model} \cref{chap:videopoet} (ABC)}\\Scalable Generative\\Multi-Modal Transformer} \\
\bottomrule
\end{tabular}
\caption[Overview of thesis structure]{\textbf{Overview of thesis structure} with involved modalities and evaluations.
Chapters in the same row are paired latent representation and generative models. 
The letters in parentheses refer to the focused component in the thesis statement: (A) integrating multiple tasks, (B) crafting high-fidelity latent representation, and (C) generating multiple modalities.}
\label{tab:org}
\end{table}

In this thesis, we strive to build multi-task models for multi-modal generation and understanding.
We start with two pixel-space prototypes for separate multi-task and multi-modal understanding setups in \cref{part:pre}.
Given the high dimensionality of visual data, we pursue concise and accurate latent representations in \cref{part:latent}.
Within these multi-modal latent spaces, we study the design of multi-task generative models in \cref{part:model}.
\cref{tab:org} presents the logical structure of this thesis, with a brief over below.
% Below we give a brief overview of the thesis, with a structure chart in \cref{tab:org}.
% In this thesis, we aim to build multi-task generative models to provide a pathway to multi-modal intelligence that can perceive, comprehend, and simulate the world.
% To achieve this goal, we divide the thesis into three parts with different focuses.
% We divide the thesis into three parts to achieve this goal.
% \cref{tab:org} provides an overview of the thesis structure.

\paragraph{\cref{part:pre}: \nameref{part:pre}.}
In the first part of this thesis, we unveil a pair of prototypes designed for multi-task and multi-modal problems that encompass \emph{video, image, and text modalities}. These prototypes showcase effective comprehension outcomes within the designated tasks, yet also underscore the need for additional exploration into generative modeling to achieve broader capabilities.

In \cref{chap:argus}, we introduce a versatile system designed to comprehend videos, achieving favorable outcomes across a variety of assessment benchmarks. This system showcases a range of capabilities, including but not limited to object detection, object tracking, foreground segmentation, activity proposal generation, and activity recognition. Its primary emphasis is on spatial-temporal activity recognition and localization, consistently delivering state-of-the-art performance across a series of benchmark scenarios. 
As a prototype of \emph{multi-task video} system, its ability to incorporate new tasks is noticeably limited, achievable solely through the integration of new modules.
In the following chapters, we will explore models adaptable to various tasks without major changes.

% \cref{chap:argus} introduces a multi-task video understanding system, which achieves dominent performance across many benchmarks.
% The system supports object detection, object tracking, foreground segmentation, activity proposal generation, activity recognition, \etc.
% Combining all of these, its core functionality is spatial-temporal activity recognition and localization, where it demonstrates dominent performance across many benchmarks.
% Despite supporting multiple tasks, its flexibility of adding new tasks is greatly limited, which can only be achieved via plugging in new modules.
% However, this cascaded design with a heuristically-crafted latent representation limits further advancement of true video understanding.

% Masked modeling is a type of generative pre-training objective that shows superior performance on language modeling with transformers.
% In \cref{chap:doc}, we adopt masked vision-language pre-training for document understanding, where the model effectively generates the masked text and pixels.
% This is a prototype of mask-based generative models with a single inference step.
% In the later parts, we will study video generation models which are trained via masked modeling and inferred with multi-step iterative decoding.

In \cref{chap:doc}, we embrace the concept of masked \emph{vision-language} pre-training to enhance document understanding.
\emph{Masked modeling} represents a form of \emph{generative pre-training} objective that benefits language modeling when applied with transformer architectures.
In our case, the model acquires valuable \emph{multi-modal} representations for tasks of visually-rich document entity retrieval, achieved by learning to recover the masked text and pixel information.
% Here, the model learns useful representation for visually-rich document entity retrieval tasks via adeptly generating concealed text and pixels. 
With a singular inference step, this model resembles a prototype for mask-based generative models.
% This stands as a prototype for mask-based generative models, following a singular inference step. 
In subsequent parts, we will delve into the realm of generation models trained using masked modeling techniques and inference through multi-step iterative decoding.

% \paragraph{Argus}
% \input{overviews/argus.tex}
% \paragraph{DocumentNet}
% \input{overviews/doc.tex}

\paragraph{\cref{part:latent}: \nameref{part:latent}.}
% While it is straightforward to build visual generative models directly in the pixel space, its high-dimensional and redundant nature makes it difficult to scale to high-resolution and/or long videos.
% To this end, most modern visual generative models operate in a learned latent space, which is bidirectionally mapped to the pixel space.
% In this part, we study the latent spaces for generative visual modeling.
% Transformer-based language models usually operate with sub-word tokens as the processing units.
% However, it is less feasible to use their direct analog of pixels for visual generative modeling with transformers.
% The challenge arises from its intricate high-dimensional and repetitive characteristics, which impede its expansion to high-resolution or lengthy videos.
% Consequently, the majority of contemporary visual generative models function within a learned latent space, which is interconnected with the pixel space through a two-way mapping. 
% In this part, we delve into an exploration of multi-modal discrete latent spaces for the purpose of generative modeling with transformers.
% In this part, we study the latent space for generative modeling.
% Due to the high-dimensional and redundant nature of the pixel space, most modern visual generative models operate in learned latent spaces.
While language models commonly function using sub-word tokens as their processing units, employing the direct equivalent of pixels for visual generative modeling with transformers presents more difficulties.
This challenge stems from the complex, high-dimensional, and repetitive nature of pixel data, which hinders the scalability of transformers to high-resolution images or lengthy videos. As a result, the prevailing approach in contemporary visual generative models involves operating within a learned latent space. This latent space is intricately connected to the pixel space through a bidirectional mapping.
In this part, we explore the concept of \emph{multi-modal latent spaces} for generative visual modeling with transformers.

% \cref{chap:3dvq} introduces a spatial-temporal vector-quantization model for video tokenization.
% Inspired by the success of various image tokenizers, we build this model with a novel architecture utilizing 3D convolutions for video modeling.
% This model presents high reconstruction fidelity at a high compression ratio, which enables the subsequent success of generative video transformers.
% This model presents a novel architecture utilizing 3D convolutions, largely inspired by the success of various image tokenizers.
% tokenize videos, inspired by the success of various image tokenizers.
% 3D VQGAN model for spatial-temporal tokenization of videos inspired by the success of VQGAN in image generation.
In \cref{chap:3dvq}, we present a \emph{spatial-temporal vector-quantization} model designed to map a video into a discrete latent space (\ie tokenization) defined by a learned codebook. Taking inspiration from the achievements of different image tokenization methods, we devise a unique architecture for this model that incorporates 3D convolutions to effectively model video data with both spatial and temporal dependencies. As a result of this design, the model achieves satisfying reconstruction fidelity even at significant compression ratios, thereby laying the foundation for the subsequent achievements of generative video transformers.

% \cref{chap:spae} proposes a groundbreaking approach that maps visual data into the latent space of a trained Large Language Model (LLM).
% By anchoring to the lexical embedding space during vector quantization, this model effectively transforms a non-linguistic modality, such as images, into a foreign language using the LLM's vocabulary.
% By organizing tokens in a coarse-to-fine structure, this interpretable multi-modal lexical representation captures both semantic meaning and visual details, which enables visual reconstruction and various multi-modal tasks.

In \cref{chap:spae}, a novel strategy is introduced, which involves the mapping of visual data into the latent space of a pre-trained \acrshort{LLM}. This model achieves its transformation by utilizing lexical token embeddings from the \acrshort{LLM} during the process of vector quantization. This mechanism adeptly converts non-linguistic modalities, like images, into a distinct language using the vocabulary of the \acrshort{LLM}.
By adopting a hierarchical arrangement of tokens from broad to intricate, this interpretable \emph{visual lexical representation} effectively encompasses both semantic significance and visual intricacies. This holistic approach facilitates visual reconstruction and empowers the performance of various multi-modal tasks.

% In \cref{chap:magvitv2}, we reflect on the lessons learned in \cref{chap:3dvq,chap:spae} and propose the design of a \emph{scalable visual token} representation learning technique with highly compressed latent spaces.
% By leveraging unprecedented vocabulary sizes with lookup-free quantization and scaled causal architectures for joint image-video tokenization, this model excels at visual \emph{compression, generation, and understanding} tasks.
% With this method, we present the first evidence showing that \acrshortpl{LLM} can outperform diffusion models on visual synthesis.
% In addition, this is the first time where a visual tokenizer designed for video generation achieves comparable results to standard codecs such as \acrshort{HEVC} and \acrshort{VVC}.

In \cref{chap:magvitv2}, we delve into an introspective examination of the insights garnered from the explorations in \cref{chap:3dvq,chap:spae}, setting the stage for introducing an innovative \emph{scalable visual token} representation learning approach. This approach marks a departure from traditional methods by integrating large vocabularies with a novel lookup-free quantization process and leveraging scaled causal architectures that facilitate the joint tokenization of images and videos. The proficiency of this model in visual \emph{generation, compression, and understanding} appears favorable against existing designs. Significantly, it presents the first evidence of \acrshortpl{LLM} surpassing diffusion models in visual synthesis tasks. Moreover, it pioneers in demonstrating that a visual tokenizer, specifically tailored for video content generation, can achieve performance on par with, if not better than, established codecs such as \acrshort{HEVC} and \acrshort{VVC}.

\paragraph{\cref{part:model}: \nameref{part:model}.}
% With the high-quality learned representations from \cref{part:latent}, we can faithfully build latent generative models that perceive, understand, and recreate the world.
% In this part, we focus on the design of data modeling techniques and task formats.
% In particular, we introduce novel approaches to enable multi-task learning with a single model, which is a key capability for general intelligence.
Harnessing the acquired high-fidelity representations detailed in \cref{part:latent}, we possess the capacity to construct latent generative models that adeptly perceive, comprehend, and replicate the intricacies of the world.
Within this section, our concentration is directed toward formulating techniques for data modeling and shaping task structures. Notably, we present methodologies tailored to facilitate multi-task learning using a solitary model. 

% In this part, we focus on the latent generative models and their task designs.

% \cref{chap:magvit} introduces a multi-task video generation model based on masked transformers.
% With the spatial-temporal vector-quantized representation from \cref{chap:3dvq}, we model a video as a sequence of visual tokens in the latent space.
% We propose an effective embedding method for masked video token modeling to facilitate multi-task learning.
% This model achieves superior video generation quality with high sampling efficiency and task flexibility. 

In \cref{chap:magvit}, we unveil a multi-task video generation model, leveraging the capabilities of \emph{masked generative transformers}. By utilizing the spatial-temporal vector-quantized representation detailed in \cref{chap:3dvq}, videos are conceptualized as sequences of visual tokens within the latent space. To enrich the landscape of multi-task learning, an effective embedding technique for masked video token modeling is introduced.
% Without any changes, a single model supports arbitrary conditional video generation tasks that take a subset of the pixels or an embedding as input.
% While demonstrating fascinating adaptability to various tasks, this model achieves exceptional video generation quality and remarkable sampling efficiency.
Remarkably, a single model, with no alterations, supports an array of conditional \emph{video generation} tasks, encompassing scenarios where input involves a subset of pixels or an embedding. This model not only exhibits an adaptability spectrum across diverse tasks but also attains a favorable level of video generation quality, alongside an efficient sampling process.

% In \cref{chap:llm}, we tackle the generation of \emph{text, image, and video} using a frozen LLM empowered with the visual lexical representation from \cref{chap:spae}.
% We propose a progressive in-context learning approach to enable frozen LLMs to perform both \emph{understanding and generation} tasks involving non-linguistic modalities such as images or videos.
% Even without updating any parameters of the LLM, it is able to perform image and video tasks such as classification, captioning, visual question answering, text-to-image, and frame prediction.
In \cref{chap:llm}, we delve into the realm of generating \emph{video, image, and text} through a frozen \Gls{LLM}, fortified by the visual lexical representation introduced in \cref{chap:spae}. Our approach introduces a progressive in-context learning methodology, empowering static \Glspl{LLM} to proficiently undertake both \emph{generation and understanding} tasks spanning non-linguistic domains, including images and videos.
Remarkably, even without any updates to the \Gls{LLM}'s parameters, it showcases prowess in image and video tasks such as classification, captioning, visual question answering, text-to-image, and frame prediction.

In \cref{chap:videopoet}, our exploration advances as we develop \emph{scalable generative multi-modal transformers} from the ground up, utilizing the scalable representation conceptualized in \cref{chap:magvitv2}. This development employs modality-specific discrete tokenization to cohesively integrate text, images, videos, and audio within a decoder-only, transformer-based framework akin to \Glspl{LLM}. By pretraining this model on a broad array of multi-modal generative tasks using the established \Gls{LLM} training methodologies, we endow the model with robust capabilities for multi-task video generation. Notably, this model represents a pioneering achievement in its ability to generate high-quality videos, complete with corresponding audio, based on a wide range of input signals.

\section*{Thesis Statement}\label{sec:stmt}
\addcontentsline{toc}{section}{Thesis Statement}

In this thesis, we build multi-task models for generating videos and other modalities under diverse conditions, as well as for understanding and compression applications.

We show the effectiveness of
\begin{enumerate}[(A)]
\item \emph{integrating multiple tasks}\\
      into a single framework for understanding and generation;
\item \emph{crafting high-fidelity latent representation} \\
      for visual data in a discrete space, optionally into text tokens; and
\item \emph{generating multiple modalities} \\
      from a shared latent space, through a unified interface, and by a single model.
\end{enumerate}
\cref{tab:org} lists the highlighted component in each chapter.

\glsresetall
\cleardoublepage
\part{Prototypes}\label{part:pre}
\paragraph{\cref{part:pre} Overview.}

\glsresetall
\cleardoublepage
\chapter{Multi-Task Video Understanding}\label{chap:argus}
\graphicspath{{papers/argus/images/}}

\paragraph{Overview.}
% Activity detection is one of the attractive computer vision tasks to exploit the video streams captured by widely installed cameras.
% Although achieving impressive performance, conventional activity detection algorithms are usually designed under certain constraints, such as using trimmed and/or object-centered video clips as inputs.
% Therefore, they failed to deal with the multi-scale multi-instance cases in real-world unconstrained video streams, which are untrimmed and have large field-of-views.
% Real-time requirements for streaming analysis also mark brute force expansion of them unfeasible.

Activity detection stands out as a captivating computer vision endeavor that capitalizes on video streams garnered from extensively deployed cameras. Despite achieving commendable results, traditional activity detection algorithms are often formulated within specific limitations. For instance, they tend to operate with trimmed or object-focused video clips as inputs. Consequently, these algorithms struggle to effectively address scenarios involving multiple scales and instances within real-world, unconstrained video streams. Such streams remain untrimmed and encompass wide field-of-views. Moreover, the necessity for real-time analysis of streaming data renders the straightforward expansion of these methods impractical.

To overcome these issues, we propose Argus++, a robust real-time \emph{multi-task} activity detection system for analyzing unconstrained video streams. 
The design of Argus++ introduces overlapping spatial-temporal cubes as an intermediate concept of activity proposals to ensure coverage and completeness of activity detection through over-sampling.
The overall system is optimized for real-time processing on standalone consumer-level hardware.
Extensive experiments on different surveillance and driving scenarios demonstrated its favorable performance in a series of activity detection benchmarks, including CVPR ActivityNet ActEV 2021, NIST ActEV SDL UF/KF, TRECVID ActEV 2020/2021, and ICCV ROAD 2021.

% We propose an action recognition system for surveillance scenarios, which wins TRECVID 2020~\cite{2020trecvidawad} Activities in Extended Video (ActEV\footnote{\url{https://actev.nist.gov}}) Challenge with a large advantage of 23.8\%~\cite{yucmu} ahead the runner up system. 
% Our system develops a dense spatial-temporal proposal generation model which collaborates with the state-of-the-art action classifiers. The proposed system utilizes multiple state-of-the-art modules and is trained on VIRAT Dataset with only released annotations. In this paper, we demonstrate the architecture and algorithms with technique details of the winner system.

\clearpage

\section{Motivation}
Nowadays, activity detection has drawn a fast-growing attention in both industry and research fields. Activity detection in extended videos \cite{corona2021meva,oh2011large} is widely applied for public safety in indoor and outdoor scenarios. Activity detection on streaming videos captured by in-vehicle cameras is applied for vision-based autonomous driving. The development of these applications brings several challenges. First, most of these systems take \textit{unconstrained} videos as input, which are recorded in large field-of-views where multi-object and multi-activity occur simultaneously and continuously over time. Second, the unconstrained videos in real world are in multiple scenarios and under multiple conditions, e.g. in dynamically changed road environments from day to night in autonomous driving \cite{singh2022road}.  Third, efficient algorithms are 
demanded for real-time processing and responding of streaming video.

Conventional activity detection works \cite{tran2018closer,feichtenhofer2020x3d,wang2016temporal,kay2017kinetics,ghadiyaram2019large} have achieved impressive performance. However, they are not suitable for real world unconstrained video understanding. Most of these works are applied under certain constraints, e.g., only for processing trimmed and/or object-centered video clips. Meanwhile, they usually are specified for certain scenarios, such as person activity, etc. Therefore, such algorithms would fail when being transferred to unconstrained videos on both efficiency and effectiveness.

Previous works \cite{rizve2021gabriella,yu2020cmu,liu2020argus} on unconstrained video analysis proposed to generate and analyze tube/tubelet proposals, which are trajectories extracted from object detection and tracking results. Tube proposal has several drawbacks. 
First, tube proposals failed to capture the trace of moving objects when cropping the proposals from the original videos. Therefore, learning the activities highly relied on trace would be difficult, e.g. 'vehicle turning right'.
Second, the tube proposals still cannot stay away from temporal activity localization to determine the existence of the activities.
Besides, most of the previous works \cite{rizve2021gabriella} utilize non-overlapping proposals, which straightforwardly cuts the tube proposals by fixed length of temporal windows. Inevitably, such methods destroy the completeness of activities. Therefore, it would result in significant degrade of performance.
Third, the objects in the tube proposal will suffer from the bounding box shift and distortion across frames, which could result in a high false alarm rate on activity detection.

To overcome the aforementioned challenges,  we propose \textit{Argus++}, an efficient robust spatial-temporal activity detection system for extended and road video activity detection. The proposed system contains four-stages: Proposal Generation, Proposal Filtering, Activity Recognition and Activity Deduplication.  The major difference between \textit{Argus++} and the former works, such as \cite{liu2020argus}, is the concept of \textit{cube} proposals. Rather than simply adapted tube proposals, i.e. cropped trajectories of detected and tracked objects, we propose to merge and crop the area of detected objects across the frames.

We summarize the contributions of this chapter as follows:
\begin{itemize}

    \item We propose \textit{Argus++}, a real-time activity detection system for unconstrained video streams, which is robust across different scenarios.
    \item We introduce overlapping spatial-temporal cubes as the core concept of activity proposals to ensure coverage and completeness of activity detection through over-sampling.
    \item The proposed system has achieved favorable performance in a large series of activity detection benchmarks, including CVPR ActivityNet ActEV 2021, NIST ActEV SDL UF/KF, TRECVID ActEV 2020/2021, and ICCV ROAD 2021.
\end{itemize}

\section{Prior Work}
\paragraph{Object detection and tracking.}
Object detection and tracking are fundamental computer vision tasks that aims to detect and track objects from images or videos. 
Image-based object detection algorithms, such as Faster R-CNN~\cite{ren2015faster} and R-FCN~\cite{dai2016r}, have demonstrated convincing performance but are often expensive to apply on every frame. 
Video-based object detection algorithms~\cite{zhu2017flow, qian2020adaptive} use optical flow guided feature aggregation to leverage motion information and reduce computation.
With the deep features extracted from the backbone convolutional network, multi-object tracking algorithms~\cite{wojke2017simple, wang2020towards} associates objects across frames based on feature similarity and location proximity.

\paragraph{Activity detection.}
In recent years, there emerged some systems designed for spatial-temporal activity detection on unconstrained videos~\cite{rizve2021gabriella,yu2020cmu,liu2020argus,chang2019mmvg, yu2019training, yu2018traffic, qian2022rethinking}. 
Generally, theses systems first generates activity proposals and then feeds them to classification models.
Since there have been a variety of video classification networks~\cite{tran2018closer, lin2019tsm, feichtenhofer2020x3d}, the major focus is on the paradigm of proposals and the generation algorithm.
In \cite{liu2020argus, chang2019mmvg}, a detection and tracking framework is employed to extract whole object tracklets as tubelets, where temporal localization is required.
In \cite{rizve2021gabriella}, an encoder-decoder network is used to generate localization masks on fixed-length clips for tubelet proposal extraction, which has varied spatial locations in different frames.

% Tube proposal has several drawbacks. 
% First, tube proposals failed to capture the trajectories of moving objects when cropping the proposals from the original videos. Therefore, learning the activities highly related to trajectory of objects would be difficult, e.g. 'vehicle turning right'.
% Second, in action recognition task, such tube proposals still require temporal activity localization in the later stage to determine the existence of the activities on video clips, which would be more difficult than classification on fix length video clips. 
% Third, the objects in the tube proposal will suffer from the bounding box shift and distortion across frame by frame processing. It will bring negative effect on visual feature extraction, which could result in a high false alarm rate on action recognition.

\section{Argus++ Activity Detection System}
\begin{sidewaysfigure}[tp]
	\centering
	\includegraphics[width=\linewidth]{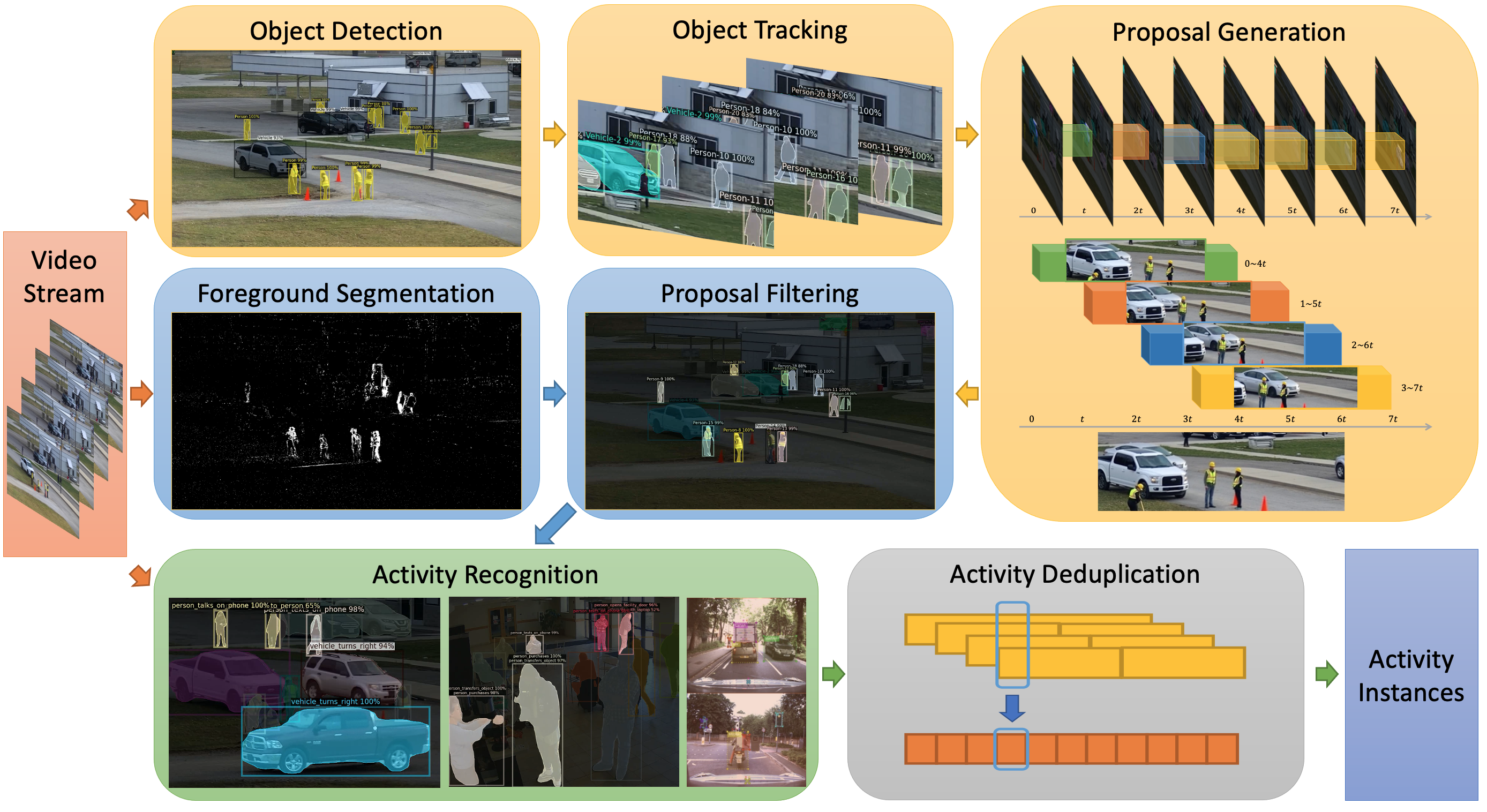}
	\caption[Architecture of Argus++]{\textbf{Architecture of Argus++}. A video stream is processed frame-by-frame through object detection and tracking to generate overlapping cube proposals. With frame-level foreground segmentation, stable proposals are filtered out. Activity recognition models determine the classification scores for each proposal. These over-sampled cubes are deduplicated to produce the final activity instances. }
	\label{fig/arch}
\end{sidewaysfigure}

% \subsection{Activity Detection in Unconstrained Video Streams}

We tackle the activity detection task in unconstrained videos which are untrimmed and with large field-of-views.
Given an untrimmed video stream $\mathcal{V}$, the system $\mathcal{S}$ should identify a set of activity instances $\mathcal{S}(\mathcal{V})=\{A_i\}$.
Each activity instance is defined by a three-tuple $A_i=(T_i, L_i, C_i)$, referring to an activity of type $C_i$ occurs at temporal window $T_i$ with spatial location $L_i$.
$L_i$ contains the precise location of $A_i$ in each frame, forming a tube in the timeline.
As such, activity detection can often be decomposed into three aspects, i.e., temporal localization ($T_i$), spatial localization ($L_i$), and action classification ($C_i$). 

Each of the three aspects poses unique challenges to the video understanding system. Due to its multi-dimensional nature, it remains hard to define and build a useful activity detection system under the strict setting. Therefore, we also evaluates with some loosened requirements.

\paragraph{Strict setting.}

All activity types are defined as atomic activities with clear temporal boundaries and spatial extents.
The evaluation metric performs bipartite matching between predictions and ground truths.

\paragraph{Loosened setting.}

Activity types are either atomic activities within a temporal window (e.g. standing up) or continuous repetitive activities that can be cut into multiple identifiable windows (e.g. walking).
The evaluation metric allows multiple non-overlapping predictions to be matched with one ground truth.

\subsection{Argus++ System}

The architecture of the proposed \textit{Argus++} system is shown in Figure \ref{fig/arch}.
To tackle the task of activity detection, we adopt an intermediate concept of \textit{spatial-temporal cube proposal} with a much simpler definition than an activity instance:
\begin{equation}
    p_i = (x_0^i, x_1^i, y_0^i, y_1^i, t_0^i, t_1^i).
\end{equation}
This six-tuple design relieves the localization precision and caters modern action classification models which works on fixed-length clips with fixed spatial window.

For an input video stream, the system first generates candidate proposals with frame-wise information such as detected objects, which will be covered in Section \ref{sec/prop_gen}.
These proposals are filtered with a background subtraction model as detailed in Section \ref{sec/prop_fil}.
Then, action recognition models described in Section \ref{sec/act_rec} are applied on the proposals to predict per-class confidence scores.
Finally, Section \ref{sec/act_ded} introduces the post-processing stage to merge and filter the proposals with scores and generate final activity instances.

\subsection{Proposal Generation} \label{sec/prop_gen}

Starting this section, we introduce each of the components of \textit{Argus++}.
The system begins by generating a set of cube proposals.
They are generated based on information from frame-level object detection with multiple object tracking methods. 
Cubes are sampled densely in the timeline with refined spatial locations.

\paragraph{Detection and tracking.}
To conduct activity recognition, we first locate the candidate objects (in most cases, person and vehicle) in the video. 
For each selected frame $F_i$, we apply an object detection model to get objects $O_i=\{o_{i, j} \mid j=1, \cdots, n_i\}$ with object types $c_{i, j}$ and bounding boxes $(x_0, x_1, y_0, y_1)_{i, j}$.
Objects are detected in a stride of every $S_\mathit{det}$ frames.
A multiple object tracking algorithm is applied on the detected objects to assign track ids to each of them as $\mathit{tr}_{i, j}$.

\begin{figure}[tp]
	\centering
	\includegraphics[width=0.4\linewidth]{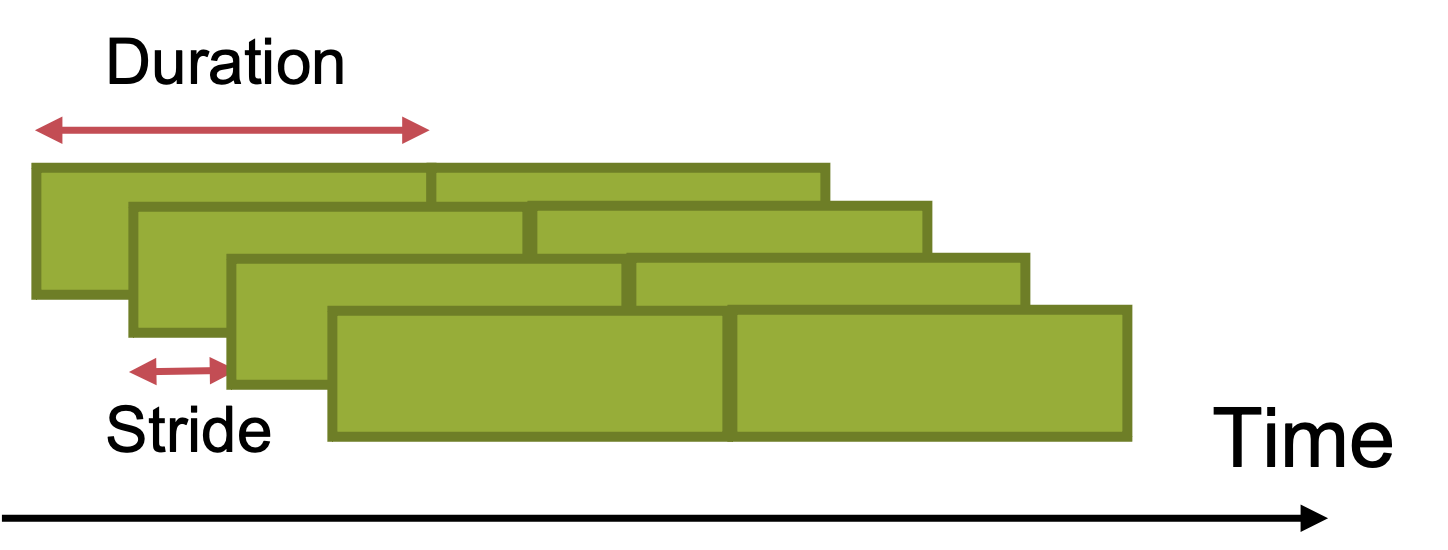}
	\caption[Dense overlapping proposals]{\textbf{Dense overlapping proposals}.}
	\label{fig/cube}
\end{figure}

\paragraph{Proposal sampling.}
To sample proposals on untrimmed videos without breaking the completeness of any activity instances, we propose a dense overlapping proposals sampling algorithm. 
As illustrated in Figure \ref{fig/cube}, this method ensures coverage of activities occurring at any time, with no hard boundaries.
Two parameters, duration $D_\mathit{prop}$ and stride $S_\mathit{prop}$, controls the sampling process.
Each proposal contains a temporal window of $D_\mathit{prop}$ frames.
New proposals are generated every $S_\mathit{prop} \le D_\mathit{prop}$ frames, possibly with overlaps. 
Generally, non-overlapping proposal system can be treated as a degraded case when $S_\mathit{prop} = D_\mathit{prop}$.

\paragraph{Proposal refinement.}
To generate proposals in a temporal window from $t_0$ to $t_1=t_0 + D_\mathit{prop}$, we select seed track ids $\mathit{Tr}_{t_c}$ from the central frame $t_c = \lfloor\frac{t_0 + t_1}{2}\rfloor$.
Their bounding boxes are enlarged as the union across the temporal window
\begin{equation}
    (x_0, x_1, y_0, y_1)_k = \bigcup(\{(x_0, x_1, y_0, y_1)_{i, j} \mid
     t_0 \le i \le t_1, \mathit{tr}_{i, j} = \mathit{tr}_{t_c, k}\}),
\end{equation}
where $k = 1, \cdots, n_{t_c}$.
This algorithm is robust through identity switch in the tracking algorithm as it uses the stable seeds from the central frame.
It also ensures the coverage of moving objects by enlarging the bounding box when it's successfully tracked.
This design is helpful for efficiency optimization by allowing a large detection stride $S_\mathit{det}$.
When later applied for activity recognition, the bounding box can be further enlarged for a fixed rate $R_{enl}$ to include spatial context and compensate for missed tracks.

\subsection{Proposal Filtering} \label{sec/prop_fil}

For now, the proposal generation pipeline applies a frame-wise object detection with slight aid of tracking information. 
The motion information of video is not yet explored.
To produce high quality proposals, we apply a proposal filtering algorithm to eliminate the proposals that are unlikely to contain activities.

\paragraph{Foreground segmentation.}
For each proposal, a foreground segmentation algorithm is implemented to generate a binary mask for every $S_{bg}$ frames for each video clip. 
We average the value of pixel masks in its cube to get its foreground score $f_i$. 
For proposals generated by object type $c$, those proposals with $f_i \le F_c$ will be filtered out.
The threshold $F_c$ is determined by allowing up to $P_\mathit{pos}$ true proposals to be filtered out.

\paragraph{Label assignment.}
To determine the above threshold and to train the activity recognition module, we need to assign labels for each generated proposal according to the ground truth activity instances.
We first convert the annotation of activity instances into the cube format, denoted as ground truth cubes, by performing dense sampling of duration $D_\mathit{prop}$ and stride $S_\mathit{prop}$ within each instance.
For each proposal, we estimate the spatial intersection-over-union (IoU) between it and ground truth cubes in the same temporal window. 
Then we follow Faster R-CNN~\cite{ren2015faster} in the assignment process:
\begin{itemize}
    \item For each ground truth cube, assign it to the proposal with the highest score above $S_{low}$.
    \item For each proposal, assign it with each ground truth cube with score above $S_{high}$.
    \item For each proposal, assign it as negative if all scores are below $S_{low}$.
\end{itemize}
$S_{high}$ and $S_{low}$ are the high and low thresholds.
Through this algorithm, each proposal may be assigned one or more positive labels, a negative label, or nothing.
Those assigned nothing are redundant detections which will not be used in classifier training.

\paragraph{Proposal evaluation.}
To measure the quality of proposals before and after the filtering, we need a method for proposal evaluation.
This can be achieved by assuming a perfect classifier in the activity recognition part, so the final metrics reflects the upper bound performance with current proposals.
To do this, we simply use the assigned labels as the classification outputs and pass through the deduplication algorithm covered later.
To further measure other properties of the generated proposals, we can only pass through a subset of them, such as only those with spatial IoU against ground truth above 0.5.

\subsection{Activity Recognition} \label{sec/act_rec}
In this section, we will elaborately introduce our action recognition modules. Given the input proposal of an activity instance $p_i$, our action recognition model $\mathbb{V}$ will give out the confidence vector $c_i$:
\begin{equation}
    \mathbb{V}(p_i) = c_i = \{c^1_i,c^2_i,...c^n_i\},
\end{equation}
where n represents the number of target actions, and $c_i \in \mathbb{R}^{n}$. Limited by GPU memory size and temporal length settings of pretrained weights, we need to select t frames out of $t_1^i-t_0^i$ samples from the activity instance. To do this, we strictly followed the sparse-sampling strategy mentioned in \cite{wang2016temporal} for both training and inference stage. To be specific, the video is evenly separated into t segments. From each segment, 1 frame will be randomly selected to generate the sampled clip. 

To transform the action recognition modules from previous multi-class task to the realm of multi-label recognition, we modified the loss function for optimization. Instead of traditional cross entropy loss, we implemented a weighted binary cross entropy loss (wBCE), in which two weight parameters are adopted, the activity-wise weight $\text{W}_a = \{w_a^1,w_a^2,...,w_a^n\}$ and the positive-negative weight $\text{W}_p = \{w_p^1,w_p^2,...,w_p^n\}$. $\text{W}_a$ balances the training samples of different activities and $\text{W}_p$ balances the positive and negative samples of a specific activity. 
With the aligned label sequence of $i^{th}$ instance represented as $Y_i=\{y_i^1,y_i^2,...,y_i^{n}\} \in \mathbb{R}^n$. The calculation of $w_a^{c}$ is derived as:
\begin{equation}
    \hat{w}_a^c =\frac{1}{\sum_{i\in [I]}y_i^c},
\end{equation}
\begin{equation}
    w_a^{c} = n \times \frac{\hat{w}_a^c}{\sum_{c\in[n]}\hat{w}_a^c}.
\end{equation}
And the derivation of $w_p^{c}$ is:
\begin{equation}
    w_p^{c} =\frac{\sum_{i\in [I]}\mathbf{1}_{y_i^c=0}}{\sum_{i\in [I]}y_i^c}.
\end{equation}
% \begin{equation}
%     w_p^{c} = \min(1000,\hat{w_p^{c}}) move to exp
% \end{equation}
In which, $[I]$ represents all input instances, and $[n]$ represent all target activities.
Compared with vanilla BCE loss, we found wBCE loss can significantly improve the final performance on the validation set.

Furthermore, we tried multiple action recognition modules and made late fusion action-wisely according to the results on the validation set. We found each classifier does show superiority on certain actions. Through the feedback from the online leaderboard,  such fusion strategy can improve the final performance with noticeable margins.

% \begin{table*}[htb]
% \centering
% \caption{NIST ActEV Self-Reported Leaderboard (SRL) 2021\protect\footnotemark Evaluation}
% \label{exp/srl21}
% \begin{tabular}{@{}lc@{}}
% \toprule
% System/Team&$\mathit{nAUDC}@0.2T_\mathit{fa}\downarrow$\\ \midrule
%     \textbf{Argus++ (Ours)} & \textbf{0.1790}   \\
%     NIST Baseline & 0.3230   \\ 
%     \bottomrule
% \end{tabular}
% \end{table*}

% \footnotetext{\url{https://actev.nist.gov/SRL\#tab_leaderboard}}
\begin{figure}[tp]
	\centering
	\includegraphics[width=\linewidth]{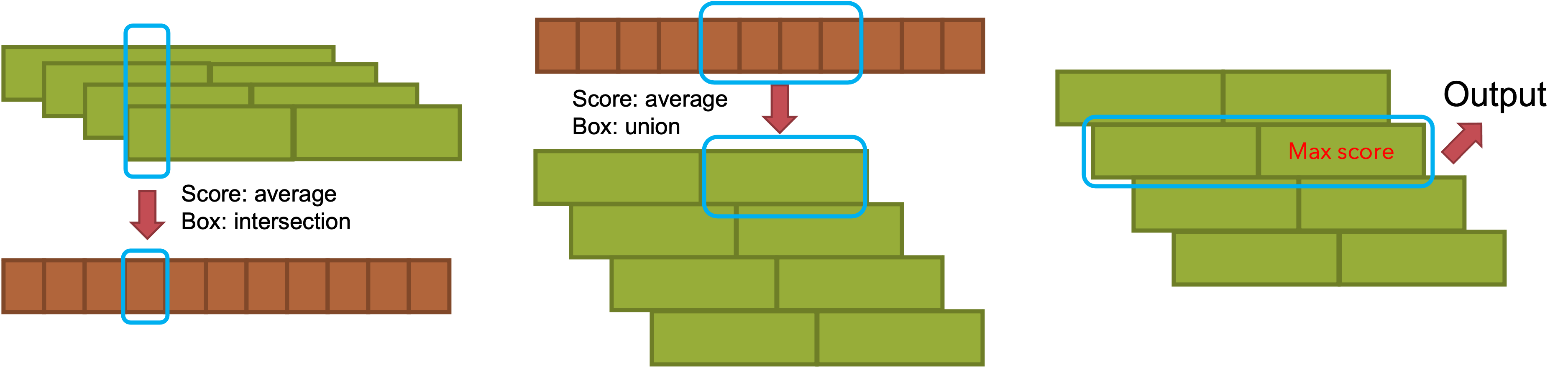}
	\caption[Deduplication algorithm for overlapping proposals]{\textbf{Deduplication algorithm for overlapping proposals}.}
	\label{fig/merge}
\end{figure}

\subsection{Activity Deduplication} \label{sec/act_ded}

\paragraph{Overlapping instances.}

As the system generates overlapping proposals, it could have duplicate predictions for some of the proposals.
This would result in a large amount of false alarms unless we deduplicate them.
Figure \ref{fig/merge} is a diagram for our deduplication algorithm which applies to each activity type with all proposals: 
\begin{enumerate}
    \item Split the overlapping cubes of duration $D_\mathit{prop}$ and stride $S_\mathit{prop}$ into non-overlapping cubes of duration $S_\mathit{prop}$.
    An output cube relies on all original cubes in the temporal window, with an averaged score and an intersected bounding box.
    \item Merge the non-overlapping cubes of duration $S_\mathit{prop}$ back into $\lfloor\frac{D_\mathit{prop}}{S_\mathit{prop}}\rfloor$ groups of non-overlapping cubes of duration $D_\mathit{prop}$.
    An output cube is merged from $\lfloor\frac{D_\mathit{prop}}{S_\mathit{prop}}\rfloor$ cubes with an averaged score and the union of bounding boxes.
    \item Select the group where the maximum score resides.
\end{enumerate}

The deduplication algorithm performs an interpolation upon the overlapping cubes.
Each group in step 3 contains information from every classification results, maximizing the information utilization.

\paragraph{Adjacent instances.}
The above deduplication process only transforms overlapping instances to non-overlapping instances with the same duration.
This would be sufficient under the \textit{Loosened Setting}, where multiple predictions are allowed for each activity.
No threshold would be needed to truncate low-confidence predictions as this happens automatically during the ground-truth matching process.

However, for the \textit{Strict Setting}, we need to further merge adjacent cubes into integrate instances.
Currently we adopt a simple yet effective algorithm, by simply merging adjacent cubes where all of them have confidence score above $S_{merg}$. 
The merged instance needs to be longer than $L_{merg}$ to be kept in the final output.

\section{Experimental Results}
\subsection{Implementation Details}

In \textit{Argus++}, we apply Mask R-CNN~\cite{he2017mask} with a ResNet-101~\cite{he2016deep} backbone from Detectron2~\cite{wu2019detectron2} pre-trained on the Microsoft COCO dataset~\cite{lin2014microsoft} as the object detector, with $S_\mathit{det}=8$. Only person, vehicle, and traffic light classes are selected.
For the tracking algorithm, we apply the work in \cite{wang2020towards} and reuse the region-of-interest feature from the ResNet backbone as in \cite{yu2020zero, qian2020electricity}.

The proposals are generated with $D_\mathit{prop}=64$ and $S_\mathit{prop}=16$.
The labels are assigned with $S_{high}=0.5$ and $S_{low}=0$.
The proposal filter is set with a tolerance of $P_\mathit{pos}=0.05$.

For activity classifiers, we adopted multiple state-of-the-art models including R(2+1)D~\cite{tran2018closer}, X3D~\cite{feichtenhofer2020x3d}, and TRM~\cite{qian2022trm}.
During training, frames are cropped with jittering~\cite{wang2016temporal} and enlarged with $R_{enl}=0.13$. For X3D and TRM, we trained modules with weights pre-trained on Kinetics~\cite{kay2017kinetics}. For R(2+1)D modules, we trained modules with weighst pre-trained on IG65M~\cite{ghadiyaram2019large}. 
We fused confidence scores from these models according to their performance on the validation set.

\subsection{Evaluation Protocols}

To measure the performance, efficiency, and generalizability of \textit{Argus++}, we evaluate it across a series of public benchmarks.
\textit{Argus++} is applied to NIST Activities in Extended Videos (ActEV) evaluations on MEVA~\cite{corona2021meva} Unknown Facility , MEVA Known Facility, and VIRAT~\cite{oh2011large} settings for surveillance activity detection. 
With slight modifications, it is also tested in the ICCV 2021 ROAD challenge for the action detection task in autonomous driving.

In the NIST evaluations, the metrics~\cite{awad2021trecvid} are designed in the \textit{Loosened Setting}, where short-duration outputs are allowed and spatial alignment is ignored.
The idea is that, after processed by the system, there will still be human reviewers to inspect the activity instances with the highest confidence scores for further usages.
The performance is thus measured by the probability of miss detection ($P_{miss}$) of activity instances within a time limit of all positive frames plus $T_\mathit{fa}$ of negative frames, where $T_\mathit{fa}$ is referred to as time-based false alarm rate.
The major metric, $\mathit{nAUDC}@0.2T_\mathit{fa}$, is an integration of $P_{miss}$ on $T_\mathit{fa} \in [0, 0.2]$.

In the ROAD challenge, the \textit{Strict Setting} is adopted by using the mean average precision (mAP) at 3D intersection-over-union (IoU) evaluation metric.
This metric does exact bipartite matching between predictions and ground truth instances, with challenging localization precision requirements.

For metrics in the following tables, $\downarrow$ means lower is better and $\uparrow$ means higher is better. For each metric, the best value is bolded and the second best is underscored. For ongoing public evaluations, the result snapshot at 11/01/2021 is presented.

\subsection{Public Benchmarks}

\begin{table*}[tp]
\centering
\begin{tabular}{@{}lccc@{}}
\toprule
System/Team&$\mathit{nAUDC}@0.2T_\mathit{fa}\downarrow$&Mean$P_{miss}@0.02T_\mathit{fa}\downarrow$&Relative Processing Time\\ \midrule
    \textbf{Argus++ (Ours)} & \textbf{0.3535} & \textbf{0.5747} & 0.576  \\
    UMD\_JHU & \underline{0.4232} & 0.6250 & 0.345  \\ 
    IBM-Purdue & 0.4238 & 0.6286 & 0.530 \\
    UCF & 0.4487 & \underline{0.5858} & 0.615 \\
    Visym Labs & 0.4906 & 0.6775 & 0.770 \\
    MINDS\_JHU & 0.6343 & 0.7791 & 0.898 \\
    \bottomrule
\end{tabular}
\caption[CVPR 2021 ActivityNet challenge ActEV SDL unknown facility evaluation]{\textbf{CVPR 2021 ActivityNet challenge ActEV SDL unknown facility evaluation results}.}
\label{exp/activitynet21}
\end{table*}

% \footnotetext{\url{http://activity-net.org/challenges/2021/challenge.html}}

\begin{table*}[tp]
\centering
\begin{tabular}{@{}lccc@{}}
\toprule
System/Team&$\mathit{nAUDC}@0.2T_\mathit{fa}\downarrow$&Mean$P_{miss}@0.02T_\mathit{fa}\downarrow$&Relative Processing Time\\ \midrule
    \textbf{Argus++ (Ours)} & \textbf{0.1635} & \textbf{0.3424} & 0.413  \\
    UCF & \underline{0.2325} & \underline{0.3793} & 0.751  \\ 
    UMD & 0.2628 & 0.4544 & 0.380 \\
    IBM-Purdue & 0.2817 & 0.4942 & 0.631 \\
    Visym Labs & 0.2835 & 0.4620 & 0.721 \\
    UMD-Columbia & 0.3055 & 0.4716 & 0.516 \\
    UMCMU & 0.3236 & 0.5297 & 0.464 \\
    Purdue & 0.3327 & 0.5853 & 0.131 \\
    MINDS\_JHU & 0.4834 & 0.6649 & 0.967 \\
    BUPT-MCPRL & 0.7985 & 0.9281 & 0.123 \\
    \bottomrule
\end{tabular}
\caption[NIST ActEV'21 SDL known facility evaluation]{\textbf{NIST ActEV'21 SDL known facility evaluation results}.}
\label{exp/sdl_kf}
\end{table*}

% \footnotetext{\url{https://actev.nist.gov/sdl\#tab_leaderboard}}

\begin{table*}[tp]
\centering
\begin{tabular}{@{}lccc@{}}
\toprule
System/Team&$\mathit{nAUDC}@0.2T_\mathit{fa}\downarrow$&Mean$P_{miss}@0.02T_\mathit{fa}\downarrow$&Relative Processing Time\\ \midrule
    \textbf{Argus++ (Ours)} & \textbf{0.3330} & \underline{0.5438} & 0.776  \\
    UCF & \underline{0.3518} & \textbf{0.5372} & 0.684  \\ 
    IBM-Purdue & 0.3533 & 0.5531 & 0.575 \\
    Visym Labs & 0.3762 & 0.5559 & 1.027 \\
    UMD & 0.3898 & 0.5938 & 0.515 \\
    UMD-Columbia & 0.4002 & 0.5975 & 0.520 \\
    UMCMU & 0.4922 & 0.6861 & 0.614 \\
    Purdue & 0.4942 & 0.7294 & 0.239 \\
    MINDS\_JHU & 0.6343 & 0.7791 & 0.898 \\
    \bottomrule
\end{tabular}
\caption[NIST ActEV'21 SDL unknown facility evaluation]{\textbf{NIST ActEV'21 SDL unknown facility evaluation results}.}
\label{exp/sdl_uf}
\end{table*}

\begin{table*}[htb]
\centering
\begin{tabular}{@{}lccc@{}}
\toprule
System/Team&$\mathit{nAUDC}@0.2T_\mathit{fa}\downarrow$&Mean  $P_{miss}@0.15T_\mathit{fa}\downarrow$&Mean $wP_{miss}@0.15R_\mathit{fa}\downarrow$ \\\midrule
    \textbf{Argus++ (Ours)} & \textbf{0.39607} & \textbf{0.30622} & \underline{0.81080}  \\
    BUPT & \underline{0.40853} & \underline{0.32489} & \textbf{0.79798}  \\ 
    UCF & 0.43059 & 0.34080 & 0.86431 \\
    M4D & 0.84658 & 0.79410 & 0.88521 \\
    TokyoTech\_AIST & 0.85159 & 0.81970 & 0.94897 \\
    Team UEC & 0.96405 & 0.95035 & 0.95670 \\
    \bottomrule
\end{tabular}
\caption[NIST TRECVID 2021 ActEV evaluation]{\textbf{NIST TRECVID 2021 ActEV evaluation results}.}
\label{exp/trecvid21}
\end{table*}

\begin{table*}[htb]
\centering
\begin{tabular}{@{}lccc@{}}
\toprule
System/Team                                     & $\mathit{nAUDC}@0.2T_\mathit{fa}\downarrow$ & Mean $P_{miss}@0.15T_\mathit{fa}\downarrow$ & Mean $wP_{miss}@0.15R_\mathit{fa}\downarrow$ \\ \midrule
\textbf{Argus++ (Ours)} & \textbf{0.42307} & \textbf{0.33241} & \textbf{0.80965} \\
UCF                                      & \underline{0.54830}                           & 0.50285                           & \underline{0.83621}                           \\
BUPT-MCPRL                               & 0.55515                           & \underline{0.48779}                           & 0.84519                           \\
TokyoTech\_AIST                          & 0.79753                           & 0.75502                           & 0.87889                           \\
CERTH-ITI                                & 0.86576                           & 0.84454                           & 0.88237                           \\
Team UEC                                 & 0.95168                           & 0.95329                           & 0.98300                           \\
Kindai\_Kobe                             & 0.96267                           & 0.95204                           & 0.93905                           \\ \bottomrule
\end{tabular}
\caption[NIST TRECVID 2020 ActEV evaluation]{\textbf{NIST TRECVID 2020 ActEV evaluation results}.}
\label{exp/trecvid20}
\end{table*}

\paragraph{ActEV sequestered-data evaluations.}

ActEV Sequestered Data Leaderboards (SDL) are platforms where a system is submitted to run on NIST's evaluation servers.
This submission format prevents access to the test data and measures the processing time with unified hardware platform~\cite{leeactev}.
For these evaluations, \textit{Argus++} was trained on MEVA, a large-scale surveillance video dataset with activity annotations of 37 types.
We used 1946 videos in its training release drop 11 as the training set and 257 videos in its KF1 release as validation set.
The optimization target is reaching better performance within 1x real-time.

Table \ref{exp/activitynet21} shows the published results from CVPR 2021 ActivityNet Challenge ActEV SDL Unknown Facility evaluation, where \textit{Argus++} has around 20\% advantage in $\mathit{nAUDC}@0.2T_\mathit{fa}$ over runner-up system.
The test set of unknown facility is captured with a different setting from MEVA, which challenges the generalization of action detection models.
Table \ref{exp/sdl_kf} shows the ongoing NIST ActEV'21 SDL Known Facility leaderboard, where \textit{Argus++} shows over 40\% advantage in $\mathit{nAUDC}@0.2T_\mathit{fa}$.
The test set of known facility shares a similar distribution with MEVA, where our system learns well and is getting nearer for real-world usages.
Table \ref{exp/sdl_uf} shows the ongoing NIST ActEV'21 SDL Unknown Facility leaderboard continued from ActivityNet, where \textit{Argus++} still holds the leading position with over 5\% advantage in $\mathit{nAUDC}@0.2T_\mathit{fa}$.

\paragraph{ActEV self-reported evaluations.}

ActEV self-reported evaluations are where only results are submitted and test data is accessible.
This currently includes the annual TRECVID ActEV evaluations on VIRAT.
% For the ongoing Self-Reported Leaderboard (SRL) on MEVA, we 
% For SRL, we follow the same training procedure as SDL.
For TRECVID, we use the official splits of VIRAT for training and validation.

% Table \ref{exp/srl21} shows results on the ongoing ActEV Self-Reported Leaderboard. Compared with the baseline system of NIST, our submission achieves 44.5\% advantage in $\mathit{nAUDC}@0.2T_\mathit{fa}$.
Table \ref{exp/trecvid21} and \ref{exp/trecvid20} shows the leaderboard of 2020~\cite{awad2021trecvid, yu2020cmu} and 2021~\cite{2021trecvidawad, yu_cmu_2021} NIST TRECVID ActEV Challenge. In 2020, our systems is 22.8\% better in $\mathit{nAUDC}@0.2T_\mathit{fa}$, 33.8\% better in Mean $P_{miss}@0.15T_\mathit{fa}$, and 3.5\% better in Mean-$wP_{miss}@0.15R_\mathit{fa}$ than the runner-up. Although the other competitors improved significantly in 2021, our system still holds the first place with noticeable margins.

\paragraph{ROAD challenge.}
Different from previous surveillance action detection benchmarks, the videos of ROAD Challenge\cite{singh2022road} are gathered from the point of view of autonomous vehicles. 
It contains 122K frames from 22 annotated videos, where each video is 8 minutes long on average. 
Totally 7K tubes of individual agents are included and each tube consists on average of approximately 80 bounding boxes linked over time. 

Table \ref{exp/road} shows the performance of our system with other competitors. Our system ranks the first with 20\% average mAP.
Although the performance is still far from satisfying in this \textit{Strict Setting}, it demonstrates the capability of \textit{Argus++} in adapting to precise 3D localization and moving camera view points.

\begin{table*}[tp]
\centering
\begin{tabular}{@{}lcccc@{}}
\toprule
System/Team    & Action@0.1 $\uparrow$ & Action@0.2 $\uparrow$ & Action@0.5 $\uparrow$ & \textbf{Average} $\uparrow$ \\ \midrule
\textbf{Argus++ (Ours)} & \textbf{28.54}      & \textbf{25.63}      & 6.98       & \textbf{20.38}   \\
THE IFY        & \underline{28.15}      & \underline{20.97}      & 6.58       & \underline{18.57}   \\
YAAAHO         & 26.81      & 20.40      & \underline{7.02}       & 18.07   \\
hyj            & 26.52      & 20.32      & \textbf{7.05}       & 17.97   \\
3D RetinaNet   & 25.70      & 19.40      & 6.47       & 17.19   \\
LeeC           & 13.64      & 9.89       & 2.23       & 8.59   \\ \bottomrule
\end{tabular}
\caption[ICCV 2021 ROAD challenge action detection]{\textbf{ICCV 2021 ROAD challenge action detection results}.}
\label{exp/road}
\end{table*}

% \footnotetext{\url{https://eval.ai/web/challenges/challenge-page/1059/leaderboard/2748}}

\subsection{Ablation Study}

\paragraph{Coverage of proposal formats.}

We analyze the coverage of dense spatial-temporal proposals and determines the best hyper-parameters for the proposal format. 
By directly use ground truth cubes as proposals, we estimate the upper bound performance of both overlapping and non-overlapping proposal formats on VIRAT validation set.
The results are shown in Table \ref{tab/overlap}, where non-overlapping proposals shows at least 6.7\% systematic errors while overlapping proposals with duration 64 and stride 16 only has 1.3\%.

\begin{table}[p]
\centering
\begin{tabular}{@{}ccccc@{}}
\toprule
Duration / Stride & 16     & 32   & 64   & 96     \\ \midrule
32                          & 0.0705 & \textit{0.1208} & - & -       \\
64                          & \textbf{0.0127} & 0.0621 & \textit{0.0673} & -  \\
96                          & 0.0275 & 0.0504 & - & \textit{0.0688} \\ \bottomrule
\end{tabular}
\caption[Proposal lower bounds]{\textbf{Lower bounds of $\mathit{nAUDC}@0.2T_\mathit{fa}$ on VIRAT validation set with different proposal formats}. Italic values are non-overlapping proposals while the others are overlapping proposals. Duration and stride are in the unit of frames.}
\label{tab/overlap}
\end{table}

\begin{table}[p]
\centering
\begin{tabular}{@{}lcc@{}}
\toprule
Name                    & Unfiltered  & Filtered   \\ \midrule
Number of Proposals         & 211271      & 62831  \\
Positive rate            & 0.1704  & \textbf{0.5204} \\
Rate of unique label & 0.4558 & 0.4415 \\
Rate of two labels & 0.4127 & 0.4252 \\
Rate of three labels & 0.1017 & 0.1060 \\ \bottomrule
\end{tabular}
\caption[Statistics of proposals]{\textbf{Statistics of proposals on VIRAT validation set}.}
\label{tab/prop_sta}
\end{table}

\begin{table*}[p]
\centering
\begin{tabular}{@{}lcccccc@{}}
\toprule
$\mathit{nAUDC}@0.2T_\mathit{fa}$     & \multicolumn{3}{c}{IoU}       & \multicolumn{3}{c}{Reference Coverage}    \\
Threshold & Average & $\ge$ 0 & $\ge$ 0.5 & Average & $\ge$ 0.5 & $\ge$ 0.9 \\  \midrule
Unfiltered Proposals & 0.2358  & 0.0772  & 0.1518    & 0.1562  & 0.1125    & 0.4211    \\
Filtered Proposals   & 0.2352  & 0.0772  & 0.1469    & 0.1563  & 0.1099    & 0.4280    \\ \bottomrule
\end{tabular}
\caption[Proposal quality metrics]{\textbf{Proposal quality metrics on VIRAT validation set}.}
\label{tab/prop_qual}
\end{table*}

\begin{table}[p]
\centering
\begin{tabular}{@{}lcc@{}}
\toprule
Proposal Filter&$\mathit{nAUDC}@0.2T_\mathit{fa}\downarrow$& Processing Time\\ \midrule
    \textbf{Enabled} & \textbf{0.4822} & 0.582  \\ 
    Disabled & 0.5176 & 0.925 \\
    \bottomrule
\end{tabular}
\caption[Effect of proposal filter]{\textbf{Effect of proposal filter on NIST ActEV'21 SDL unknown facility micro set}.}
\label{exp/sdl_prop}
\end{table}

\paragraph{Performance of proposal filtering.}

We examine the quality of the proposals with and without the filter, as shown in Table \ref{tab/prop_sta} and \ref{tab/prop_qual}.
With the proposal evaluation procedure introduced in Section \ref{sec/prop_fil}, the proposals are further filtered by IoU with reference and coverage of reference at levels from 0, 0.1, to 0.9 to calculate partial results.

With the dense cube proposals, the best $\mathit{nAUDC}@0.2T_\mathit{fa}$ we can achieve with a ideal classifier is 0.08, as indicated in the IoU $\ge 0$ column.
The IoU and reference coverage bounded scores are used to measure the spatial matching quality of proposals, as the $\mathit{nAUDC}@0.2T_\mathit{fa}$ does not consider spatial alignments.
We can see that even with a condition of IoU $\ge 0.5$, our proposal can achieve up to 0.15, which indicates the spatial preciseness.
The proposal filter is also proved effective, which removed 70\% of original proposals without dropping the recall level.

The effect of the proposal filter is also evaluate on the SDL, as shown in Table \ref{exp/sdl_prop}.
It not only reduces processing time from 0.925 to 0.582, but also improves $\mathit{nAUDC}@0.2T_\mathit{fa}$ due to reduced false alarms.

\section{Summary}
In this chapter, we proposed Argus++, a robust real-time \emph{multi-task} activity detection system for analyzing unconstrained video streams. We introduced overlapping spatial-temporal cubes as an intermediate concept of activity proposals to ensure coverage and completeness of activity detection through over-sampling. The proposed system is able to process unconstrained videos with robust performance across multiple scenarios and real-time effiency on consumer-level hardware. 
Extensive experiments on different surveillance and driving scenarios demonstrated its favorable performance in a series of activity detection benchmarks, including CVPR ActivityNet ActEV 2021, NIST ActEV SDL UF/KF, TRECVID ActEV 2020/2021, and ICCV ROAD 2021.

Future works are suggested to focus on extending the current system to more applications, such as action detection in UAV captured videos, first-person human activity understanding, etc. The proposed system could also be extended to end-to-end frameworks for better performance.

\glsresetall
\cleardoublepage
\chapter{Masked Multi-Modal Pre-Training}\label{chap:doc}
\graphicspath{{papers/documentnet/figures/}}

\renewcommand{\dataname}{DocumentNet}
\renewcommand{\modelname}{UniFormer}
\renewcommand{\taskname}{\acrshort{VDER}}

\paragraph{Overview.}
Document understanding tasks, in particular, \Gls{VDER}, have gained significant attention in recent years thanks to their broad applications in enterprise AI.
However, publicly available data have been scarce for these tasks due to strict privacy constraints and high annotation costs.
% as document images often contain personally identifiable information, publicly available data have been scarce, not only because of privacy constraints but also the costs of acquiring annotations.
To make things worse, the non-overlapping entity spaces from different datasets hinder the knowledge transfer between document types.
In this chapter, we propose a method to collect massive-scale and weakly labeled data from the web to benefit the training of \taskname{} models. 
% Such a method will generate a huge amount of document image data to compensate for the lack of training data in many VDER settings. Moreover, 
The collected dataset, named \dataname{}, does not depend on specific document types or entity sets, making it universally applicable to all \taskname{} tasks.
The current \dataname{} consists of 30M documents spanning nearly 400 document types organized in a four-level ontology.
Empowered by \dataname{}, we present a lightweight \emph{multi-modal} architecture named \modelname{}, which can learn a unified representation from text, layout, and image crops without needing extra visual pretraining.
Experiments on a set of broadly adopted \taskname{} tasks show significant improvements when \dataname{} is incorporated into the pre-training.
%  for both classic and few-shot learning settings.
With the recent emergence of \Glspl{LLM}, \dataname{} provides a large data source to extend their multi-modal capabilities for \taskname{}.
% as VDER tasks are increasingly being studied by large language models (LLM), we believe that such a dataset can potentially benefit the training of LLMs especially those with multi-modal capabilities.

\clearpage

\section{Motivation}
% VDER tasks
Document understanding is one of the most error-prone and tedious tasks many people have to handle every day.
Advancements in machine learning techniques have made it possible to automate such tasks.
In a typical Visually-rich Document Entity Retrieval (VDER) task, pieces of information are retrieved from the document based on a set of pre-defined entity types, known as the \textit{schema}.
For example, ``amount'', ``date'', and ``item name'' are major parts of an invoice schema.
% are some of the main entities that constitute the schema. 
% The goal of VDER models is to correctly extract the text in the documents for each corresponding entity.

% Challenges of VDER tasks, 1) lack of raw data, 2) expensive to annotate, 3) datasets are not mutually beneficial to each other 
The current setup of VDER tasks presents several unique challenges for acquiring sufficient training data.
First, the availability of raw document images is greatly limited due to privacy constraints.
Real-world documents, such as a driver's license or a bank statement, often contain personally identifiable information and are subject to access controls.
% raw document images are difficult to acquire, even those unannotated ones. Human-generated documents, whether a driver's license, an invoice, or a bank statement, often contain sensitive information and thus with limited availability. 
% Access to these documents is highly restricted, and are often protected from being used beyond the data providers. 
Second, detailed annotation is costly and typically requires intensive training for experienced human annotators.
\Eg, it takes deep domain knowledge to correctly label different fields in complex tax forms.
% Second, the annotation process of such documents is expensive and time-consuming as it often involves intensive training for experienced human labelers. 
% Take the tax form as an example, the annotators often have to be provided with deep context before the annotation quality can reach a satisfactory level. 
Finally, knowledge sharing between various types of documents is constrained by inconsistent label spaces and contextual logic.
For example, the entity sets (\ie, schema) could be mutually exclusive, or the same entity type could take different semantic meanings in different contexts.
% as each type of document can contain a completely different set of entities (i.e., schema), transferring knowledge from one task to the other becomes difficult as the label spaces might differ. 
% Even for documents with the same entity name, it might require completely different retrieving logic given the context of different document types. 
% In the past, various approaches have been proposed to mitigate each challenge. 
% [desired length at here]
% For example, using 

A number of models have been proposed for VDER tasks with various success~\cite{huang2022layoutlmv3,lee2022formnet,appalaraju2021docformer,gu2021unidoc}.
To tackle the aforementioned challenges, most prior works initialize from a language model followed by BERT-style~\cite{devlin2019bert} pre-training on document datasets with additional layout and visual features.
However, even the largest dataset currently in use, \ie, IIT-CDIP~\cite{lewis2006building} dataset, has a limited size and only reflects a subset of document types.
% So far,  as a pre-training dataset along with language model pre-training such as BERT~\cite{devlin2019bert} or RoBERTa~\cite{liu2019roberta} has been adopted to compensate the lack of data when training VDER models. However, the released dataset has a limited size and only reflects a limited number of document types.

% how web data can help 1) raw data acquired from the web2) no need to annotate. 3) can benefit tasks regardless of the schema being used
% the dataset we have
In this chapter, we introduce the method of building the \dataname{} dataset, which enables massive-scale pre-training for VDER modeling.
\dataname{} is collected over the Internet using a pre-defined ontology, which spans hundreds of document types with a four-level hierarchy.
Experiments demonstrated that \dataname{} is the key to advancing the performance on the commonly used FUNSD~\cite{jaume2019funsd}, CORD~\cite{park2019cord}, and RVL-CDIP~\cite{lewis2006building} benchmarks.
%  in both classic and few-shot setups.
More recently, LLMs \cite{openai2023gpt, googlepalm2} have shown great potential for VDER tasks given their reasoning capabilities.
\dataname{} provides massive-scale multimodal data to boost the performance of LLMs for document understanding.

\begin{table*}[t]
\centering
% \begin{adjustbox}{max width=\textwidth}
\begin{tabular}{@{}lccccc@{}}
% {@{}l@{\hspace{4pt}}c@{\hspace{4pt}}c@{\hspace{4pt}}c@{\hspace{4pt}}c@{\hspace{4pt}}c@{}}
\toprule
Dataset & \makecell{\#Samples$\uparrow$} & \makecell{Ontology} & \makecell{Diverse\\Domains} & \makecell{High-quality\\OCR} & \makecell{Annotation} \\ \midrule
FUNSD~\cite{jaume2019funsd}  & 199 &       &      & & E=3  \\
Kleister-NDA~\cite{stanislawek2021kleister}   &  540 &       &      & \cmark & E=4 \\ 
VRDU-Ad-buy~\cite{wang2022benchmark}   & 641 &       &      & \cmark & E=14 \\
SROIE~\cite{huang2019icdar2019}    & 973  &  &      & & E=4 \\
CORD~\cite{park2019cord}  &  1K &       &      & & E=30 \\
DeepForm~\cite{borchmann2021due}   &  1.1K &       &      & \cmark & E=5  \\ 
VRDU-Registration~\cite{wang2022benchmark}   &  1.9K &       &      & \cmark & E=6 \\ 
Kleister-Charity~\cite{stanislawek2021kleister}   &  2.7K  &       &      & \cmark & E=8 \\
DocVQA~\cite{mathew2021docvqa} & 12.8K & & & \cmark & Q \\ \hdashline
CC-PDF~\cite{powalski2021going} & 350K & & \cmark & & \\
PubLayNet~\cite{zhong2019publaynet} & 358K & & \cmark&  & B=5 \\
RVL-CDIP~\cite{lewis2006building}   &  400K  &    & \cmark  &   & T=16 \\ 
UCSF-IDL~\cite{powalski2021going} & 480K & & \cmark & & \\
IIT-CDIP~\cite{lewis2006building}   &  11.4M  &  & \cmark     &  \\
% Harvard Caselaw\footnotemark & 40M & & & & & \xmark \\ 
\hdashline
\color{mygray}ImageNet~\cite{deng2009imagenet} & \color{mygray}1.3M images & \color{mygray}\cmark & \color{mygray}- & \color{mygray}- & \color{mygray} T=1K \\
\color{mygray}ActivityNet~\cite{caba2015activitynet} & \color{mygray}20K videos & \color{mygray}\cmark & \color{mygray}- & \color{mygray}- & \color{mygray} T=200 \\ \midrule
\emph{\dataname-v1 (ours)}  & {9.9M}      &  \cmark     &   \cmark   & \cmark & \textbf{T=398, E=6} \\
\emph{\dataname-v2 (ours)}  & \textbf{30M}      &  \cmark     &   \cmark   & \cmark & \textbf{T=398, E=6} \\ \bottomrule
\end{tabular}
% \end{adjustbox}
% \vspace{-2mm}
\caption[Comparison between \dataname{} dataset and existing document datasets]{\textbf{Comparison between the proposed \dataname{} dataset and existing document understanding datasets}. Datasets from other areas also built with ontology are listed in \textcolor{mygray}{gray}. Annotation includes sample type (T), bounding box (B), entity (E), and question (Q), where the value refers to the number of classes.}
\label{tab:datasets}
% \vspace{-4mm}
\end{table*}

% \footnotetext{{\url{https://dataverse.harvard.edu/dataverse/caselawaccess}}.}
% \footnotetext{\url{http://www.industrydocuments.ucsf.edu/}}

\section{Prior Work}
% \subsection{Datasets}

Tab.~\ref{tab:datasets} provides an overview of relevant document datasets.
We divide them into three groups as introduced below.

% \vspace{-2mm}
\paragraph{Single-domain document datasets.}
Many small document datasets with entity-span annotations have been used for tasks such as entity extraction.
They contain less than 100k pages from a single domain.
Newer datasets come with high-quality OCR annotation thanks to the advantage of relevant tools, while older ones, such as FUNSD~\cite{jaume2019funsd}, often contain OCR errors.
These datasets do not contain sufficient samples for the pre-training of a large model.

% \vspace{-2mm}
\paragraph{Large document datasets.}
A few larger datasets contain over 100k pages from different domains. 
However, they usually do not contain OCR annotations or entity-level labels.
IIT-CDIP~\cite{lewis2006building} has been the largest dataset commonly used for pre-training of document understanding models.
% Due to historical reasons, the image quality of IIT-CDIP is often unsatisfactory.
Although these datasets are large, their image quality and annotation completeness are often unsatisfactory.
To complement them, we collect high-quality document images from the Internet to build the \dataname{} datasets with rich OCR and entity annotations, and demonstrate their effectiveness in document model pre-training.

% \vspace{-2mm}
\paragraph{Ontology-based datasets.}
Large labeled datasets are usually collected following an ontology.
ImageNet~\cite{deng2009imagenet} for image recognition is built upon the synsets of WordNet~\cite{miller1998wordnet}.
ActivityNet~\cite{caba2015activitynet} for activity recognition adopts an activity taxonomy with four levels. % of granularity and 200 classes.
To the best of our knowledge, \dataname{} is the first large-scale document dataset built upon a well-defined ontology.

% \vspace{-2mm}
\paragraph{Pretrained document models.}
A variety of pretrained document models have emerged, including LayoutLM~\cite{xu2020layoutlm},  UDoc~\cite{gu2021unidoc}, LayoutLMv2~\cite{xu2021layoutlmv2}, TILT~\cite{powalski2021going}, BROS~\cite{hong2022bros}, DocFormer~\cite{appalaraju2021docformer}, SelfDoc~\cite{li2021selfdoc}, LayoutLMv3~\cite{huang2022layoutlmv3}, \etc.
% App.~\ref{app:related_model} provides detailed comparisons of their designs.

% \paragraph{Pre-training Data.}
% Almost all of the models mentioned above are pretrained on the IIT-CDIP dataset or its subset, as detailed in Table \ref{tab:sota}.
% As an exception, TILT~\cite{powalski2021going} is pretrained on a collection of RVL-CDIP, UCSF-IDL, and CC-PDF.

% XDoc~\cite{chen2022xdoc} IIT-CDIP

% UDOP~\cite{tang2022unifying} IIT-CDIP

\section{\dataname{} Dataset}
\begin{figure*}[tp]
  \centering
\begin{tabular}{@{}c@{}c@{}c@{}c@{}}
\includegraphics[width=0.24\textwidth,clip]{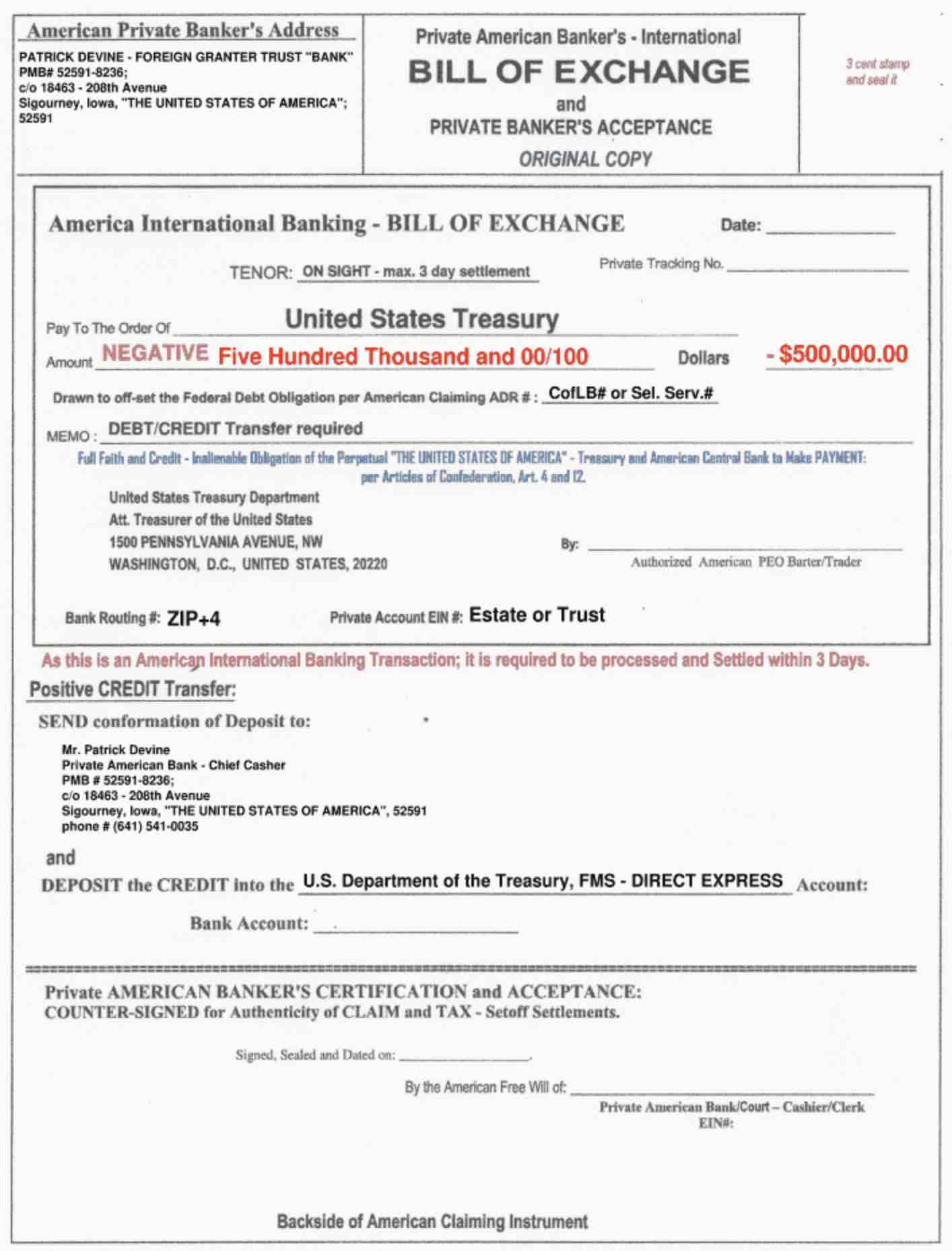} & 
\includegraphics[width=0.24\textwidth,clip]{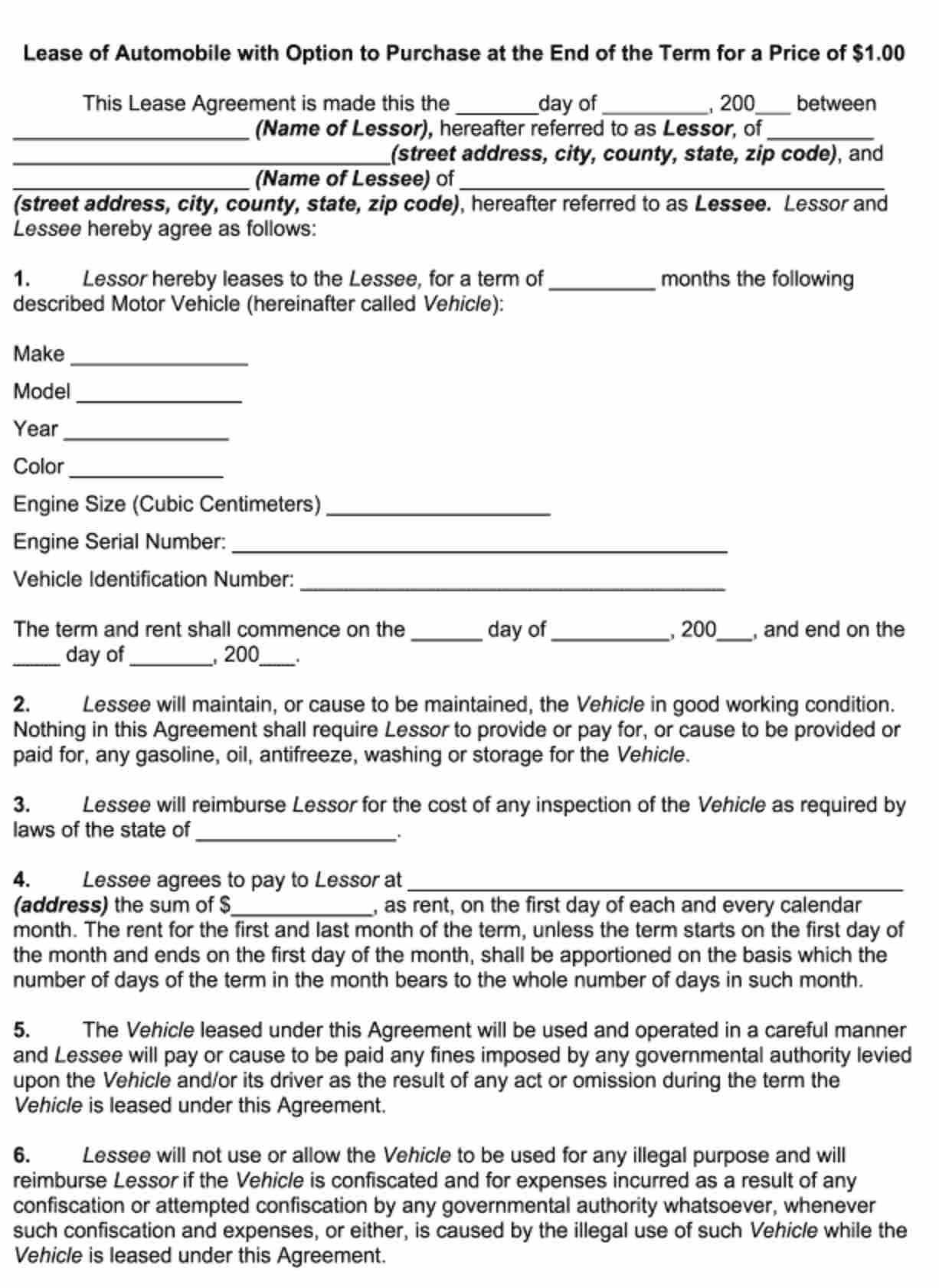} & 
\includegraphics[width=0.24\textwidth,clip]{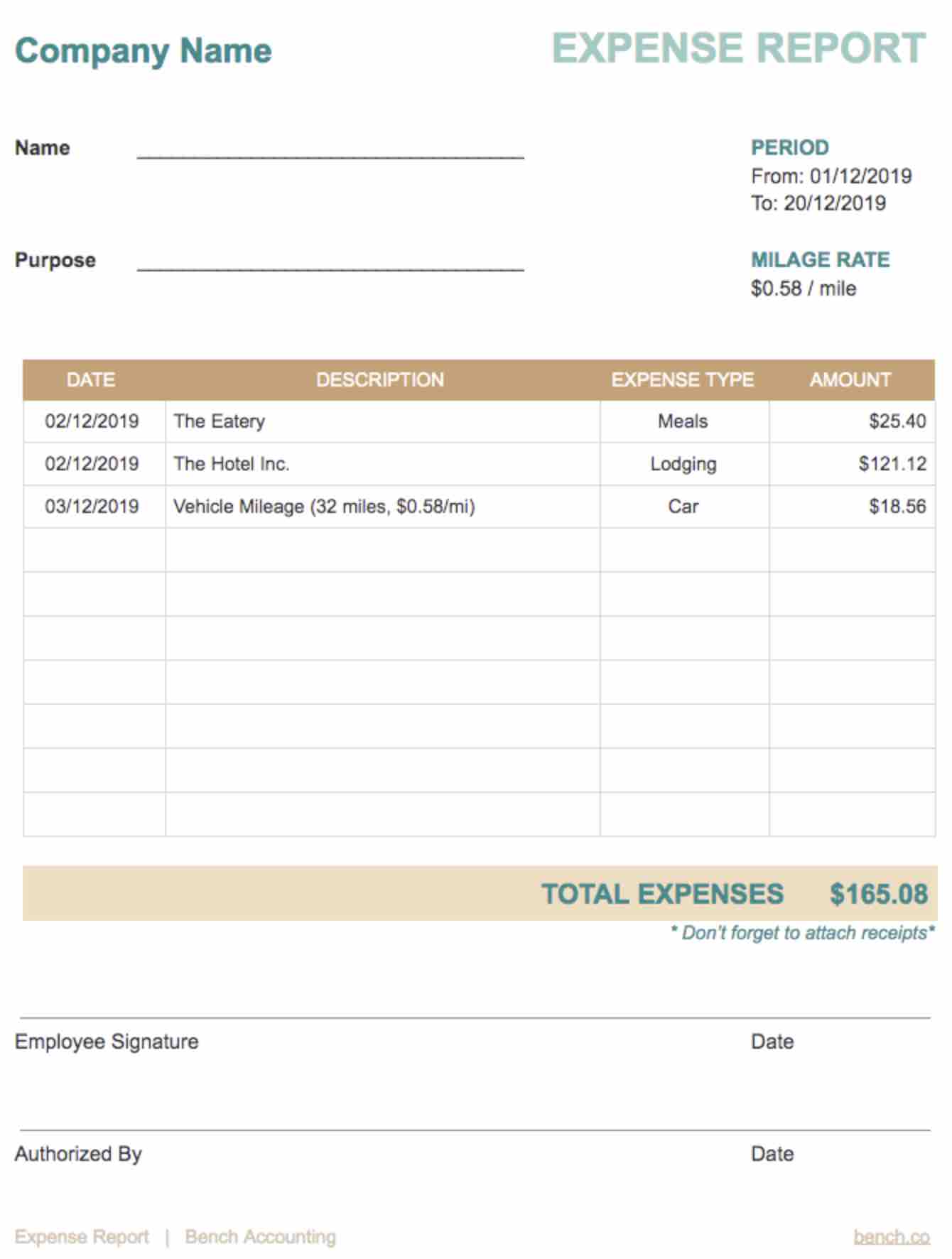} & 
\includegraphics[width=0.24\textwidth,clip]{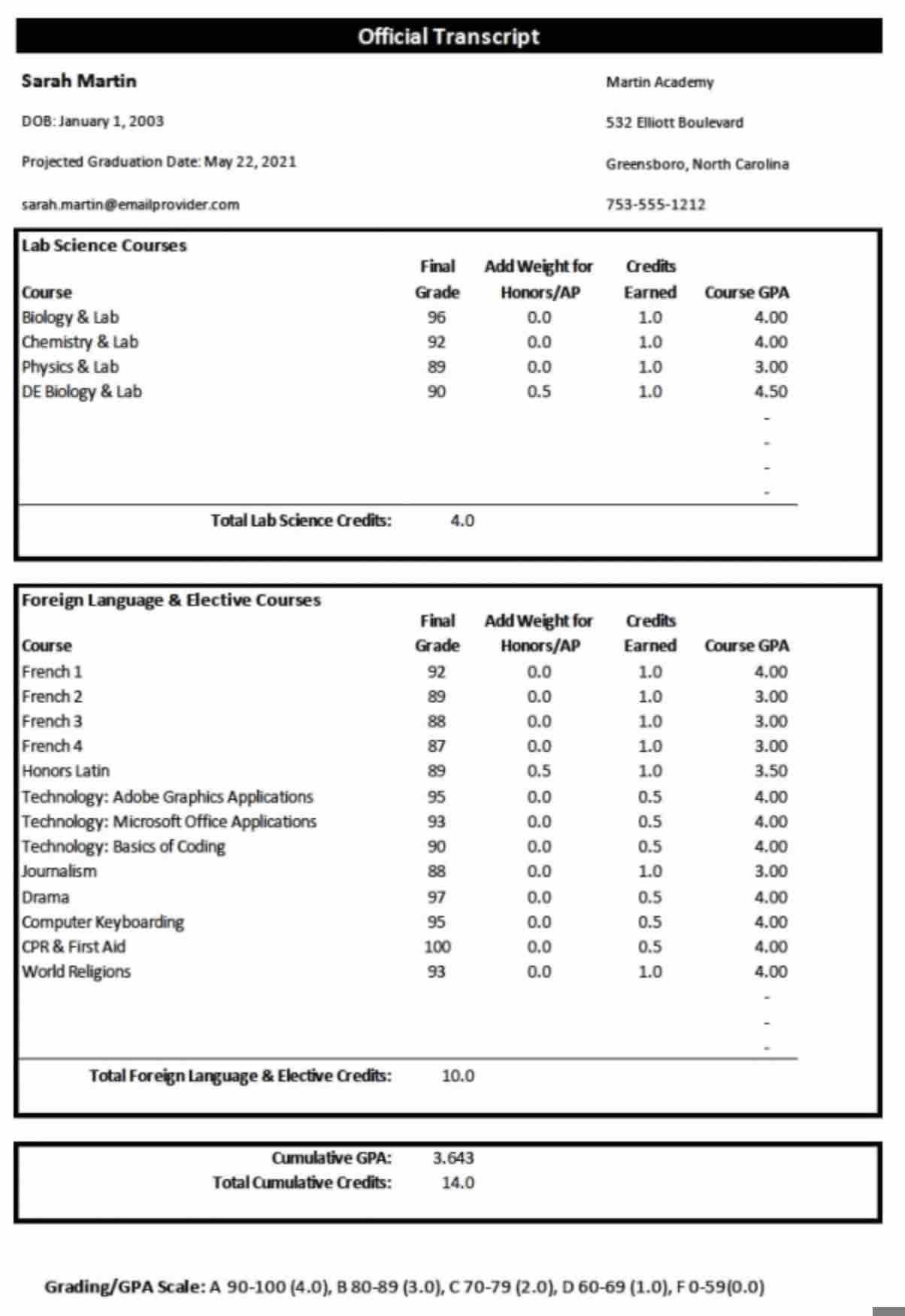} \\
Financial & Legal & Business & Education
\end{tabular}
\caption[Exemplar documents of each of the four top-level hierarchies]{\textbf{Exemplar documents of each of the four top-level hierarchies}. 
Images are downloaded via keyword searching using a commercial search engine. All images are for demonstration purposes only and do not contain real transactions or personal information. \label{fig:example}
}
\end{figure*}

\begin{figure*}[tp]
  \includegraphics[width=\textwidth,trim={0 0.1cm 0 0.2cm},clip]{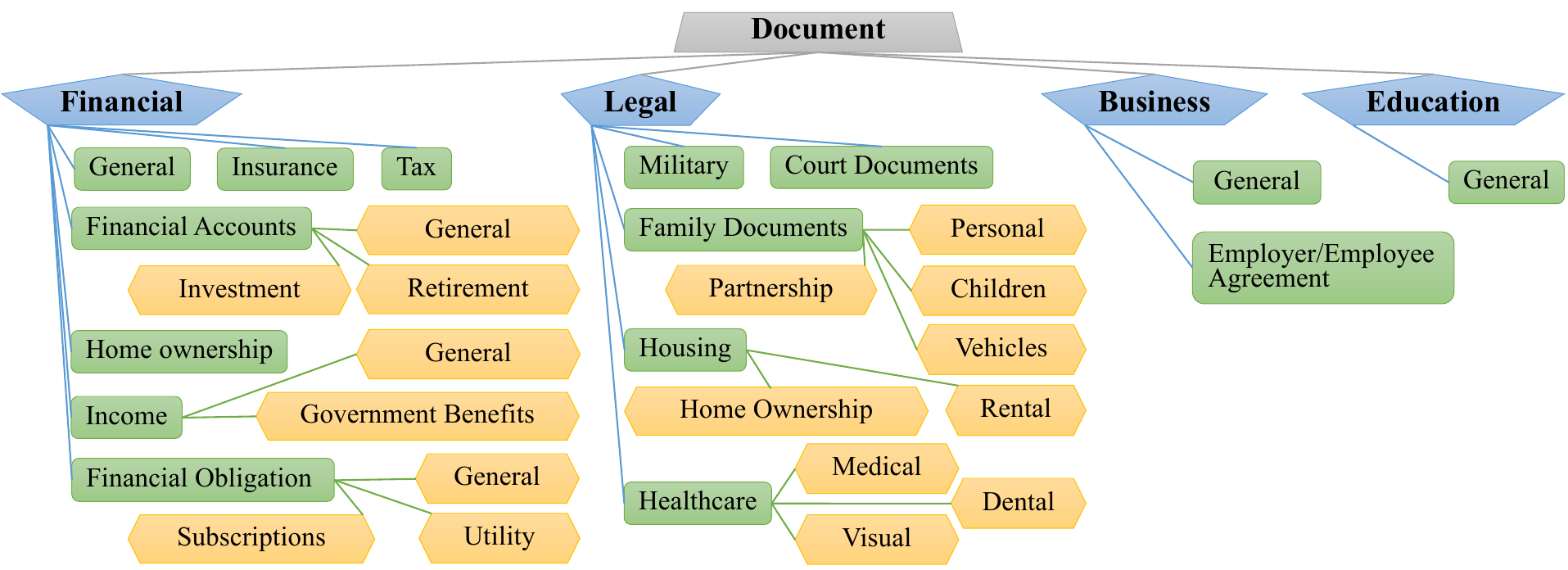}
  \caption[Document ontology tree stub]{\textbf{Document ontology tree stub}, based on which the proposed \dataname{} datasets are collected. We create a document ontology with about 400 search keywords hierarchically connected by three intermediate layers. 
  % See Appendix \ref{app:ontology} for the full ontology and the corresponding search keywords we used.
  }
  \label{fig:ontology}
\end{figure*}

Blindly crawling the Web for images may seem easy, but it is not a practical solution since most images on the Web are not relevant to document types.
We need a scalable pipeline to only select the concerned images.
Broadly, this is achievable via a nearest-neighbor search of relevant keywords in a text-image joint embedding space.
First, we design a set of query keywords, \ie, the document ontology, and encode them into the embedding space of general Web images. 
Further, a nearest-neighbor algorithm retrieves the top-K semantically closest images to each query keyword. 
Finally, a deduplication step consolidates all retrieved images across all query keywords. 
Fig.~\ref{fig:example} illustrates several exemplar documents retrieved using our provided keywords. 
% We illustrate more details for each step in the following subsections.

% \vspace{-2mm}
\paragraph{Ontology creation.}
Each text string in the ontology list serves as a seed to retrieve the most relevant images from the general Web image pool. An ideal ontology list should therefore cover a broad spectrum of query keywords across and within the concerned downstream application domains. Although algorithmic or generative approaches may exist, we manually curated about 400 document-related query keywords that cover domains of finance, business, personal affairs, legal affairs, tax, education, \etc, as illustrated in \cref{fig:ontology}.
% The full ontology hierarchy and keyword list are provided in App.~\ref{app:ontology}.

% \begin{figure}[tp]
%     \centering
%     \includegraphics[width=\linewidth, trim={0.2cm 0 0.2cm 0}, clip]{dataset_distance_histo.png}
%     % \vspace{-6mm}
%     \caption{\textcolor{blue}{Mean} and \textcolor{ForestGreen}{standard deviation} of the dot-product distance between the retrieved 30M document images and each query keyword. A distance of $1.0$ indicates the closest semantic relevance.}
%     \label{fig:dataset_distance_histo}
%     % \vspace{-6mm}
% \end{figure}

\begin{figure}[tp]
\centering
\begin{subfigure}{0.58\linewidth}
\raisebox{1cm}{
\includegraphics[width=\linewidth, trim={0.2cm 0 0.2cm 0}, clip]{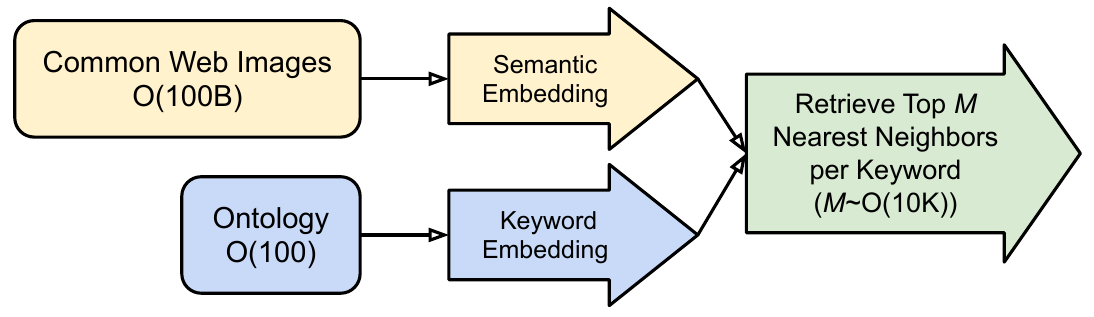}
}
\caption{\textbf{Data collection pipeline}.}
\label{fig:data_pipeline}
\end{subfigure}
\begin{subfigure}{0.40\linewidth}
\includegraphics[width=\linewidth, trim={0.2cm 0 0.2cm 0}, clip]{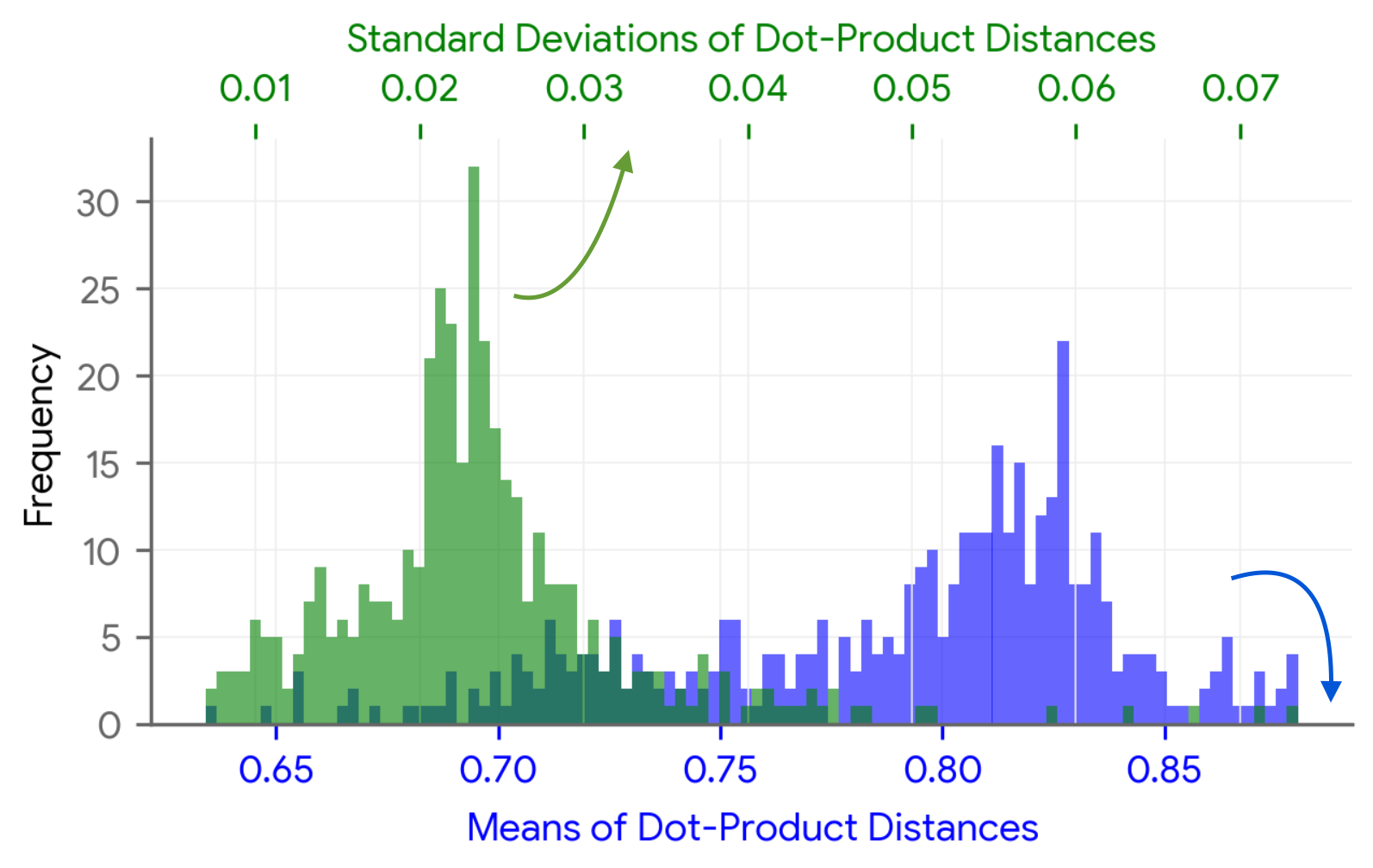}
\caption{\textbf{\textcolor{blue}{Mean} and \textcolor{ForestGreen}{standard deviation} of the dot-product distance} between the retrieved 30M document images and each query keyword. A distance of $1.0$ indicates the closest semantic relevance.}
\label{fig:dataset_distance_histo}
\end{subfigure}
\caption[Data collection pipeline and statistics]{\textbf{Data collection pipeline and statistics}.}
\end{figure}

% \vspace{-2mm}
\paragraph{Image retrieval from ontology.}
To retrieve only the most relevant document images out of the hundreds of billions of general Web images, we leverage a highly efficient nearest neighbor pipeline by defining the distance metric as the dot product distance between the semantic feature vectors of the image and each of the target query keywords. Here we refer to Graph-RISE~\cite{juan2020imageembed} for the semantic image embedding, and all query keywords are encoded into the same feature space as the images.
Empirically, we pick the top 10k nearest neighbors for each query keyword. Note that the same image might be retrieved via multiple semantically similar keywords, so a de-duplication step is needed afterwards. We summarize the main pipeline steps in Fig.~\ref{fig:data_pipeline}.
Fig.~\ref{fig:dataset_distance_histo} shows statistical insights of the retrieved 30M document images with the mean and standard deviation histogram over each of the query keywords.
The majority of the retrieved images are with mean distance values greater than $0.8$ and standard deviations no more than $0.03$, indicating high relevance to the document ontology.

% \vspace{-2mm}
\paragraph{OCR and annotation.}
\label{sec:ocr}
The retrieved images are fed into an OCR engine to generate a text sequence in reading order.
We apply a text tagging model to weakly annotate the text segments of each sequence into 6 classes, including \textit{email addresses, mail addresses, prices, dates, phone numbers,} and \textit{person names}. 
% The annotation could be omitted if the underlined text segment does not fit into any class.
Albeit noisy, these classification labels provide additional supervision for pre-training.

% \vspace{-2mm}
\paragraph{Post-processing and open-source tools.}
We adopt some heuristic-based filtering to improve sample quality.
For example, we remove samples where the overall OCR result is poor due to blurry or noisy images.
Some proprietary tools are used for scalable processing during the construction of \dataname{}, but open-source alternatives are readily available.
\Eg, CLIP~\cite{radford2021learning} for text-image embedding, Google ScaNN~\cite{avq_2020} for scalable nearest-neighbor search, Google Cloud OCR~(\url{https://cloud.google.com/vision/docs/ocr}), and Google Cloud NLP~(\url{https://cloud.google.com/natural-language/docs/reference/rest/v1/Entity#type}) for text tagging.
% ,  For example, to name a few, the text-image embedding is available in CLIP~\cite{radford2021learning}; the scalable nearest neighbor search module can be found in ; the OCR engine may be found in Google Cloud OCR~(\url{https://cloud.google.com/vision/docs/ocr}); the sequence classifier is available in .
% For example, we remove all images that are too blurry or have too much noise. We also remove all images that are not in the desired format.

With all of the above steps, we have obtained a dataset of high-quality document images that are closely relevant to our query ontology. 
This dataset contains multiple modalities, including the image pixels, the OCR characters, the layout coordinates, and the segment tags.
% This dataset is ready for a variety of downstream tasks, such as document classification, document summarization, and question answering.

\section{\modelname{} Model}
\begin{table}[tp]
\centering
\begin{adjustbox}{max width=\linewidth}
\begin{tabular}{@{}lll@{}}
\toprule
 & Task                                       & Target Modality    \\
\midrule
MMLM & \makecell[l]{Multimodal Masked Language Modeling} & OCR characters     \\
MCM & Masked Crop Modeling            & Image pixels    \\
TT & Token Tagging                 & Segment tags \\
\bottomrule
\end{tabular}
\end{adjustbox}
% \vspace{-2mm}
\caption[\modelname{} pre-training objectives and corresponding target modalities]{\textbf{\modelname{} pre-training objectives and corresponding target modalities}.}
\label{tab:pretrain}
% \vspace{-4mm}
\end{table}

In this section, we detail our \modelname{} model architecture and setups for pretraining and fine-tuning for \taskname{}.
\modelname{} takes advantage of all the modalities available in \dataname{}, with the pre-training objectives listed in \ref{tab:pretrain}.

\subsection{Multimodal Tokenization}

Let $\rmD \in \sR^{H \times W \times 3}$ be a visually-rich document image with height $H$ and width $W$.
We obtain a sequence of characters by applying OCR on the document image.
The characters are accompanied by their bounding box coordinates.
Then we perform a multimodal tokenization process as follows.

With a pre-defined text tokenizer, we first tokenize the character sequence into a sequence of text tokens $\rvc$.
$\rvp$ represents the 1D position of the tokens ranging from $0$ to $|\rvc|-1$.
For each token $\ervc_i$, we obtain its bounding box $\ervb_i = (x_0, y_0, x_1, y_1)_i$ by taking the union of the bounding boxes of its characters.
We enlarge the bounding box by a context ratio $r$ on each side and obtain the corresponding visual image crop $\ervv_i$ for each token from $\rmD$.

\subsection{\modelname{} Architecture}

\begin{figure*}[tp]
  \centering
  \includegraphics[width=\linewidth,trim={0 0 1.2cm 0},clip]{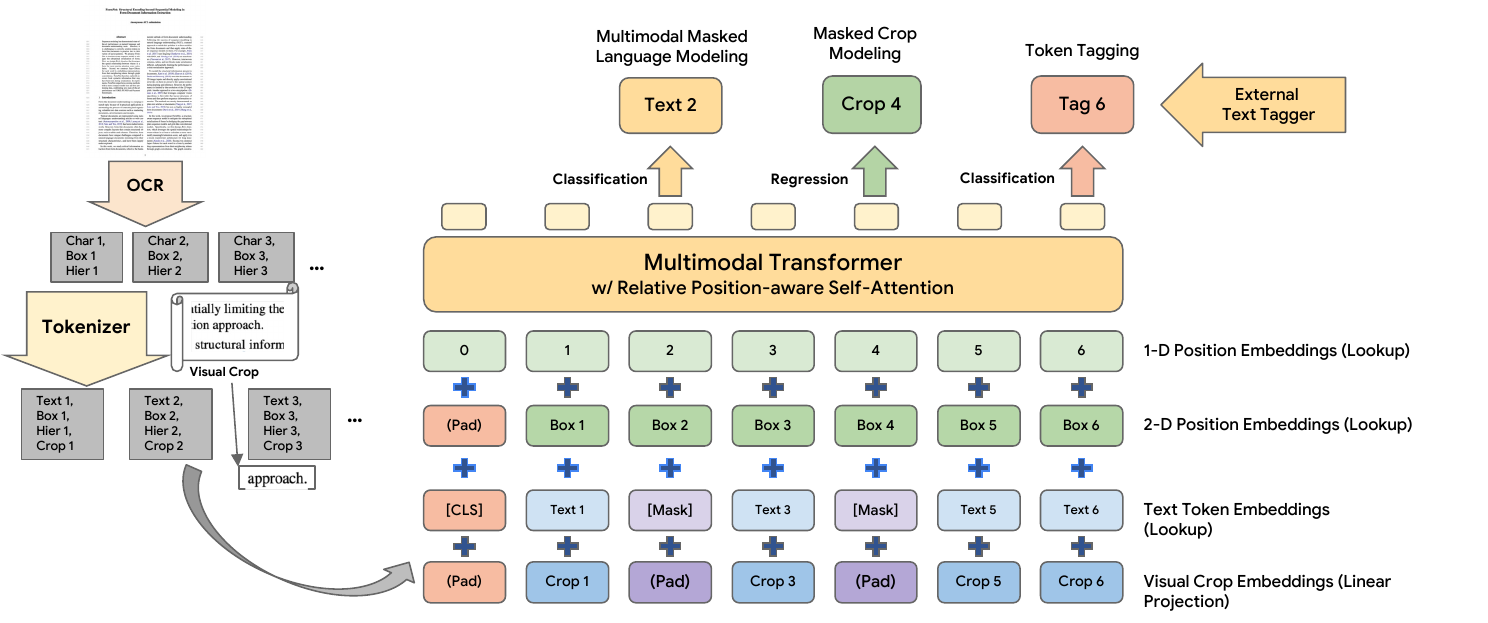}
  % \vspace{-6mm}
  \caption[\modelname{} pre-training pipeline]{\textbf{\modelname{} pre-training pipeline}. The multimodal tokenization process (left) outputs tokens with aligned image crops. The \modelname{} model (right) learns a unified token representation with three objectives (top).}
  \label{fig:pretrain_arch}
  % \vspace{-4mm}
\end{figure*}

Fig. \ref{fig:pretrain_arch} illustrates the model architecture for our proposed \modelname{}.
\modelname{} is built upon BERT~\cite{devlin2019bert} and utilizes its tokenizer and pretrained weights.
The input for each token consists of a text embedding and a 1D position embedding for $\ervp$.

Following LayoutLM~\cite{xu2020layoutlm}, we add 2D position embeddings $x_0$, $y_0$, $x_1$, $y_1$, $w$, $h$, where $w = x_1 - x_0$ and $h = y_1 - y_0$.
These embeddings are used to represent the spatial location of each token.
All the embeddings mentioned above are obtained from trainable lookup tables.

Following LayoutLMv2~\cite{xu2021layoutlmv2}, \modelname{} adopts relative position-aware self-attention layers by adding biases to the attention scores according to relative 1D locations $\triangle \ervp$ and relative 2D locations $\triangle \frac{x_0 + x_1}{2}$, $\triangle \frac{y_0 + y_1}{2}$.

\paragraph{Image Crop Input}

\begin{figure}[tp]
\centering
\begin{subfigure}[b]{0.36\linewidth}
    \centering
    \includegraphics[width=\linewidth, clip]{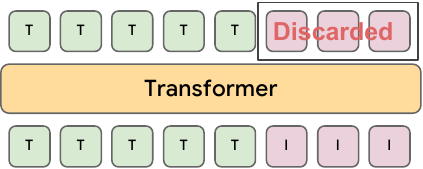}
\end{subfigure}\hspace*{2cm}%
\begin{subfigure}[b]{0.20\linewidth}
    \centering
    \includegraphics[width=\linewidth, clip]{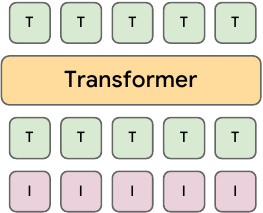}
\end{subfigure}
\caption[Unaligned vs. aligned visual features]{\textbf{Unaligned (left) vs. aligned (right) visual features}. The unaligned visual features result in a longer sequence but are usually discarded in downstream tasks. T: Text, I: Image.}
\label{fig:align}
\end{figure}

To model visual information, we add a crop embedding by linearly projecting the flattened pixels in the image crop, following ViT~\cite{dosovitskiy2020image}.
Different from prior works using either uniform patches~\cite{huang2022layoutlmv3}, regional features~\cite{li2021selfdoc,gu2021unidoc}, or global features~\cite{appalaraju2021docformer}, our multimodal tokenization and linear embedding of image crops has the following advantages: 
\begin{itemize}
    \item It eliminates the separate preprocessing for the visual modality, such as feature extraction with a pretrained CNN~\cite{xu2021layoutlmv2} or manually defined patches~\cite{huang2022layoutlmv3}.
    \item It obtains an aligned partition of the visual information with the text tokens, encouraging better cross-modal interaction.
    \item It eliminates the need for separate visual tokens as in \cite{xu2021layoutlmv2,huang2022layoutlmv3}, resulting in a shorter token sequence and better efficiency, as shown in Fig. \ref{fig:align}.
    \item It provides a unified joint representation for text and visual modalities in document modeling with semantic-level granularity.
\end{itemize}

\subsection{Pretraining}
During pretraining, we adopt the following objectives on a \modelname{} parameterized by $\theta$.
For each objective, we use a separate head upon the last attention layer.
Let $\rho$ denote the always available input embeddings, including the 1D and 2D positions.

\paragraph{Multimodal Masked Language Modeling (MMLM)}
We randomly select 15\%~\cite{devlin2019bert} of the tokens, denoted as $\gM$, to mask and predict the language modality.
In the masked language input $\overline{\rvc}$, 80\% of the masked tokens are replaced with a special \mask{} token, while another 10\% are replaced with a random token and the remaining 10\% are kept as is.
In the masked crop input $\overline{\rvp}$, crops for all masked tokens are replaced with an empty image.
The language prediction is formulated as a multi-class classification problem with the cross-entropy loss as
\begin{equation}
  \Ls_\mathit{MMLM} = \mathop{\E}\limits_{\rmD} \Big[ \sum_{\ervc_i \in \gM} -\log p_\theta (\ervc_i \mid [\overline{\rvc}, \overline{\rvv}, \rho]) \Big].
\end{equation}

\begin{figure*}
    \centering
    \includegraphics[width=\linewidth,trim={0 0 0.8cm 0},clip]{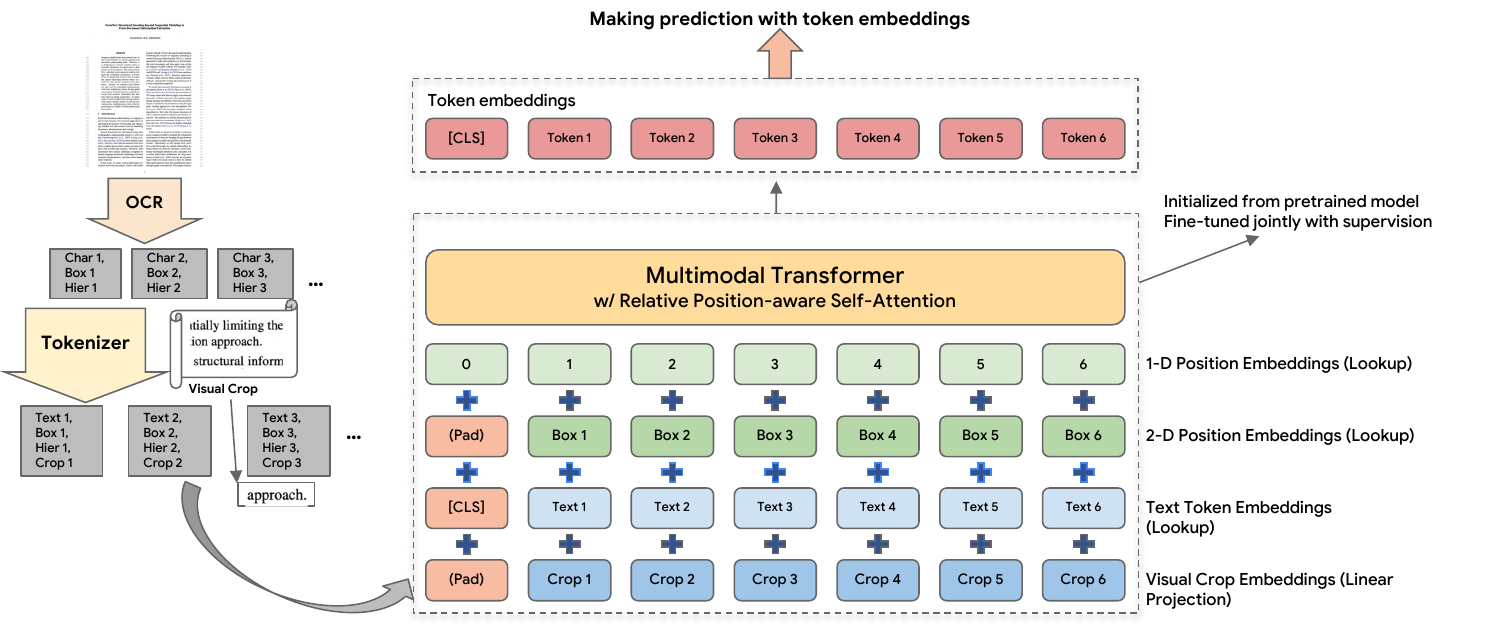}
    \caption[\modelname{} finetuning pipeline]{\textbf{\modelname{} finetuning pipeline}.}
    \label{fig:finetune_arch}
\end{figure*}

\paragraph{Masked Crop Modeling (MCM)}
We also predict the visual modality by reconstructing the image crops for the masked tokens in MMLM, in a way similar to MAE~\cite{he2022masked}.
It is formulated as a regression problem with a linear layer over flattened pixels.
The MCM loss is defined as 
\begin{equation}
    \Ls_\mathit{MCM} = \mathop{\E}\limits_{\rmD} \Big[ \sum_{\ervc_i \in \gM} \left\|\hat{\ervv}_i - \ervv_i\right\|_2^2 \Big],
\end{equation}
where $\hat{\rvv} = f_\theta(\overline{\rvc}, \overline{\rvv}, \rho])$.

\paragraph{Token Tagging (TT)}
We add an extra pretraining task by predicting the tags $\rvt$ for each token in an unmasked sequence.
The tags are extracted from an external text tagger as described in Sec. \ref{sec:ocr}.
Since each token may have multiple tags, it is formulated as a multi-label classification problem with the binary cross-entropy loss as 
\begin{equation}
\begin{aligned}
  \Ls_\mathit{TT} &= \mathop{\E}\limits_{\rmD} \Big[ \sum_{i, k} -\ervt_{i, k} \log p_\theta (\ervt_{i, k} \mid [\rvc, \rvv, \rho]) \\
  &-(1 - \ervt_{i, k}) \log (1 -  p_\theta (\ervt_{i, k} \mid [\rvc, \rvv, \rho])))  \Big],
\end{aligned}
\end{equation}
where $k=1,2,\cdots,K$ refers to the $K$ types of tags.

\paragraph{Pretraining Loss}
The overall pretraining objective is given as 
\begin{equation}
    \Ls_\mathit{pretrain} = \Ls_\mathit{MMLM} + \alpha \Ls_\mathit{MCM} + \beta \Ls_\mathit{TT},
\end{equation}
where $\alpha, \beta$ are the corresponding loss weights.

\subsection{Finetuning}

Fig. \ref{fig:finetune_arch} illustrates the pipeline for the finetuning of \modelname{}.
During finetuning, no tokens are masked.
We adopt the following two tasks in finetuning.

\paragraph{Entity Extraction}
Entity extraction is formulated as a sequence tagging problem.
The ground-truth entity spans are converted into a sequence of BIO tags $\rve$ over all tokens.
The BIO tagging is formulated as follows:
$\rve$ is initialized with all $\gO$ tags which indicates ``Other" refering to background tokens.
For each entity span with type $\gT$, start position $i$ and end position $j$ (both inclusive), we assign 
\begin{align}
    \erve_i &= \gT_\text{Begin}, \\
    \erve_{i+1} &= ... = \erve_{j} = \gT_\text{Intermediate}.
\end{align}
The prediction of BIO tags is modeled as a multi-class classification problem with the objective as
\begin{equation}
  \Ls_\mathit{EE} = \mathop{\E}\limits_{\rmD} \Big[ \sum_{i} -\log p_\theta (\erve_i \mid [\rvc, \rvv, \rho]) \Big].
\end{equation}

\paragraph{Document Classification}
We use the embedding of the starting \cls token for document classification.
The logits are predicted with an MLP head on top of the \cls embedding.
Let $l$ be the correct class, the objective is
\begin{equation}
  \Ls_\mathit{DC} = \mathop{\E}\limits_{\rmD} \Big[ -\log p_\theta (l \mid [\rvc, \rvv, \rho]) \Big].
\end{equation}

\section{Experimental Results}
\begin{table*}[tp]
\centering
\begin{adjustbox}{max width=\textwidth}
\begin{tabular}{@{}llllccc@{}}
% {@{}l@{\hspace{8pt}}cl@{\hspace{8pt}}c@{\hspace{8pt}}c@{\hspace{8pt}}c@{\hspace{8pt}}c@{\hspace{8pt}}c@{}}
\toprule
\multirow{2}{*}{Model} & \multirow{2}{*}{Inputs}  & \makecell[l]{Pre-training\\Data} & \makecell[l]{Pre-training\\Objectives} & {\makecell{FUNSD\\Entity F1$\uparrow$}}  & {\makecell{CORD\\Entity F1$\uparrow$}} & {\makecell{RVL-CDIP\\Accuracy$\uparrow$}}  \\ \midrule

BERT & T  & -  & MLM & 60.26 & 89.68 & 89.81 \\

LayoutLM & T + L & IIT-CDIP & MVLM & 78.66 & 94.72 & 91.78 \\ \midrule

\makecell[l]{\emph{\modelname{}}} & T + L + C & IIT-CDIP                & MMLM        & 80.63 & 95.17 & 93.47   \\ \hdashline
\multirow{3}{*}{\makecell[l]{\emph{\modelname{}}}} & \multirow{3}{*}{\makecell{T + L + C}} &  \multirow{3}{*}{\makecell[l]{IIT-CDIP\\\ \ + \emph{\dataname-v1}}}   & MMLM   & 82.61 & 95.91 & 94.86   \\
& &          & MMLM + MCM & 83.45 & 96.08 & 95.15   \\ %\cmidrule(lr){3-7}
 % &  & IIT-CDIP               & \multirow{2}{*}{\makecell[l]{MMLM + MCM + TT}} & 83.46 & 96.22 & 94.54   \\
% &  &  \dataname-v1      &    & 80.18 & 95.55 & 94.38   \\
 &  &   &    MMLM + MCM + TT    & \textbf{84.18} & \textbf{96.45} & \textbf{95.34}   \\
\bottomrule
\end{tabular}
\end{adjustbox}
% \vspace{-2mm}
\caption[Ablation studies on three document understanding benchmarks]{\textbf{Ablation studies on three document understanding benchmarks} regarding pretraining datasets, pretraining objectives, and model architectures. Input modalities include text (T), layout (L), and crop (C). Metrics are entity F1 and classification accuracy $\times 100$.}
\label{tab:ablation}
% \vspace{-4mm}
\end{table*}

We pre-train \modelname{} on \dataname{} and evaluate on three document understanding benchmarks.
% two settings:
% % experimented our method on two settings: 
% (1) the classic VDER setting with the full split of train and test;
% (2) the few-shot VDER setting where we have meta-train and meta-test task sets with each task containing a set of samples that satisfies the $N$-way $K$-shot setting. 

% \paragraph{Protocol}
% \vspace{-2mm}
\subsection{Implementation Details}
\paragraph{Pre-training.}
We initialize our \modelname{} with BERT weights using the uncased vocabulary.
The models are pre-trained using the Adam optimizer~\cite{kingma2014adam}.
We adopt a cosine learning rate schedule with linear warmup during the first 2\% steps and a peak learning rate of $10^{-4}$.
% The loss weights are $\alpha=1$ and $\beta=30$.
We use 20\% of the samples for the token tagging pre-training task.
% When the token tagging pre-training task is enabled, 20\% of the samples are used for this task.
The models are trained for 500K steps with a batch size of 2048 on 128 TPUv3 devices.

% \paragraph{Masked Crop Modeling}
% Figure \ref{fig:mcm} visualizes the ground truth and predictions in the MCM training. 
% With a simple linear projection into the embeddings and another simple linear projection into flattened output pixels, our \modelname{} successfully reconstructs the token image crops.
% As we have observed in language modeling, ambiguity exists for masked crop modeling as well.
% The semantic structure does not change despite the prediction of the first token becomes ``30" instead of ``31".

% \begin{figure}[tp]
%     \centering
%     \includegraphics[width=\linewidth, clip]{mcm.pdf}
%     \caption{Masked Crop Modeling ground truth (top) and predictions (bottom) given surrounding text and token crops.}
%     \label{fig:mcm}
% \end{figure}

% \subsection{Classic VDER Setting}
\paragraph{Downstream tasks.}
We evaluate the performance of pre-trained \modelname{} models on three commonly used benchmarks: entity extraction on FUNSD and CORD, and document classification on RVL-CDIP.
FUNSD contains 199 documents with 149 for training and 49 for evaluation. It is labeled with 3 entity types, i.e., header, question, and answer.
CORD contains 1000 documents with 800 for training, 100 for validation, and 100 for testing.
It is labeled with 30 entity types for receipts, such as menu name, price, \etc.
RVL-CDIP contains 400K documents in 16 classes, with 320K for training, 40K for validation, and 40K for testing.
% \paragraph{Downstream Tasks.}

% Detailed setups are provided in App. \ref{app:classic}.

% \vspace{-2mm}
\paragraph{Finetuning.}
For entity extraction on FUNSD and CORD, we add a multi-class classification head on top of all text tokens to perform BIO tagging.
We fine-tune with a peak learning rate of $5\times10^{-5}$, following a schedule of linear warm-up in the first 10\% steps and then linear decay. Dropout with 0.1 probability is applied in the head layers. \modelname{} is fine-tuned for 1000 steps with a batch size of 32 on FUNSD and 256 on CORD.
For document classification on RVL-CDIP, we add a multi-class classification head on top of the [CLS] token. We fine-tune with a constant learning rate of $10^{-5}$ for 15000 steps with a batch size of 2048.

\begin{table*}[t]
\centering
\begin{adjustbox}{max width=\textwidth}
\begin{tabular}{@{}l@{\hspace{8pt}}l@{\hspace{8pt}}c@{\hspace{8pt}}l@{\hspace{8pt}}c@{\hspace{8pt}}c@{\hspace{8pt}}c@{}}
\toprule
Model & Initialization & \makecell{Total\\Parameters} & \makecell{Pretrain\\Data Source}           & \makecell{FUNSD\\Entity F1$\uparrow$}  & \makecell{CORD\\Entity F1$\uparrow$} & \makecell{RVL-CDIP\\Accuracy$\uparrow$}  \\ \midrule

% BERT~\cite{devlin2019bert} & & 110M & - & - & 60.26 & 89.68 & 89.81 \\
% \multicolumn{2}{l}{\color{mygray}RoBERTa~\cite{liu2019roberta}} & \color{mygray}125M & \color{mygray}- & \color{mygray}- & \color{mygray}66.48 & \color{mygray}93.54 & \color{mygray}90.06 \\

\multirow{2}{*}{\makecell[l]{LayoutLM}} & BERT & 113M & IIT-CDIP  & 78.66 & 94.72 & 91.78 \\
 & \makecell[l]{BERT + ResNet-101} & 160M &  IIT-CDIP  & 79.27 & - & 94.42 \\
\makecell[l]{UDoc} & \makecell[l]{BERT + ResNet-50} & 272M & IIT-CDIP  & - & - & 95.05 \\
\makecell[l]{LayoutLMv2} & \makecell[l]{UniLM + ResNeXt-101} & 200M & IIT-CDIP  & 82.76 & 94.95 & 95.25 \\
\makecell[l]{TILT} & \makecell[l]{T5 + U-Net} & 230M & \makecell[l]{RVL-CDIP + \\\ \ \ UCSF-IDL + \\\ \ \ CC-PDF}  & - & 95.11 & 95.25 \\
\makecell[l]{BROS} & BERT & 110M & IIT-CDIP  & 83.05 & 95.73 & - \\
\makecell[l]{DocFormer} & \makecell[l]{LayoutLM + ResNet-50} & 183M & IIT-CDIP  & 83.34 & 96.33 & 96.17 \\
\makecell[l]{SelfDoc} & \makecell[l]{BERT + ResNeXt-101} & 137M & RVL-CDIP  & 83.36 & - & 92.81 \\
% \color{mygray}XYLayoutLM & \color{mygray}LayoutXLM & \color{mygray}- & \color{mygray}- & \color{mygray}- & \color{mygray}83.35 & \color{mygray}- & \color{mygray}- \\
% \color{mygray}\makecell[l]{FormNet~\cite{lee2022formnet}} & \color{mygray}- & \color{mygray}345M & \color{mygray}- & \color{mygray}700K & \color{mygray}84.69 & \color{mygray}97.28 & \color{mygray}- \\
% \color{mygray}\makecell[l]{LiLT~\cite{wang2022lilt}} & \color{mygray}RoBERTa & \color{mygray}131M & \color{mygray}IIT-CDIP & \color{mygray}11M & \color{mygray}88.41 & \color{mygray}96.07 & \color{mygray}- \\
\makecell[l]{LayoutLMv3}$^*$ & \makecell[l]{RoBERTa} & 126M & IIT-CDIP  & - & 96.11 & 95.00 \\
% \color{mygray}\makecell[l]{LayoutLMv3~\cite{huang2022layoutlmv3}} & \color{mygray}\makecell[l]{RoBERTa + DiT Tokenizer$^*$} & \color{mygray}133M & \color{mygray}IIT-CDIP & \color{mygray}11M & - & \color{mygray}96.56 & \color{mygray}95.44 \\
\midrule
\makecell[l]{\emph{\modelname{}}} & BERT & 115M & \makecell[l]{IIT-CDIP +\\ \ \ \ \emph{\dataname-v1}}          & \textbf{84.18} & \textbf{96.45} & 95.34   \\ 
% \midrule
\bottomrule
\end{tabular}
\end{adjustbox}
% \vspace{-2mm}
\caption[Comparison with state-of-the-art document pretraining approaches]{Comparison with state-of-the-art document pretraining approaches on three document understanding benchmarks. $^*$ denotes a variant that does not use its proprietary tokenizer in pre-training. Metrics are entity F1 and classification accuracy $\times 100$.}
\label{tab:sota}
% \vspace{-2mm}
\end{table*}

% \vspace{-2mm}
\subsection{Quantitative Evaluation}
\paragraph{Ablation Studies.}
Table \ref{tab:ablation} lists the ablation results for pre-training data, pre-training objectives, and model design.
Compared to LayoutLM, our unified embedding of the visual modality and MMLM pre-training results in a much stronger baseline.
Adding our \dataname{} into the pre-training leads to a significant performance boost across all three tasks.
Further incorporating MCM and TT pre-training objectives to fully leverage \dataname{} yields consistent improvements, where the entity extraction tasks benefit more from TT and the document classification task gains more from MCM.
% Increasing the pre-training data size from 11M to 39M and introducing the MCM and TT tasks further boost the performance.
% Regarding \dataname{}, we can see that pre-training only on \dataname-v1 can already yield a reasonable performance.
% pre-training jointly on IIT-CDIP and \dataname-v1 is evidently better than pre-training on IIT-CDIP only, especially for the FUNSD and RVL-CDIP benchmarks.

% \vspace{-2mm}
\paragraph{Comparisons with state-of-the-art.}
We compare the performance on the three benchmarks with state-of-the-art approaches in Table \ref{tab:sota}.
As shown, most prior methods use stronger language or image initialization compared to our lightweight \modelname{}, but all of them are only pre-trained on datasets no larger than IIT-CDIP.
Although \modelname{} is only using 115M parameters and BERT initialization,
% As suggested by LayoutLM~\cite{xu2020layoutlm}, the language initialization has a significant impact on the final model performance.
% Therefore, we emphasis models based on BERT-base for fair comparisons and list the others for reference.
% As shown, despite not using any pre-trained visual models, our approach outperforms all BERT-based document models.
it outperforms all baseline approaches after pre-training on our \dataname{} dataset, with FUNSD entity F1 84.18\%, CORD entity F1 96.45\%, and RVL-CDIP accuracy 95.34\%.

\section{Summary}
In this chapter, we proposed a \emph{multi-modal} method to use massive and noisy web data to benefit the training of VDER models. Our approach has the benefits of providing a large amount of document data with little cost compared to usual data collection processes in the VDER domain. Our experiments demonstrated significantly boosted performance for document understanding tasks. There are a number of areas that would warranty extensions or future work. First, a systematic study on the exact keywords and strategies of collecting such a data that would optimize the model outcome is yet to be studied. The methods proposed in this chapter is merely a starting point for methods along this direction. Secondly, architecture changes that specifically targets the proposed methods of massive and noisy data collecting remains an open research question. One observation we had when examining the data is that many of them contains empty forms while others have filled in content. Models that can explicitly take advantage of both formats should further boost the performance of the model.

% \chapter{Audio Latent Diffusion}

\glsresetall
\cleardoublepage
\part{Multi-Modal Latent Spaces}\label{part:latent}
\paragraph{\cref{part:latent} Overview.}

\glsresetall
\cleardoublepage
\chapter{Spatial-Temporal Vector-Quantized Representation}\label{chap:3dvq}
\graphicspath{{papers/magvit/figures/}}

\renewcommand{\modelname}{MAGVIT}
\renewcommand{\methodname}{COMMIT}
\renewcommand{\mask}{\texttt{[MASK]}\xspace}

\paragraph{Overview.}
% In this chapter, we introduce an innovative spatial-temporal vector-quantization model meticulously crafted for the precise tokenization of videos. Drawing insights from the successes attained by diverse image tokenization techniques, we embark on the creation of a distinctive architectural framework tailored for this model. This novel architecture ingeniously integrates 3D convolutions, harnessing their prowess to adeptly capture the intricate interplay of spatial and temporal dependencies inherent in video data.

% The fruit of this endeavor is a model that not only meets but exceeds expectations. Its capacity to maintain exceptional reconstruction fidelity, even when subjected to substantial compression ratios, is a testament to its inherent potency. This achievement, in turn, serves as a cornerstone, laying the pivotal groundwork upon which the subsequent accomplishments of generative video transformers rest.

In this chapter, we unveil an innovative spatial-temporal vector-quantization model meticulously crafted for the purpose of video tokenization \emph{latent representation}. Drawing inspiration from the notable accomplishments attained by diverse image tokenization techniques, we embark on the creation of a distinctive architecture tailored to this model. This architecture ingeniously integrates 3D convolutions, enabling the model to adeptly capture the intricate interplay of spatial and temporal dependencies inherent in video data.
As a testament to our innovative approach, the model achieves a level of reconstruction fidelity that stands out even when operating under significant compression ratios. This achievement not only demonstrates the efficacy of our spatial-temporal vector-quantization model but also lays a robust foundation for the subsequent strides made in the realm of generative video transformers. Through this pivotal chapter, we pave the way for the exciting possibilities that lie ahead in the fusion of advanced video tokenization and generative transformer techniques.

\clearpage

\section{Motivation}

% vision transformers~\cite{weissenborn2019scaling, rakhimov2021latent, nash2022transframer}

Transformers~\cite{vaswani2017attention} have demonstrated strong modeling capabilities for language~\cite{devlin2019bert}, where each sample usually consists of less than 1k text tokens.
On the contrary, visual modalities are in much higher dimensional spaces.
For example, a 256$\times$256 RGB image has nearly 200k pixels, and a 16-frame clip of 128$\times$128 has almost 1M.
With super long sequences, the quadratic computation cost and the long-term dependency limit the performance of pixel-space generative modeling with transformers~\cite{chen2020generative} beyond 64$\times$64.

To achieve feasible generation at higher resolutions and better fidelity, 
%  and image~\cite{wang2022image} modalities.
recent generative image transformers such as DALL·E~\cite{ramesh2021zero} and others~\cite{esser2021taming,yu2022scaling,chang2022maskgit,ding2021cogview}
operate in a compressed discrete latent space produced by variants of Vector-Quantized Variational AutoEncoders (VQ-VAE)~\cite{van2017neural}.
% we propose an efficient
The VQ-VAE encoder maps an input image into a spatially downsampled feature map.
A vector quantizer (VQ) discretizes the feature map with a learned codebook, followed by a decoder reconstructing the input image.

Inspired by the successful practice for images, we design a 3D-VQ model to tokenize a video into a low-dimensional spatial-temporal manifold~\cite{yan2021videogpt, ge2022long}.
Instead of modeling each frame independently, we introduce spatial-temporal downsampling to achieve higher compression ratios by exploiting the temporal redundancy while preserving high reconstruction quality.
This way, generative transformers can model longer videos with more compact representation.

The only existing 3D-VQ model before our work is a shallow model from TATS~\cite{ge2022long}.
We present comprehensive comparisons of the model architectures.
Quantitative and qualitative evaluations show that our deep network design and advanced training strategy produce much higher visual quality.
This successful design fosters the strong video generation results of MAGVIT~\cite{yu2022magvit} presented in \cref{chap:magvit}.

% Contribution: We introduce a spatial-temporal video quantization model design with high reconstruction fidelity.

% First, we learn a 3D vector-quantized (VQ) autoencoder to quantize a video into discrete tokens. 

% \section{Prior Work}
% The state-of-the-art model TATS~\cite{ge2022long} uses two hierarchical transformers to reduce the computation for long video generation, with tokens learned by a 3D-VQGAN~\cite{esser2021taming}. 
% %that adds a GAN loss and a perceptual loss to the VQ-VAE.
% Unlike prior works, we introduce a non-autoregressive transformer with higher efficiency and flexibility.

\section{MAGVIT 3D-VQ Model}
\subsection{Model Design}
\label{sec:mth_tok}
Our video VQ autoencoder is built upon the image VQGAN~\cite{esser2021taming}.
Let $\rmV \in \sR^{T \times H \times W \times 3}$ be a video clip of $T$ frames.
The VQ encoder tokenizes the video as $f_\gT: \rmV \rightarrow \rvz \in \sZ^{N}$, where $\sZ$ is the codebook.
%
% For example, a video clip with 16 frames and 128$\times$128 resolution may be quantized to 4$\times$16$\times$16 tokens. 
%
The decoder $f_\gT^{-1}$ maps the latent tokens back to video pixels.

The VQ autoencoder is a crucial module as it not only sets a quality bound for the generation but also determines the token sequence length, hence affecting generation efficiency. 
Existing methods apply VQ encoders either on each frame independently (2D-VQ)~\cite{le2021ccvs, gupta2022maskvit} or on a supervoxel (3D-VQ)~\cite{yan2021videogpt, ge2022long}.
% \lu{try to add more references}. \lijun{Only these two used 3D so far.}
We propose different designs that facilitate \modelname{} to perform favorably against other VQ models for video (see \cref{tab:token}). 
%   \lu{we need to have the comparison in the main paper}

% \vspace{-5mm}
\paragraph{3D architecture.}
%Empirically, we find VQGAN preforms significantly better with 3D than 2D convolutions~\cite{tran2015learning} on videos. 
%
We design a 3D-VQ network architecture to model the temporal dynamics as follows.
The encoder and decoder of VQGAN consist of cascaded residual blocks~\cite{he2016deep} interleaved by downsampling (average pooling) and upsampling (resizing plus convolution) layers.
We expand all 2D convolutions to 3D convolutions with a temporal axis.
%To in video, we build a 3D-VQ model by replacing the 2D convolutions in VQGAN with 3D convolutions~\cite{tran2015learning}.
% We change the strided-convolution into x for downsampling~\cite{zhang2019making} and strided-deconvolution into nearest resizing plus convolution for upsampling~\cite{odena2016deconvolution}.
As the overall downsampling rate is usually different between temporal and spatial dimensions,
%(\eg, 16 / 4 = 4 and 128 / 16 = 8 in the above example),
we use both 3D and 2D downsampling layers, where the 3D ones appear in the shallower layers of the encoder.
The decoder mirrors the encoder with 2D upsampling layers in the first few blocks, followed by 3D ones.
% \cref{app:vq_arch}
% Appendix A.1
% illustrates the detailed architecture. 
% \lu{For the submission, cref will stop working as there will be two seperate documents.} \lijun{We can still compile the full document and then split the pdf file, so the section id is correct but the hyperlink does not work.}
% of our proposed 3D-VQ.
%
Note that a token is not only correlated to its corresponding supervoxel
% of size $\frac{T}{T'} \times \frac{H}{H'} \times \frac{W}{W'}$, 
but depends on other patches due to the non-local receptive field.

% \vspace{-4mm}
\paragraph{Inflation and padding.}
We initialize our 3D-VQ with weights from a 2D-VQ in a matching architecture to transfer learned spatial relationships~\cite{carreira2017quo}, known as 3D inflation.
We use inflation on small datasets such as UCF-101~\cite{soomro2012ucf101}.
We use a central inflation method for the convolution layers, where the corresponding 2D kernel fills in the temporally central slice of a zero-filled 3D kernel.
The parameters of the other layers are directly copied.
%
%All other parameters including the codebook are unchanged.
%
% This keeps the 3D network output of a static video the same as the 2D network output of a single frame, while encouraging the learning of motion into the 3D kernels.
%
To improve token consistency for the same content at different locations~\cite{ge2022long}, we replace the \texttt{same} (zero) padding in the convolution layers with \texttt{reflect} padding, which pads with non-zero values. 
%
% \lu{lijun is this the same as TATS? if so need a reference.} \lijun{added reference, it's not exactly the same as Jose's experiments pointed to the best choice of reflect rather than replicate}

% \vspace{-4mm}
\paragraph{Training.}
We apply the image perceptual loss~\cite{esser2021taming} on each frame. 
The LeCam regularization~\cite{tseng2021regularizing} is added to the GAN loss to improve the training stability. 
We adopt the discriminator architecture from StyleGAN~\cite{karras2019style} and inflate it to 3D.
%, with weights initialized by 3D inflation. 
%
%MH: do not simply say "change" (low novelty)
%With this change, our model is trained stably with GAN loss from the beginning, without needing the VQGAN two-stage training.
%
With these components, unlike VQGAN, our model is trained stably with GAN loss from the beginning.
%, without needing the VQGAN two-stage training.

%or vision transformer backbones~\cite{vim2021}, in which each token is represented as an integer index in a codebook. 
% \lu{we achieve this by [1 sentence]. As illustrate in \cref{fig:pipeline}, explain the high-level idea of your method [3-5] sentences. Here try to stay with natural language for better understanding. Below you gonna do formal description.}.

\begin{sidewaysfigure}
\centering
\begin{subfigure}[t]{0.490\linewidth}
\vskip 0pt 
\includegraphics[width=\linewidth]{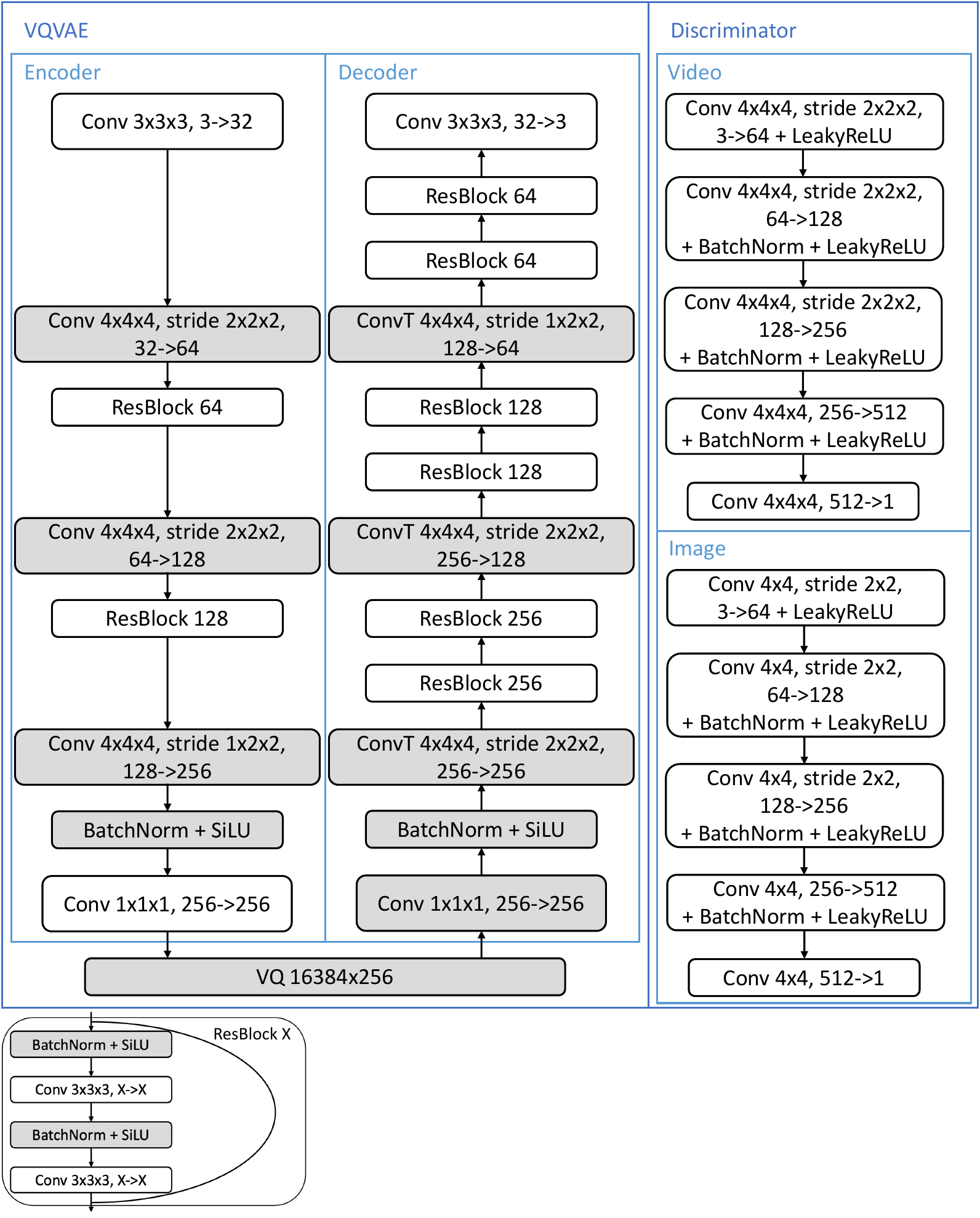}
\caption{TATS~\cite{ge2022long} 3D-VQ. (32M+14M parameters)}
\end{subfigure}
\begin{subfigure}[t]{0.490\linewidth}
\vskip 0pt 
\includegraphics[width=\linewidth]{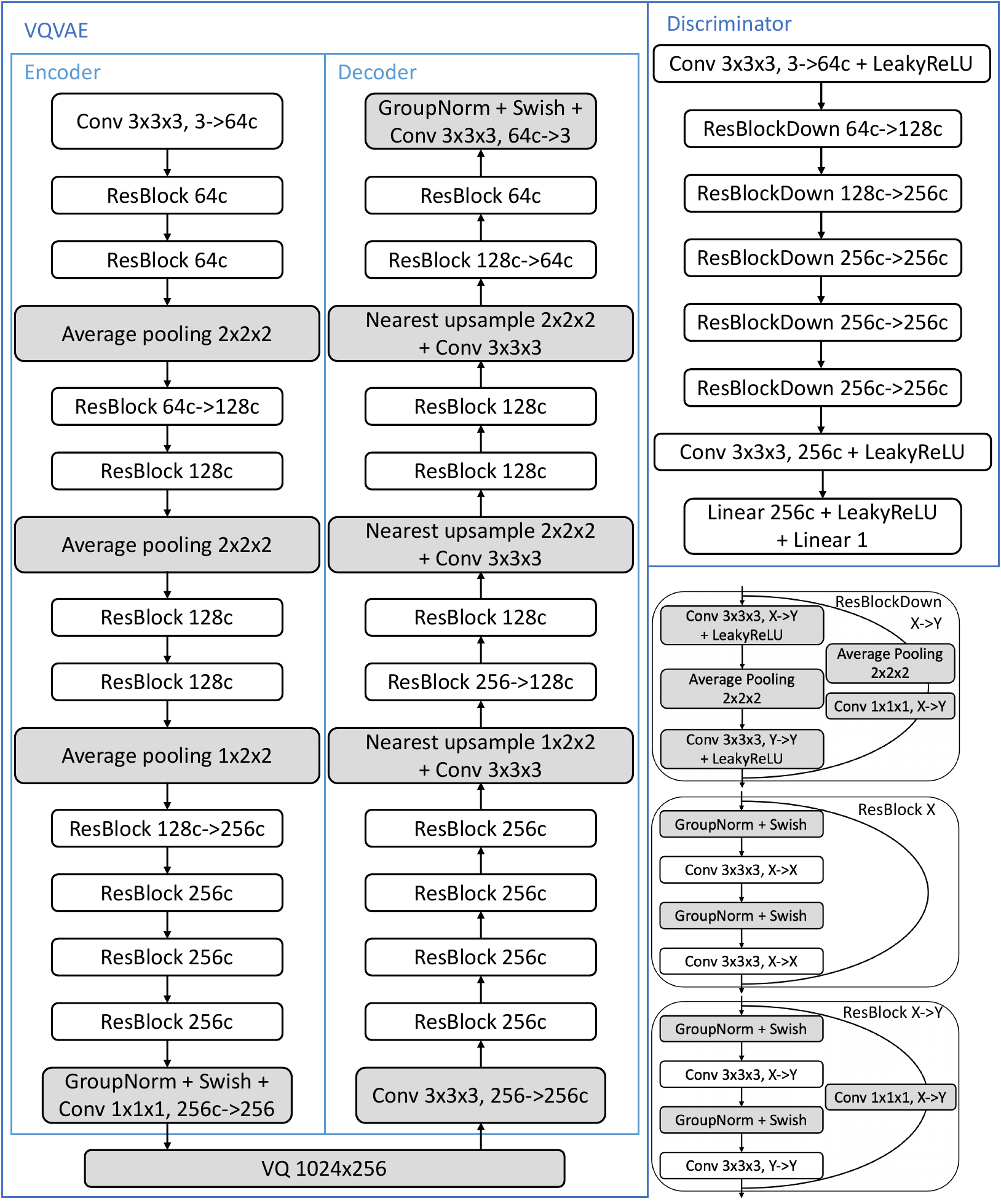}
\caption{\emph{\modelname{}} (ours) 3D-VQ.\\(41M+15M parameters at $c=1$ (B), 158M+61M at $c=2$ (L))}
\end{subfigure}
\caption[Comparison of 3D-VQ model architectures between \modelname{} and TATS]{\textbf{Comparison of 3D-VQ model architectures between \modelname{} and the TATS}~\cite{ge2022long}. 
We highlight the blocks with major differences in \textcolor{gray}{gray} background and detail their design differences in \cref{app:vq_arch}.
We train the models to quantize 16-frame clips of 128$\times$128 resolution into $4\times16\times16$ tokens.
The number of parameters in parentheses are broken down between VQVAE and discriminators.}
\label{fig:vq_arch}
\end{sidewaysfigure}

\subsection{Architecture Comparison}
\label{app:vq_arch}

We design two variants of the \modelname{} 3D-VQ module, \ie, the base (B) with 41M parameters and the large (L) with 158M parameters, excluding the discriminators.

\cref{fig:vq_arch} shows the architectures of the \modelname{} 3D-VQ module and compares it with the 3D-VQ module in TATS~\cite{ge2022long} which held the previous state-of-the-art for video generation. Compared with TATS, the major design choices in \modelname{} 3D-VQ are listed below. 
%design with the prior best 3D-VQ in TATS~\cite{ge2022long}.
% The encoder and decoder consist of cascaded residual blocks~\cite{he2016deep} interleaved by downsampling and upsampling layers.
%The major differences of our 3D-VQ design from TATS include:
\begin{itemize}
\item Average pooling, instead of strided convolution, is used for down-sampling.
\item Nearest resizing and convolution
%instead of the strided deconvolution, 
are used for up-sampling.
\item We use spatial down- and up-sampling layers near the latent space and spatial-temporal down- and up-sampling layers near the pixel space, resulting in mirrored encoder-decoder architecture.
\item A single deeper 3D discriminator is designed rather than two shallow discriminators for 2D and 3D separately.
\item We quantize into a much smaller vocabulary of 1,024 as compared to 16,384.
\item We use group normalization~\cite{wu2018group} instead of batch normalization~\cite{ioffe2015batch} and Swish~\cite{ramachandran2018searching} activation function instead of SiLU~\cite{hendrycks2016gaussian}.
\item We use the LeCAM regularization~\cite{tseng2021regularizing} to improve the training stability and quality.
\end{itemize}

\section{Experimental Results}
\subsection{Implementation Details}
\label{app:train}

We follow the same learning recipe across all datasets, with the only variation in the number of training epochs.
The implementation is available at 
Here are the training details for the 3D-VQ model:

\begin{multicols}{2}
\begin{itemize}
\item Video: 16 frames, frame stride 1, 128$\times$128 resolution. (64$\times$64 resolution for BAIR)
\item Base channels: 64 for B, 128 for L.
\item VQVAE channel multipliers: 1, 2, 2, 4. (1, 2, 4 for 64$\times$64 resolution).
\item Discriminator channel multipliers: 2, 4, 4, 4, 4. (2, 4, 4, 4 for 64$\times$64 resolution)
\item Latent shape: 4$\times$16$\times$16.
\item Vocabulary size: 1,024.
\item Embedding dimension: 256.
\item Initialization: central inflation from a 2D-VQ trained on ImageNet with this setup.
\item Peak learning rate: $10^{-4}$.
\item Learning rate schedule: linear warm up and cosine decay.
\item Optimizer: Adam with $\beta_1=0$ and $\beta_2=0.99$.
%\item Commitment cost: $0.25$.
\item Generator loss type: Non-saturating.
\item Generator adversarial loss weight: $0.1$.
\item Perceptual loss weight: $0.1$.
\item Discriminator gradient penalty: r1 with cost $10$.
\item EMA model decay rate: $0.999$.
\item Batch size: 128 for B, 256 for L.
\item Speed: 0.41 steps/sec on 16 TPU-v2 chips for B, 0.56 steps/sec on 32 TPU-v4 chips for L.
\end{itemize}
\end{multicols}

% The hardware requirement is 16 TPU-v2 chips for \modelname{}-B and 32 TPU-v4 chips for \modelname{}-L.
% The training speed is 0.4 steps/sec
Using more hardware resources can speed up the training.
We train \modelname{} 3D-VQ models for each dataset separately.
The training epochs for each dataset are listed in \cref{tab:train_step1}.

\begin{table}[tp]
\centering
\begin{tabular}{@{}lcc@{}}
\toprule
Dataset & B & L \\ \midrule
UCF-101 & 500 & 2000 \\
BAIR & 400 & 800  \\ 
Kinetics-600 & 45 & 180 \\
SSv2 & 135 & 400  \\
nuScenes & 1280 & 5120  \\
Objectron & 1000 & 2000  \\
Web12M & 5 & 20  \\ \bottomrule
\end{tabular}
\caption[Training epochs of \modelname{} 3D-VQ for each dataset]{\textbf{Training epochs of \modelname{} 3D-VQ for each dataset.}}
\label{tab:train_step1}
\end{table}

% \begin{table}[tp]
% \centering
% \caption{Comparison between the architecture, initialization, and scale for the tokenizer. FVD and IS values are from the reconstruction results on the UCF-101 training set.}
% \label{tab:token}
% \scriptsize
% \begin{tabular}{@{}lcccc@{}}
% \toprule
% Method & \makecell{ImageNet\\Initialization} & \# Params & FVD$\downarrow$ & IS$\uparrow$    \\ \midrule
% 2D-B   & No             & 14M       & 328 & 72.66 \\
% 2D-B   & Yes            & 14M       & 235 & 81.78 \\
% 2D-L   & No             & 53M       & 240 & 80.89 \\
% 2D-L   & Yes            & 53M       & 216 & 82.60 \\ \midrule
% 3D-B   & No             & 41M       & 127 & 82.10 \\
% 3D-B   & Average        & 41M       & 103 & 84.79 \\
% 3D-B   & Central        & 41M       & 58  & 86.99 \\
% 3D-L   & Average        & 158M      & 54  & 87.32 \\
% 3D-L   & Central        & 158M      & \textbf{39}  & \textbf{88.15} \\ \bottomrule
% \end{tabular}
% \end{table}

\begin{table}[tp]
\centering
% \scriptsize
% \begin{tabular}{@{}lcccc@{}}
% \toprule
% Init. & 2D-VQ-B (14M) & 2D-VQ-L (53M) & 3D-VQ-B (41M) & 3D-VQ-L (158M) \\ \midrule
% No & 328 / 72.7 & 240 / 80.9 & 127 / 82.1   & 45 / 87.1                                                     \\
% Yes & 235 / 81.8 & 216 / 82.6  %& 103 / 84.8   & 54 / 87.3 \\
% % Yes / Central & -  & - 
% & 58 / 87.0   & \textbf{25} / \textbf{88.9} \\ \bottomrule
% \end{tabular}
\begin{tabular}{@{}lcccccc@{}}%{@{}l@{\hspace{8pt}}c@{\hspace{5pt}}c@{\hspace{8pt}}c@{\hspace{5pt}}c@{\hspace{8pt}}c@{\hspace{5pt}}c@{}}
\toprule
\multirow{2}{*}{Tokenizer} & \multicolumn{2}{l}{From Scratch} & \multicolumn{4}{l}{ImageNet Initialization} \\
& FVD$\downarrow$ & IS$\uparrow$ & FVD$\downarrow$ & IS$\uparrow$ & FVD$\downarrow$ & IS$\uparrow$ \\ \midrule
MaskGIT~\cite{chang2022maskgit} 2D-VQ 
% & 14M & 328 / 72.7 & 235 / 81.8 \\
%  & 53+24M
 & 240 & 80.9 & 216 & 82.6& \multicolumn{2}{c}{-} \\
TATS~\cite{ge2022long} 3D-VQ & 
% 32+14M & 
162 & 80.6 & \multicolumn{2}{c}{-}& \multicolumn{2}{c}{-} \\ \midrule
& & & \multicolumn{2}{c}{Average} & \multicolumn{2}{c}{Central} \\
\emph{\modelname{}} 3D-VQ-B (ours) & 
% 41+15M &
127 & 82.1 & 103 & 84.8 & 58 & 87.0 \\ 
% & 158+61M
\emph{\modelname{}} 3D-VQ-L (ours)
& 45 & 87.1 & 35 & 88.3 & \textbf{25} & \textbf{88.9} \\
\bottomrule
\end{tabular}
% \vspace{-1mm}
\caption[Comparison of tokenizer architectures and initialization methods]{\textbf{Comparison of tokenizer architectures and initialization methods} on UCF-101 training set reconstruction results.
% The size of model parameters are in parentheses. 
% The values are reconstruction FVD and IS, different from the generation metrics.
The 2D-VQ compresses by 8$\times$8 spatially and the 3D-VQ compresses by 4$\times$8$\times$8 spatial-temporally.
% The number of parameters is broken down as VQVAE + discriminator.
% The cells are formatted as reconstruction FVD$\downarrow$ / IS$\uparrow$.
% $^*$ marks the values obtained from released models.
}
\label{tab:token}
% \vspace{-4mm}
\end{table}

\begin{table*}[tp]
\centering
% \scriptsize
\resizebox{1.0\textwidth}{!}{%
\begin{tabular}{lccccccccc}
\toprule
\multirow{2}{*}{VQ Tokenizer} & \multicolumn{3}{c}{From Scratch} & \multicolumn{6}{c}{ImageNet Initialization} \\
& PSNR$\uparrow$ & SSIM$\uparrow$ & LPIPS$\downarrow$ & PSNR$\uparrow$ & SSIM$\uparrow$  & LPIPS$\downarrow$ & PSNR$\uparrow$ & SSIM$\uparrow$ & LPIPS$\downarrow$ \\ \midrule
MaskGIT 2D  &
 21.4 & 0.667 & 0.139 & 21.5 & 0.685 & 0.114 & \multicolumn{3}{c}{-} \\
% TATS 3D & 
% 23.2 & 0.784 &  & \multicolumn{3}{c}{-}& \multicolumn{3}{c}{-} \\ 
\midrule
& & & & \multicolumn{3}{c}{Average} & \multicolumn{3}{c}{Central} \\
% \emph{\modelname{}} 3D-VQ-B & 
% 20.7 & 0.640 & 0.143 & 21.1 & 0.662 & 0.123 & 21.3 & 0.673 & 0.113 \\ 
\emph{\modelname{}} 3D-L &
21.8 & 0.690 & 0.113 & 21.9 & 0.697 & 0.103 & \textbf{22.0} & \textbf{0.701} & \textbf{0.099} \\
\bottomrule
\end{tabular}
}%
\caption[Image quality metrics of different tokenizers]{\textbf{Image quality metrics of different tokenizers} on UCF-101 training set reconstruction.
}
\label{tab:token_imquality}
\end{table*}

\subsection{Quantitative Evaluations}
% \lu{Here we should include 1) a baseline of existing 3D VQGAN architecture. 2) a baseline of existing 2D VQGAN achitecture (MAskGIT), 2) different padding comparison. 3) different inflations methods. Let's talk about this on Monday.}
%So far, we have verified \modelname{}'s generation quality on several datasets in terms of FVD and IS.
%Aside from the above-described quantitative results, 
We evaluate the design options of our \modelname{} 3D-VQ model.
\cref{tab:token} lists the reconstruction FVD and IS metrics on the UCF-101 training set, which provides an upper bound for generation quality.
In addition, we report the image quality metrics (PSNR, SSIM, LPIPS) for the VQGAN reconstruction in  \cref{tab:token_imquality}. 
% , which are different from the generation metrics as they measure the intermediate quantization.
% Nevertheless, reconstruction quality bounds the generation quality.
%and compare it to the existing 2D or 3D VQ architecture
We compare the proposed 3D architecture with existing 2D~\cite{chang2022maskgit} and 3D~\cite{ge2022long} VQ architectures. 
We train the MaskGIT~\cite{chang2022maskgit} 2D-VQ and our 3D-VQ with the same protocol and evaluate the official TATS~\cite{ge2022long} 3D-VQ model.
We compare two inflation methods for our 3D-VQ model, \ie, average~\cite{carreira2017quo} and central inflation.
% We implement these models and train them under the same protocol as \modelname{}.
%the tokenizer architectures and initialization, using reconstruction FVD and IS metrics on the UCF-101 training set.
% We compare the VQ architecture and initialization techniques at \cref{app:vq}.
%and 3D tokenizers with different initialization methods on the UCF-101 dataset, as shown in \cref{tab:token}.
%With an input of $16\times 128\times128$, the 2D models quantize it into $16\times16\times16$ tokens while the 3D models quantize to $4\times 16\times16$.

The results show the following. First, 3D-VQ models, despite producing a higher compression rate, show better video reconstruction quality than 2D-VQ, even with fewer parameters. 
Second, the proposed VQ performs favorably against baseline architectures with a similar size and gets much better with a larger model.
Third, ImageNet~\cite{deng2009imagenet} initialization boosts the performance for 2D and 3D models, where the central inflation outperforms the average inflation.
The results demonstrate the excellent reconstruction fidelity of our tokenizer design.
% , which enables the high generation quality of \modelname{}. 
%For the 3D inflation methods, central inflation appears much better than average inflation as it encourages dynamic motion.
%As we usually expect, increasing the model scale can improve the performance.

% \paragraph{Tokenizer reconstruction.}
% We report the image quality metrics (PSNR, SSIM, LPIPS) for the VQGAN reconstruction in  \cref{tab:token_imquality}. 
% We compare MAGVIT 3D against the baseline MaskGIT 2D to highlight our 3D design while keeping the remaining components the same. 
% As shown, the results in \cref{tab:token_imquality} are consistent with the findings by FVD in Tab. 6.

% \subsection{High-Fidelity Tokenization}

\subsection{Qualitative Evaluations}

% The quantitative comparison of the 3D-VQ from TATS and \modelname{} is presented in \cref{tab:im_quality}. ??
% In addition, 
We qualitatively compare the reconstruction quality of \modelname{} 3D-VQ and the baseline models on UCF-101. 
In addition, we present \modelname{}'s high-quality reconstruction on example YouTube videos with more samples at \url{https://magvit.cs.cmu.edu}.

\begin{figure*}[p]
\centering
\begin{subfigure}[t]{\linewidth}
\centering
\includegraphics[width=0.72\linewidth]{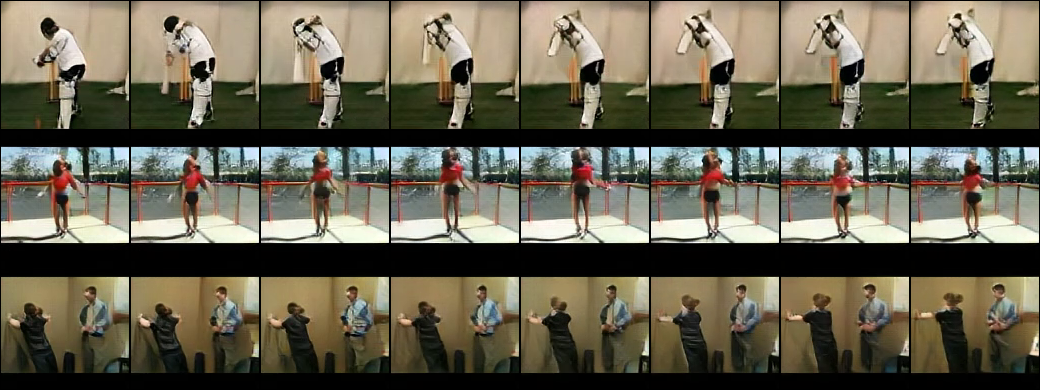}
\caption{MaskGIT~\cite{chang2022maskgit} 2D-VQ
% (FVD=240, IS=80.9)
}
\end{subfigure}
\hfill
\begin{subfigure}[t]{\linewidth}
\centering
\includegraphics[width=0.72\linewidth]{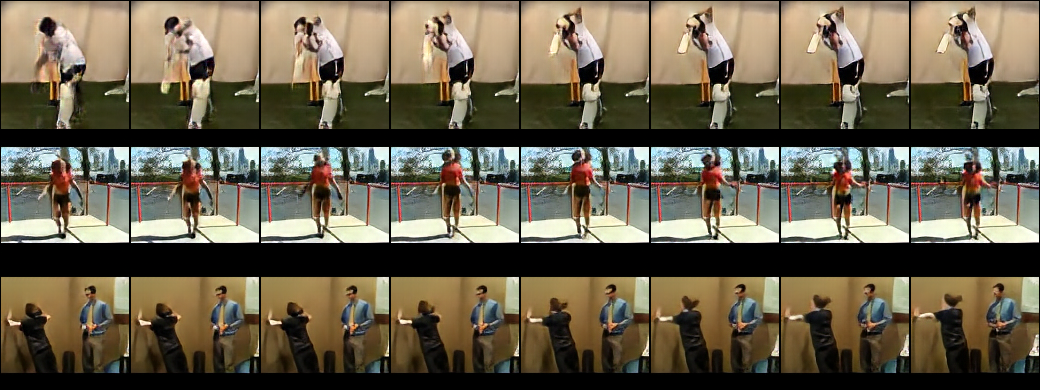}
\caption{TATS~\cite{ge2022long} 3D-VQ
% (FVD=162, IS=80.6)
}
\end{subfigure}
\hfill
\begin{subfigure}[t]{\linewidth}
\centering
\includegraphics[width=0.72\linewidth]{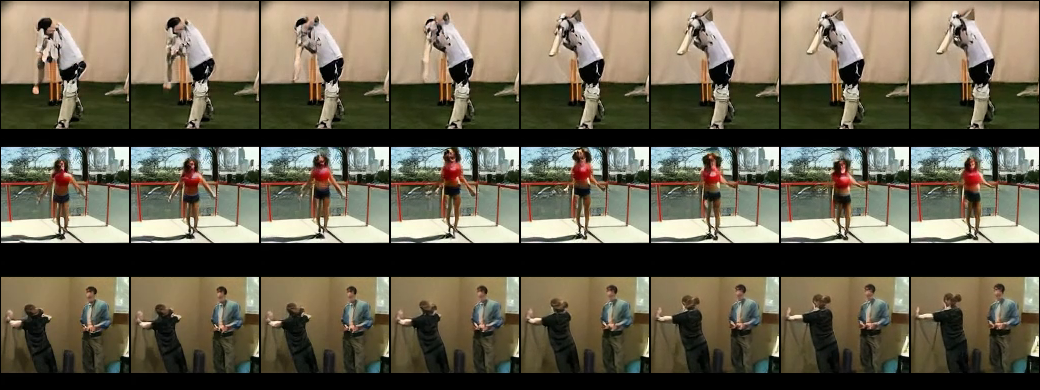}
\caption{\modelname{} 3D-VQ-L (ours)
% , FVD=25, IS=88.9)
}
\end{subfigure}
\hfill
\begin{subfigure}[t]{\linewidth}
\centering
\includegraphics[width=0.72\linewidth]{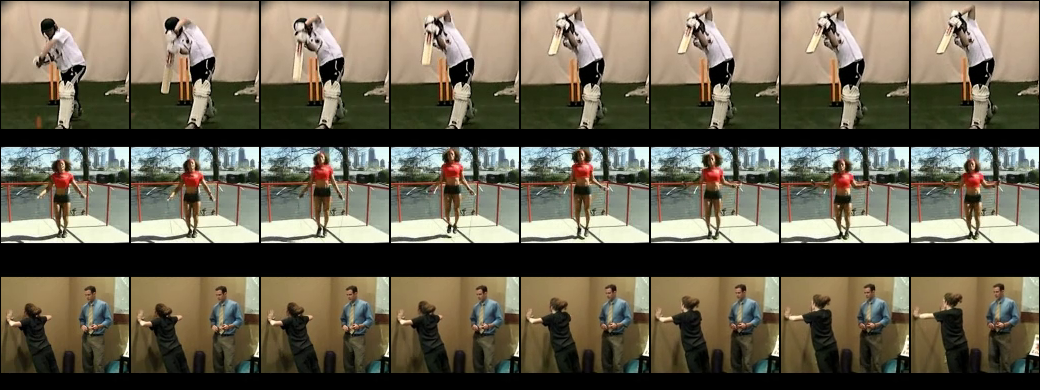}
\caption{Real
}
\end{subfigure}
\caption[Comparison of tokenizers on UCF-101 training set reconstruction]{\textbf{Comparison of tokenizers on UCF-101 training set reconstruction}.
Videos are reconstructed at 16 frames 64$\times$64 resolution 25 fps and shown at 12.5 fps, with the ground truth in (d). 
% MaskGIT 2D-VQ produces a reasonable image quality, but falls short of frame consistency which causes significant flickering when played as a video (\eg, the curtain color in the first row and the wall color in the third row).
% TATS 3D-VQ has a better temporal consistency but loses details for moving objects (\eg, the woman's belly in the second row).
% In contrast, our 3D VQ produces consistent frames with greater details reconstructed for both static and moving pixels.
% The reconstruction FVD and IS are listed in parenthese.
}
\label{fig:ucf_token}
\end{figure*}

\paragraph{Comparison of reconstruction quality.}
\cref{fig:ucf_token} compares the reconstruction quality of the three VQ tokenizers on the UCF-101, including the 2D-VQ from MaskGIT~\cite{chang2022maskgit}, the 3D-VQ from TATS~\cite{ge2022long}, and \modelname{} 3D-VQ, where the videos are taken from the UCF-101 training set. We obtain the TATS model from their official release at \url{https://songweige.github.io/projects/tats/}. We train the MaskGIT 2D-VQ and \modelname{} 3D-VQ using the same protocol on the UCF-101 dataset.

%evaluate the official TATS 3D-VQ model.
We can see that the MaskGIT 2D-VQ produces a reasonable image quality, but falls short of frame consistency which causes significant flickering when played as a video (\eg, the curtain color in the first row and the wall color in the third row).
TATS 3D-VQ has a better temporal consistency but loses details for moving objects (\eg, the woman's belly in the second row).
In contrast, our 3D VQ produces consistent frames with greater details reconstructed for both static and moving pixels.

\paragraph{Scalable tokenization.}
Since the tokenizers are trained in an unsupervised manner, they exhibit remarkable generalization performances and can be scaled to big data as no labels are required. To demonstrate this, we train a large \modelname{} 3D-VQ on the large YouTube-8M~\cite{abu2016youtube} dataset while ignoring the labels, and use the model to quantize randomly sampled videos on YouTube.

\begin{sidewaysfigure}[p]
\centering
\includegraphics[width=\linewidth,trim={0 13.7cm 0 0},clip,angle=180]{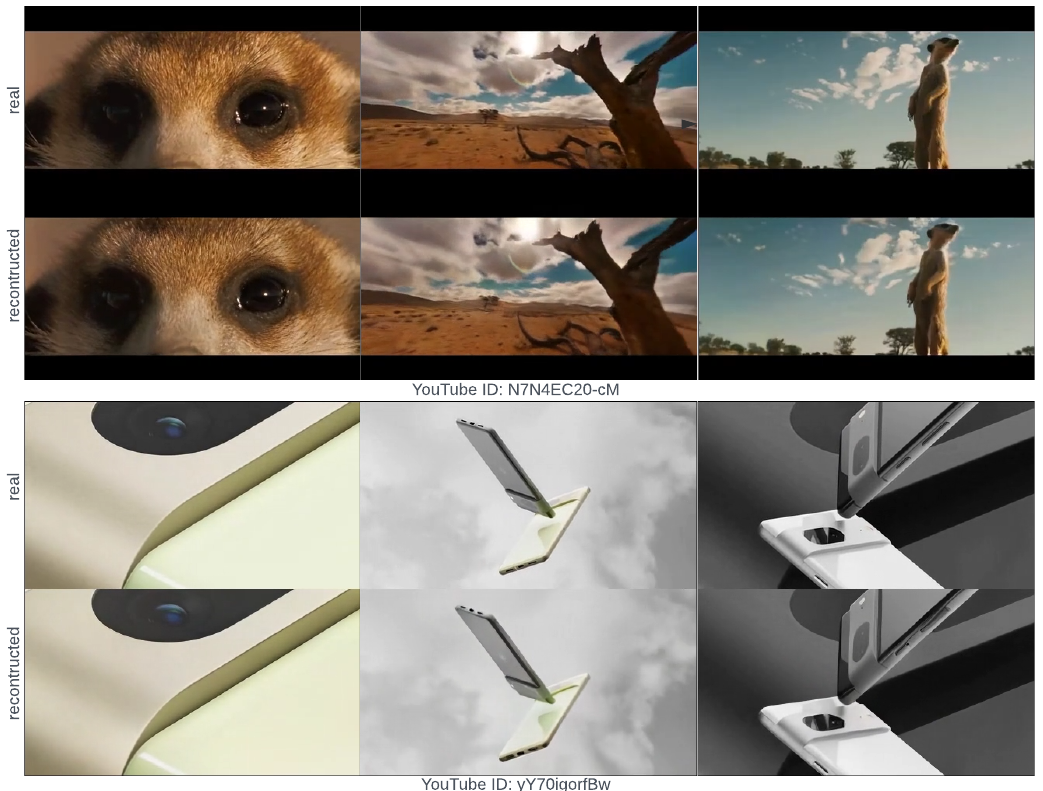}
\caption[High-fidelity reconstruction with scalable spatial-temporal resolution]{\textbf{Our 3D-VQ model produces high reconstruction fidelity with scalable spatial-temporal resolution.} For each group, the top row contains real YouTube videos and the bottom row shows the reconstructed videos from the discrete tokens. 
% The original videos are in 240p (240 $\times$ 432) resolution with $N$ frames.
% Our 3D-VQ model represents the video as $\frac{N}{4}$ $\times$30$\times$ 54 discrete tokens with a codebook of size 1024, representing a total compression rate of 614.4.
% Despite such high compression, the reconstructed results show stunning details and are almost indistinguishable from the real videos.
}
\label{fig:vq_sample1}
\end{sidewaysfigure}

\begin{sidewaysfigure}[p]
\centering
\includegraphics[width=\linewidth,trim={0 0 0 13.7cm},clip]{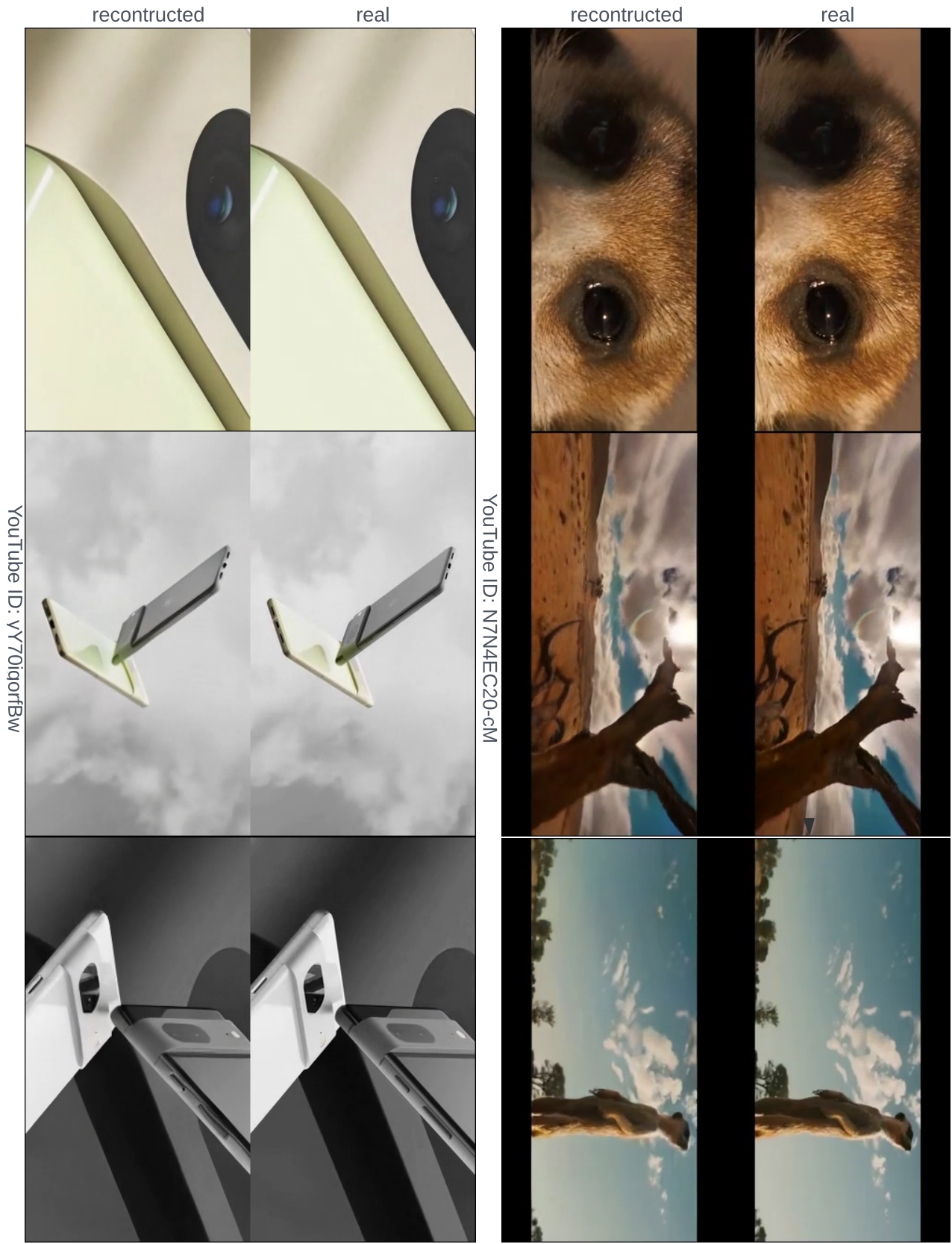}
\caption[High-fidelity reconstruction with scalable spatial-temporal resolution]{\textbf{Our 3D-VQ model produces high reconstruction fidelity with scalable spatial-temporal resolution}. For each group, the top row contains real YouTube videos and the bottom row shows the reconstructed videos from the discrete tokens.
}
\label{fig:vq_sample2}
\end{sidewaysfigure}

\cref{fig:vq_sample1,fig:vq_sample2} show the original and reconstructed videos from YouTube at 240p (240 $\times$ 432) resolution with arbitrary lengths (\eg 4,096 frames).
Although the tokenizer is only trained with 16-frame 128$\times$128 videos, it produces high reconstruction fidelity for high spatial-temporal resolutions that are unseen in training. Our 3D-VQ model compresses the video by a factor of 4 temporally, by 8$\times$8 spatially, and by 2.4 (24 bits $\rightarrow$ 10 bits) per element. %, yielding a 614.4$\times$ compression rate.
Our 3D-VQ model represents a $N$-frame video as $\frac{N}{4}$ $\times$30$\times$ 54 discrete tokens with a codebook of size 1024, representing a total compression rate of 614.4.
Despite such high compression, the reconstructed results show stunning details and are almost indistinguishable from the real videos.


\section{Summary}

In this chapter, we introduce the \modelname{} 3D-VQ model, which tokenizes a video into a low-dimensional spatial-temporal \emph{latent representation} with high reconstruction quality.
Quantitative and qualitative evaluations show the favorable performance of our model compared to previous best video tokenizers.
This successful design presents a compact but comprehensive latent space for video generation, which fosters the strong video generation results of MAGVIT~\cite{yu2022magvit} presented in \cref{chap:magvit}.

\glsresetall
\cleardoublepage
\chapter{Visual Lexical Representation}\label{chap:spae}
\graphicspath{{papers/spae/}}

\renewcommand{\modelname}{SPAE} % Semantical Pyramid AutoEncoder

\paragraph{Overview.}
% In this chapter, we propose a revolutionary method to map non-linguistic modalities, such as images, into the token space of a frozen Large Language Model (LLM).
% We introduce the Semantic Pyramid AutoEncoder (SPAE) for creating a multi-modal lexical representation for images and videos, thereby enabling the cross-modal capabilities of LLMs trained solely on text data.
% SPAE converts visual content into interpretable lexical tokens derived from the LLM's vocabulary. These tokens capture both semantic meaning and detailed information, enabling visual reconstruction and facilitating various multimodal tasks.
In this chapter, a revolutionary approach is presented for the mapping of non-linguistic modalities, like images, onto the token space of an immutable \Gls{LLM}. The innovation is encapsulated in the form of the \Gls{SPAE}, which is designed to learn a \emph{multi-modal} lexical \emph{latent representation} for images and videos. By doing so, it empowers the cross-modal proficiencies of LLMs that have been exclusively trained on textual data.
The core functionality of \Gls{SPAE} lies in its capacity to transmute visual content into intelligible lexical tokens that are harnessed from the vocabulary of the \Gls{LLM}. These tokens possess the dual capability of encapsulating both semantic significance and appearance details. As a result, they not only facilitate visual reconstruction but also serve as catalysts for a range of diverse multi-modal tasks.

% In contrast to typical VQ-VAE methods, SPAE's encoder maps to a discrete latent space of interpretable words. SPAE tokens exhibit a pyramid structure with multiple scales. The upper pyramid layers encode semantic concepts, while the lower layers prioritize appearance representations for fine image details. This design allows token length adjustments for different tasks, using fewer tokens for comprehension tasks and more tokens for generation tasks. This approach enhances the LLM's capability to understand and generate multimodal content.
\clearpage

\section{Motivation}
%Large Language Models (LLMs) have made significant advances in solving a wide range of NLP tasks, while expanding their capabilities beyond language into other modalities.
%However, little success has been made in the generation of another modality with an LLM itself.
% Large Language Models (LLMs) capture rich conceptual knowledge about the world in their lexical embeddings, from which certain inference can be carried out. 
%
% As such, we should be able to leverage frozen LLMs to perform visual tasks when the appropriate visual representations can be constructed. 
%
Large Language Models (LLMs) have made significant advances in solving a wide range of NLP tasks, while expanding their capabilities beyond language into other modalities.
To facilitate LLMs for such cross-modal tasks,
we propose to learn a vector quantizer to map an image, or some other non-linguistic (``foreign'') modality, 
to the token space of a frozen LLM.
% This effectively translates the image into a language that the LLM can comprehend, enabling us to leverage the generative abilities of the LLM to perform conditional image understanding and generation tasks without having to train on image-text pairs.

In this chapter, we %do just that and
introduce Semantic Pyramid AutoEncoder (SPAE) for a multimodal lexical representation for images and videos.
%  for enabling frozen LLMs to perform both understanding and generation tasks involving non-linguistic modalities such as images or videos. 
%
SPAE converts between raw pixels and interpretable lexical tokens (or words) extracted from the language model's vocabulary.
The resulting tokens capture both the semantic meaning and the fine-grained details needed for visual reconstruction, effectively translating the visual content into a language comprehensible to the LLM, and empowering it to perform a wide array of multimodal tasks.

\begin{figure}[tp]
%{r}{0.5\textwidth}
% \vspace{-5mm}
\centering
\includegraphics[width=\columnwidth,trim={0 0 0 0},clip]{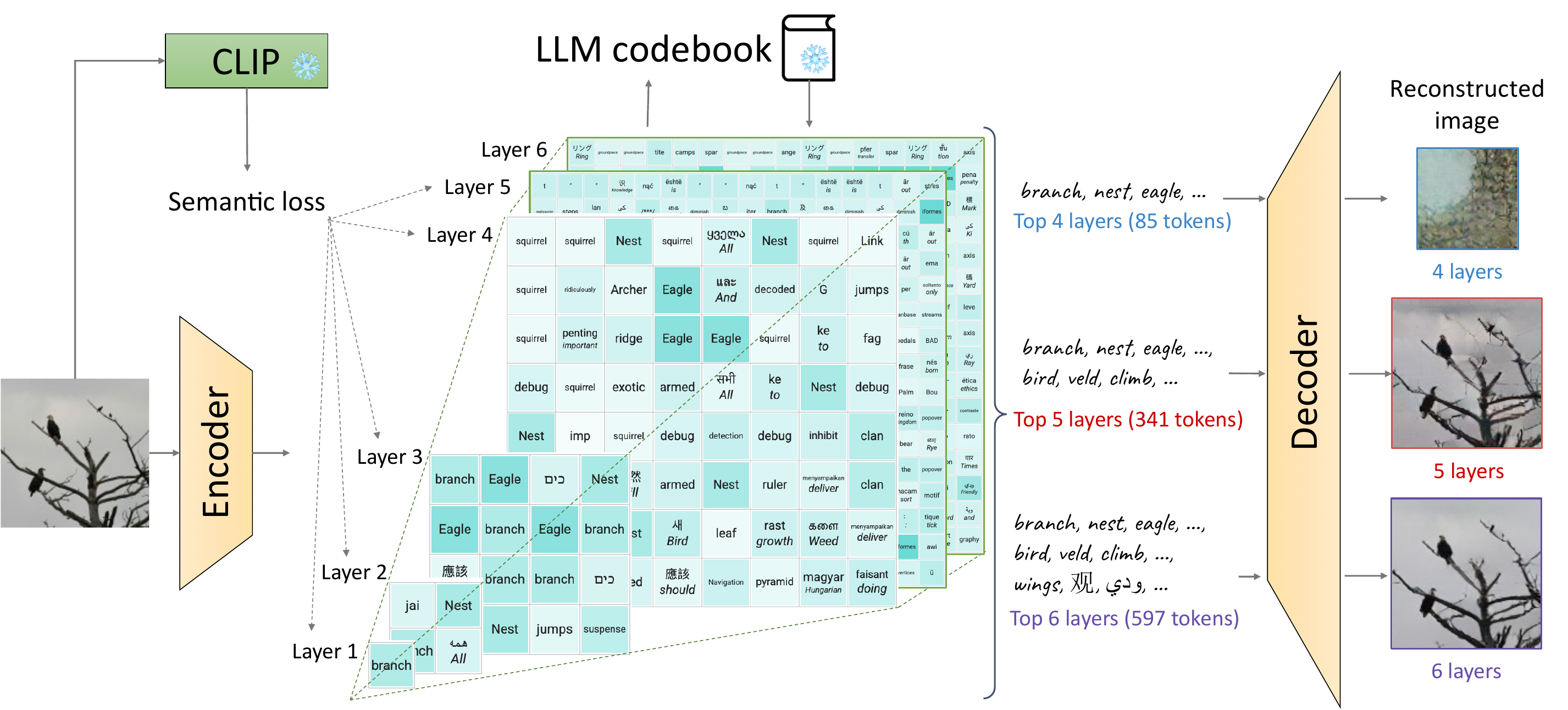}
\caption[Framework of the proposed \modelname{} model]{\textbf{Framework of the proposed \modelname{} model}.
An image is encoded into a pyramid of lexical tokens capturing semantic concepts and appearance details necessary for reconstruction.
% \mhy{I think you are working on this fiugre. After that, you need to refer to it in Section 1. Good to have a figure in page 2 (as early as possible)}
}
\label{fig:framework}
% \vspace{-4mm}
\end{figure}

% We introduce a novel Semantic Pyramid AutoEncoder (\modelname{}) that produces a lexical word sequence that (1) carries rich semantics, and (2) retains fine details for signal reconstruction. 
In contrast to the majority of VQ-VAE approaches~\cite{van2017neural}, our encoder maps to an interpretable discrete latent space, \ie, words. As depicted in \cref{fig:framework}, \modelname{} tokens have a multi-scale representation arranged in a pyramid structure.
The upper layers of the pyramid comprise semantic-central concepts, while the lower layers prioritize appearance representations that captures the fine details for image reconstruction.
This design enables us to dynamically adjust the token length to accommodate various tasks, such as using fewer tokens for understanding tasks and more tokens for generation tasks.

%without requiring any updates to the LLM's parameters. 
%
%By learning image tokens using the LLM's vocabulary, the visual content is effectively translated into a language comprehensible to the LLM,
%
% Our approach is validated through in-context learning experiments with frozen PaLM 2 and GPT 3.5 on a diverse set of image understanding and generation tasks.
% Our method marks the first successful attempt to enable a frozen LLM to generate image content while surpassing state-of-the-art performance in image understanding tasks, under the same setting, by over 25\%. 

% The proposed method is LLM-agnostic and has been demonstrated with PaLM-2/Bard and GPT-3.5.

\section{Prior Work}
% \lijun{ready, checked with grammarly}

% \vspace{-3mm}

% \vspace{-4mm}
\paragraph{Tokenization via vector quantization.}
VQ-VAE~\cite{van2017neural} compresses data into a discrete latent space defined by a codebook via vector quantization.
VQGAN~\cite{esser2021taming} enhances the reconstruction quality with adversarial and perceptual objectives.
These discrete latent quantities, often referred to as \emph{tokens}, are widely used to learn generative transformer models for image~\cite{rombach2022high,chang2022maskgit}, video~\cite{yu2022magvit,ge2022long,villegas2022phenaki}, image-video~\cite{yu2023language}, and audio~\cite{borsos2022audiolm,defossez2022high}.%\enlargethispage*{0.1ex}
Our \modelname{} model is built upon the VQGAN framework and applicable to different modalities.
% \vspace{-4mm}
\paragraph{Tokenization into lexical representations.}
The codebooks in typical VQGANs are learned jointly with the encoder and decoder stacks, which are not directly interpretable via natural languages.
LQAE~\cite{liu2023language} replaces the learned codebook with frozen word embeddings from BERT~\cite{devlin2019bert} to connect with an English vocabulary.
However, the LQAE tokens seldom contain semantic concepts in an image, and the reconstruction quality is worse than that with a learned codebook.
Our \modelname{} quantizes an input sample into semantically related tokens in a multilingual vocabulary while preserving the high reconstruction quality of a VQGAN for generative tasks.
In addition, \modelname{} tokens are organized in a  multi-layer coarse-to-fine pyramid for flexible usage in different tasks.

% MAGVLT~\cite{kim2023magvlt} for both image and text generation in a single transformer -> it's not an LLM
% \vspace{-4mm}

\section{SPAE: Semantic Pyramid AutoEncoder}

% % \vspace{-3mm}
Our goal is to model an image, or some other non-linguistic modality (\eg, video or audio), as a language sequence that LLMs can comprehend.
%enable multimodal understanding and generation with a frozen LLM.
\emph{Semantic Pyramid AutoEncoder} (\modelname{}) generates a lexical word sequence with dynamically adjustable length that carries rich semantics and retains fine details for signal reconstruction.
To work with a frozen LLM via in-context learning, we introduce a progressive in-context denoising method to facilitate image generation. 
We use the image modality in this section to introduce our \modelname{} model in 2D, and
later showcase the results of a 3D variant with the video modality in our experiments.

\subsection{Model Architecture}
% \label{sec:tokenizer}
% % \vspace{-3mm}

Our \modelname{} model extends the VQ-VAE~\cite{van2017neural} framework, which comprises an encoder, a quantizer, and a decoder. The CNN encoder maps an image $\rmI \in \sR^{H \times W \times 3}$ into continuous embeddings $\rmZ \in \sR^{h \times w \times c}$. Each element $\rvz \in \rmZ$ is then passed through the quantizer, which assigns it to the closest entry in a codebook, resulting in the quantized embedding. Let $\hat{\rmZ}$ represent the quantized embeddings for the entire image. The CNN decoder receives $\hat{\rmZ}$ as input and generates the reconstructed image $\hat{\rmI}$.
%Each element in is spatial-aware due to the convolution  or positional embedding in the decoder.
Below we highlight the design differences in \modelname{}.

%given its high reconstruction fidelity across different modalities.

As illustrated in~\cref{fig:framework}, \modelname{} generates lexical tokens arranged in a pyramid structure, which contains semantic concepts in the upper layers and appearance with progressively refined details in the lower layers.
We introduce a semantic loss to encourage the usage of conceptually relevant tokens.

%\lu{figure does not draw the progressive part and semantic meaning.}

% % \vspace{-4mm}
\paragraph{Frozen language codebook.}
%A standard VQGAN model jointly learns the VQ codebook with the encoder and decoder stacks through a dictionary learning objective~\cite{van2017neural}.
%Therefore, tokens from a VQGAN are not directly interpretable in natural languages.
To generate lexical tokens, we utilize a pretrained LLM codebook $\sC = \{(k, \rve(k)) \mid k \in \sT\}$ and freeze it during training, where $\sT$ is a subset of the LLM vocabulary.
Here, $\rve(\cdot)$ produces the text embedding for a sub-word $k$
% where $k$ denotes the integer index for the the sub-word token and  is the text embedding 
which may be obtained from any layer of the LLM.
% either from the LLM's input embeddings or the embeddings in subsequent layers. 
Since the codebook is aligned with the language vocabulary, we use the terms ``token'' and ``word'' interchangeably.

% To generate lexical tokens, we utilize a pretrained Language Model (LLM) codebook denoted as $\sC = {(k, \rve(k)) \mid k \in \sT}$. During training, we keep the codebook frozen. Here, $k$ represents the integer index associated with a sub-word token, and $\rve(\cdot)$ corresponds to the text embedding for that sub-word token.
%as in LQAE~\cite{liu2023language} 

% Lu: add this to the supplementary materials
% \begin{wrapfigure}{r}{0.4\textwidth}
% % % \vspace{-5mm}
% \centering
% \includegraphics[width=0.4\columnwidth,trim={0 0 0 0},clip]{figures/dilation_sampler.pdf}
% % % \vspace{-5mm}
% \caption{Dilation subsampler.}
% \label{fig:dilation}
% % % \vspace{-5mm}
% \end{wrapfigure}

\begin{figure}[tp]
% \vspace{-7mm}
\centering
\includegraphics[width=0.5\textwidth,trim={0 0 0 0},clip]{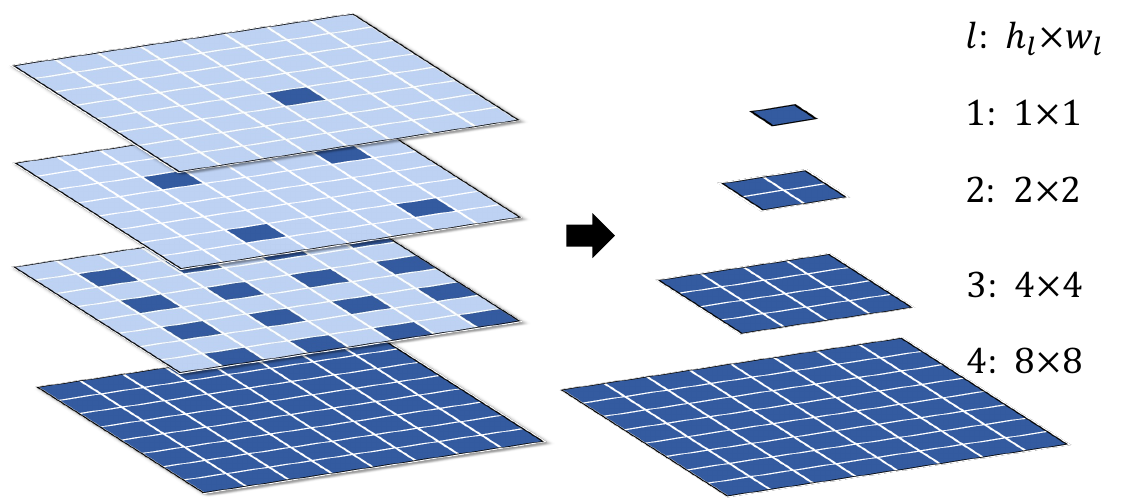}
% \vspace{-2mm}
\caption[Dilation subsampler visualization]{\textbf{Dilation subsampler visualization}.}
% \vspace{-10mm}
\label{fig:dilation}
\end{figure}

% % \vspace{-4mm}
\paragraph{Token pyramid.}
% \modelname{} encodes an image into embeddings $\rmZ \in \sR^{h_D \times w_D \times c}$
The \modelname{} quantizer produces $D$ layers of tokens where the tokens at layer $l$ are denoted as $\rvk_l  \in \sT^{h_l \times w_l}$.
%
%The embeddings are quantized into $D$ layers of tokens $\{\rvk_l  \in \sT^{h_l \times w_l} \mid l \in [1, 2, \hdots, D]\}$, 
Prior works use Residual Quantization (RQ) to generate multi-layer tokens~\cite{lee2022autoregressive,zeghidour2021soundstream}. 
In these methods, tokens from all layers have uniform shapes and do not carry specific semantic meanings. 
In contrast, we propose a pyramid token structure by enforcing the constraint $h_l \le h_{l+1} \land w_l \le w_{l+1}$. 
The pyramid structure is purposefully designed to concentrate semantics within the upper layers of the pyramid.
This design allows for representing semantic concepts with notably fewer tokens,
%  using a significantly reduced number of tokens, 
\eg, as few as five tokens for understanding tasks. 
The high token efficiency stems from the pyramid structure, as a conventional layer without pyramid structures needs a minimum of $hw$ tokens (\eg, 256) to represent the image. Token efficiency is crucial for in-context learning as it enables the accommodation of more examples within the context.
% RQ does not have this challenge as its tokens do not carry any inherent semantic meaning.
% \kevin
% {It is unclear if the spatial arrangment of tokens is related to the spatial structure of the image, or is just a bag of tokens at each layer.
% I think what you are doing is partitioning all the discrete tokens into a set of sets,
% with $D$ outer sets,
% and each inner set for each layer $l$.
% }
% \lijun{Based on empirical results, tokens from convolution-based VQVAE have spatial correspondence with the image patches, but I think this is irrelevant to what we are trying to describe here.
% The tokens are in an ordered list of sets, with each set for each layer $l$, and deeper layers could depend on the previous ones, both in RQ and SAQ.
% } \lu{I think this is implied as each layer follows the VQVAE paper whose decoder uses CNNs or ViTs with positional embedding.}
% \kevinn{Oh, I see. Just mention you use CNN encoder to get a spatial continuous feature map, or VIT with positional embedding. That will clarify things.} % done
%layers in monotonically increasing sizes. This structure concentrates high level concepts in the shallow layers with fewer tokens, while appearance details are preserved in deeper layers.
\cref{fig:dilation} illustrates the formation of
the pyramid with 
a dilation subsampler $\rmP(l)$, which selects the positions for quantization at layer $l$ as
% % \vspace{-2mm}
\begin{equation}
    \rmP(l) = \{(h'i - \left\lceil\frac{h'}{2}\right\rceil + 1, w'j - \left\lceil\frac{w'}{2}\right\rceil + 1) \mid (i, j) \in ([1, h_l] \times [1, w_l]) \cap \sZ^2  \},
\end{equation}
% % % \vspace{-4mm}
where $h'=\frac{h_D}{h_l}$, and $w'=\frac{w_D}{w_l}$ are the downsample ratios.

\begin{figure}[tp]
    % \vspace{-5mm}
\centering
\includegraphics[width=0.5\textwidth,trim={0 0 0 0},clip]{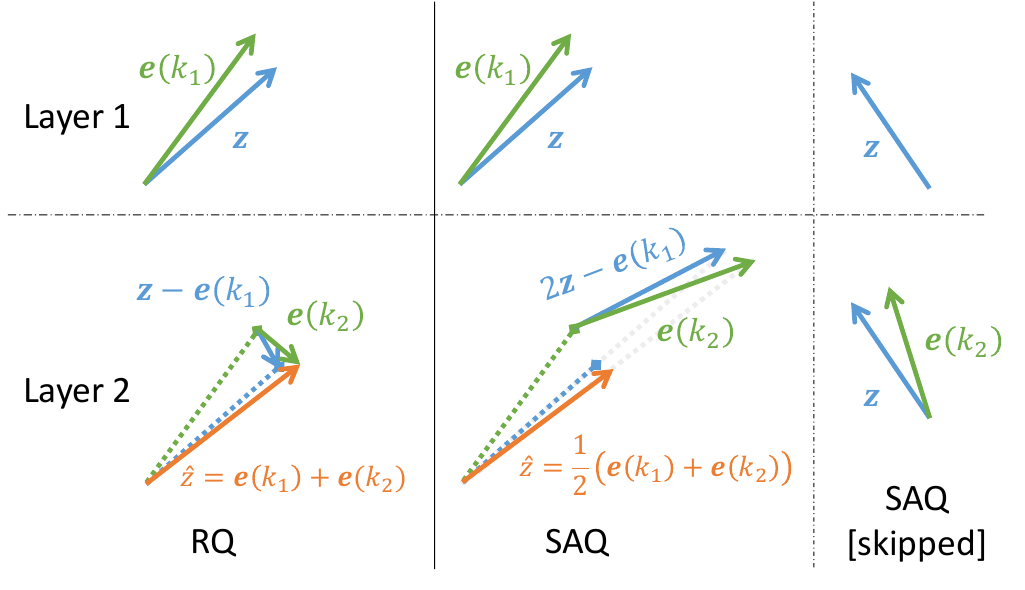}
\caption[Comparison between RQ and SAQ]{\textbf{Comparison between RQ and SAQ}. We show a 2-layer quantization process in a 2-dimensional space as an example. At layer $l$, we use \textcolor{MidnightBlue}{blue for the current remainder embeddings $\rvz_l$}, \textcolor{YellowGreen}{green for current post-quantization embeddings $\rve(k_l)$}, and \textcolor{orange}{orange for the reconstructed embeddings up to layer $l$ as $\hat{\rvz}_{\le l}$.}}
\label{fig:saq}
% \vspace{-10mm}
\end{figure}

% \begin{wrapfigure}{r}{0.4\textwidth}
% % % \vspace{-5mm}
% \centering
% \includegraphics[width=0.4\columnwidth,trim={0 0.4cm 0 0.4cm},clip]{figures/saq_vs_rq.pdf}
% % % \vspace{-5mm}
% \caption{Comparison between RQ and SAQ.}
% % % \vspace{-3mm}
% \label{fig:saq}
% \end{wrapfigure}
% % % \vspace{-4mm}
\paragraph{Streaming average quantization.}
% For each latent position, the token at layer $l$ is quantized from
% \begin{equation}
%     \rvz_l = \rvz - \textstyle \sum_{i=1}^{l-1}\rve(k_i)
% \end{equation}
% The embedding up to layer $l$ is reconstructed as
% \begin{equation}
%     \rvz'_l = \textstyle \sum_{i=1}^{l}\rve(k_i)
% \end{equation}
% \kevin{
% This is unclear.
% I thought you learned a bag of codewords for the entire image? Where does spatial locality come in? 
% }
For each embedding $\rvz$ at position $(x, y)$, we obtain its discrete tokens $k_l$ sequentially from layer $1$ to $D$.
At layer $l$, if $(x, y) \in \rmP(l)$, the quantizer assigns discrete token
% we obtain its discrete token at layer $l$ by 
$k_l = \argmin_{k \in \sT}\|\rvz_l - \rve(k)\|_2^2$, where $\rvz_l$ is the current layer embedding, calculated from
% % \vspace{-1mm}
\begin{equation}
\label{eq:svq_encoder}
    \rvz_l = \rvz + \textstyle \sum_{i=1}^{l-1}\1_{(x, y) \in \rmP(i)}(\rvz - \rve(k_i)).
\end{equation}
% \kevin{Should this be $\hat{z}_{lxy}$ on LHS?} \lijun{This is pre-quantization embedding, but I agree with the suggestion below to define a reconstruction function for clarity}
% \lu{clarify it should be $\rvz_{(l, x, y)}$}
% where $\1$ is the indicator function. 
% \kevinn{What is difference between $z_{lxy}$
% and $z_{xy}$?}
% \lijun{$z_{lxy}$ is the remainder embedding to be quantized at layer $l$. $z_{xy}$ is the original embedding output, which is also $z_{0, x, y}$. Updated for better clarity.}
The quantized embedding reconstructed with the first $l$ layers is given by the average of the existing token embeddings as
% \kevinn{How does $\hat{z}$ relate to $z$?}
%decoder reconstructs taking the input of $\hat{\rmZ}(l)$, up to layer $l$,
% % \vspace{-2mm}
\begin{equation}
    \hat{\rvz}_{\le l} = \frac{\sum_{i=1}^{l}\1_{(x, y) \in \rmP(i)}\rve(k_i)}{\sum_{i=1}^{l}\1_{(x, y) \in \rmP(i)}}.
\end{equation}
Using the input of $\hat{\rmZ}_{\le l}$ from tokens up to layer $l$, the decoder can progressively reconstruct the image with dynamic token lengths, resulting in gradually improved quality with refined appearance details.
%
% \kevin{It would be clearer to define the reconstruciton as a function $\hat{z} =  r(z) = r'(k(z))$, which takes the continuous encoding $z$, maps it to discrete codewords $k(z)$,
% and then reconstructs by doing spatial averaging.
% }\lu{added the reconstruction when introducing VQVAE.}
%
We term this approach as \emph{Streaming Average Quantization} (SAQ) due to its resemblance to computing the average on streaming data, where $\hat{\rvz}_{\le l+1} = \hat{\rvz}_{\le l} + \frac{1}{\hat{l} + 1}\rve(k_{l+1}), \hat{l}=\sum_{i=1}^{l}\1_{(x, y) \in \rmP(i)}$.
% as shown in \cref{eq:svq_encoder} to compute the unnormalized average embedding in each layer.
% \kevin{What does streaming mean?}\lu{added}
\cref{fig:saq} compares our proposed SAQ with Residual Quantization (RQ)~\cite{lee2022autoregressive,zeghidour2021soundstream}.
RQ is applicable but yields worse results in this context, as revealed by our ablation studies.
This can be attributed to (1) varying scales of embeddings in residual layers, potentially dividing the codebook into multiple parts, and (2) misalignment in the summation of word embeddings, which undermines learning semantically meaningful tokens in later layers.
At layer 2, the SAQ remainder embedding $\rvz_2 = 2\rvz - \rve(k_1)$ is at a more similar scale to $\rvz$, compared to the RQ remainder $\rvz - \rve(k_1)$.
We find that the scale consistency promotes better utilization of the frozen language codebook despite a large number of layers being used.
% An average of word embeddings also leads to better semantic alignment than a summation.
% With variable layer sizes, quantization in the first few layers may be skipped for those positions not selected by the subsampler.
Due to the pyramid structure, quantization in the first few layers may be skipped for those positions not selected by the dilation subsampler. Considering the scale consistency across quantization layers, the use of SAQ is more appropriate in this case.

% to produce multi-layer tokens in the pyramid structure.
% RQ is sub-optimal for \modelname{} with a frozen language codebook, in part because
% (1) RQ negatively affects the code utilization rate because embeddings in different residual layers are at different scales, which could divide the codebook into multiple parts;

% SAQ takes a streaming average of the embeddings, which dynamically simulates a bag-of-words representation.
% Variable layer shapes are supported by skipping early quantization steps when a position is only present in later layers.

% % \vspace{-4mm}
\subsection{Training Objectives}
\paragraph{Semantic loss.}
We encourage the semantic similarity between the image $\rmI$ and each lexical token $k$ denoted by $s(\rmI, k)$.
During training, we build per-layer candidate token pools as
% % \vspace{-1mm}
\begin{equation}
    \rmC_l(\rmI) = \{k \in \sT \mid s(\rmI, k) \ge \rho_l\},
\end{equation}
where $\rho_l$ is a threshold.
We set $\rho_l \ge \rho_{l+1}$ to allow deeper layers to have a larger pool of candidate tokens while sacrificing some semantics.

% Using image-text pairs~\cite{liu2022cross} could lead to an ideal definition of $s(\rmI, k)$, but it limits the utilization of large-scale unpaired data.
To define the similarity score,
we employ a pretrained CLIP model~\cite{radford2021learning}.
%to capture similarity and train \modelname{} on unpaired image (and video) data without text descriptions.
In more details,
let $f_\gI$ and $f_\gT$ be a pair of image and text CLIP embedding functions.
We precompute the text feature for each token $k \in \sT$ as
% % \vspace{-2mm}
\begin{equation}
    f'_\gT(k) = \frac{1}{|\rvp|} \textstyle \sum_{i=1}^{|\rvp|} f_\gT(\ervp_i(k)),
\end{equation}
where $\rvp$ is a list of prompt templates, such as \texttt{"a photo of ..."}.
During training, we extract the image feature $f_\gI(\rmI)$ and compute the dot-product similarity as $\rvs'(\rmI, k) = f_\gI(\rmI) \cdot f'_\gT(k)$.
The similarity score is then normalized to account for the varying scales across different images.
%scale of the similarity scores varies by image, so we use the relative score as the final measurement:
% % % \vspace{-1mm}
\begin{equation}
    \rvs(\rmI, k) = \frac{\rvs'(\rmI, k) - \textstyle \min_j \rvs'(\rmI, j)}{\max_j \rvs'(\rmI, j) - \min_j \rvs'(\rmI, j)}. \label{eq:relative_clip}
\end{equation}

We define the semantic loss 
for the encoder parameters $\theta_e$ 
% that generate $\rmZ$ 
as
% % \vspace{-2mm}
\begin{equation}
    \gL_\mathrm{semantic}(\theta_e ;\rmI) = \mathop{\E}\limits_{l \in [1, D']} \ \ \mathop{\E}\limits_{\rvz_l} \ \ \mathop{\E}\limits_{c \in \rmC_l(\rmI)} -\log \frac{\exp(-\|(\rvz_l - \rve(c)\|^2_2)}{\sum_{k \in \sT} \exp(-\|\rvz_l - \rve(k)\|^2_2)},
\end{equation}
where we randomly sample semantically
similar target codes $c$ for each layer embedding in the first $D'$ layers.
% This  semantic loss is applied to the upper layers of the token pyramid. 
%when $\rmC_l(\rmI)$ is non-empty.

% % \vspace{-4mm}
\paragraph{Appearance loss.}
Using an improved objective from~\cite{yu2022magvit}, the appearance loss is calculated:
% % % \vspace{-2mm}
\begin{equation}
% \begin{small}
    \gL_\mathrm{appearance}(\theta_e, \theta_d; \rmI) \!\!=\!\! \|\rmI - \hat{\rmI}\|_2^2 + \beta \textstyle \sum_{l=1}^D\|\rmZ - \text{sg}(\hat{\rmZ}_{\le l})\|_2^2 + \lambda \gL_\mathrm{GAN} + \eta \gL_\mathrm{Perceptual} + \phi \gL_\mathrm{LeCAM},
% \end{small}
\end{equation}
% \kevin{How is $\hat{I}$ computed?
% Why do you care about reconstruction accuracy of $\hat{Z}$?
% }\lu{$\hat{I}$ is the decoded image. $\|\rmZ - \hat{\rmZ}_l\|_2^2$ is in the VQVAE paper's Eq. (3) the third loss. lijun you need to put a sg here $\|\rmZ - \text{sg}(\hat{\rmZ}_l)\|_2^2$??}
where $\gL_\mathit{GAN}$, $\gL_\mathit{Perceptual}$, and $\gL_\mathit{LeCAM}$ are the VQGAN~\cite{ge2022long}, perceptual~\cite{johnson2016perceptual}, and LeCAM~\cite{tseng2021regularizing} losses. 
In addition, $\text{sg}(x)$ is the stop-gradient operation.
The appearance loss is applied to both the encoder $\theta_e$ and decoder parameters $\theta_d$, excluding the frozen codebook embedding.
% \kevinn{You need to give the equaiton for $\hat{I}$}. % defined in paragraph 1 of sec 3.1, which would be decoder(quantizer(encoder(I)))

\paragraph{Overall objective.}
To stabilize the training and balance between appearance and semantics, we add a dynamic weight for the semantic guidance loss as $w = \text{sg}\Big(\frac{\gL_\mathrm{appearance}(\rmI)}{\gL_\mathrm{semantic}(\rmI)}\Big)$.
The total training loss excluding the GAN discriminator is
% % \vspace{-3mm}
\begin{equation}
    \gL_\mathrm{\modelname{}}(\theta_e, \theta_q) = \mathop{\E}\limits_{\rmI} \Big[\gL_\mathrm{appearance}(\theta_e, \theta_q; \rmI) + \alpha w \gL_\mathrm{semantic}(\theta_e; \rmI)\Big].
\end{equation}
% \kevin{This is confusing. You put an sg on the first term but not the second, so you will still end up taking gradients wrt both losses.
% }
% During decoding, quantized embeddings $\rmX' \in \sR^{h_k \times w_k}$ can be constructed with the first $k \le d$ levels of tokens, which is then decoded into the pixel space.

% \subsubsection{Loss}

% It is worth noting that \modelname{} is trained standalone without back propagation through any language model, which is different from LQAE and Frozen~\cite{tsimpoukelli2021multimodal}.

\section{Experimental Results}
\subsection{Experimental Settings}

% \vspace{-3mm}
\paragraph{Language codebooks.}
To verify the compatibility with different LLMs, we train two variants of \modelname{}, namely \modelname{}$_\text{PaLM}$ and \modelname{}$_\text{GPT}$.
The \modelname{}$_\text{PaLM}$ codebook is taken from the input embedding layer of a PaLM 2-S checkpoint with a 
65k vocabulary of the most frequent sentence pieces.
\modelname{}$_\text{GPT}$ uses a byte-pair encoding vocabulary with 99k UTF-8 tokens
(\url{https://github.com/openai/tiktoken}),
%\footnote{\url{ https://github.com/openai/tiktoken}}, 
where we obtain the contextual token embeddings from OpenAI \texttt{text-embedding-ada-002}
%\footnote{
(\url{ https://platform.openai.com/docs/models/embeddings}).

\paragraph{Image \modelname{}.}
An image SPAE encodes a 128$\times$128 image into 16$\times$16 embeddings.
Following the VQGAN~\cite{esser2021taming} architecture, we use 128 base filters with channel multipliers [1, 2, 2, 4] and 2 residual blocks at each scale, which results in 59M parameters in total.
The embeddings are quantized into a token pyramid of 6 layers where each layer has $2^k \times 2^k$ tokens and $k=[0,1,2,3,4,4]$. 
We apply semantic guidance loss to the first five layers, with thresholds of 0.98, 0.95, 0.9, 0.85, and 0.8.
The CLIP with a ViT-L/14~\cite{dosovitskiy2020image} vision backbone is used.
We use 80 prompt templates from the zero-shot ImageNet classification setup to precompute the CLIP text embeddings for the vocabulary.
In addition, we use the Adam~\cite{kingma2014adam} optimizer with loss weights $\alpha=1, \beta=0.33, \lambda=0.1, \eta=0.1, \phi=10^{-4}$ and a learning rate of $10^{-4}$ following a linear warmup/cooldown and root square decay schedule.
Following the prior work~\cite{liu2023language}, \modelname{} is trained on the ImageNet ILSVRC2012~\cite{deng2009imagenet} dataset.
We train with a batch size of 256 for 450k steps, which takes 1.4k TPUv3-hours.

\paragraph{Image \modelname{}-8.}
In addition to the primary \modelname{} model with six pyramid layers, we also train an \modelname{}-8 model with eight layers to conduct a more in-depth analysis of the coarse-to-fine reconstruction process.
The two extra layers each contain 16$\times$16 tokens.
The semantic loss is still applied on the first 5 layers as in the primary model.
%So the last four equal-sized layers at the end to capture appearance details.

\paragraph{Video \modelname{}.}
We initialize a video \modelname{} by VQGAN inflation~\cite{yu2022magvit} from a pretrained image \modelname{}, which encodes 16 frames at 128$\times$128 resolution into 4$\times$16$\times$16 embeddings.
A video \modelname{} consists of 176M parameters.
The pyramid layers contain 1$\times$1$\times$1, 1$\times$2$\times$2, 1$\times$4$\times$4, 2$\times$8$\times$8, 4$\times$16$\times$16, and 4$\times$16$\times$16 tokens.
The video embedding is obtained as the average CLIP embedding for all frames.
The model is trained on the Kinetics-600~\cite{carreira2018short} dataset which contains 384k videos.
We train with a batch size of 512 for 130k steps, which takes 5.8k TPUv4-hours.

\subsection{Quantitative Evaluations}

%\paragraph{Appearance and semantic metrics.}
We compare the image and video reconstruction quality using the tokens produced by \modelname{} and the VQGAN baseline used in state-of-the-art image~\cite{chang2022maskgit,lezama2023discrete,chang2023muse} and video generation~\cite{yu2022magvit}. We use FID~\cite{heusel2017gans}, Inception Score (IS)~\cite{salimans2016improved}, and LPIPS~\cite{zhang2018unreasonable} to compare with the image VQGAN from MaskGIT~\cite{chang2022maskgit} on the ImageNet validation set, and FVD~\cite{unterthiner2018towards} to compare the 3D-VQGAN from MAGVIT~\cite{yu2022magvit} on the  Kinetics-600 validation set. To quantify the semantics, we compute the CLIP and relative CLIP scores (\cref{eq:relative_clip}), both averaged across all lexical tokens.

\begin{table}[tp]
% \vspace{-3mm}
\centering
\resizebox{1.0\textwidth}{!}{%
\begin{tabular}{@{}lccccccccc@{}}
\toprule
\multirow{2}{*}{Model} & \# Layers & \multirow{2}{*}{FID$\downarrow$}  & \multirow{2}{*}{IS$\uparrow$}     & \multirow{2}{*}{LPIPS$\downarrow$} & \multirow{2}{*}{CLIP$\uparrow$}   & Classification \\ 
& : \# Tokens & & & & & Accuracy$\uparrow$ \\ \midrule
VQGAN  & 1: 256 & 5.48 & 119.69 & 0.13  & n/a & 19.6       \\ \midrule
\ \ \ \ + frozen codebook & 1: 256 & 7.44 & 101.39 & 0.17  & 0.1464 & 19.5   \\
\ \ \ \ \ \ \ \ + semantic guidance & 1: 256 & 5.17 & 124.41 & 0.13  & 0.1518 & 46.2   \\ \hdashline
\multirow{2}{*}{\ \ \ \ \ \ \ \ \ \ \ \ + 2-layer RQ} & 1: 256 & 11.94 & 89.01 & 0.22 & 0.1595 & 56.2   \\
& 2: 512 & 6.05 & 113.93 & 0.15  & 0.1547 & -   \\  \hdashline
\multirow{2}{*}{\ \ \ \ \ \ \ \ \ \ \ \ + 2-layer SAQ} & 1: 256 & 12.30 & 93.33 & 0.21 & 0.1613 & 56.6 \\
& 2: 512 & 5.08 & 125.27 & 0.14  & 0.1595 & -   \\ \midrule
\multirow{6}{*}{\makecell[l]{\ \ \ \ \ \ \ \ \ \ \ \ + 6-layer pyramid SAQ\\\emph{\modelname{}} (ours)}} & 1: 1 & - & - & - & \textbf{0.1879} & 52.0 \\
& 2: 5 & - & - & - & 0.1868 & 64.2 \\
& 3: 21 & - & - & - & 0.1815 & \textbf{65.1} \\
& 4: 85 & - & - & - & 0.1711 & 58.5 \\
& 5: 341 & 9.49 & 109.46 & 0.17 & 0.1604 & 46.3 \\
& 6: 597 & \textbf{4.41} & \textbf{133.03} & \textbf{0.12}  & 0.1577 & -   \\ \hdashline
\ \ \ \ \ \ \ \ \ \ \ \ \ \ \ \ \textcolor{mygray}{no per-layer thresholds} & \textcolor{mygray}{6: 597} & \textcolor{mygray}{4.33} & \textcolor{mygray}{122.25} & \textcolor{mygray}{0.11} & \textcolor{mygray}{0.1650} & \textcolor{mygray}{59.4 (layer 3)} \\
\ \ \ \ \ \ \ \ \ \ \ \ \ \ \ \ \textcolor{mygray}{no dynamic semantic weight} & \textcolor{mygray}{6: 597} & \textcolor{mygray}{9.00} & \textcolor{mygray}{85.14} & \textcolor{mygray}{0.19} & \textcolor{mygray}{0.1847} & \textcolor{mygray}{65.1 (layer 3)} \\
\ \ \ \ \ \ \ \ \ \ \ \ \ \ \ \ \textcolor{mygray}{no perceptual loss} & \textcolor{mygray}{6: 597} & \textcolor{mygray}{40.47} & \textcolor{mygray}{33.41} & \textcolor{mygray}{0.20} & \textcolor{mygray}{0.1994} & \textcolor{mygray}{69.5 (layer 3)} \\ \midrule
\multirow{8}{*}{\makecell{\emph{\modelname{}-8} (ours)}} & 1: 1 & - & - & - & \textbf{0.2051} & - \\
& 2: 5 & - & - & - & 0.2046 & - \\
& 3: 21 & - & - & - & 0.2012 & - \\
& 4: 85 & - & - & - & 0.1896 & - \\
& 5: 341 & 43.42 & 49.78 & 0.32 & 0.1709 & - \\
& 6: 597 & 8.93 & 116.12 & 0.18 & 0.1667 & - \\
& 7: 853 & 4.78 & 135.01 & 0.13 & 0.1647 & - \\
& 8: 1109 & \textbf{3.89} & \textbf{140.55} & \textbf{0.11}  & 0.1634 & - \\
\bottomrule
\end{tabular}
}%
\caption[Comparison of reconstruction and semantic relevance for image tokenization]{\textbf{Comparison of reconstruction quality and semantic relevance for image tokenization}. We compare \modelname{} and VQGAN baselines used in state-of-the-art image~\cite{chang2022maskgit,lezama2023discrete,chang2023muse} generation models with ablation studies on codebook, training objective, quantization method, and token structure at 128$\times$128 resolution. 
Classification accuracy is evaluated under the mini-ImageNet 5-way 1-shot setup (\cf \cref{sec:eval}).
%comparison between each \modelname{} layer and VQGANs in MaskGIT~\cite{chang2022maskgit} and MAGVIT~\cite{yu2022magvit}.
}
\label{tab:vq_ablation_full}
% \vspace{-4mm}
\end{table}

\begin{table}[tp]
% \vspace{-mm}

\centering
% \resizebox{\textwidth}{!}{
\begin{tabular}{@{}lllccc@{}}
\toprule
Method & Dataset & \makecell{\# Layers\\: \# Tokens} & FID$\downarrow$ & IS$\uparrow$ & LPIPS$\downarrow$ \\ \midrule
    VQGAN (LDM~\cite{rombach2022high}) & OpenImages & 1: 256 & 5.15 & 144.55 & - \\
\emph{\modelname{}} (ours) & ImageNet-21k & 6: 597 & 3.08 & 173.79 & 0.19   \\
% MaskGIT~\cite{chang2022maskgit} 2D-VQ  & 1: 256 & 5.48 & 119.69 & 0.13  & -      & -        \\ 
% MAGVIT~\cite{yu2022magvit} 3D-VQ & 1: 1024 & 3.79 & - & - & - & - \\ \midrule
\bottomrule
\end{tabular}
% }
% \vspace{-4mm}
\caption[Comparison of reconstruction quality with scalability]{\textbf{Comparison of reconstruction quality with scalability} between \modelname{} and VQGAN baselines on 256$\times$256 images. 
%comparison between each \modelname{} layer and VQGANs in MaskGIT~\cite{chang2022maskgit} and MAGVIT~\cite{yu2022magvit}.
}
\label{tab:recons_quality_256}
\end{table}

\begin{table}[tp]
% \vspace{-mm}

\centering
% \resizebox{\textwidth}{!}{
\begin{tabular}{@{}llccc@{}}
\toprule
Method &  \makecell{\# Layers\\: \# Tokens} & FVD$\downarrow$ & CLIP$\uparrow$ & \makecell{Relative\\CLIP$\uparrow$} \\ \midrule
 VQGAN & 1: 1024 & 6.79 & n/a & n/a \\ \midrule
\multirow{6}{*}{\emph{\modelname{}} (ours)} & 1: 1 & - & \textbf{0.2061} & \textbf{0.8425} \\
& 2: 5 & - & 0.2056 & 0.8402   \\
& 3: 21 & - & 0.2032 & 0.8286 \\
& 4: 149 & - & 0.1896 & 0.7620  \\
%& 5: 341 & 9.49 & 109.46 & 0.17 & 0.1604 & 0.5914 & 5: 1173 & - & 0.1670 & 0.6531\\
 & 5: 1173 & 52.28 & 0.1670 & 0.6531\\
 & 6: 2197 & \textbf{6.35} & 0.1635 & 0.6367\\
% MaskGIT~\cite{chang2022maskgit} 2D-VQ  & 1: 256 & 5.48 & 119.69 & 0.13  & -      & -        \\ 
% MAGVIT~\cite{yu2022magvit} 3D-VQ & 1: 1024 & 3.79 & - & - & - & - \\ \midrule
\bottomrule
\end{tabular}
% }
% \vspace{-4mm}
\caption[Comparison of reconstruction and semantic relevance for video tokenization]{\textbf{Comparison of reconstruction quality and semantic relevance for video tokenization} between \modelname{} and the VQGAN baselines used in state-of-the-art video~\cite{yu2022magvit} generation models. 
%comparison between each \modelname{} layer and VQGANs in MaskGIT~\cite{chang2022maskgit} and MAGVIT~\cite{yu2022magvit}.
}
\label{tab:token_quality}
\end{table}

\begin{figure*}[tp]
\centering
% \vspace{-4mm}
\begin{subfigure}{0.49\linewidth}
\includegraphics[width=\linewidth]{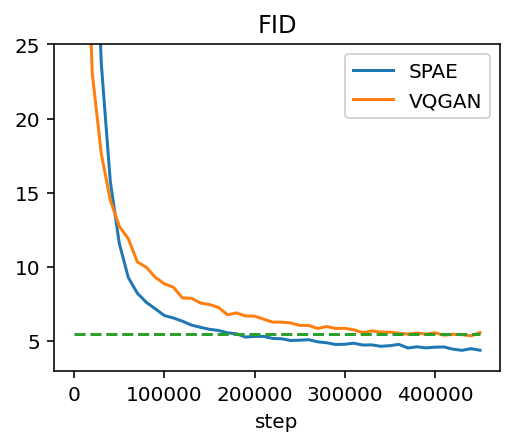}
\end{subfigure}
\begin{subfigure}{0.49\linewidth}
\includegraphics[width=\linewidth]{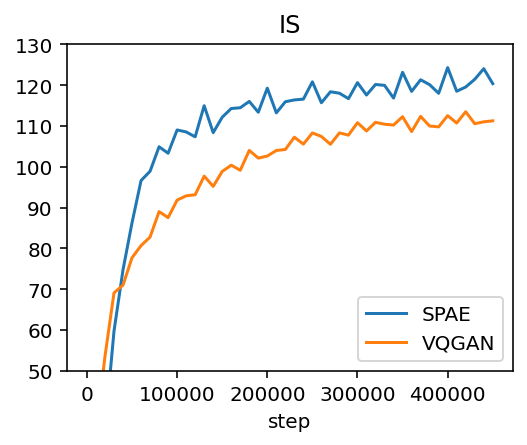}
\end{subfigure}
\begin{subfigure}{0.49\linewidth}
\includegraphics[width=\linewidth]{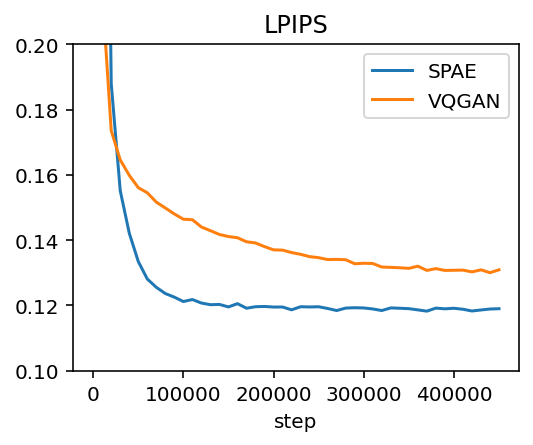}
\end{subfigure}
\begin{subfigure}{0.49\linewidth}
\includegraphics[width=\linewidth]{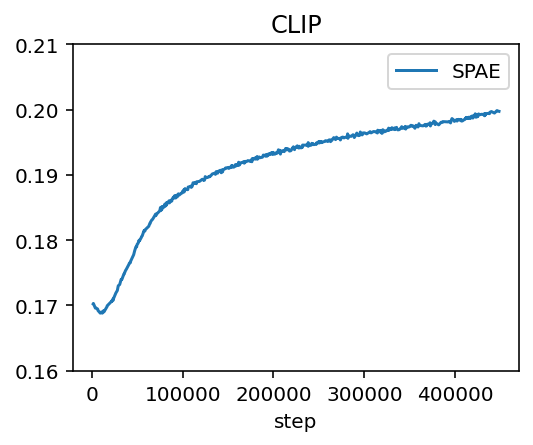}
\end{subfigure}
% \vspace{-4mm}
\caption[Training curves of SPAE in comparison to VQGAN]{\textbf{Training curves of SPAE in comparison to VQGAN}.}
\label{fig:train}
\end{figure*}

\paragraph{Comparison to VQGAN.}
The results are presented in \cref{tab:vq_ablation_full} for image models and in \cref{tab:token_quality} for video models. 
% \lu{ref the reconstruction figure.} 
Unlike VQGAN tokens, which lack specific semantic meaning, \modelname{} tokens demonstrate high semantic CLIP scores, more evident in the lower layers. As the number of layers increases, more tokens are utilized, resulting in improved reconstruction quality. This flexibility allows for dynamic adjustment of the token length to accommodate various tasks, such as using fewer tokens for understanding tasks.
While \modelname{} may have more lossy reconstruction compared to VQGAN when using a similar number of tokens, this is compensated by going into deeper layers, as shown in the layer 6 of \modelname{} in \cref{tab:vq_ablation_full}.
In \cref{fig:train}, we show the training curves of FID, IS, LPIPS, and CLIP score of SPAE and VQGAN. As shown, within 40\% of the training steps, SPAE shows better FID than the final VQGAN checkpoint. The CLIP score keeps improving as the training proceeds, while the LPIPS saturates quite early.
In \cref{tab:recons_quality_256}, we showcase the scalability of our model by training on the ImageNet-21k dataset with 13M images and list the comparable variant from LDM~\cite{rombach2022high} as a reference.

\begin{figure}[tp]
\centering
\includegraphics[width=\columnwidth]{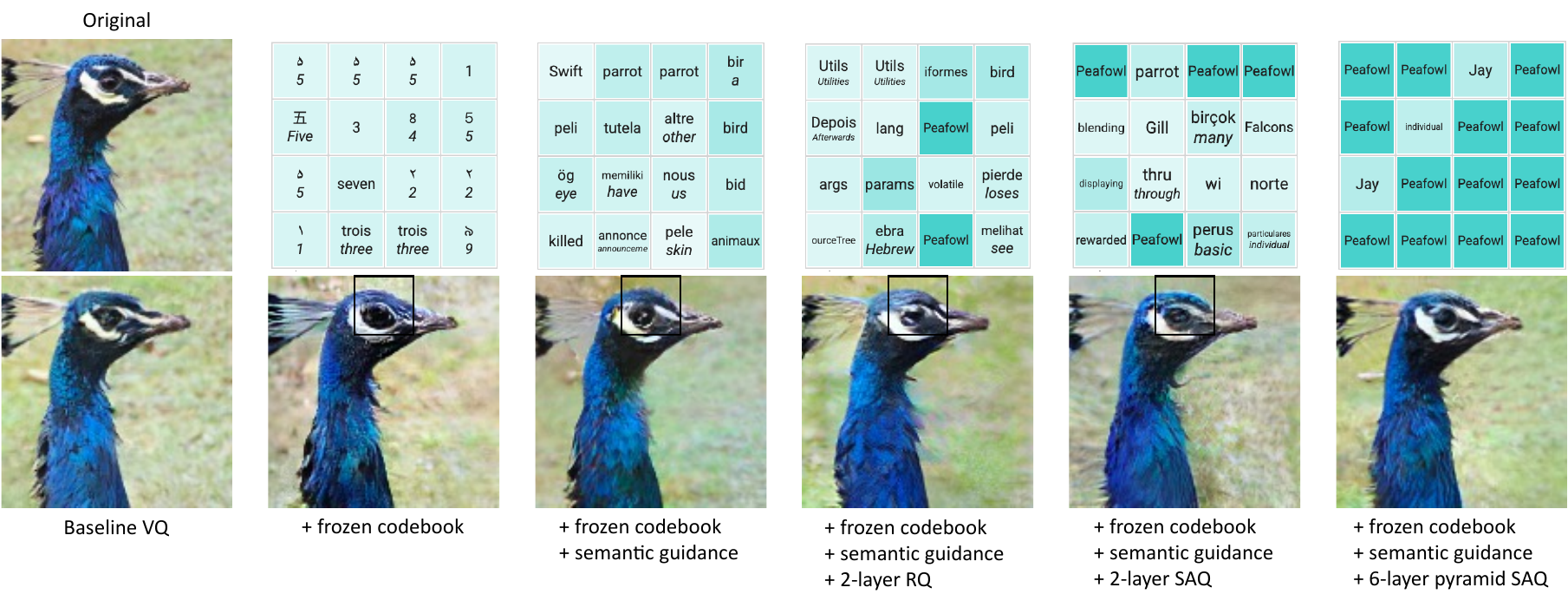}
\caption[Ablation examples with reconstructed image and semantic tokens]{\textbf{Ablation examples with reconstructed image and semantic tokens} for models listed in \cref{tab:vq_ablation_full}. For non-pyramid tokens, we show a 4$\times$4 crop from the first layer corresponding to the region indicated by the black box. For pyramid tokens, we use the third layer which consists of 4$\times$4 tokens. We use darker cells to show tokens with higher CLIP similarity to the original image. For non-English sub-word tokens, we show automatic translation for reference in italic fonts below the original token.}
\label{fig:ablation}
% \vspace{-1mm}
\end{figure}

\paragraph{Ablation studies.}

% The results in the top half of \cref{tab:vq_ablation_full} and \cref{fig:ablation} verify the effectiveness of the proposed designs in \modelname{}, as evaluated by reconstruction quality (FID, IS, LPIPS) and semantic similarity (CLIP and Relative CLIP).
% We have the following findings.
% %
% First, simply using a frozen codebook negatively affects the reconstruction results, but with semantic guidance it performs comparably with the original VQGAN while producing meaningful lexical words.
% %
% Second, RQ hurts reconstruction quality with a frozen codebook, which is different from the standard setup~\cite{lee2022autoregressive} where the codebook is learned.
% %
% Third, SAQ improves both quality and semantic similarity, where the pyramid adds flexibility for dynamic adjustment of the token length that can balance high-level semantics and low-level appearance details.

The results in \cref{tab:vq_ablation_full} and \cref{fig:ablation} verify the effectiveness of the proposed designs in \modelname{}, as evaluated by reconstruction quality (FID, IS, LPIPS) and semantic relevance (CLIP score, few-shot classification accuracy).
We have the following findings.
First, simply using a frozen codebook negatively affects the reconstruction results, but with semantic guidance it performs comparably with the original VQGAN while producing meaningful lexical words.
Second, RQ hurts reconstruction quality with a frozen codebook. This is different from RQ's standard setup~\cite{lee2022autoregressive} where the codebook is learned.
Third, SAQ improves both quality and semantic similarity, where the pyramid enables representation with much fewer tokens.
% adds flexibility for dynamic adjustment of the token length that facilitates representation with much fewer tokens.
This allows for accommodating more examples within the fixed and constrained in-context length.
% This enables superior understanding performance and the vital capability to accommodation of more examples for in-context learning.
% This enables SPAE to use significantly fewer tokens to represent an image for the understanding task while achieving superior performance.
%
Finally, per-layer semantic thresholds benefit understanding and the dynamic semantic loss weight helps reconstruction.
The perceptual loss leverages a trained network with access to classification labels, but removing it results in a surprising improvement in classification accuracy while greatly hurting the reconstruction.

\begin{figure}[tp]
% \vspace{-5mm}
\centering
\includegraphics[width=\columnwidth,trim={0 0 0 0},clip]{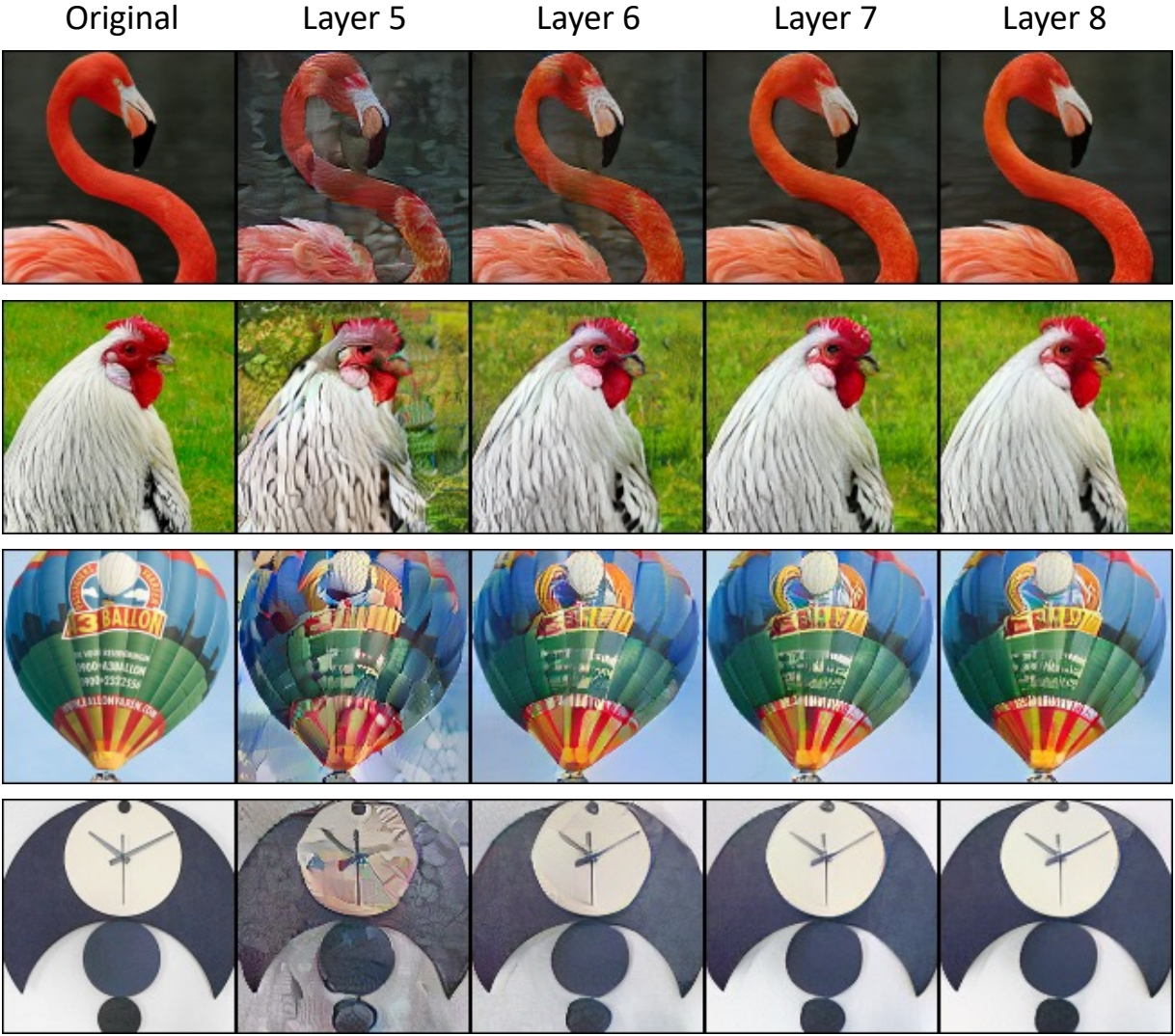}
\caption[Examples of coarse-to-fine image reconstruction]{\textbf{Examples of coarse-to-fine image reconstruction} by \modelname{}-8. The top 5 layers reconstruct a noisy image. The appearance details gradually get refined as more token layers are aggregated by the streaming average quantization process.}
\label{fig:reconstruction}
% \vspace{-4mm}
\end{figure}

\paragraph{Token quality with more \modelname{} layers.}
The bottom section of \cref{tab:vq_ablation_full} shows the per-layer reconstruction quality and semantic relevance of tokens from the \modelname{}-8 model in comparison to the default model.
With more token layers, the model gains larger capacity for both semantic and appearance, where the appearance gets pushed into deeper layers.
At layer 1 to 6, \modelname{}-8 yields consistently higher CLIP scores than \modelname{}.
At the last three layers, \modelname{}-8 also has better reconstruction quality than the last two layers of \modelname{}.
These results suggest the potential of better reconstruction quality and semantic relevance from using more token layers.
\cref{fig:reconstruction} shows the coarse-to-fine image reconstruction by the last four layers of \modelname{}-8.

% \vspace{-4mm}

\begin{figure}[tp]
    % \vspace{-3mm}
\centering
\includegraphics[width=\columnwidth,trim={0 0 0 0},clip]{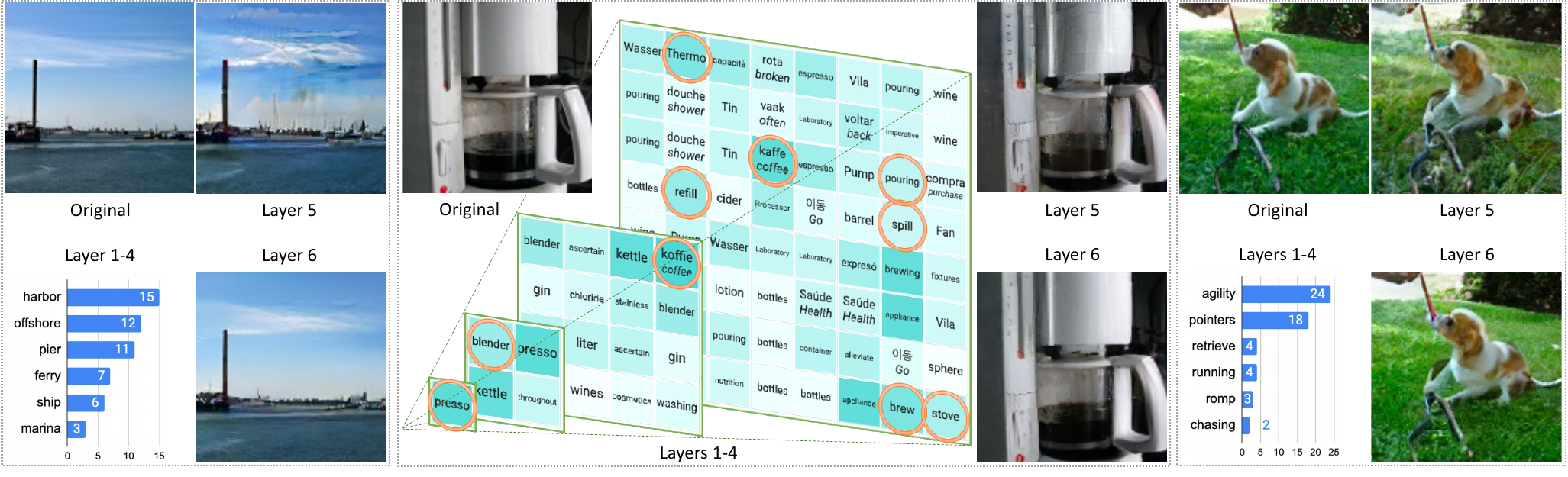}
% \vspace{-4mm}
\caption[Examples of pyramid image tokenization and reconstruction]{\textbf{Examples of pyramid image tokenization and reconstruction} by a 6-layer \modelname{}. 
We show the raw pyramid or histogram of most frequent tokens for the first four layers, and reconstructed images from layer 5 and 6.
In the pyramid, we use darker cells to show tokens with higher CLIP similarity to the original image. For non-English sub-word tokens, we show automatic translation for reference in italic fonts below the original token. Circled tokens are mentioned in \cref{sec:eval}. 
% See full pyramid visualizations in the Appendix.%\lu{change all Lx to Layer x.}
% \lijun{To update the bottom part with histogram format, and potentially more samples.}
}
%\lu{need to redraw this}
\label{fig:pyramid_ex}
% \vspace{-4mm}
\end{figure}

\subsection{Qualitative Evaluations.}
\begin{figure}[p]
    % \vspace{-5mm}
\centering
\begin{subfigure}{0.89\textwidth}
\includegraphics[width=\columnwidth,trim={0 0 0 0},clip]{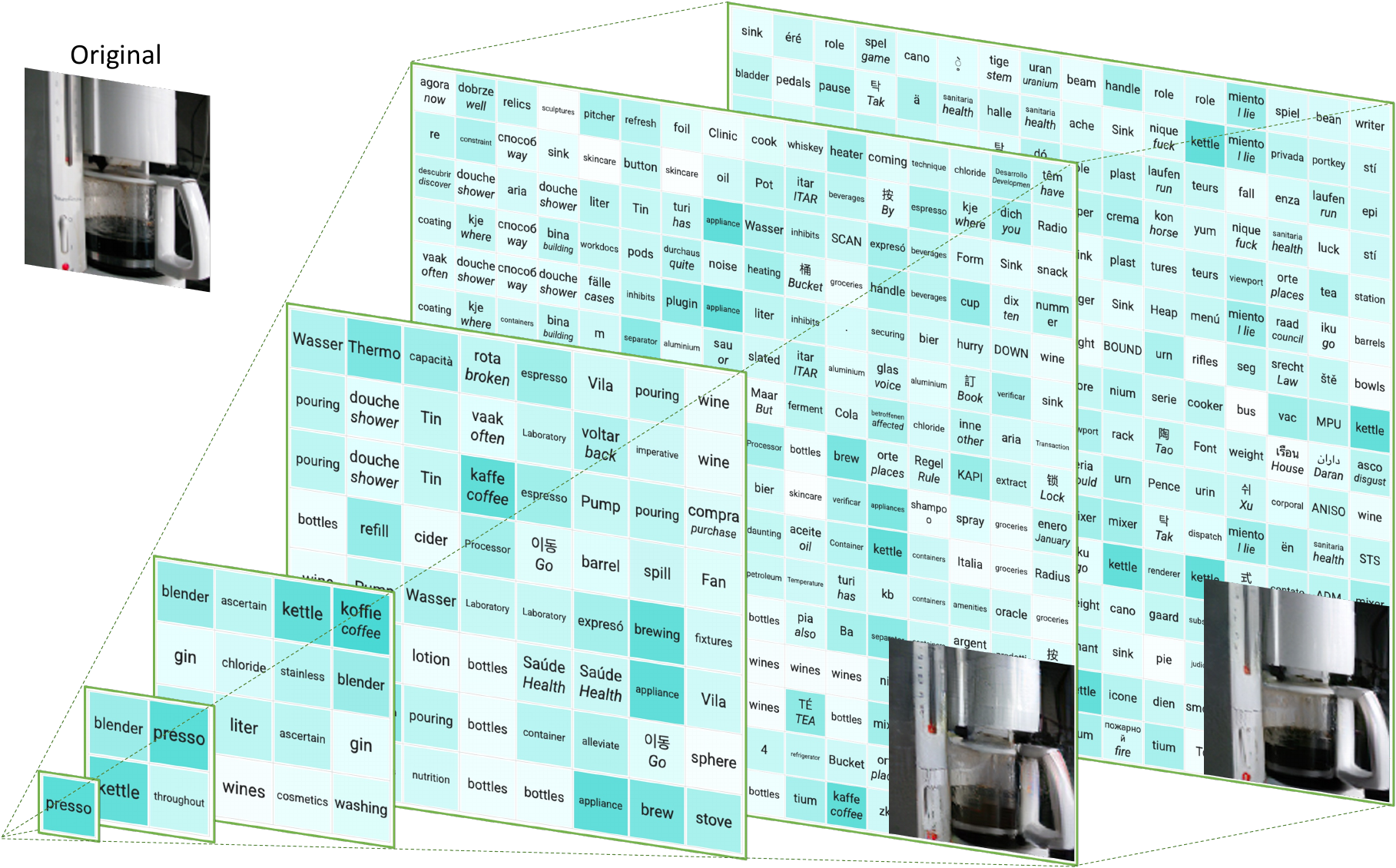}
\caption{Many keywords are present to describe various aspects from the stove to the thermometer.}
\end{subfigure}
\begin{subfigure}{0.89\textwidth}
\includegraphics[width=\columnwidth,trim={0 0 0 0},clip]{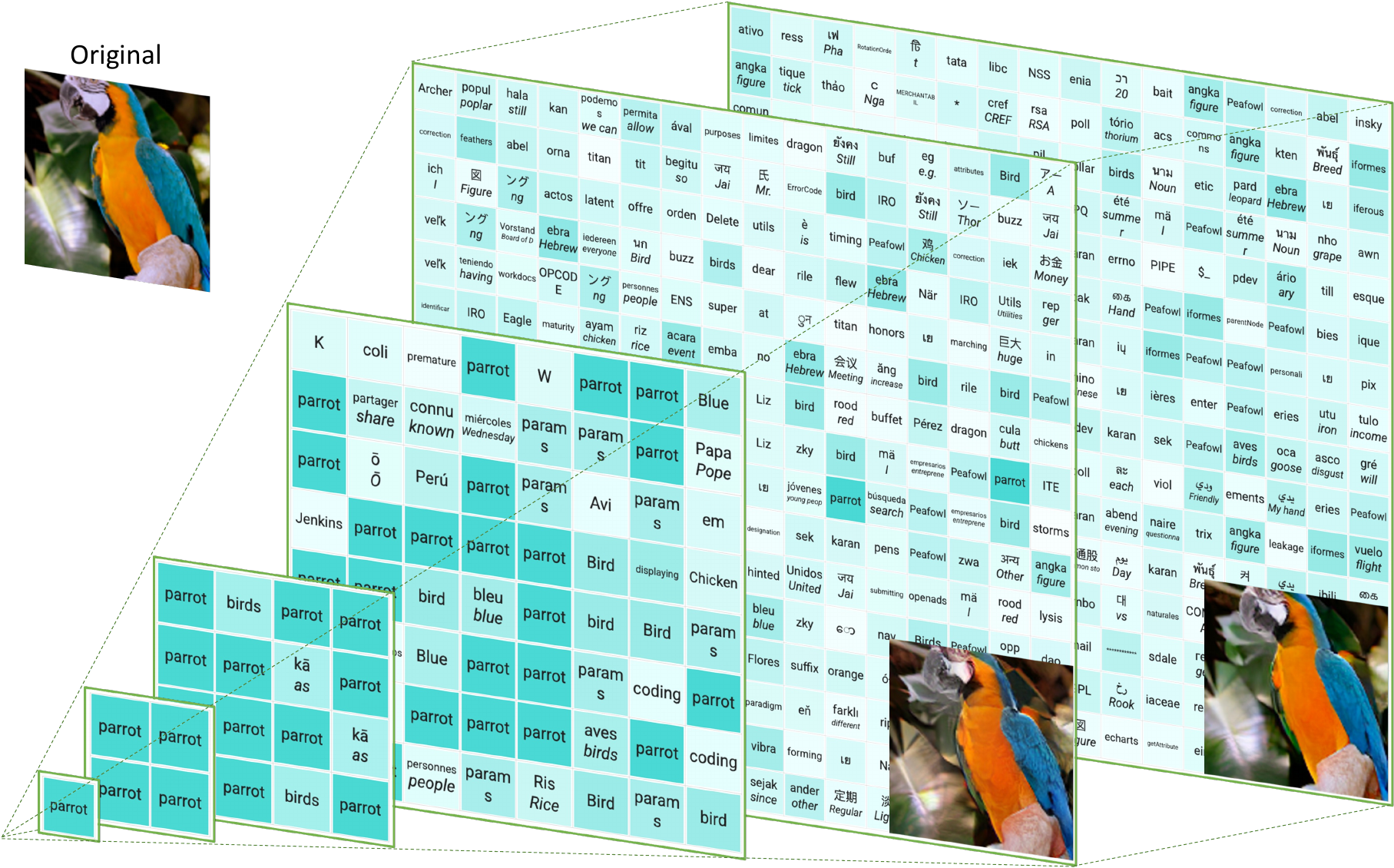}
\caption{A single unified concept is repeatedly highlighted.}
\end{subfigure}
% \vspace{-4mm}
\caption[Examples of pyramid image tokenization and reconstruction]{\textbf{Examples of pyramid image tokenization and reconstruction} by a 6-layer \modelname{}.
For visualization purposes only, we use darker cells to show tokens with higher CLIP scores regarding the original image. For non-English sub-word tokens, we show automatic translation for reference in italic fonts below the original token. We show tokens in all six layers, along with reconstructed images from the last two layers.}
\label{fig:pyramid}
% \vspace{-4mm}
\end{figure}

We visualize the tokens produced by \modelname{} in \cref{fig:pyramid_ex,fig:pyramid}, where 
we show the raw pyramid or histogram of tokens with top frequencies for the first four layers, with reconstructed images from layer 5 and 6. We have the following findings.

First, the \modelname{} tokens  are organized in a pyramid structure, with every layer comprising semantically related tokens to the image. The few tokens in the top layers seem to capture the primary theme of the image. For instance, in \cref{fig:pyramid_ex}, the token \texttt{presso} (highlighted in orange) represents the espresso machine and other tokens like \texttt{blender} refer to related regions. Layer 3 and Layer 4 reveal additional details about localized objects. For example, the token \texttt{Thermo} refers to the thermometer in the top-left region, while \texttt{stove} appears in the bottom-right area. In addition to nouns, related verbs also show up, including \texttt{pouring, refill, spill}, and \texttt{brew}.

Second, it is worth noting that the CLIP model has an English-only vocabulary.
However, thanks to the multilingual vocabularies and embeddings from the LLM, \modelname{}'s semantic guidance is able to map to similar concepts in other languages, such as \texttt{koffie} in Dutch and \texttt{kaffe} in Danish as corresponding terms to the concept of coffee.

Third, similar to RQ tokens~\cite{lee2022autoregressive}, \modelname{} tokens can reconstruct the image with progressively refined details when more layers, and thus tokens, are utilized. \cref{fig:pyramid_ex} shows Layer 5 begins to produce a reasonable reconstruction while Layer 6 further enhances the level of detail and smoothness.

\cref{fig:pyramid} shows full tokenization and reconstruction samples by a 6-layer \modelname{} from ImageNet validation set.
Key concepts are captured in the first few layers, whereas the later layers focus on the visual appearance.
In the coffee machine example, many keywords are present to describe various aspects from the stove to the thermometer.
In the parrot case, a single unified concept is repeatedly highlighted.

% In addition, \cref{fig:ablation} visualizes the tokenization for the same image with the ablation models used in \cref{tab:vq_ablation_full}.
% \cref{fig:reconstruction} shows the coarse-to-fine image reconstruction by the last four layers of \modelname{}-8.

\section{Summary}
In this chapter, we introduce the Semantic Pyramid AutoEncoder (SPAE) to map an image, or some other non-linguistic modality, to the token space of a frozen LLM.
We use \modelname{} to produce a multi-scale \emph{multi-modal} lexical \emph{latent representation} for images and videos.
The resulting tokens capture both the semantic meaning and the fine-grained details needed for visual reconstruction, effectively translating the visual content into a language comprehensible to the LLM.
In \cref{chap:llm}, we will use \modelname{} to empower LLMs to perform a wide array of multimodal tasks.

\glsresetall
\cleardoublepage
\chapter{Scalable Visual Token Representation} \label{chap:magvitv2}

\graphicspath{{papers/magvitv2/}}

\newcommand{\quantizername}{LFQ}
\renewcommand{\modelname}{MAGVIT-v2}
\newcommand{\webpage}{{\small \url{https://magvit.cs.cmu.edu/v2}}}
\newcommand{\cl}[1]{{#1}}

\paragraph{Overview.}
While \Glspl{LLM} are the dominant models for generative tasks in language, they do not perform as well as diffusion models on image and video generation.
% their performance for image and video generation is not as good as falls behind that of diffusion models.
To effectively use \Glspl{LLM} for visual generation, one crucial component is the visual tokenizer that maps pixel-space inputs to discrete tokens appropriate for \Gls{LLM} learning.
% One crucial component for using LLMs effectively in visual generation is obtaining a good visual tokenizer, mapping pixel-space inputs to discrete visual tokens, such that the resulting tokens are appropriate for LLM modeling. 
In this chapter, we introduce \acrshort{MAGVIT}-v2, a video tokenizer designed to generate concise and expressive tokens as \emph{multi-modal latent representation} for both videos and images using a common token vocabulary.
% Our tokenizer utilizes a novel quantization approach and supports tokenization of both images and videos using a shared vocabulary
%that eliminates the need for codebook lookup, generating binary codes directly. Furthermore, architectural modifications lead to not only quality enhancements but also tokenization of both images and videos using a shared vocabulary.
Equipped with this new tokenizer, we show that \Glspl{LLM} outperform diffusion models on standard image and video generation benchmarks including ImageNet and Kinetics. 
In addition, we demonstrate that our tokenizer surpasses the previously top-performing video tokenizer on two more tasks: (1) video compression comparable to the next-generation video codec (\acrshort{VVC}) according to human evaluations, and (2) learning effective representations for action recognition tasks.
% serving as an effective pre-training target for video action recognition.
%In addition to generation qiality, we show two potential brought by the proposed tokens: it produces favoarble compression than the
% The proposed tokenizer significantly outperforms previous video tokenizers and constitutes the first neural model to outperform the next-generation video codec H.266 in human evaluation. Furthermore, we demonstrate that the obtained representations are effective for self-supervised pre-training for video classification.

\clearpage

% \section{Proposed Work}
% \input{overviews/lq_proposal.tex}

\section{Motivation} \label{sec:magvitv2_intro}

% Generative AI is popular.
% There exist several different models for content generation: 

%The language model, commonly referred to as LM or LLM, is the de facto model for natural language modeling
Large transformer-based language models, commonly referred to as LMs or LLMs, are the de facto models for natural language generation~\cite{openai2023gpt,googlepalm2,touvron2023llama}. Over time, LMs have expanded their capabilities to generate content in various modalities, asserting their dominance in other domains like audio~\cite{agostinelli2023musiclm}, speech~\cite{rubenstein2023audiopalm}, code generation~\cite{li2023starcoder}, medical applications~\cite{singhal2023towards} and robotics~\cite{brohan2023rt}.

% LMs have established its leading role in various domains beyond natural language such as .
%\lu{add more audio generation chapter}.
% The modality (\eg, audio or image) is mapped into discrete tokens, which are treated as language words within the transformer architecture.

LMs are capable of generating images and videos. To do so, the image pixels are mapped into a sequence of discrete tokens by a visual tokenizer (\cf \cref{sec:magvitv2_background}). These tokens are then fed into the LM transformer, as if they were lexical words, for generative modeling. Despite notable advancements in employing LMs for visual generation~\cite{esser2021taming,chang2022maskgit}, LMs still do not perform as well as diffusion models~\cite{rombach2022high}.
% operate on sequences of visual tokens obtained by some form of vector-quantized autoencoder (\cref{sec:background})
% In the case of visual (image or video) generation, the best performing language models operate on sequences of visual tokens obtained by some form of vector-quantized autoencoder (\cref{sec:background}). Although there has been significant progress in using LMs under this setting \cite{esser2021taming,chang2022maskgit,yu2022magvit}, LMs still trail behind diffusion models~\cite{rombach2022high}. 
For instance, when evaluating on the ImageNet dataset, a gold standard benchmark for image generation, the best language model~\cite{lee2022draft} underperforms 
% in comparison to
the diffusion model~\cite{gao2023masked} by a substantial 48\% margin (FID 3.41 \vs 1.79 when generating images at the 256$\times$256 resolution).

\emph{Why do language models lag behind diffusion models in visual generation?} This chapter suggests that a primary reason is the lack of a good visual representation, resembling our natural language system, for effectively modeling the visual world.
% Unlike natural language, humans have not yet developed an optimal vocabulary for our visual world. The challenges of learning such visual vocabulary significantly limit the generative capabilities of the LLM, which otherwise excel in language, graph, and various other domains.
To substantiate this hypothesis, this chapter shows that, when utilizing a good visual tokenizer, the masked language model~\cite{devlin2019bert,chang2022maskgit,yu2022magvit} surpasses the state-of-the-art diffusion models  in terms of both generation fidelity and efficiency across image and video benchmarks, given the same training data, comparable model size, and training budget. To the best of our knowledge, this provides the first evidence that language models beat diffusion models on the hallmark ImageNet benchmark. 

% To the best of our knowledge, our results provide the first evidence suggesting that the language model can outperform the diffusion model on ImageNet given the same training data, comparable model size, and training budget. This progress is attributed to the use of a good visual tokenizer. 

It is worth emphasizing that our intention is not to assert whether the language model is superior to others, but to promote the exploration of visual tokenization methods for LLMs.
A fundamental difference of LLMs from other models, such as diffusion models, is that LLMs utilize a discrete latent format: tokens obtained from a visual tokenizer.
%MH: argue -> show (argue is a bit aggressive and confrontional)
We show that the values of these discrete visual tokens should not be overlooked considering their distinct advantages as follows. 

\begin{enumerate}
    \item \textbf{Compatibility with LLMs.} The main advantage of a token representation is that it shares the same form as language tokens, making it straightforward to leverage the optimizations our community has developed over many years for LLMs. This includes faster training and inference speeds~\cite{shazeer2019fast,lester2021power}, advancements in model infrastructure~\cite{dao2022flashattention,du2022glam}, learning recipes for model scaling~\cite{brown2020language,chowdhery2022palm}, and GPU/TPU optimization, among other innovations. Unifying vision and language by the same token space could set the stage for a true multimodal LLM that can understand, generate, and reason within our visual environment.
%as illustrated by AudioPaLM~\cite{rubenstein2023audiopalm} in the audio domain. 
% the biggest benefit of token representation resembeles of lanuage, rendering it straightforward to leverage all the optimization methods, the community has been deleopved for years, for the LLMs include training and inference speed~\cite{shazeer2019fast,ainslie2023gqa}, , even hardware GPU/TPU among others.
%\lu{yong} Could you please provide some references for well-known efforts to enhance the training and inference. Works have shown joint learning of understanding and generation suggest feasible~\cite{rubenstein2023audiopalm,yu2023scaling}.
%MQA and GQA~\cite{shazeer2019fast,ainslie2023gqa} introduce architectural modifications aimed at boosting the training and inference speed of the Transformer model. Some other initiatives focus on refining fine-tuning efficiency by incorporating slight parameter adjustments to a single model to cater to various scenarios~\cite{hu2021lora,lester2021power}.
% efficiency of large language models (LLMs), including optimizations for GPU/TPU hardware utilization in LLMs? [here we do not consider acceleration technique such as distillation]
\item \textbf{Compressed representation.} The discrete token may offer a fresh perspective on video compression. The 
visual tokens can serve as a new video compression format to reduce disk storage and bandwidth during internet transfers. Unlike compressed RGB pixels, these tokens can be fed directly into generative models, bypassing the conventional decompression and latent encoding steps. This allows for faster processing in generative video applications, especially beneficial in edge computing cases. 
\item \textbf{Visual understanding benefits}. 
Prior research has shown that the discrete tokens are valuable as a pre-training target in self-supervised representation learning, as discussed in BEiT~\cite{bao2021beit} and BEVT~\cite{wang2022bevt}. 
Additionally, research finds that using tokens as the model inputs improves the robustness and generalization~\cite{mao2021discrete}.

\end{enumerate}

% However, the quality of transformer (language model and masked token model) is lagging behind denoising diffusion on image and video generation. For example, on the standard ImageNet benchmark, the best diffusion gets xx\% better FID=?? versus FID=6.8. On the other hand, LLM has been a de-facto model in the field of language modeling, and has established its leading role audio generation~\cite{}. What's missing? 

In this chapter, we introduce \modelname{}, a video tokenizer designed to map videos (and images) into compact discrete tokens. Our model is built on the state-of-the-art video tokenizer, MAGVIT~\cite{yu2022magvit}, within the VQ-VAE framework~\cite{van2017neural}. We propose two new techniques. First, a novel lookup-free quantization method enables the learning of a large vocabulary that is able to improve generation quality of the language model.
Second, through extensive empirical analyses, we have identified modifications to the tokenizer that not only enhance generation quality but also enable the tokenization of both images and videos using a shared vocabulary.
%.
%enables the learning of a large vocabulary that is able to
% his new method eliminates conventional need for codebook lookup and produces binary codes directly. 
%By virtual of this design, the generation quality of the language model (LM) significantly enhances over a large vocabulary, a property not seen in existing VQ-VAE approaches. 

% Our primary focus is on video tokenization, which challenges in spatial-temporal compression.
% that can effectively map video (and image) into compact tokens. We focus on the video tokenization which is challenging to the spatial-temporal compression. 
%\lu{explains the new designs}; \lu{two novel designs tokenizer}. 

% enables the masked language model to yield higher fidelity compared to the state-of-the-art diffusion model on the ImageNet benchmark, using a similar model size and training budget. It also significantly outperforms the diffusion model as well as the prior best tokenizer on the standard video benchmark. 

We empirically demonstrate that our model outperforms the previously top-performing video tokenizer, MAGVIT, in three key areas. First, our model significantly improves the generation quality of MAGVIT, establishing the state of the art on the common image and video benchmarks. Second, user studies indicate that its compression quality exceeds that of MAGVIT and the current video compression standard, HEVC~\cite{sullivan2012overview}. Moreover, it is on par with the next-generation video codec, VVC~\cite{vvc}.
Finally, we show that, compared to MAGVIT, our new tokens are stronger for video understanding tasks across two setups and three datasets.
%Kinetics~\cite{carreira2018short} and Something-Something V2~\cite{goyal2017something}. 
The main contributions of this chapter are:
%\vspace{-2mm}
\begin{itemize}%[nosep, leftmargin=*]
\item A new video tokenizer that outperforms the previously best-performing video tokenizer in three areas: visual generation, video compression, and action recognition.
\item A novel lookup-free quantization approach that enables improving the visual generation quality of language models by learning a large vocabulary.
\item To the best of our knowledge, the first evidence suggesting that a language model can outperform diffusion models on ImageNet when provided with the same training data, an equivalent model size, and a similar training budget.
\item A video compressor with better quality than HEVC and VVC, at similar bit rates, according to user studies. To our knowledge, this is the first successful attempt of a visual tokenizer designed for video generation to achieve comparable results to standard codecs.
% Generation: we achieve state-of-the-art video and image generation with token-based transformers that can seamlessly utilize existing LLM infrastructure. This also marks the first time where token-based transformers perform favorably against diffusion models on the important ImageNet benchmarks.
% \item Compression: we accomplish high-fidelity video reconstruction better than prior best video tokenizer ~\cite{yu2022magvit}. In addition, we present the first neural model that outperforms next-generation video codec H.266 in human evaluation.
% \item Understanding: we demonstrate the effectiveness of using tokens for pretraining which benefits video classification performance.
\end{itemize}

\section{Background} \label{sec:magvitv2_background}
%\vspace{-4mm}
\paragraph{Language Model (LM) for visual generation.}
LMs have been extended to generate images and videos. A visual tokenizer $f$ is used to first map visual inputs into a sequence of discrete tokens. A video  $\rmV \in \sR^{T \times H \times W \times 3}$ (or image when $T=1$) is tokenized into a discrete representation $\rmX = f(\rmV) \in \{1, 2, \cdots, K\}^{T' \times H' \times W'}$, where $K$ is the codebook (vocabulary) size of the visual tokenizer. $\rmX$ is flattened into a 1D token sequence obtained using raster scan ordering and then fed into an LM transformer for generative modeling. 

Two types of LMs are commonly used for visual generation. The \emph{Autoregressive LM (AR-LM)} includes ImageGPT~\cite{chen2020generative}, DALL-E~\cite{ramesh2021zero}, Parti~\cite{yu2022scaling}, \etc. 
An AR-LM predicts the next token given the previous tokens along with additional conditioning information $\rvc$ using a categorical distribution for $p_\theta(\ervx_i \mid \rvx_{<i}; \rvc)$. During inference, AR-LMs use the standard autoregressive decoding over the tokens. Finally, the tokens are converted back to pixels by a decoder associated with the visual tokenizer.

%A new sequence can be directly sampled from the model by $\hat{\rvx}_i \sim p_\theta(\rvx_i | \hat{\rvx}_0, \ldots, \hat{\rvx}_{i-1}, \rvc)$.

The \emph{Masked LM (MLM)} is another type of language model for visual generation, such as: MaskGIT~\cite{chang2022maskgit}, MAGVIT~\cite{yu2022magvit}, Phenaki~\cite{villegas2022phenaki}, and MUSE~\cite{chang2023muse}, among others. An MLM is trained using a masked token objective \cite{devlin2019bert}, where some tokens in the sequence are randomly masked and need to be predicted given the observed tokens. Let $\rvm \in \{0,1\}^n$ be a random binary sequence where $\rvm^\top \vone \in [0, n-1]$. The MLM learns $p_\theta(\ervx_i \mid \{\ervx_j: \ervm_j=1, \forall j\}; \rvc)$ for all $i$ where $\rvm_i=0$. To generate a video or image during inference, the MLM uses the non-autoregressive decoding algorithms
%\cite{ghazvininejad2019mask} with adaptations tailored 
for images and videos~\cite{chang2022maskgit,yu2022magvit}. 
The decoding starts with a fully masked sequence, which is iteratively filled by repeating two steps: (1) sample the whole sequence $\hat{\rvx}^{(t)}$ from $p_\theta$ given the non-masked tokens from the previous step, (2) re-mask the $\lfloor\lambda(t)\cdot n\rfloor$ tokens in $\hat{\rvx}^{(t)}$ with the lowest probability, following a decreasing masking ratio schedule $\lambda(t)$, according to timestamp $t$.
% For image algorithm details, refer to \cite{chang2022maskgit}; for video, see \cite{yu2022magvit}. 

%\vspace{-4mm}
\paragraph{Denoising Diffusion Models (DDM).}
DDMs \cite{sohl2015deep,song2019generative} are regarded as the state-of-the-art in visual generation due to their high-quality image~\cite{dhariwal2021diffusion,ho2022imagen,dhariwal2021diffusion,saharia2022photorealistic} and video generation~\cite{ho2022video,ho2022imagen,singer2022make}. For instance, DDPM~\cite{ho2020denoising} learns a denoising process parameterized as conditional Gaussian distributions over image pixels. 
Recently, diffusion models and language models have displayed a significant overlap, where DDMs diffuse over latents rather than raw pixels. These latents are obtained using models similar to the visual tokenizer used by LMs. In fact, the very first latent in diffusion, proposed by \cite{rombach2022high}, is derived from a visual tokenizer. 
Binary latents for image modeling are also used in \cite{wang2023binary}, where the diffusion process is parameterized with Bernoulli distributions.
Later studies have identified advantages in substituting the U-Net~\cite{ronneberger2015u} denoising backbone with a Transformer \cite{peebles2022scalable,jabri2023scalable} or a hybrid of both \cite{hoogeboom2023simple}, making the distinctions between diffusion and language models in visual generation more blurred.
% Additionally, the diffusion model's architecture has been shifting from the U-Net to the transformer architecture~\cite{peebles2022scalable}. Consequently, the boundaries between diffusion and language models in visual generation have become less distinct. 
%Specifically, both DDM and MLM emulate a gradual corruption process on the latents and generate samples through iterative denoising restoration.
Yet, a fundamental difference between DDMs and LMs lies in the latent format, \ie, continuous \vs discrete. We have discussed the benefits of having discrete tokens in \cref{sec:magvitv2_intro} and 
%: compatibility with language models, a compact representation, and a video pretraining target \lijun{todo}. We 
will show that the proposed tokenizer improves in these aspects.
\paragraph{Visual tokenization.}
% \label{sec:background_tokenizer}
Visual tokenization plays an essential role in mapping pixels into a discrete representation suitable for generative modeling. VQ-VAE \cite{van2017neural} is a cornerstone work in image tokenization. A VQ-VAE model consists of a convolutional neural network (CNN) encoder, a vector-quantization (VQ) bottleneck, and a CNN decoder.
Given a video $\rmV \in \sR^{T \times H \times W \times 3}$, the VQ-VAE's encoder $E$ produces latent embeddings $\rmZ = E(\rmV) \in \sR^{T'\times H'\times W'\times d}$. 
Each embedding vector $\rvz \in \sR^d$ in $\rmZ$ is then passed through the vector quantizer $q$, which assigns it to the closest entry $\rvc \in \sR^d$ in the learned codebook embedding $\rmC \in \sR^{K \times d}$:
% The individual embeddings $\rvx \in \sR^d$ in each latent position are vector-quantized by nearest neighbor search over the learned codebook $\rmC=\sR^{K \times d}$,
% \lijun{is $\rvx$ the index or the post-quantization embedding? based on the wording in sec 3, embedding seems more suitable. Then I think we do not need $\rvc_\rvx$ and can just use $\rvx$ in the loss}
\begin{equation}
%\vspace{-2mm}
    q(\rvz) = \rvc_i, \text{ where } i=\underset{j \in \{1,2,\cdots,K\}}{\argmin}\Vert \rvz - \rvc_j \Vert_2.
\end{equation}
To get discrete tokens, we drop the embedding dimension and represent $\rmZ$ by its indices $\rmX \in \{1, 2, \cdots, K\}^{T' \times H' \times W'}$. For decoding, embeddings of all image tokens are given as input to the decoder $D$ to reconstruct the input $\hat{\rmV} = D(\rmZ)$. 
%The standard version of the VQVAE is trained with the following loss optimizing jointly the encoder, decoder and codebook:
% \begin{equation}
% \mathcal{L}_{\text{vqvae}} = \| \rmV - \hat{\rmV} \|_2^2 + \| \text{sg}(\rmZ) - \rvc_\rmZ \|_2^2 + \beta \| \rmZ - \text{sg}(\rvc_\rmZ) \|_2^2 
% \end{equation}
% where $\beta$ is a hyperparameter balancing the codebook and encoder updates, and $\rvc_\rmZ$ denotes the concatenation of all quantized vectors. The straight-through gradient estimation is used, denoted by the stop gradient operation $\text{sg}$, to copy the gradients directly from the decoder to the encoder.
Following VQ-VAE, VQGAN~\cite{esser2021taming} introduces an adversarial loss and feature-level perceptual losses to enhance the image quality.

Video tokenization is more challenging since it requires modeling the visual dynamics within the compressed spatial-temporal latent space. 
Early studies on video tokenization treat frames as independent images with no temporal compression~\cite{wu2021n,gupta2022maskvit}. Later research \cite{yan2021videogpt,ge2022long,yu2022magvit} integrates 3D CNNs to tokenize spatial-temporal volumes. 
The state of the art in video tokenization is MAGVIT~\cite{yu2022magvit}, which introduces a better 3D architecture, an inflation technique for initialization using image pre-training, and robust training losses. With MAGVIT, the LMs achieve leading generation quality across multiple video benchmarks. However, MAGVIT struggles to tokenize images and often results in noticeable flickering in longer videos.
Beyond the CNN-based models discussed above, additional models have been proposed. ViT-VQGAN \cite{yu2021vector} introduces transformer blocks as a substitute for CNNs for image tokenization. C-ViViT \cite{villegas2022phenaki} further extends this idea for video tokenization. 
Despite these advances in vector quantization (VQ), the codebook learned by previous VQ models is relatively small (\eg, $8$k) due to the difficulty in improving the generation quality with larger vocabularies. In contrast, our tokenizer can induce a large vocabulary (\eg, $262$k) that can be effectively modeled by an LM, leading to enhanced image and video generation quality. 

% Video tokenization is more challenging and VQGAN has been adapted to meet this purpose~\cite{ge2022long,villegas2022phenaki,yu2022magvit}. The state of the art in video tokenization is MAGVIT~\cite{yu2022magvit}, which introduces a better 3D architecture, an inflation technique for initialization using image pre-training, and robust training losses. With MAGVIT, the LMs achieve leading generation quality across multiple video benchmarks. However, MAGVIT struggles to tokenize images and often results in noticeable flickering in longer videos.
%In the video domain, the encoder and decoder can be implemented using 3D CNNs 

% In this work we build upon the state-of-the-art video tokenizer MAGVIT from \cite{yu2022magvit}. MAGVIT introduces several innovations to video tokenization, obtaining high reconstruction quality and a discrete latent space that produces state-of-the-art video generation when coupled with a masked language model.
% Introduce the MAGVIT in more details.

% Introduce VQ-VAE and follow-up works.

\paragraph{Text-to-\{image, video\}.}
Text-to-image and text-to-video generation has garnered significant rapid advancements using both language models~\cite{yu2023scaling,chang2023muse} and diffusion models~\cite{ho2022imagen,blattmann2023align,singer2022make,ge2023preserve,ramesh2022hierarchical}. Although diffusion models, such as Midjourney, are considered the top performers in these tasks, it is unclear whether their advantage stems from the model, data, or some other unidentified factors. Indeed, it is challenging to scientifically compare these text-to-image models as they are trained on varied datasets, with some even being proprietary data, under inconsistent training conditions. To facilitate a fairer comparison, this chapter prioritizes using the ImageNet and Kinetics benchmarks.

\section{MAGVIT-v2 Video Tokenizer}

%
%The complexity of modeling spatial-temporal dynamics motivated our focus on video tokenization rather than on image tokenization. 

We introduce a new \emph{video tokenizer} designed to map the spatial-temporal dynamics from a visual scene into compact discrete tokens suitable for language models. 
\cl{Compared with image generation, video generation still faces substantial challenges in generating consistent and realistic motion. We are interested in exploring the capabilities of language models in tackling this unsolved challenge. Therefore, we focus on a video tokenizer that can effectively represent video for generative modeling.}
Our approach builds upon the state-of-the-art video tokenizer, MAGVIT, as detailed in \cite{yu2022magvit}. 
This section highlights two new designs: a lookup-free quantizer and a collection of enhancements to the tokenizer model.
%\vspace{-2mm}
\subsection{Lookup-Free Quantizer}
Although the community has made great progress in developing VQ-VAEs, the relationship between improvements in the reconstruction quality and subsequent generation quality is still not well understood.
A common misconception is that improving reconstruction equates to improving the generation of the language model.
% For example, enlarging the VQ vocabulary could result in better reconstruction in \cref{fig:curve}, but it does not translate to an improvement of generation in \cref{fig:curve_gen} in many cases~\cite{yu2022scaling}.
For example, enlarging the vocabulary can improve reconstruction quality. However, such improvement only extends to generation when the vocabulary size is small, and a very large vocabulary can actually hurt the performance of the language model. 

\begin{figure}[tp]%{r}{0.5\textwidth}
%\vspace{-4mm}
    \centering
    % \begin{subfigure}{0.43\linewidth}
    \includegraphics[width=0.7\linewidth,trim={0 0 0 0},clip]{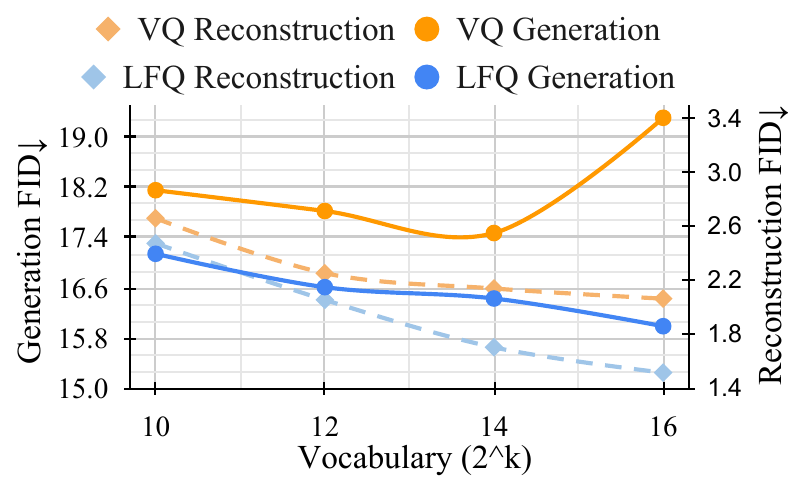}
    % \caption{Reconstruction quality.}
    % \label{fig:curve_recons}
    % \end{subfigure}
    % \begin{subfigure}{0.5\linewidth}
    % \includegraphics[width=\linewidth,trim={0 0 0 1cm},clip]{gen_curve.pdf}
    % \caption{Generation quality.}
    % \label{fig:curve_gen}
    % \end{subfigure}
    %\vspace{-6mm}
    \caption[Reconstruction and generation quality curves]{\textbf{Reconstruction and generation quality curves} in FID on ImageNet when scaling the tokenizer's vocabulary size with Vector Quantization (VQ) and Lookup-Free Quantization (LFQ). 
    Comparison is done at 128$\times$128 resolution using an MLM with 306-372M parameters.
    %  using a tokenizer with 52M and
    % \lu{Size of LM and tokenizer}
    % \lijun{todo}  \lu{change the color green to a warm color.}
    % The area of the dots indicate the parameter size for the tokenizers in (a) and language models in (b). VQ uses fixed 256 embedding dimensions, while VQ (lower embed) uses $2^{18-k}$ dimensions.
    }
    \label{fig:motivation}
    %\vspace{-2mm}
\end{figure}

As illustrated by the dashed curves in \cref{fig:motivation}, the reconstruction FID, indicated by the right $y$-axis (where a lower value is better), improves as the vocabulary size (the $x$-axis) increases. The orange solid curve in \cref{fig:motivation} represents the LM's generation quality (the left $y$-axis). The generation FID initially improves but deteriorates for larger vocabulary. This may shed light on why the vocabulary size of most language models for visual generation is around 1-8k~\cite{esser2021taming,villegas2022phenaki}, which is significantly smaller than the size of natural language vocabulary, \ie over 200k.

A simple trick for training a larger codebook involves decreasing the code embedding dimension when increasing the vocabulary size~\cite{yu2021vector}.
This trick captures the intuition of limiting the representational capacity of individual tokens, which in turn facilitates learning over the distribution of a large vocabulary.

% ~\cite{rakhimov2021latent,le2021ccvs}

%, \ie increasing $K$ while reducing  in the vocabulary $\rmC^{K\times d}$.

%\vspace{-4mm}
\paragraph{Lookup-Free Quantization (\quantizername{}).}

Motivated by the above observation, we reduce the VQ-VAE codebook's embedding dimension to zero. Formally, the codebook $\rmC \in \sR^{K \times d}$ is replaced with an integer set $\sC$ where $|\sC| = K$.
% reduced to an integer set $\sC = \{1, 2, \cdots, K\}$. 
Recall that in VQ-VAE models, the quantizer must look up all $K$ $d$-dimensional embeddings in the codebook, where $d$ is typically $256$, when computing the closest codebook entry to the encoder output. This new design eliminates the need for such embedding lookup entirely hence we call it \emph{lookup-free quantization (\quantizername{})}.
We found that \quantizername{} can grow the vocabulary size in a way benefiting the generation quality of language models. As shown by the blue curves in \cref{fig:motivation}, both reconstruction and generation consistently improves as the vocabulary size increases -- a property not observed in current VQ-VAE methods.

While various \quantizername{} methods are available, this chapter discusses a straightforward variant that assumes independent codebook dimensions and binary latents. 
Specifically, the latent space of \quantizername{} is decomposed as the Cartesian product of single-dimensional variables, as $\sC = \bigtimes_{i=1}^{\log_2{K}}C_i$.
%Lookup-free quantization can scale to larger vocabulary in a way that benefits the generation of language model. We set the vocabulary size to $2^{18}$ comparable to that of PaLM to leverage the instrcture and learning receipes built for LLMs.
Given a feature vector $\rvz \in \sR^{\log_2{K}}$, each dimension of the quantized representation $q(\rvz)$ is obtained from:
\begin{equation}
% %\vspace{-1mm}
    q(\ervz_i) = C_{i, j}, \text{ where } j = \argmin_k \|\ervz_i - C_{i, k}\|,
\end{equation}
%Depending on the values in $C_i$, we have various ways to compute the $\argmin$. However,
where $C_{i, j}$ is the $j$-th value in $C_i$.
With $C_i = \{-1, 1\}$, the $\argmin$ can be computed by the sign function as 
% We call it \emph{lookup-free binary quantization (LBQ)}:
\begin{equation}
% %\vspace{-1mm}
    q(\ervz_i) = \sign(\ervz_i) = -\1\{\ervz_i \le 0\} + \1\{\ervz_i > 0\}.
\end{equation}
% which can be easily done via hashing or rounding according to the values in $C_i$.
% where $\sC_i$ is the predefined ordered list of integers for the $i$-th dimension.
% % The effective size of the vitual codebook becomes $|\sC| = \prod_{i=1}^N |C_i|$.
% To get the token index in $\sC$, we have
With \quantizername{}, the token index for $q(\rvz)$ is given by:
\begin{equation}
%\vspace{-1mm}
\mathit{Index}(\rvz) = \sum_{i=1}^{\log_2{K}} \argmin_k \|\ervz_i - C_{i, k}\| \prod_{b=0}^{i-1} |C_b| = \sum_{i=1}^{\log_2{K}} 2^{i-1} \1\{\ervz_i > 0\}, \label{eq:tokenid}
\end{equation}
where $|C_0|=1$ sets the virtual basis.

% where $N=\log{K}$ and all $\sC_i = \{-1, 1\}$. 
% As a result, we have
% \begin{equation}
%     \mathit{Id}(\rvx) = \sum_{i=1}^N 2^{i-1} \1\{x_i > 0\} \label{eq:tokenid}
% \end{equation}
% The number of tokens  $N=\log_2{K}$ 
% In addition, $|\sC| = 2^N$ and $B=N$.
% The overall space and time complexity is just $O(B)$.

We add an entropy penalty during training to encourage codebook utilization:
% Some VQ models, \eg in \cite{chang2022maskgit}, involves an entropy penalty~\cite{jansen2020coincidence} during training to encourage the codebook utilization.
\begin{align}
    \gL_\mathit{entropy} &= \E[H(q(\rvz))] - H[\E(q(\rvz))].
    % = \E \big[\sum_{i=1}^{\log_2{K}} H(\ervq (\ervx_i))\big] - H[\E(\rvq(\rvx))]
    % H(\rvq(\rvx)) &= \sum_{i=1}^{\log_2{K}} H(\ervq_i (\ervx_i)) 
    \label{eq:entropy}
\end{align}
This penalty is inspired by a similar loss used in image VQGAN model~\cite{chang2022maskgit}, which is also found in entropy-based clustering~\cite{jansen2020coincidence}. In \quantizername{}, given the independence among dimensions, we rewrite $H(q(\rvz)) = \sum_{i=1}^{\log_2{K}} H(q (\ervz_i))$ .
The $H[\E(q(\rvz))]$ term can be approximated with sub-groups of dimensions for $K>2^{18}$ where direct estimation is memory bound. 

% It prevents unhealthy initial states of the model where some bits are never activated.
% leads to a healthy initial state of the model where all bits are utilized.
% With the decomposed latent space in \quantizername{}, the first term can be directly decomposed and computed efficiently as $H(\rvq(\rvx)) = \sum_{i=1}^N H(\ervq_i (\ervx_i))$.

We note that there are various other variants of \quantizername{}, \eg, opting for the multivariant over the binary codebook $C_i$ or employing other quantization techniques such as \cite{agustsson2019generative}.
%or using mutivaint codebook instead of the binary code. 
As the first to introduce this concept, we focus on the simplest form with independent binary dimensions, which shows promising improvements. Other \quantizername{} methods merit further research. 
% as $H[\E(\rvq(\rvx)] \approx \E_\mathit{\rvg}\big[H[\E(\rvq_\rvg(\rvx_\rvg)] \big]$.
    % , \rvg \in \{(1, 2, ..., n), (n+1, n+2, ..., 2n), ... (N-n+1, N-n+2, ..., N), (1, n+1, ..., N-n+1), ..., (n, 2n, ..., N)  \}
% The tokens produced by VQ are meaningless integers over a categorical distribution defined by the embeddings in the codebook, which downstream models usually do not have access to.
% On the other hand, LQ tokens represent the embedding by themselves and have a well defined distance metric.

% Generative transformers~\cite{ramesh2021zero,esser2021taming,chang2022maskgit,yu2022magvit} ususally adopt a two-stage framework including a VQ-based tokenizer and a transformer operating in the token space.

% Unless specially designed, the transformer is agnostic to the VQ codebook but only tries to fit the latent distribution of meaningless integers.

% usually operate in the discrete latent space learned by VQ.
% One approach we have found to work best is binary quantization.

\cl{In addition to the entropy penalty (\cref{eq:entropy}), an \quantizername{}-based tokenizer is trained using the standard combination of \emph{reconstruction},  \emph{GAN}, \emph{perceptual}, and \emph{commitment} losses~\cite{esser2021taming}, excluding the inapplicable codebook loss. Following \cite{yu2022magvit}, we use LeCAM regularization~\cite{tseng2021regularizing} for improved stability. 
% For the latents, besides the entropy penalty~(\cref{eq:entropy}), we also use the \emph{commitment} loss~\cite{van2017neural},  $\mathcal{L}_{comm} = \|\rvz - \text{stopgrad}(q(\rvz))\|_2^2$.
}

\subsection{Visual Tokenizer Model Improvement}\label{sec:tokenizer_improvement}
\begin{figure}[tp]
%\vspace{-6mm}
    \centering
    \begin{subfigure}{1\textwidth} % no visible
    \refstepcounter{subfigure}\label{fig:arch:a}
    \refstepcounter{subfigure}\label{fig:arch:b}
    \refstepcounter{subfigure}\label{fig:arch:c}
    \end{subfigure}%
    \includegraphics[width=\linewidth,trim={0 0 0 0},clip]{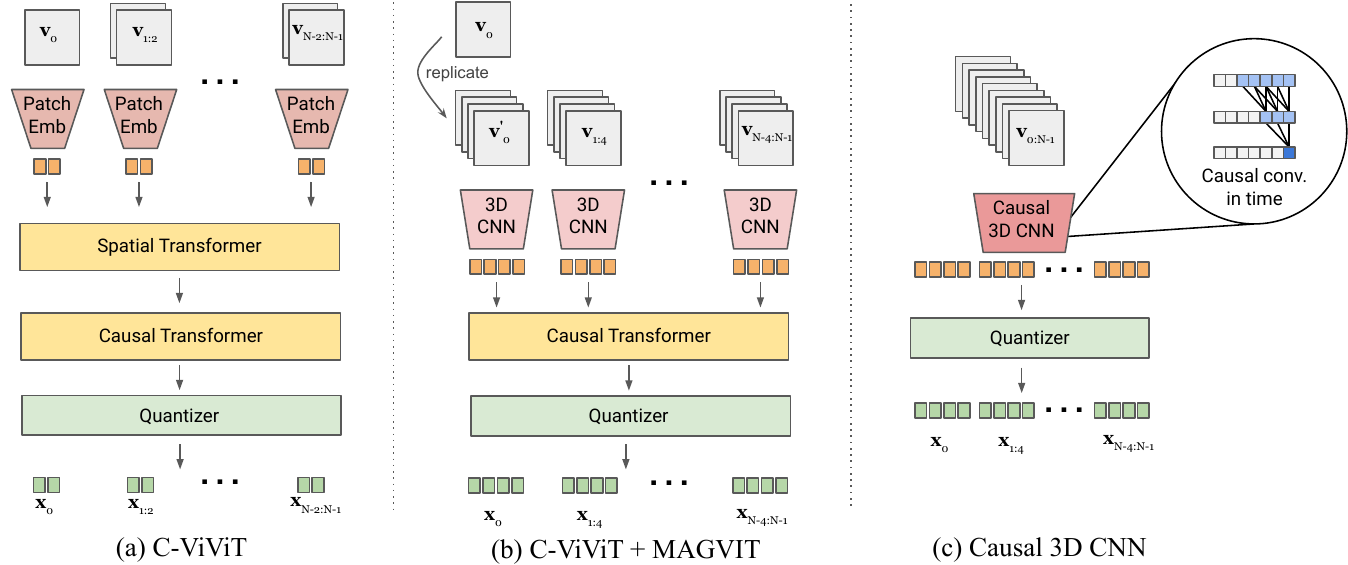}
    %\vspace{-2mm}
    \caption[Causal tokenizer architecture comparison]{\textbf{Causal tokenizer architecture comparison}.
    % \textbf{(a)} C-ViViT. \textbf{(b)} Hybrid of C-ViViT and MAGVIT, where the 3D CNN processes blocks of 4 frames, connected by a causal transformer.
    % \textbf{(c)} \modelname{}: causal 3D CNN using causal convolutions in the temporal dimension. 
    The decoders, which are omitted from the figure, employ an architecture that is symmetric to the encoder.
    % \cl{\scriptsize See detailed architecture diagram in the Appendix.}
    }
    \label{fig:arch}
    %\vspace{-6mm}
\end{figure}

%The devil is in the details.

% \lu{could this be a story: we need to tokenize image and video into the same codebook -> need to tokenize image seperately -> need to get causal design for image -> pros and cons of such design.}

% To fully leverage the enhanced modeling capability brought by the new quantizer, we introduce a new tokenizer model design that unifies the tokenization of images and variable length videos, with causal temporal dependency enforced.

% new designs for the encoder, decoder, and discriminator modules in the tokenizer.
% In particular, the new model 

% \lijun{causal 3d cnn figure}

%\vspace{-2mm}
\paragraph{Joint image-video tokenization.}
A desirable feature of visual tokenization is the capability to tokenize images and videos using a shared codebook. However, the MAGVIT tokenizer, which utilizes the 3D CNN, faces challenges in tokenizing images due to the temporal receptive field. %While replicating an image multiple times to form a static video is possible, this approach introduces redundant code and can potentially alter the training data distribution, resulting in reduced motion.

To build a joint image-video tokenizer, a new design is needed. We begin our discussion by revisiting an existing method C-ViViT~\cite{villegas2022phenaki}. As depicted in \cref{fig:arch:a}, C-ViViT employs full spatial transformer blocks combined with causal temporal transformer blocks. This approach performs reasonably well but has two drawbacks. First, unlike CNNs, the positional embeddings makes it difficult to tokenize spatial resolutions that were not seen during training. Second, empirically we found that 3D CNNs perform better than spatial transformer and produce tokens with better spatial causality of the corresponding patch.
% the self-attention creates a global dependency among tokens. Though such dependency also exists in CNN, it exhibits more globally in the self-attention across all tokens. Hence, a change to a local pixel patch can influence many tokens that are further away,  posing challenges to video editing.
% \jose{doesnt this happen in 3dcnn with enough receptive field as well?}\lijun{I think it remains local spatially, e.g., for an encoder with 16 conv layers on 128x128} 

To tackle these drawbacks, we explore two plausible designs. \cref{fig:arch:b} combines C-ViViT and MAGVIT. Assuming a temporal compression ratio of 4, a 3D CNN processes blocks of 4 frames followed by a causal transformer. In \cref{fig:arch:c}, we use the temporally causal 3D convolution to replace the regular 3D CNN. Specifically, the temporal padding scheme for a regular 3D convolution layer with kernel size $(k_t, k_h, k_w)$ includes $\lfloor\frac{k_t-1}{2}\rfloor$ frames before and $\lfloor\frac{k_t}{2}\rfloor$ frames after the input frames.
In contrast, a causal 3D convolution layer pads with $k_t - 1$ frames before the input and nothing after, so that the output for each frame only depends on the previous frames.
% In this way, at all layers of a deep network, the feature of each frame only depends on the previous frames.
In consequence, the first frame is always independent of other frames, allowing the model to tokenize single images.

Temporal convolutional subsampling with stride $s$ is sufficient for $s\times$ down-sampling by mapping $1 + s \times t$ frames into $1 + t$.
After a regular $s \times$ up-sampling, we drop the first $s - 1$ resulting frames, which maps  $1 + t$ frames into $1 + s \times t$ and allows for the tokenization of a single image. \cref{tab:magvitv2_ablation_arch} empirically compares the designs in \cref{fig:arch}, and we find that the causal 3D CNN performs the best.
\paragraph{Architecture modifications.}
% \lijun{we have ablation results for each design}
In addition to using causal 3D CNN layers, we made several other architectural modifications to improve upon the MAGVIT model. First, we change the encoder downsamplers from average pooling into strided convolutions to leverage learned kernels, and replace the decoder upsamplers from
% First, the average pooling layers in the encoder are replaced with strided convolutions to leverage learned kernels. Second, for the decoder's upsampling layers, the previous method of 
nearest resizing followed by convolution with a depth-to-space operator.
Second, we defer the temporal downsampling from the first few encoder blocks to the last ones.
In addition, the downsampling layer in the discriminator now utilizes 3D blur pooling~\cite{zhang2019making} to encourage shift invariance. 
Finally, we add one adaptive group normalization layer before the residual blocks at each resolution in the decoder to pass in the quantized latents as the control signal following StyleGAN~\cite{karras2019style}.
\cref{tab:magvitv2_ablation_imagenet,tab:magvitv2_ablation_ucf101} empirically verify these designs.
%We also double the number of residual blocks at each resolution to increase the model parameters.

% Besides the pure convolution layers, we revisit the designs of other model components to maximize the performance.
% In particular, we change the average pooling layers in the encoder into strided convolutions to leverage learned kernels.

\begin{figure}[t]
    %\vspace{-6mm}
        \centering
        \includegraphics[width=\linewidth]{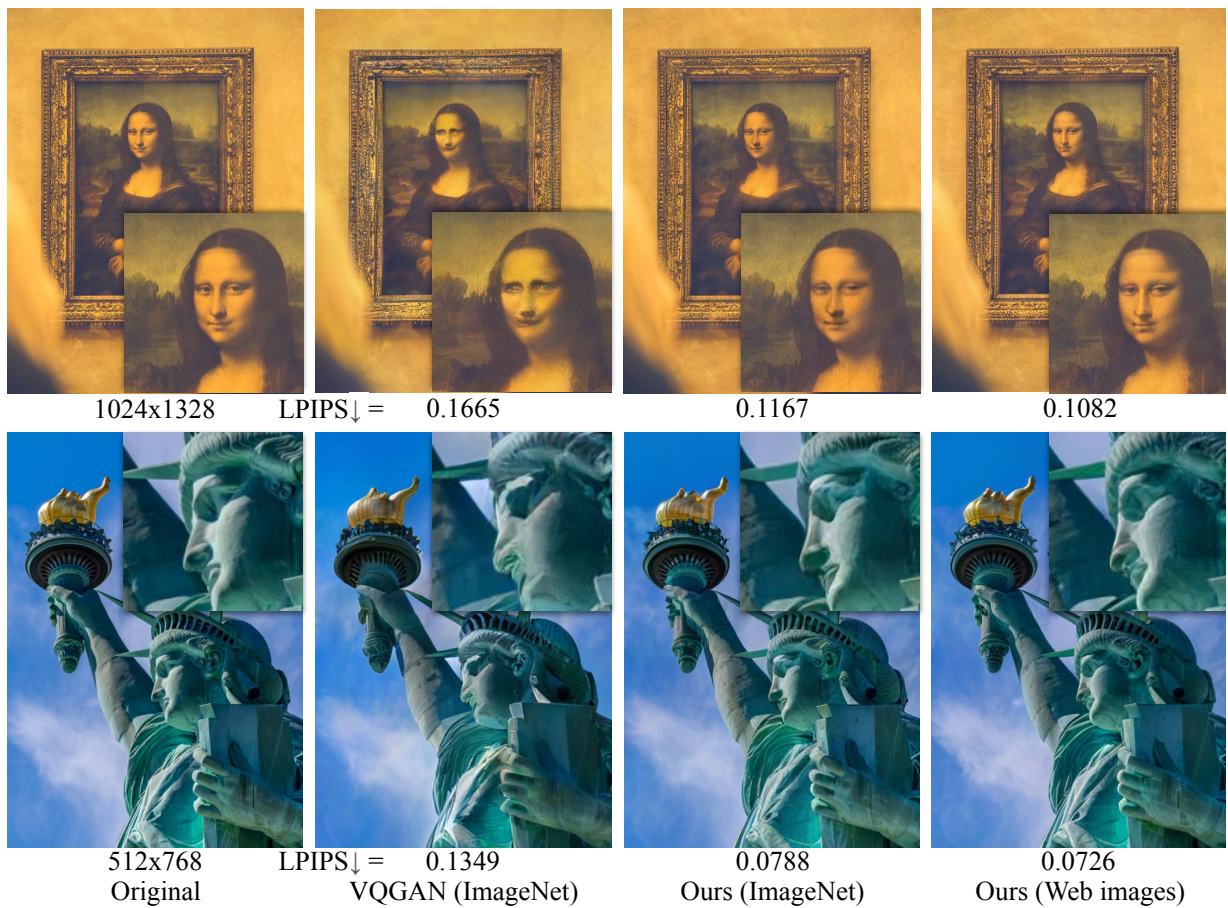}
        %\vspace{-4mm}
        \caption[Image reconstruction samples with different tokenizers]{\textbf{Image reconstruction samples with different tokenizers}. We compare the VQGAN used in MaskGIT~\cite{chang2022maskgit} with two of our models trained on ImageNet and web images~\cite{chen2022pali}. {Original images are by Eric TERRADE and Barth Bailey on Unsplash}.}
        % https://unsplash.com/photos/0WQOCx1g8hw
        % https://unsplash.com/photos/s0di82cRiUQ
        \label{fig:magvitv2_reconstruction}
        %\vspace{-4mm}
    \end{figure}

%\lu{any change in inflation method? This is new in MAGVIT and we can }

%\vspace{-5mm}
\paragraph{Token factorization for efficient prediction.}
The output tokens can be fed into language models to generate videos. To assist smaller transformers predicting in a large vocabulary, we can factorize the \quantizername{} token's latent space into equal subspaces. For instance, rather than predicting using a codebook of size $2^{18}$, we can predict in two concatenated codebooks, each of size $2^9$. We embed each subspace token separately and use their embedding summation as the token embedding for the transformer input. \cl{We find  it beneficial to use weight tying~\cite{press2017using}, a common technique in language modeling, which involves sharing the weights between the embedding and softmax layers.
For the output layer with a factorized vocabulary}, we use the embedding matrix for each subspace to obtain logits with seperate prediction heads. 

\section{Experimental Results}
This section empirically verifies the proposed tokenizer across three distinct tasks: video and image generation, video compression, and action recognition.
\cref{fig:magvitv2_reconstruction} visually compares the reconstruction quality of our tokenizer with prior works.
More qualitative samples are shown at \webpage{}.

%\vspace{-2mm}
\subsection{Experimental Setups}
%\vspace{-2mm}
\paragraph{Datasets.} 
We use Kinetics-600 (K600)~\cite{carreira2018short} and UCF-101~\cite{soomro2012ucf101} for video generation experiments, along with ImageNet~\cite{deng2009imagenet} for image generaton.
In addition, MCL-JCV~\cite{wang2016mcl} is used as the testbed for video compression, with Kinetics-400 (K400)~\cite{kay2017kinetics} and SSv2~\cite{goyal2017something} for video understanding. 
% For compression evaluation we used the dataset: MCL-JCV~\cite{wang2016mcl}.
%\vspace{-4mm}
\paragraph{Implementation details.}
We follow the tokenizer training setting and hyperparameters in~\cite{yu2022magvit}, unless stated otherwise. \quantizername{} is used, which eliminates the codebook embedding, to increase the default codebook size to $K=2^{18}$. 
%that is comparable to that of PaLM 2~\cite{googlepalm2}.
The weight of $\gL_\mathit{entropy}$ follows an annealing schedule with a $3\times$ higher starting point and linearly decays to a fixed value of $0.1$ within 2k steps.
%We also $5\times$ upweight the $\ell_2$ reconstruction loss to encourage better fidelity.
% We defer details regarding the evaluation setup of each subsection to the Appendix.
% \lu{talk about factorization of two codebooks}
% The same MLM as in \cite{yu2022magvit} is used.
% \cite{yu2022magvit} inflates a 2D convolution layers into a regular 3D one
% by filling in the temporally central slice of a zero-filled kernel.
% For the causal 3D layers, we adapt the inflation strategy to fill in the temporally last slice to correspond to the causal padding scheme.
% In addition, we disable the inflation for the discriminator and train it from scratch for better stability.
% We train our model using the infrastructure \url{https://github.com/google-research/magvit}
% \lu{add all the tables; it's fine some numbers are missing}.

\begin{figure}[tp]
    \centering
    \includegraphics[width=\linewidth]{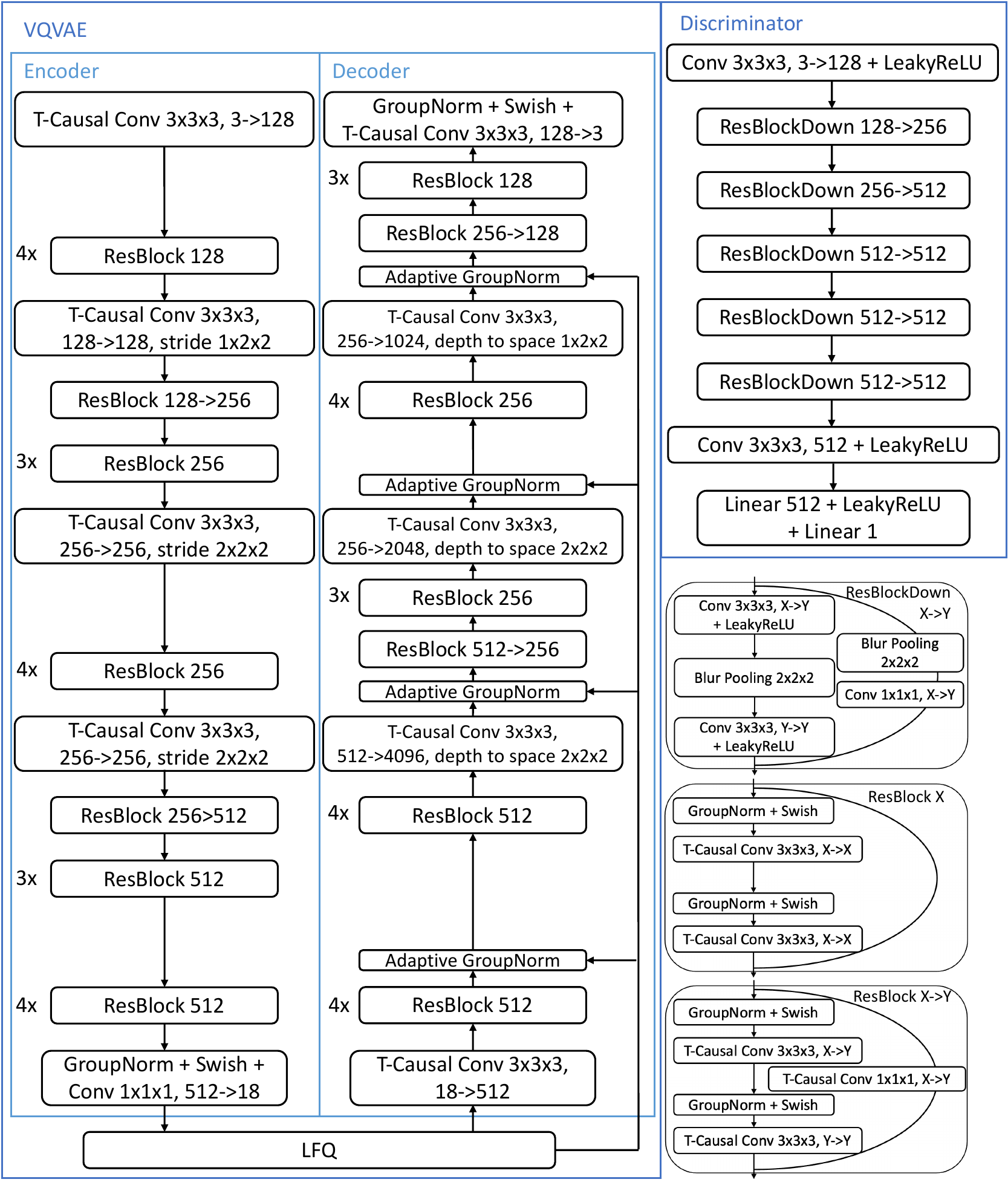}
    \caption[\modelname{} tokenizer architecture]{\cl{\textbf{\modelname{} tokenizer architecture}. T-Causal Conv refers to temporally causal convolution.}}
    \label{fig:full_arch}
\end{figure}
\cref{fig:full_arch} illustrates the architecture of our proposed \modelname{}.
We provide detailed training hyperparameters for our models as listed below:

\begin{itemize}%[nosep, leftmargin=*]
    \item Video input: $17$ frames, frame stride $1$, $128\times128$ resolution.
    \item Base channels: $128$.
    \item VQVAE channel multipliers: $1, 2, 2, 4$. 
    \item Discriminator channel multipliers: $2, 4, 4, 4, 4$.
    \item Number of residual blocks: $4$.
    \item Latent shape: $5\times16\times16$.
    \item Vocabulary size:  $2^{18}$.
    \item Initialization: central inflation from a 2D model trained on ImageNet with this setup.
    \item Entropy loss weight: $0.1$.
    \item Entropy loss annealing steps: $2000$.
    \item Entropy loss annealing factor: $3$.
    \item Reconstruction loss weight: $5.0$.
    \item Generator loss type: Non-saturating.
    \item Generator adversarial loss weight: $0.1$.
    \item Discriminator gradient penalty: r1 with cost $10$.
    \item Perceptual loss weight: $0.1$.
    \item Commitment loss weight: $0.25$.
    \item LeCAM weight: $0.001$.
    \item Peak learning rate: $10^{-4}$.
    \item Learning rate schedule: linear warm up and cosine decay.
    \item Optimizer: Adam with $\beta_1=0$ and $\beta_2=0.99$.
    \item EMA model decay rate: $0.999$.
    \item Batch size: $256$.
    % \item Speed: 1.0 steps/sec on 256 TPU-v5e chips.
    \end{itemize}
    % \item Transformer:
    % \begin{itemize}[nosep, leftmargin=*]
    % \item Number of heads: $16$.
    % \item Number of layers: $24$.
    % \item Hidden size: $1024$.
    % \item MLP dimension: $4096$.
    % \item Vocabulary factorization groups: $2$.
    % \item Sequence length: $1281$.
    % \item Hidden dropout rate: $0.1$.
    % \item Attention dropout rate: $0.1$.
    % \item Mask rate schedule: cosine.
    % \item Peak learning rate: $10^{-4}$.
    % \item Learning rate schedule: linear warm up and cosine decay.
    % \item Optimizer: Adam with $\beta_1=0.9$ and $\beta_2=0.96$.
    % \item Weight decay $0.045$.
    % \item Label smoothing: $10^{-4}$.
    % \item Max gradient norm: $1$.
    % \item Batch size: $256$.
    % \item Speed: 3.7 steps/sec on 256 TPU-v5e chips.
% \end{itemize}
% \end{itemize}

%\vspace{-2mm}
\subsection{Visual Generation}\label{sec:gen}
%\vspace{-2mm}
The masked language model (MLM)~\cite{devlin2019bert} is used in image and video generation. To verify the tokenizer, we employ the same MLM transformers in MAGVIT~\cite{yu2022magvit}. \cl{We select the MLM due to its competitive performance on benchmark datasets~\cite{yu2022magvit,lezama2023discrete}.
% In the Appendix, we also show that an autoregressive language model (AR-LM) coupled with the proposed tokenizer outperforms the prior work MAGVIT.
}
As we use a smaller MLM ($\sim$300M parameters) with a large codebook ($2^{18}\approx$262K), the token factorization as discussed in \cref{sec:tokenizer_improvement} is applied using two heads with each predicting from a codebook of size $2^9$.

%\vspace{-4mm}
\paragraph{Video generation.}
We consider two standard video benchmarks,
UCF-101 for class-conditional generation and K600 for frame prediction with 5-frame condition.
FVD~\cite{unterthiner2018towards} is used as our primary evaluation metric. 

We inflate an image tokenizer trained at 128$\times$128 for video modeling.
Different from the inflation in \cite{yu2022magvit}, we fill in the temporally last slice to correspond to the causal padding scheme. In addition, we disable the inflation for the discriminator and train it from scratch for better stability.
We train the causal video tokenizer on Kinetics-600 training set for 190 epochs with batch size 256.
This tokenizer is also used in subsequent evaluations of video compression and action recognition.

With the causal tokenizer producing 5$\times$16$\times$16 tokens for a 17$\times$128$\times$128 clip, the first 2$\times$16$\times$16 tokens are provided as the condition of the first 5 frames, per the standard setup of Kinetics-600 frame prediction benchmark.
We train the MLM transformer following \cite{yu2022magvit} with token factorization for 360 epochs with batch size 256.
The model is sampled with a cosine schedule using temperature 32.

\begin{table}[tp]
\centering
%\vspace{-6mm}
%\vspace{-2mm}
% \resizebox{\linewidth}{!}{%
\centering
\begin{tabular}{@{}llcccc@{}}
\toprule
Type & Method   & K600 FVD$\downarrow$  & UCF FVD$\downarrow$  & \#Params  & \#Steps  \\ \midrule
% \multicolumn{5}{l}{\textit{Pixel-Space Generative Models}} \\
% ARLM & Video Transformer~\cite{weissenborn2019scaling} & 170.0\mytiny{$\pm$5.0} & & 373M & 197k  \\
% GAN & DVD-GAN-FP~\cite{clark2019adversarial}  & 69.1\mytiny{$\pm$1.2} & & & 1                    \\
GAN & TrIVD-GAN-FP~\cite{luc2020transformation}  & 25.7\mytiny{$\pm$0.7} &  & & 1                \\
% Diffusion & RaMViD~\cite{hoppe2022diffusion}   &  17.5\mytiny{$\pm$1.1} & & 308M & 1000  \\
Diffusion & Video Diffusion~\cite{ho2022video}  & 16.2\mytiny{$\pm$0.3} & & 1.1B &   256        \\
Diffusion & RIN~\cite{jabri2023scalable} & 10.8 & & 411M & 1000 \\
\hdashline
% \multicolumn{6}{l}{\textit{Latent-Space Generative Models}} & Latent bpp\\
% ARLM + VQ & LVT~\cite{rakhimov2021latent} & 224.7 &  & 50M  & 16k  \\
% ARLM + VQ & CogVideo$^*$~\cite{hong2022cogvideo}  & 109.2 & 626 & 9.4B   & 64k     \\
% ARLM + VQ & CCVS~\cite{le2021ccvs}   & 55.0\mytiny{$\pm$1.0} & 386\mytiny{$\pm$18} & 361M   & 1024 \\
AR-LM + VQ & TATS~\cite{ge2022long}  &  & 332\mytiny{$\pm$18} & 321M & 1024 \\
MLM + VQ & Phenaki~\cite{villegas2022phenaki} & 36.4\mytiny{$\pm$0.2} & & 227M   & 48 \\
% ARLM + DCT & Transframer~\cite{nash2022transframer}  & 25.4  &  & 662M &  $<$4k      \\
% T MAGVIT-B-FP        & 24.5\mytiny{$\pm$0.9}     & 43.3\mytiny{$\pm$1.8} & 88M & 0.039 & 12 \\
MLM + VQ & MAGVIT~\cite{yu2022magvit}        & 9.9\mytiny{$\pm$0.3} & 76\mytiny{$\pm$2} & 306M   & 12  \\ 
\midrule
\cl{MLM + \quantizername{}} & \cl{non-causal baseline} & \cl{11.6\mytiny{$\pm$0.6}} & & \cl{307M} & \cl{12} \\
% \cl{AR-LM + \quantizername{}} &  & & \cl{?} & \cl{840M} & \cl{1280} \\
\multirow{2}{*}{MLM + \quantizername{}} & \multirow{2}{*}{\emph{\modelname{} (ours)}} & 5.2\mytiny{$\pm$0.2} &  & \multirow{2}{*}{307M} & 12  \\
 & & \textbf{4.3\mytiny{$\pm$0.1}}  & \textbf{58\mytiny{$\pm$3}} & & 24  \\
\bottomrule
\end{tabular}
% %\vspace{-2mm}
% }
\caption[Video generation results]{\textbf{Video generation results}: frame prediction on Kinetics-600 and class-conditional generation on UCF-101. We adopt the evaluation protocol of MAGVIT.
% \emph{Method types}: AutoRegressive Language Model (ARLM), Masked Language Model (MLM), Generative Adversarial Network (GAN), continuous Diffusion (Diff.). 
% \emph{Latent types}: \quantizerfullname{} (\quantizername{}), Vector Quantization (VQ), Variational AutoEncoder (VAE), Discrete Cosine Transformation (DCT). 
% We report the bits per pixel (bpp) of the representation used in each latent-space model.
}
\label{tab:gen_k600}
\end{table}

\begin{figure}[tp]
    %\vspace{-4mm}
        \centering
        \includegraphics[width=\linewidth,trim={0 0 0 0},clip]{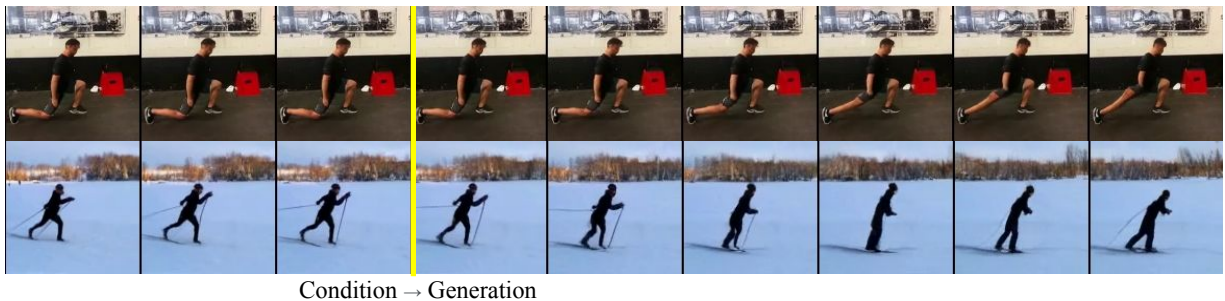}
        %\vspace{-7mm}
        \caption[Frame prediction samples on Kinetics-600]{\textbf{Frame prediction samples on Kinetics-600}.}
        \label{fig:magvitv2_k600}
        %\vspace{-2mm}
    \end{figure}

% We trained the video tokenizer on Kinetics-600 training set for 190 epochs with batch size 256, then trained the MLM transformer following \cite{yu2022magvit} with token factorization for 360 epochs with batch size 256. 
% Further implementation details are described in the appendix. 
\cref{tab:gen_k600} shows that our model surpasses all prior arts in both benchmarks.
Specifically, it outperforms the previous best model MAGVIT by a large margin, while using the same MLM transformer backbone. 
\cl{In addition, it significantly outperforms the non-causal baseline on frame prediction, highlighting the contribution of the causal tokenizer.}
These results demonstrate the essential role of a good visual tokenizer in enabling LMs to generate high-quality videos.
\cref{fig:magvitv2_k600} shows qualitative samples from the model.

\begin{table}[tp]
% {-2mm}
% {-2mm}
% \resizebox{\linewidth}{!}{%
\centering
\begin{tabular}{@{}llcccccc@{}}
\toprule
\multirow{2}{*}{Type}  & \multirow{2}{*}{Method}  & \multicolumn{2}{c}{w/o guidance} & \multicolumn{2}{c}{w/ guidance}  & \multirow{2}{*}{\#Params}  & \multirow{2}{*}{\#Steps}  \\
& & FID$\downarrow$ & IS$\uparrow$ & FID$\downarrow$ & IS$\uparrow$ \\ \midrule
% \multicolumn{6}{l}{\textbf{512 $\times$ 512 resolution}} \\
% \multicolumn{6}{l}{\textit{Pixel-Space Generative Models}} \\
% ADM & & 23.24 & 58.06 & 7.72 & 172.71 \\
GAN & StyleGAN-XL~\cite{sauer2022stylegan} & & & 2.41 & 267.8 & 168M & 1 \\
Diff. + VAE$^*$ & DiT-XL/2~\cite{peebles2022scalable} & 12.03 & 105.3 & 3.04 & 240.8 & 675M  & 250  \\
Diffusion & ADM+Upsample~\cite{dhariwal2021diffusion}  & 9.96  & 121.8 & 3.85 & 221.7 & 731M & 2000 \\
% GAN & BigGAN-deep~\cite{brock2018large} & 8.43 & 177.9 & & & 160M  & 1 \\
Diffusion & RIN~\cite{jabri2023scalable} & 3.95 & 216.0 & & & 320M  & 1000 \\
Diffusion & simple diffusion~\cite{hoogeboom2023simple} & 3.54 & 205.3 & 3.02 & 248.7 & 2B & 512\\
Diffusion & VDM++~\cite{kingma2023vdm}  & 2.99 & 232.2 & 2.65 & 278.1  & 2B & 512 \\
\hdashline
% \multicolumn{8}{l}{\textit{Latent-Space Generative Models}} & Latent bpp \\

MLM + VQ & MaskGIT~\cite{chang2022maskgit} & 7.32 & 156.0 & & & 227M & 12  \\
% MLM + VQ & Token-Critic~\cite{lezama2022improved} & 6.80 & 182.1 &  &  & 368M & 36  \\
MLM + VQ & DPC+Upsample~\cite{lezama2023discrete}  & 3.62 & 249.4 & & & 619M & 72    \\
% MLM + VQ & DPC-light(5)+U~\cite{lezama2023discrete}  & 3.62 & 249.40 & & & 619M & 72 & 0.039   \\
\midrule
\multirow{2}{*}{MLM + \quantizername{}} & \multirow{2}{*}{\emph{\modelname{} (ours)}} & 4.61 & 192.4 & & & \multirow{2}{*}{307M} & 12 \\
& & 3.07 & 213.1 & \textbf{1.91} & \textbf{324.3} &  & 64 \\
% & \emph{\modelname{} (ours)} (40 bits) &  & & &  & 312M & & 0.039 \\
\bottomrule
\end{tabular}
% }
% {-6mm}
\caption[Image generation results at 512$\times$512]{\textbf{Image generation results}: class-conditional generation on ImageNet 512$\times$512.
% \lu{we can remove GAN.} \lijun{recent diffusion papers all kept those, esp. StyleGAN-XL which is very strong}
Guidance indicates the classifier-free diffusion guidance~\cite{ho2021classifier}.
$^*$ indicates usage of extra training data.
We adopt the evaluation protocol and implementation of ADM.
}
\label{tab:gen_in512}
\end{table}
\begin{table}[t]
\centering

\resizebox{\linewidth}{!}{%
\begin{tabular}{@{}l@{\hspace{5pt}}l@{\hspace{4pt}}c@{\hspace{2pt}}c@{\hspace{3pt}}c@{\hspace{2pt}}c@{\hspace{2pt}}c@{\hspace{2pt}}c@{}}
\toprule
\multirow{2}{*}{Type} & \multirow{2}{*}{Method}  & \multicolumn{2}{c}{w/o guidance} & \multicolumn{2}{c}{w/ guidance}  & \multirow{2}{*}{\# Params}  & \multirow{2}{*}{Steps} \\
& & FID$\downarrow$ & IS$\uparrow$ & FID$\downarrow$ & IS$\uparrow$ \\ \midrule
% \multicolumn{6}{l}{\textbf{512 $\times$ 512 resolution}} \\
% % ADM & & 23.24 & 58.06 & 7.72 & 172.71 \\
% $[C]$ ADM-U  & 9.96  & 121.78 & 3.85 & 221.72 & 731M & 128$^2$ pixel & 2000 \\
% $[G]$ BigGAN-deep & 8.43 & 177.90 & & & 160M & & 24 & 1 \\
% $[C]$ RIN & 3.95 & 216.00 & & & 320M & & 24 & 1000 \\
% $[C]$ Simple diffusion & 3.54 & 205.30 & 3.02 & 248.70 & & & 24 &\\
% $[C]$ VDM++ & 2.99 & 232.20 & 2.65 & 278.10  & & & 24 &\\
% $[G]$ StyleGAN-XL & & & 2.41 & 267.75 & 168M & & 1 \\
% \hdashline
% $[C]$ DiT-XL/2 & 12.03 & 105.25 & 3.04 & 240.82 & 675M & 64$^2\times$(VAE 4)$^*$ & 250 \\
% $[T]$ MaskGIT  & 7.32 & 156.00 & & & 227M & 32$^2\times$(VQ 1024) & 12 \\
% % Token-Critic & 32$\times$32$\times$(VQ 1024) & 6.80 & 182.10  \\
% $[D]$ DPC  & 6.09 & 228.10 & & & 454M & 32$^2\times$(VQ 1024) & 180  \\
% \hdashline
% $[T]$ \emph{\modelname} (Ours)  & 3.51 & 235.47 & \textbf{1.92} & \textbf{325.02} & 307M & 16$^2\times$(\quantizername{} $2^{18}$) & 64 \\
% % $[T]$ \emph{\modelname}-XL (Ours) & & & & & 635M & 16$^2\times$(\quantizername{} $2^{18}$) \\
% \midrule
% \multicolumn{6}{l}{\textbf{256 $\times$ 256 resolution}} \\
% ADM &  & 10.94 & 100.98 & 4.59 & 186.70 \\
% \multicolumn{5}{l}{\textit{Pixel-Space Generative Models}} \\
GAN & BigGAN-deep~\cite{brock2018large} & 6.95 & 171.4 & & & 160M & 1 \\
GAN & StyleGAN-XL~\cite{sauer2022stylegan} & & & 2.30& 265.1 & 166M & 1 \\ 
\hdashline
Diff. + VAE$^*$ & LDM-4~\cite{rombach2022high} & 10.56  & 103.5 & 3.60 & 247.7 & 400M & 250 \\
Diff. + VAE$^*$ & DiT-XL/2~\cite{peebles2022scalable}  & 9.62  & 121.5 & 2.27  & 278.2 & 675M  & 250 \\
Diff. + BAE & Binary latent diffusion~\cite{wang2023binary} & 8.21 & 162.3 & & & 172M & 64  \\
Diffusion & ADM+Upsample~\cite{dhariwal2021diffusion} & 7.49 & 127.5 & 3.94  & 215.8 & 608M & 2000 \\
Diff. + VAE$^*$ & MDT~\cite{gao2023masked}  & 6.23  & 143.0 & 1.79  & 283.0 & 676M & 250  \\
Diff. + VAE$^*$ & MaskDiT~\cite{zheng2023fast} & 5.69  & 178.0 & 2.28  & 276.6 & 736M & 40 \\ 
Diffusion & CDM~\cite{ho2022cascaded}  & 4.88 & 158.7 & & & & 8100 \\
Diffusion & RIN~\cite{jabri2023scalable} & 3.42 & 182.0 & & & 410M &  1000 \\
Diffusion & simple diffusion~\cite{hoogeboom2023simple} & 2.77 & 211.8 & 2.44 & 256.3 & 2B & 512 \\
Diffusion & VDM++~\cite{kingma2023vdm} & 2.40 & 225.3 & 2.12 & 267.7 & 2B & 512 \\
\hdashline
% \multicolumn{8}{l}{\textit{Latent-Space Generative Models}} & Latent bpp \\
AR-LM + VQ & VQGAN~\cite{esser2021taming} & 15.78 & 78.3  & & & 1.4B & 256 \\
MLM + VQ & MaskGIT~\cite{chang2022maskgit}  & 6.18  & 182.1 & & & 227M & 8  \\
MLM + VQ & Token-Critic~\cite{lezama2022improved} & 4.69 & 174.5 & & & 368M & 36 \\
MLM + VQ & Contextual RQ-Transformer~\cite{lee2022draft} & 3.41 & 224.6 & & & 1.4B & 72 \\
MLM + VQ & DPC~\cite{lezama2023discrete}  & 4.45 & 244.8 & & & 454M & 180 \\
% ADM & - & 10.94 & 6.02 & 100.98 & 0.69 & 0.63 \\
\midrule
MLM + \quantizername{} & \emph{\modelname{} (ours)} & 3.65 & 200.5 & \textbf{1.78} & \textbf{319.4} & 307M & 64  \\
% $[T]$ \emph{\modelname}-XL (Ours)  & $<$3 & & $<$2 & & 635M & 16$^2\times$(\quantizername{} $2^{18}$) \\
\bottomrule
\end{tabular}
}
\caption[Image generation results at 256$\times$256]{\textbf{Image generation results}: class-conditional generation on ImageNet 256$\times$256.
% \lu{we can remove GAN.} \lijun{recent diffusion papers all kept those, esp. StyleGAN-XL which is very strong}
Guidance indicates the classifier-free diffusion guidance~\cite{ho2021classifier}.
$^*$ indicates usage of extra training data.
We adopt the evaluation protocol and implementation of ADM.
}
% \caption{\textbf{Class-conditional image generation on ImageNet 256$\times$256}. 
% % \emph{Method types}: AutoRegressive Language Model (ARLM), Masked Language Model (MLM), Generative Adversarial Network (GAN), continuous Diffusion (Diff.). 
% % \emph{Latent types}: \quantizerfullname{} (\quantizername{}), Vector Quantization (VQ), Variational AutoEncoder (VAE), Binary AutoEncoder (BAE).
% % We report the bits per pixel (bpp) of the representation used in each latent-space models.
% Guidance indicates the classifier-free diffusion guidance~\cite{ho2021classifier}.
% $^*$ indicates usage of extra training data.
% We adopt the evaluation protocol and implementation of ADM.
% }
\label{tab:gen_in256}
% \end{subtable}
\end{table}
% \textbf{}

%\vspace{-4mm}
\paragraph{Image generation on ImageNet.}
We evaluate \modelname{} on image generation under the standard ImageNet class-conditional setting.
We present results for resolution 512$\times$512 and 256$\times$256. FID~\cite{heusel2017gans} and Inception Score (IS)~\cite{salimans2016improved} are used as evaluation metrics. 

We set up two image tokenizers to downsample by 16$\times$ and 32$\times$, where they are used for generation at 256$\times$256 and 512$\times$512, respectively.
In both cases, an image is represented as 16$\times$16 tokens.
% to map an image into 16$\times$16 tokens.
% This represents a 32$\times$ downsample for the 512$\times$512 version and 16$\times$ for 256$\times$256.
% with 32$\times$ and 16$\times$ downsample ratio, respectively, for generation at 512$\times$512 and 256$\times$256 resolutions, where an image is always represented as 16$\times$16 tokens.
% The image tokenizer maps an image, either 256$\time$256 or 512$\times$512 resolution, into 16$\times$16 tokens.
We train them on the ImageNet training set for 270 epochs using a batch size of 256, both with 256$\times$256 images.
With this tokenizer we train a Masked Language Model following \cite{yu2022magvit}, using the token factorization described in \cref{sec:tokenizer_improvement}.
We train for 1080 epochs in accordance with the prior best model MDT~\cite{gao2023masked}, with batch size 1024 for better efficiency.
For preprocessing and data augmentation, we randomly crop 80-100\% of an image while keeping the aspect ratio, followed by random horizontal flipping.
% Data augmentation including random flipping 
% Following the literature on ImageNet, we do not apply data augmentation. 
% We resize the image keeping its aspect ratio and then proceed with a central crop.
The class label is dropped for 10\% of the training batches to enable classifier-free guidance~\cite{ho2021classifier}.
For unguided generation, we use temperature 30 for 512$\times$512 and 15 for 256$\times$256 in the non-autoregressive decoding.
For guided generation, we adopt the guidance schedule from \cite{gao2023masked} with temperature scaling~\cite{lezama2023discrete}, where we use guidance scale 25 with temperature 15.

\begin{figure}[t]
% %\vspace{-4mm}
    \centering
    \includegraphics[width=\linewidth]{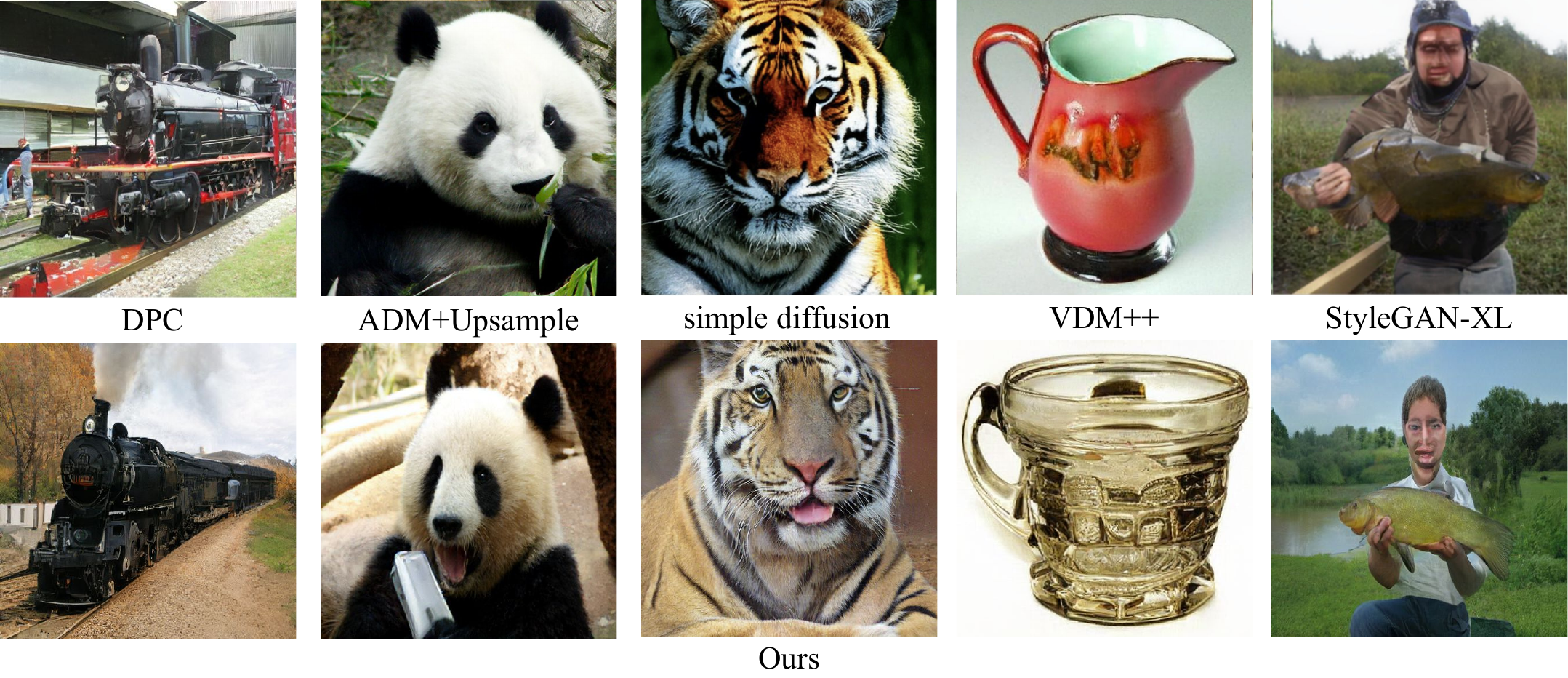}
    %\vspace{-7mm}
    \caption[Class-conditional generation samples on ImageNet 512$\times$512]{\textbf{Class-conditional generation samples on ImageNet 512$\times$512}. We compare with each of the previous works with a random sample from the same image class.}
    \label{fig:imagenet}
    %\vspace{-6mm}
\end{figure}

As shown in \cref{tab:gen_in512,tab:gen_in256}, our model surpasses the best performing diffusion models both in sampling quality (FID and IS) and inference-time efficiency (sampling steps). 
It is worth noting that all the models compared are trained using the same ImageNet training data, with a comparable model size and training budget. Therefore, the performance primarily evaluates the model's capabilities. The masked language model, equipped with our tokenizer, exhibits a notable improvement in FID over the best diffusion model baseline at 512$\times$512 (FID=1.91 \vs 2.65, 28\%$\downarrow$). While this margin narrows at 256$\times$256 resolution, the MLM uses a 50\% reduced model size and needs much fewer decoding steps (\eg, 64 \vs 250) to get the image generation quality.
% It is noteworthy that all the compared models are trained on the same training data, using comparable modelsize, and training budget. So the performance reflects is a test of the model itself. Our model shows significant FID improvement over the best diffusion on 512X512 (FID=1.91 vs. ).
% Albeit the margin is smaller, our model uses 50\% smaller model and much smaller decoding step to generate the image.
% \jose{128x128?}  \lijun{no 128}
% We trained the tokenizer on the ImageNet training set for 270 epochs using a batch size of 256. With this tokenizer we trained a Masked Language Model following \cite{chang2022maskgit}, with the token factorization described in Section~\ref{sec:tokenizer_improvement}, for 1080 epochs and with batch size 1024. We refer to the appendix for further implementation details. \jose{describe augmentations?} 
Qualitative samples in comparison with other models are shown in \cref{fig:imagenet}.

% \cref{} shows the ImageNet generation results. Our model is compared with   
% 256: appendix

% \subsection{Video Generation and Future Frame Generation}

% %\vspace{-4mm}

% \setlength\intextsep{20pt}
\begin{figure}[p]
    \includegraphics[width=\textwidth]{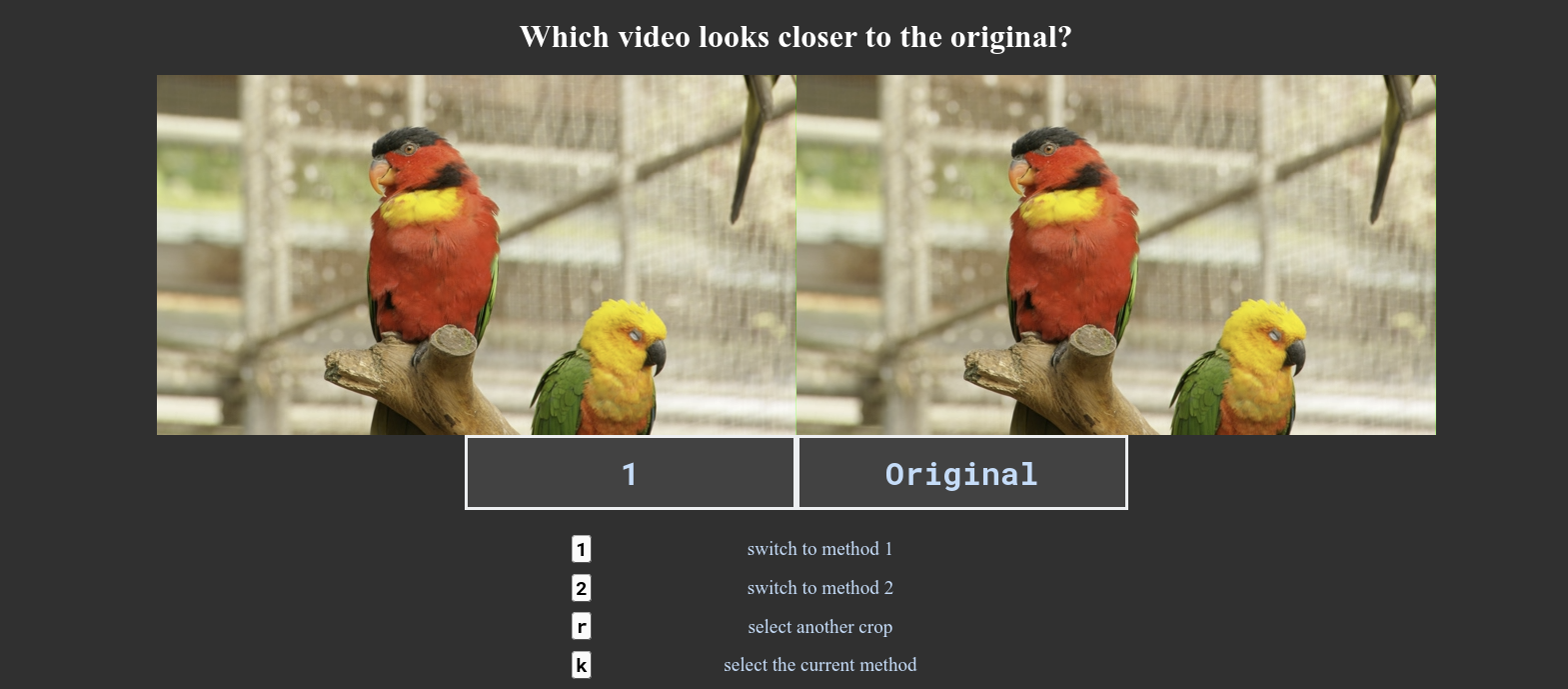}
    \caption[Rating interface for subjective compression evaluation]{\textbf{Rating interface for subjective compression evaluation}.}
    \label{fig:user_rater}
\end{figure}
\begin{figure}[tp]%{r}{0.45\textwidth}
%\vspace{-2mm}
    \centering
    %\begin{subfigure}{.48\linewidth}
    % Figure generated by this colab:
    % https://colab.corp.google.com/drive/1jpXXVPyztiyUPFxrfkdpubZ0NmEJAAW3?resourcekey=0-VKb_gdWn7Gp4Sy4e5MRdNg
    \includegraphics[width=0.6\linewidth]{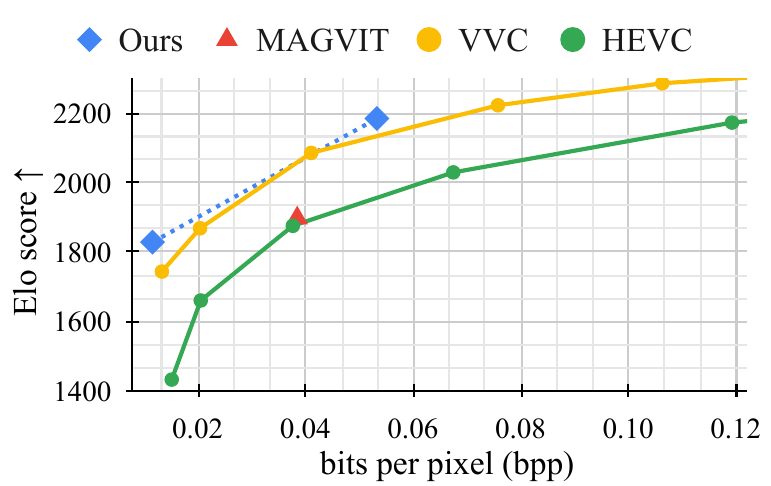}
    %\end{subfigure}
    %\begin{subfigure}{0.48\linewidth}
    %\includegraphics[width=\linewidth]{figures/psnr-codec-curves.pdf}
    %\end{subfigure}
    %\vspace{-6mm}
    \caption[Video compression rater study]{\textbf{Video compression rater study}.
    % In a subjective rater study assessing compression quality, our model is preferred to other video models including MAGVIT, HEVC (H.265), and VVC (H.266). We calculate Elo scores based on pairwise preferences to quantify the relative visual quality between the models (higher is better).
    }
    \label{fig:compression}
    %\vspace{-28mm}
\end{figure}

\begin{figure}[tp]
    \centering
    \begin{subfigure}{0.32\linewidth}
    \includegraphics[width=\linewidth]{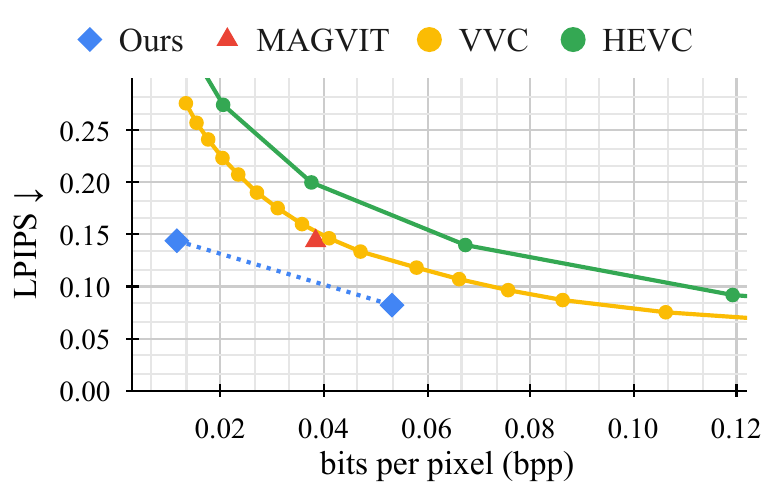}
    \caption{\cl{LPIPS$\downarrow$}}
    \end{subfigure}
    \begin{subfigure}{0.32\linewidth}
    \includegraphics[width=\linewidth]{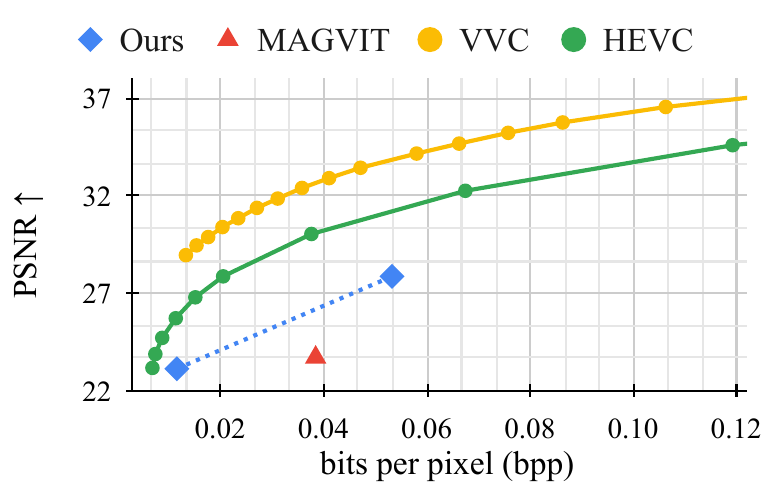}
    \caption{\cl{PSNR$\uparrow$}}
    \end{subfigure}
    \begin{subfigure}{0.32\linewidth}
    \includegraphics[width=\linewidth]{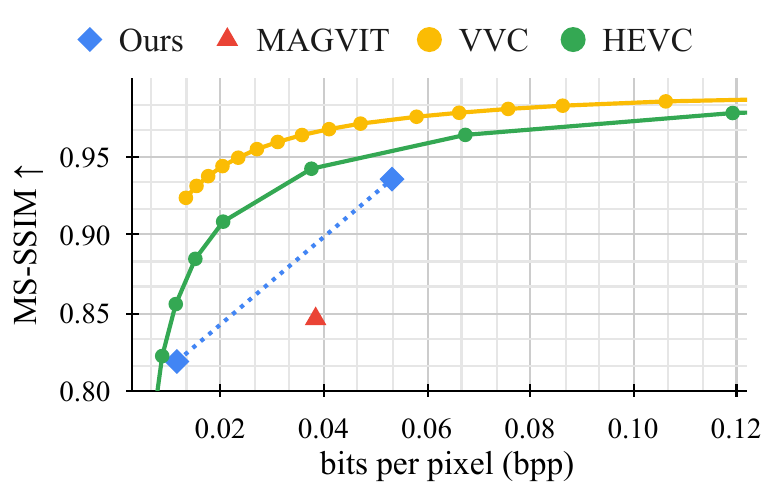}
    \caption{\cl{MS-SSIM$\uparrow$}}
    \end{subfigure}
    \caption[Video compression metrics]{\cl{\textbf{Video compression metrics}, supplementary to \cref{tab:compression}.} }
    \label{fig:compression_metrics}
\end{figure}

%\vspace{-8mm}
\subsection{Video Compression}

We conduct a subjective rater study to assess the compression quality of \modelname{}. The study is conducted on the 30 videos of the MCL-JCV dataset, resized to a resolution of 640$\times$360.
Sixteen raters are engaged, each providing responses to an average of roughly 800 pairwise-preference questions.
% conducted using 16 raters, with an average of approximately 800 pairwise-preference questions per rater.
% The questions were presented with an interface that parallels the one used for the Challenge on Learned Image Compression (http://compression.cc/), extended to comparing videos. Raters were not allowed to pause the videos.
% The study was conducted on the 24 videos of the MCL-JCV dataset (https://mcl.usc.edu/mcl-jcv-dataset/), scaled down to a resolution of 640x360 pixels.
% \newpage

To rate the quality of the different methods, we use a two-alternative forced choice rating methodology~\cite{fechner1860elemente}. As this methodology produces a sequence of binary decisions, we calculate Elo scores~\cite{elo1978rating} based on pairwise preferences to quantify the relative visual quality between the models. The study was conducted on the 30 videos of the MCL-JCV dataset~\cite{wang2016mcl}, scaled down to a resolution of 640$\times$360 pixels.
Sixteen raters are engaged, each providing responses to an average of roughly 800 pairwise-preference questions. The questions are presented with an interface that parallels the one used for the Challenge on Learned Image Compression ({\small \url{http://compression.cc/}}), extended to comparing videos, as shown in \cref{fig:user_rater}. Raters are instructed to compare the two videos and are not allowed to pause the videos.

We calculate Elo scores~\cite{elo1978rating} based on pairwise preferences to quantify the relative visual quality between the models. The study compares our model with MAGVIT as well as the current video compression standard HEVC (H.265) video codec~\cite{sullivan2012overview} and the next-generation codec VVC (H.266)~\cite{vvc}. As shown in \cref{fig:compression}, raters prefer our model to the compared methods at multiple bit rates.

\begin{table}[tp]%{r}{0.45\textwidth}
%{-10mm}
    \centering
    %{-2mm}
    % \resizebox{\linewidth}{!}{%
    \begin{tabular}{lcccc}
    \toprule
    Method  & LPIPS$\downarrow$ & PSNR$\uparrow$ & MS-SSIM$\uparrow$ \\
    \midrule
    HEVC~\cite{sullivan2012overview} & 0.199& 30.10 & 0.943 \\ 
    VVC~\cite{vvc} & 0.153 & \textbf{32.65} & \textbf{0.966} \\
    \midrule
    MAGVIT~\cite{yu2022magvit} & 0.144 & 23.70 & 0.846 \\ 
    \emph{\modelname{} (ours)} & \textbf{0.104} & 26.18 & 0.894 \\
    \bottomrule
    \end{tabular}
    % }
    %{-8mm}
    \caption[Video compression metrics]{\textbf{Video compression metrics}.
    % We compare the compression quality of our approach using three common quantitative metrics (PSNR, MS-SSIM, and LPIPS) at 0.0384 bpp, the bit rate of MAGVIT. The results show that our model outperforms MAGVIT on all metrics, and it outperforms all methods on LPIPS. Since PSNR and MS-SSIM correlate relatively poorly with subjective assessments of visual quality, it is not surprising that our method scores worse on these distortion metrics than HEVC and VVC.
    }
    \label{tab:compression}
\end{table}

% magvit v1
% {'LPIPS': 0.14433361030154757,
%  "MS-SSIM/sRGB/R'G'B'": 0.8455671715630425,
%  "PSNR/sRGB/R'G'B'": 23.695795461697045}

% magvitv2-58219767
% {'LPIPS': 0.1441961204952664,
%  "MS-SSIM/sRGB/R'G'B'": 0.8187397972106935,
%  "PSNR/sRGB/R'G'B'": 23.121115535481767}

% magvitv2-59508633
% {'LPIPS': 0.08232382029427424,
%  "MS-SSIM/sRGB/R'G'B'": 0.9356586151970757,
%  "PSNR/sRGB/R'G'B'": 27.856519574652776}
%\lu{add a sentence to give a bit more context on HEVC abd VCC}.
% Here we will add a figure comparing MAGVIT, HEVC, VCC compression quality in user studies.
We also compare the compression quality using common distortion metrics (LPIPS, PSNR, and MS-SSIM). 
\cl{\cref{tab:compression} compares at 0.0384 bpp, the bit rate of MAGVIT, with full curves in \cref{fig:compression_metrics}}.
The results show that our model outperforms MAGVIT on all metrics, and it outperforms all methods on LPIPS, a metric which correlates more closely with subjective quality assessments.
%\cl{At the same bit rate, standard codecs 
% based on block-level Fourier-related transformations may capture local details more accurately than neural models but introduce block artifacts, which can result in higher PSNR and MS-SSIM due to their lack of realistic-ness measurement~\cite{agustsson2019generative}.}
% According to \cite{agustsson2019generative}, PSNR and MS-SSIM focus on local details but cannot measure realistic-ness, where
%
\cl{
At equal bit rates, standard codecs may render local details
 more accurately than neural models but  also introduce block
 artifacts,  detrimental to  perceptual quality yet not captured by PSNR and MS-SSIM~\cite{agustsson2019generative}.
}
%\cl{Future research is needed to improve the efficiency of the tokenizer to match that of standard codecs.}
% \cl{Further research is needed to implement a CPU version of the tokenizer matching the compute efficiency of standard codecs.}
\cl{Despite promising results with TPUs, further research is needed to adapt our model to run efficiently on CPUs like standard codecs.}

%Since PSNR and MS-SSIM correlate relatively poorly with subjective assessments of visual quality, it is not unexpected that our method scores worse on these distortion metrics than HEVC and VVC.

% \newfloatcommand{capbtabbox}{table}[][\FBwidth]

% \begin{figure}
% \begin{floatrow}
% \ffigbox{%
%   \includegraphics[width=\linewidth]{figures/elo-codec-curves.pdf}%
% }{%
%   \caption{In a human rater study of compression quality, our model is preferred to other video models including MAGVIT, HEVC (H.265), and VVC (H.266), as shown by relative Elo scores.}
%   }
% \capbtabbox{%
%     \begin{tabular}{lcccc}
%     \toprule
%       & PSNR$\uparrow$ & MS-SSIM$\uparrow$ & LPIPS$\downarrow$\\
%     \midrule
%     HEVC & 30.15 & 0.94 & 0.20 \\ 
%     VVC & 32.72 & 0.97 & 0.15\\ 
%     MAGVIT\\ 
%     Ours\\
%     \bottomrule
%     \end{tabular}

% }{%
%   \caption{Quantitative metrics averaged over the MCL-JCV video dataset targeting an average bpp of 0.0391.}%
% }
% \end{floatrow}
% \end{figure}

% %\vspace{-2mm}
% \setlength\intextsep{20pt}
\begin{table}[tp]%{r}{0.5\textwidth}
% %{-4mm}
\centering
%{-2mm}
% \resizebox{\linewidth}{!}{%
\begin{tabular}{@{}l@{\hspace{5pt}}c@{\hspace{10pt}}c@{\hspace{5pt}}c@{\hspace{5pt}}c@{}}
\toprule
\makecell[r]{Token as transformer's:} & Output & \multicolumn{3}{c}{Input}  \\
Tokenizer & SSv2 & SSv2 & K400 & K600 \\ \midrule
% VQVAE in \cite{wang2022bevt} & 67.10 & - & - & - \\
% 3D VQVAE \lu{in BEVT?} \lijun{it's ours} & 64.13 & 41.27 & 44.44 & 45.67 \\
3D VQ-VAE & 64.13 & 41.27 & 44.44 & 45.67 \\
% MAGVIT~\cite{yu2022magvit} & 67.22 & 57.34 / 82.80 & 72.29 / 89.56 & 74.65 / 91.22 \\
MAGVIT~\cite{yu2022magvit} & 67.22 & 57.34 & 72.29 & 74.65 \\
\emph{\modelname{} (ours)} & \textbf{67.38} & \textbf{62.40} & \textbf{75.34} & \textbf{77.93} \\ \midrule
\textcolor{gray}{Raw pixel} & \textcolor{gray}{\cl{64.83}} & \textcolor{gray}{63.08} & \textcolor{gray}{76.13} & \textcolor{gray}{78.92} \\ 
\textcolor{gray}{\cl{HoG descriptor~\cite{wei2022masked}}} & \textcolor{gray}{\cl{65.86}} & \textcolor{gray}{\cl{n/a}} & \textcolor{gray}{\cl{n/a}} & \textcolor{gray}{\cl{n/a}} \\
\bottomrule
\end{tabular}
% }
%{-14mm}
\caption[Video action recognition performance]{\textbf{Video action recognition performance} (classification accuracy$\uparrow$ $\times$100).}
\label{tab:understanding}
\end{table}
%{-8mm}
% \begin{table}[tp]
%     \centering
%     \caption{\textbf{Video action recognition performance} (classification accuracy $\times 100$).}
%     \label{tab:understanding}
%     \begin{tabular}{lccccccc}
%     \toprule
%     & Codebook & $\gL_\mathit{GAN}$ & $\gL_\mathit{Perc.}$ & $\gL_\mathit{Comm.}$ & $\gL_\mathit{Entr.}$ & SSv2 Acc.$\uparrow$ & K400 Acc.$\uparrow$ \\ \midrule
%     \multicolumn{4}{l}{Target: \textbf{MAGVIT}~\cite{yu2022magvit}} \\
%     (VQGAN) & 1024 & \checkmark & \checkmark & \checkmark & \checkmark & \\ 
%     (VQVAE) & 1024 & & & \checkmark & \checkmark \\ 
%     \midrule
%     \multicolumn{4}{l}{Target: \textbf{\modelname{}} (ours)} \\
%     (\quantizername GAN) & 1024 & \checkmark & \checkmark & \checkmark & \checkmark & \\ 
%     & 4096 & \checkmark & \checkmark & \checkmark & \checkmark & \\ 
%     & 16384 & \checkmark & \checkmark & \checkmark & \checkmark & \\ 
%     & 65536 & \checkmark & \checkmark & \checkmark & \checkmark & \\ \hdashline
%      & 16384 & & \checkmark & \checkmark & \checkmark & \\ 
%     (\quantizername VAE) & 16384 &  &  & \checkmark & \checkmark & \\ 
%     & 16384 & & &  & \checkmark & \\ 
%     (L2 only) & 16384 & & &  &  & \\ \bottomrule
%     \end{tabular}
% \end{table}
\leavevmode
\subsection{Video Understanding}
%\vspace{-2mm}
In this subsection, we assess the tokenizer's capability to learn a video understanding model for action recognition. Two setups are examined: (1) using tokens as prediction targets for the transformer's output, and (2) using tokens as the input to the transformer. For the former setup, we use a similar architecture following the BEVT~\cite{wang2022bevt} pre-training. For the tokens as inputs, to work with the ViViT backbone~\cite{arnab2021vivit}, we detokenize the tokens to pixels before feeding them to \cl{frozen} ViViT transformers \cl{trained on raw pixels}.

%A good tokenizer would excel in both settings.

\paragraph{Tokens as prediction targets.}
BEiT~\cite{bao2021beit} and BEVT~\cite{wang2022bevt} class of models pretrain visual encoders on pixel inputs by predicting tokens as targets in a masked-modeling framework, and demonstrate state-of-the-art downstream results. We use a simplified BEVT pre-training setup to test the effectiveness of our video tokens as targets for masked modeling. The main difference is that we drop the image-stream from pre-training and only use the video stream and for this reason, we also drop the multiple decoders completely and adopt an encoder-only architecture similar to BEiT. Detailed pre-training and fine-tuning setup is presented in \cref{tab:bevt_experiment_setup}. 
% In \cref{tab:understanding} of the main paper, we show that our video tokens are effective targets for masked modeling based video understanding.

\paragraph{Tokens as inputs.}
We show that we can re-use video understanding models trained on pixels using our video tokens as input, with very minimal performance drop. For this experiment, we train a factorized variant of the ViViT model \cite{arnab2021vivit} on pixels, and evaluate it on de-tokenized pixels from our model. We use the same hyper-parameters as used in \cite{arnab2021vivit} with a Base sized model operating on 32 frames of inputs at 224p resolution. For the Kinetics-600 experiment, we use the same hyper-parameters as the Kinetics-400 experiments.

\begin{table}[tp]
	\centering
    %  \setlength\tabcolsep{4pt}
    % \resizebox{0.7\linewidth}{!}{
		\begin{tabular}	{l|l|l}
			\toprule
			\textbf{Config} & \textbf{SSv2 Pre-Training} & \textbf{SSv2 Fine-tuning} \\
			\midrule
    inputs & pixels & pixels \\
    input size & 16 $\times$ 224 $\times$ 224 $\times$ 3 & 16 $\times$ 224 $\times$ 224 $\times$ 3 \\
    targets & tokens & classes \\
    encoder & ViT-B & ViT-B \\
    decoder & linear & linear \\
    masking & block-tube \cite{wang2022bevt} & none \\
    masking ratio & 0.75 & 0.0 \\
    mask temporal length & 16 & 0 \\
	batch size & 1024  & 512  \\
	training epochs & 800 & 50  \\
	ViT sequence length & 8 $\times$ 16 $\times$ 16  & 8 $\times$ 16 $\times$ 16 \\
% 	optimizer
	\textbf{optimization} \\
	optimizer & AdamW  & AdamW \\ 
% 	base learning rate & 2.5e-1  \\
	optimizer momentum & 0.9 & 0.9 \\
	layer decay & 0.75  & 0.75 \\ 
	weight decay & 0.05 & 0.05 \\
	learning rate schedule & cosine decay  & cosine decay \\
	warmup epochs & 40  & 5 \\
	\textbf{data augmentations} \\
	random horizontal flip & true & false \\
	label smoothing & 0.1 & 0.1 \\
	mixup & none & 0.8 \\
	cutmix & none & 1.0 \\
	droppath & 0.0 & 0.1 \\
	dropout & 0.1 & 0.0 \\
	random color augmentation & false & false \\
		\bottomrule
		\end{tabular}
% 		}
\caption[Experimental configurations with tokens as targets]{\textbf{Experimental configurations with tokens as targets}.}
	\label{tab:bevt_experiment_setup}
\end{table}

\cref{tab:understanding} shows that \modelname{} outperforms the previous best MAGVIT in these evaluations. Specifically, when using the decoded tokens as input, the performance approaches that of the model trained with ground-truth pixels using the same ViViT backbone. While these numbers are still worse than the state-of-the-art in action recognition, they represent solid improvements credited to the new tokenizer.
% Here we will compare BEVT training using different tokenizers other than the proposed one.

%\vspace{-2mm}
\subsection{Ablation Study}
%\vspace{-2mm}
In \cref{fig:motivation}, we have ablated \quantizername{} \vs VQ and the vocabulary size.
In \cref{tab:magvitv2_ablation}, we validate the key designs proposed in \cref{sec:tokenizer_improvement}. Specifically, \cref{tab:magvitv2_ablation_arch} compares the architecture illustrated in \cref{fig:arch}; \cref{tab:magvitv2_ablation_imagenet} and \cref{tab:magvitv2_ablation_ucf101} verify the \quantizername{} and other improvements on ImageNet and UCF-101, respectively.

\cref{tab:magvitv2_ablation_lfq} verifies the entropy penalty $\gL_\mathit{entropy}$ introduced in \cref{eq:entropy}, which plays an important role under the LFQ setup to achieve better reconstruction and better codebook utilization.
When using binary latents, certain LFQ dimensions may become dead if the dataset is not complexed enough for the large vocabulary size.
In such cases, $\gL_\mathit{entropy}$ is beneficial to ensure the activation of all bits, which enforces a discrete uniform prior in the latent space.
Multi-variate latents are less likely to have dead dimensions, since fewer dimensions are used at the same vocabulary size.
A potential alternative to $\gL_\mathit{entropy}$ is applying normalization before quantization, such that the continuous latents are centered around $0$.
The specific choice of normalization is left as future work.
% $\gL_\mathit{entropy}$ is similar to the KL divergence term in VAEs~\cite{kingma2013auto}, which enforces a prior distribution in the latent space.
% The main difference is that VAE applies KL on independent dimensions, where as the entropy penalty is jointly applied on all dimensions.

% vocabulary size: figure 1
% vq vs lq: figure 1
% table 4
% group transformer
% sq: appendix

\begin{table}[tp]
%{-4mm}
\centering
% \scriptsize

%{-2mm}
\begin{subtable}{\linewidth}
\centering

\begin{tabular}{@{}lccc@{}}
\toprule
 & \#Params & FID$\downarrow$ & FVD$\downarrow$ \\ \midrule
\textcolor{gray}{MAGVIT} & \textcolor{gray}{39M} & \textcolor{gray}{n/a} & \textcolor{gray}{107.15}  \\
C-ViViT & 90M & 28.02 & 437.54 \\
C-ViViT + MAGVIT & 67M & 13.52  & 316.70 \\
\makecell[l]{\emph{\modelname{}}: Causal 3D CNN} & 58M & \textbf{7.06} & \textbf{96.33} \\ \bottomrule
\end{tabular}
\caption{\label{tab:magvitv2_ablation_arch}Causal architectures on UCF-101. FID is calculated on the first frame.}
\vspace{5mm}
\end{subtable}
% \hfill

\begin{subtable}{\linewidth}
\centering

% \resizebox{\linewidth}{!}{%
\begin{tabular}{@{}lcc@{}}
\toprule
 & FID$\downarrow$ & LPIPS$\downarrow$ \\ \midrule
MAGVIT & 2.65 & 0.1292 \\
+ LFQ & 2.48 & 0.1182 \\
+ large vocabulary & 1.34 & 0.0821 \\
+ up/downsampler & 1.21 & 0.0790 \\
% + $\gL_{reconstruction}$ weight &  &  \\
+ deeper model & 1.20 & 0.0686 \\
+ adaptive normalization & \textbf{1.15} & \textbf{0.0685} \\ \bottomrule
\end{tabular}
% }
\caption{\label{tab:magvitv2_ablation_imagenet}Image tokenization on ImageNet 128$\times$128.}
\vspace{5mm}
\end{subtable}
% \hfill
\begin{subtable}{\linewidth}
\centering

% \resizebox{\linewidth}{!}{%
\begin{tabular}{@{}lcc@{}}
\toprule
 & FVD$\downarrow$ & LPIPS$\downarrow$ \\ \midrule
MAGVIT & 24.55 & 0.0988 \\
+ LFQ \& large vocabulary & 16.12 & 0.0694 \\
+ up/downsampler & 15.37 & 0.0678 \\
+ late temporal downsample & 11.11 & 0.0653 \\
+ deeper model & 8.90 & 0.0542 \\
+ 3D blur pooling & \textbf{8.62} & \textbf{0.0537} \\ 
% + causal 3D CNN & \textbf{} & \textbf{} \\
\bottomrule 
\end{tabular}
% }
\caption{\label{tab:magvitv2_ablation_ucf101}Video tokenization on UCF-101.}
\vspace{5mm}
\end{subtable}

\begin{subtable}{\linewidth}
\centering

% \resizebox{\linewidth}{!}{%
\begin{tabular}{@{}lccccc@{}}
\toprule
$\gL_\mathit{entropy}$  & L1$\downarrow$ & L2$\downarrow$ & Utilization$\uparrow$ & Entropy$\uparrow$ \\ \midrule
& 0.174 & 0.052 & 0.219 & 3.12 \\
\checkmark & \textbf{0.147} & \textbf{0.039} & \textbf{1.000} & \textbf{7.18} \\
% VQ  & - & 0.177 & 0.060 & 0.973 & 7.68 \\ \hdashline
% FSQ & - & 0.172 & 0.056 & 1.000 & 7.28 \\ \midrule
% \multirow{3}{*}{LFQ}  & 0.025 & 0.  & 0. &  & 4.67 \\
%  & 0.050 & 0.194 & 0.069 & 1.000 & 7.87 \\
%   & 0.100 & 0.198 & 0.072 & 1.000 & 7.91 \\
\bottomrule 
\end{tabular}
% }
\caption{\label{tab:magvitv2_ablation_lfq}Entropy penalty on CIFAR10 with a $2^8$ LFQ vocabulary.}
\end{subtable}

\caption[Ablation study verifying key design choices]{\textbf{Ablation study verifying key design choices}.}
\label{tab:magvitv2_ablation}
%{-4mm}
\end{table}

\section{Summary}

We introduce \modelname{}, a novel video tokenizer that exploits lookup-free quantization along with architectural advancements to tokenize images and video with a shared vocabulary. The experiments show that our tokenizer outperforms the previously leading video tokenizer across three areas: visual generation, video compression, and action recognition in videos. Our results suggest that a good visual tokenizer is key for enabling language models to excel in image and video generation. These results demonstrate the great capabilities of LMs in visual generation, and advocate for further exploration of advanced visual tokenization methods as \emph{multi-modal latent representation} designed for LLMs.
%We have demonstrated that Language Models outperform diffusion models for visual generation when equipped with a good visual tokenizer mapping pixel-space inputs to discrete tokens in a suitable vocabulary. To this end, we 
%\modelname{} outperforms previous video tokenizers in reconstruction and compression quality. When coupled with a Masked Language Model, it outperforms all previous methods, including diffusion models, on standard image and video generation benchmarks including ImageNet and Kinetics.

\glsresetall
\cleardoublepage
\part{Multi-Task Generative Models}\label{part:model}
\paragraph{\cref{part:model} Overview.}

% \chapter{Continuous-Time Discrete Diffusion Models}

\glsresetall
\cleardoublepage
\chapter{Masked Generative Video Transformer}\label{chap:magvit}
\graphicspath{{papers/magvit/figures/}}

\renewcommand{\modelname}{MAGVIT}
\renewcommand{\methodname}{COMMIT}
\renewcommand{\mask}{\texttt{[MASK]}\xspace}

\paragraph{Overview.}
In this chapter, we introduce the \Gls{MAGVIT} to tackle various video synthesis tasks with a single model.
%
%MH: this sentence is redudant as you already mention in the previous sentence. %Lijun: agreed
%We train a single \modelname{} model and apply it to multiple video generation tasks at inference time.
% To achieve this goal, 
With the 3D tokenizer introduced in \cref{chap:3dvq}, we propose an embedding method for masked video token modeling to facilitate \emph{multi-task} learning.
We conduct extensive experiments to demonstrate the quality, efficiency, and flexibility of \Gls{MAGVIT}.
Our experiments show that (i) \Gls{MAGVIT} performs favorably against state-of-the-art approaches and establishes the best-published \acrshort{FVD} on three video generation benchmarks, including the challenging Kinetics-600. 
(ii) \Gls{MAGVIT} outperforms existing methods in inference time by  two orders of magnitude against diffusion models and by 60$\times$ against autoregressive models. 
(iii) A single \Gls{MAGVIT} model supports ten diverse generation tasks and generalizes across videos from different visual domains.
%
%We will open source our framework and models.
The source code and trained models are released to the public at \url{https://magvit.cs.cmu.edu}.

\clearpage

\begin{figure*}[p]
\begin{subfigure}{\linewidth}
\centering
\includegraphics[width=0.64\textwidth, trim={0.2cm 0 17.4cm 0},clip]{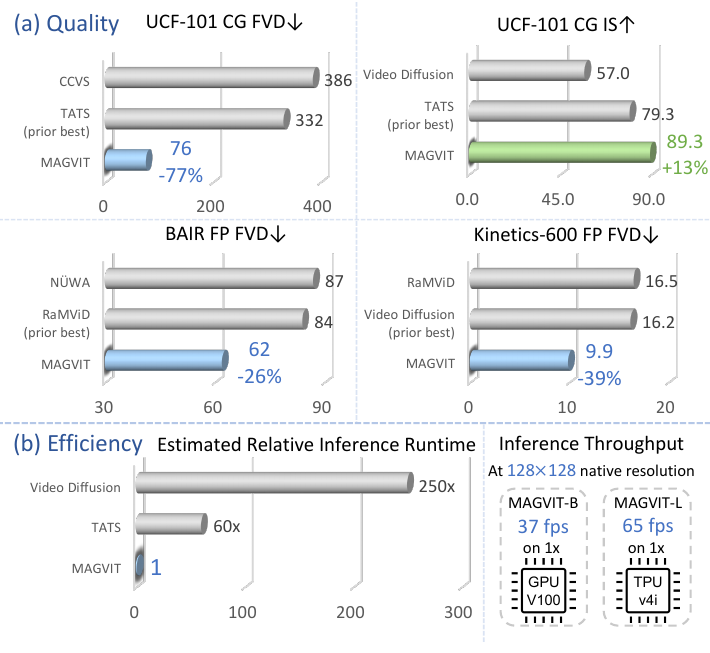}
\end{subfigure}
\begin{subfigure}{\linewidth}
\centering
\includegraphics[width=0.8\textwidth, trim={12.3cm 0 0 0},clip]{teaser}
\end{subfigure}
\caption[Overview of the video generation quality, efficiency, and flexibility of \modelname{}]{\textbf{Overview of the video generation \emph{quality}, \emph{efficiency}, and \emph{flexibility} of the proposed \modelname{} model}.
%
%MH: in the text, try not to say "improves from previous best"... as it lacks
%\modelname{} largely improves from previous best performance on two video generation tasks across three common benchmarks.
(a) \modelname{} achieves the state-of-the-art FVD~\cite{unterthiner2018towards} and Inception Score (IS)~\cite{saito2020train} on two video generation tasks and three benchmarks, in comparison with prior best diffusion models (RaMViD~\cite{hoppe2022diffusion}, Video Diffusion~\cite{ho2022video}) and autoregressive models (CCVS~\cite{le2021ccvs}, TATS~\cite{ge2022long}, {N\"UWA}~\cite{wu2021n}).
% \lu{add the compared method reference here}
%
%Meanwhile, it runs orders-of-magnitude faster than previous methods, enabling real-time neural video synthesis.
%
(b) It is two orders of magnitude faster than diffusion models and 60$\times$ faster than autoregressive models.  %enabling real-time neural video synthesis.
%
%MH: sometimes you say 8+ (here and beginning of introduction) and sometimes 10 (end of introduction), which is which? stick to one number. %Lijun: aligned to 10
(c) A single \modelname{} model accommodates different generation tasks, ranging from class-conditional generation to dynamic inpainting of a moving object.
%References: CCVS~\cite{le2021ccvs}, TATS~\cite{ge2022long}, Video Diffusion~\cite{ho2022video}, N\"UWA~\cite{wu2021n}, RaMViD~\cite{hoppe2022diffusion}.
}
\label{fig:teaser}
\end{figure*}

\section{Motivation}

%brief intro to video creation
%MH: not sure whether we want to say north star or not (which is used internally but not widely in the vision community)
%Guided by the ``north star'' to generate and manipulate video content on-demand, the computer vision community has made substantial progress in the past years~\cite{tulyakov2018mocogan}. 
%Recently, our community is witnessing a paradigm shift from the conventional generative adversarial network (GAN) models~\cite{vondrick2016generating,tulyakov2018mocogan,saito2017temporal,clark2019adversarial,luc2020transformation} to modern models such as vision transformers~\cite{weissenborn2019scaling, rakhimov2021latent, nash2022transframer}.
Recent years have witnessed significant advances in image and video content creation based on learning frameworks ranging from
generative adversarial networks (GANs)~\cite{vondrick2016generating,tulyakov2018mocogan,saito2017temporal,clark2019adversarial,luc2020transformation}, 
%MH: add some diffusion models for image and video generation including stable diffusion models
%Lijun: added image stable diffusion and video diffusion works. Note that stable diffusion (rombach2022high) is the only image generation cited in this whole sentence. @Lu Do we want to only talk about video here or also add image gan and transformers?
%Lu: let's focus on video.
diffusion models~\cite{rombach2022high,ho2022video,voleti2022masked,hoppe2022diffusion,gu2022vector},
to vision transformers~\cite{weissenborn2019scaling, rakhimov2021latent, nash2022transframer}. 
Inspired by the recent success of generative image transformers such as DALL·E~\cite{ramesh2021zero} and other approaches~\cite{esser2021taming,yu2022scaling,chang2022maskgit,ding2021cogview}, we propose an efficient and effective video generation model by leveraging masked token modeling and multi-task learning.

We introduce the MAsked Generative VIdeo Transformer (\emph{\modelname{}}) for multi-task video generation. 
Specifically, we build and train a single \modelname{} model to perform a variety of diverse video generation tasks and demonstrate the model's efficiency, effectiveness, and flexibility against state-of-the-art approaches.
\cref{fig:teaser}(a) shows the quality metrics of \modelname{} on a few benchmarks with efficiency comparisons in (b), and generated examples under different task setups such as frame prediction/interpolation, out/in-painting, and class conditional generation in (c).

%Building upon the image synthesis transformers~\cite{chang2022maskgit,lezama2022improved},
%MH: no need to cite this as it would inadvertently give false impression that this is a "follow-up" work (i.e., little novelty). 
%Following the prior works on image transformers~\cite{chang2022maskgit,lezama2022improved},
%
\modelname{} models a video as a sequence of visual tokens in the latent space and learns to predict masked tokens with BERT~\cite{devlin2019bert}.
%
% There are two main modules in the proposed framework. 
%
% First, we design a 3D quantization model to tokenize a video, with high fidelity, into a low-dimensional spatial-temporal manifold~\cite{yan2021videogpt, ge2022long}.
We adopt the 3D quantization model introduced in \cref{chap:3dvq} and
propose an effective \emph{masked token modeling} (MTM) scheme for multi-task video generation.
Unlike conventional MTM in image understanding~\cite{wang2022image} or image/video synthesis~\cite{chang2022maskgit,gupta2022maskvit,han2022show}, we present an embedding method to model a video condition using a multivariate mask and show its efficacy in training. 

We conduct extensive experiments to demonstrate the quality, efficiency, and flexibility of \modelname{} against state-of-the-art approaches. 
Specifically, we show that \modelname{} performs favorably on two video generation tasks
% , \ie class-conditional generation and frame prediction, on
across three benchmark datasets, including UCF-101~\cite{soomro2012ucf101}, BAIR Robot Pushing~\cite{ebert2017self,unterthiner2018towards}, and Kinetics-600~\cite{carreira2018short}.
%
% significantly improves the state-of-the-art generation quality on three common video benchmarks: UCF-101~\cite{}, BAIR Robot Pushing~\cite{}, and
For the class-conditional generation task on UCF-101, \modelname{}
% outperforms state-of-the-art approaches and 
reduces state-of-the-art  FVD~\cite{unterthiner2018towards} 
from $332$~\cite{ge2022long} to $76$ (\mbox{$\downarrow\!77\%$}). % relative improvement). 
For the frame prediction task, \modelname{} performs best in terms of FVD on BAIR ($84$  \cite{hoppe2022diffusion} $\!\rightarrow\! 62, \downarrow\!26\%)$ and  Kinetics-600 ($16$ \cite{ho2022video}  $\!\rightarrow\! 9.9, \downarrow\!38\%)$.

%HERE
Aside from the visual quality, \modelname{}'s video synthesis is highly efficient. 
For instance, \modelname{} generates a 16-frame 128$\times$128 video clip in $12$ steps, which takes $0.25$ seconds on a single TPUv4i~\cite{jouppi2021ten} device. 
On a V100 GPU, a base variant of \modelname{} runs at $37$ frame-per-second (fps) at 128$\times$128 resolution.
% resolution using mini-batch. \lijun{for gpu mini-batch doesn't affect speed much, the result is measured at batch size = 1}
% \lu{here we should describe the speed for generating a mini-batch.} \lijun{it's measured at batch size = 1}
%
When compared at the same resolution, \modelname{} is two orders of magnitude faster than the video diffusion model~\cite{ho2022video}.
% for generating video at the same resolution. 
%
%For example, video diffusion takes $\sim\!\!256$ steps with a 3D-UNet.
In addition, \modelname{} is $60$ times faster than the autoregressive video transformer~\cite{ge2022long} and $4$-$16$ times more efficient than the contemporary non-autoregressive video transformer~\cite{gupta2022maskvit}.
%
%Compared to the recent non-autoregressive transformers~\cite{gupta2022maskvit}, \modelname{} is $4$-$16$ times more efficient than the 

We show that \modelname{} is flexible and robust for multiple video generation tasks with a single trained model, including frame interpolation, class-conditional frame prediction, inpainting, and outpainting, etc.
In addition, \modelname{} learns to synthesize videos with complex scenes and motion contents from diverse and distinct visual domains, including 
actions with objects~\cite{goyal2017something}, autonomous driving~\cite{caesar2020nuscenes}, and object-centric videos from multiple views~\cite{ahmadyan2021objectron}.
%
%
%including frame interpolation, class-conditional frame prediction, inpainting, and outpainting.
%To illustrate the generalization of \modelname{}, 
%We show generation results on video benchmarks of motions from distinct visual domains including actions with objects~\cite{goyal2017something}, autonomous driving~\cite{caesar2020nuscenes}, and object centric videos of multiple views~\cite{ahmadyan2021objectron}.
%
%MH: it is not necessary to say this in  introduction. 
%It is noteworthy that on the video inpainting and outpainting tasks, \modelname{} is still worse than the best model tailored to each task. %
%
%It is not our intention to compare to these dedicated models but instead to demonstrate a generic model for video generation and manipulation.
%
%Neither is our attention to compare .
%But to demonstrate the 
%reasonable results even though the generic model is not designed for such tasks.
%Admittedly, limitations for example notble gap to image genertaion quality and resolution. On specific tasks of video inpainting 
% Also trains faster when compared to DVD-GAN~\cite{luc2020transformation} autoregressive transformer~\cite{weissenborn2019scaling} which uses up to 512 TPU v4 accelerators.
% Our base model can be conveniently trained using 64 TPU v2 accelerators, and 

The main contributions of this chapter are:
% To our knowledge, this appears to be the first work to embed condition into mask in video synthesis.
\begin{itemize}%[nosep, leftmargin=*]
%MH: sometimes you say 8+ (Figure 1, abstract and intro), and sometimes you say 10 tasks. Stick to one number and make it consistent. 
    %\item To the best of our knowledge, we present the first efficient non-autoregressive transformer for multi-task video generation. We show that a trained model can perform ten tasks at inference time.
    \item To the best of our knowledge, we present the first masked multi-task transformer for efficient video generation and manipulation. We show that a trained model can perform ten different tasks at inference time.
    % \item We introduce a spatial-temporal video quantization model design with high reconstruction fidelity.
    \item We propose an effective embedding method with diverse masks for numerous video generation tasks. 
    \item We show that \modelname{} achieves the best-published fidelity on three widely-used benchmarks, including  UCF-101, BAIR Robot Pushing, and Kinetics-600 datasets.
    % class-conditional generation on and frame prediction on
    %MH: ablation studies should not be counted as "contributions". 
\end{itemize}
%MH: already mentioned in abstract
%We will open source our framework and models to facilitate video generation research.

\section{Prior Work}
% \vspace{-1mm}
\paragraph{GAN-based approaches.}

% Early works on video synthesis employ GAN-based models. 
% Different from image GANs, video GAN research has been focusing on finding the network architecture that can encode temporal dynamics into the generator at scale. 
% For example, VGAN~\cite{vondrick2016generating} proposes a two-stream 3D convolution to disentangle the foreground and background of a scene. 
% TGAN~\cite{saito2017temporal} and MoCoGAN~\cite{tulyakov2018mocogan} decompose motion and content into a sequence of motion codes and a fixed content code. 
% DVD-GAN~\cite{clark2019adversarial} extends the image BigGAN~\cite{brock2018large} with dual discriminators for spatial and temporal guidance.
% DIGAN~\cite{yu2022generating} leveraged an implicit neural representation as the generator output for a dense encoding of the video.
% Despite GANs' early success, the issues with their training instability and lack of generation diversity have limited their performance.
Early success in video synthesis has been made by GAN models~\cite{vondrick2016generating, saito2017temporal, tulyakov2018mocogan, clark2019adversarial, brock2018large, yu2022generating,brooks2022generating,skorokhodov2022stylegan,gupta2022rv,acharya2018towards,kahembwe2020lower,tian2021good}. 
%However, training instability and lack of generation diversity~\cite{chang2022maskgit} have limited their performance.
Training instability and lack of generation diversity~\cite{chang2022maskgit} are known issues of GAN models.
%\lu{do not criticize GAN methods directly.}

% \vspace{-4mm}
\paragraph{Autoregressive transformers.}
Inspired by the success of GPT~\cite{brown2020language}, autoregressive transformers have been adapted for image~\cite{esser2021taming,ramesh2021zero,yu2022scaling,ding2021cogview,chen2020generative} and video generation~\cite{weissenborn2019scaling,babaeizadeh2021fitvid,wu2021n,hong2022cogvideo}.
A focus for video is autoregressive modeling of visual dynamics. Studies have switched from modeling the raw pixels~\cite{oord2016pixelrnn,chen2020generative} to the discrete codes in a latent space~\cite{rakhimov2021latent,yan2021videogpt}.
% lies in the design of the autoregressive element and the complexity brought by the resulting sequence length.
% \cite{weissenborn2019scaling} models a video as a sequence of sub-scaled 3D volumes with block-local attention and reports promising results on two frame prediction benchmarks.
% FitVid~\cite{babaeizadeh2021fitvid} uses frame-level autoregressive LSTM inside a variational auto-encoder to overfit in the pixel level.
%Recently, studies have switched from generation in the raw pixel space to latent spaces~\cite{rakhimov2021latent}.
%For example, VideoGPT~\cite{yan2021videogpt} learns discrete tokens using a VQ-VAE model then learns an autoregressive transformer to model the flattened tokens.
The state-of-the-art model TATS~\cite{ge2022long} uses two hierarchical transformers to reduce the computation for long video generation, with tokens learned by a 3D-VQGAN~\cite{esser2021taming}. 
%that adds a GAN loss and a perceptual loss to the VQ-VAE.
Unlike prior works, we introduce a non-autoregressive transformer with higher efficiency and flexibility.
%Autoregressive approaches demonstrate promising generation quality, but its decoding process becomes a major efficiency drawback for modeling the long video token sequences.

% \vspace{-4mm}
\paragraph{Non-autoregressive transformers.}
% MaskViT~\cite{gupta2022maskvit} adopts a bidirectional transformer for frame prediction, which enjoys the efficiency benefits but trades off the quality.
Concurrently, a few methods use non-autoregressive transformers for image synthesis~\cite{chang2022maskgit,zhang2021m6,lezama2022improved,sohn2022visual,sun2022score}. \cref{sec:magvit_background} reviews a state-of-the-art model called MaskGIT~\cite{chang2022maskgit}. Compared with these approaches~\cite{gupta2022maskvit, han2022show}, we present an embedding mask to model multi-task video conditions with better quality.

%n embedding method to model a video condition using a multivariate mask and verify its efficacy in training.
%for multi-task video generation

% \vspace{-4mm}
\paragraph{Diffusion models.}
Diffusion models have recently received much attention for image synthesis.
For example, the state-of-the-art video diffusion model~\cite{ho2022video} extends the image denoising diffusion model~\cite{sohl2015deep,austin2021structured,song2019generative,tzen2019neural,ho2020denoising} by incorporating 3D U-Net~\cite{cciccek20163d} architectures and joint training on both images and videos.
Despite its high-quality, sampling speed is a bottleneck hindering the application of diffusion models in video synthesis. We show a different solution to train a highly-efficient model that offers compelling quality.  

% \vspace{-4mm}
\paragraph{Multi-task video synthesis.} 
%While multi-task learning is often utilized in recent foundation models~\cite{wang2022image,nash2022transframer}, 
Multi-task video synthesis~\cite{nash2022transframer, han2022show, wu2021n} is yet to be well-studied.  
Transframer~\cite{nash2022transframer} is the closest to our work, which adopts an image-level representation for autoregressive modeling of tasks based on frame prediction. 
%To our knowledge, 
We present an efficient non-autoregressive multi-task transformer, and verify the quality and efficiency on ten video generation tasks.
%we present the first masked multi-task transformer for efficient video generation and manipulation. We show that a trained model can perform ten tasks at inference time.

% \vspace{-4mm}
\paragraph{Text-to-video.} All of our models are trained only on public benchmarks, except the Web video model. We leave the text-to-video task as future work. As shown in recent works
~\cite{ho2022imagen,villegas2022phenaki,singer2022make}, training such models requires large, and sometimes non-public, datasets of paired texts and images. 

% \paragraph{Multi-modal Foundation Model}
% \lijun{@lu}
% \huiwen{I am happy to help with drafting this section. @lijun @lu, what do you think?}
% \lijun{Cool, thanks! Feel free to go ahead. I think we primarily want to introduce some existing multi-task models for comparison here}
% The majority of aforementioned works are task-specific for frame prediction or class-conditional generation. 
% Few works have shown the multi-task generation ability \cite{nash2022transframer}.

\section{Preliminaries: Masked Image Synthesis}\label{sec:magvit_background}

% \subsection{Two-stage Synthesis by Vision Transformer}
% Generally, transformer-based models generate images in two stages~\cite{Ramesh21dalle,esser2021taming,chang2022maskgit}. First, the image is quantized into a grid of discrete tokens by a Vector-Quantized (VQ) autoencoder built upon VAE~\cite{Oord17vqvae,razavi2019generating}, GAN~\cite{esser2021taming}, or vision transformer backbones~\cite{vim2021}, in which each token is represented as an integer index in a codebook. In the second stage, an autoregressive transformer decoder~\cite{chen2020imagegpt} is learned on the flattened token sequence to generate image tokens sequentially based on the previously generated result (\ie autoregressive decoding). In the end, the generated codes are mapped to pixel space using the decoder obtained from the first stage.

%This section briefly reviews the related works on masked image transformers. 
%MH: not a good way to state your mehtod. 
%Our work is built upon the two-stage image synthesis of the non-autoregressive transformer models~\cite{ramesh2021zero,esser2021taming,chang2022maskgit,lezama2022improved}. 
%
%
The proposed video generation framework is based on a two-stage image synthesis process~\cite{ramesh2021zero,esser2021taming} with non-autoregressive transformers~\cite{chang2022maskgit,lezama2022improved}.
In the first stage, an image is quantized and flattened into a sequence of discrete tokens by a Vector-Quantized (VQ) auto-encoder~\cite{van2017neural,esser2021taming,yu2021vector}. 
In the second stage, masked token modeling (MTM) is used to train a transformer model~\cite{gu2022vector,chang2022maskgit} on the tokens.
Let $\rmI \in \sR^{H \times W \times 3}$ be an image and $\rvz \in \sZ^N$ denote the corresponding token sequence of length $N$.

% For instance, 
We take MaskGIT~\cite{chang2022maskgit} as an example.
In the second stage, it applies a binary mask 
% $\rvm \in \{\keep, \mask \}^{N}$ 
$\ervm_i \in \{x\!\rightarrow\!x, x\!\rightarrow\!\mask\}$
to each token to build a corrupted sequence 
% $\overline{\rvz}=\maskfn(\rvz, \rvm)$.
$\overline{\rvz}=\rvm(\rvz)$.
% , where \lu{I do not think we need this equation.}
% \begin{equation}
% \label{eq:apply_mask}
% \overline{\evz}_i = \mathbbm{1}\{m_{i}\,{=}\,\keep\}\evz_i + \mathbbm{1}\{m_{i}\,{\ne}\,\keep\}\evm_i
% \end{equation}
%
Condition inputs, such as class labels, are incorporated as the prefix tokens $\rvc$.
A BERT~\cite{devlin2019bert}  parameterized by $\theta$ is learned to predict the masked tokens in the input sequence $[\rvc, \overline{\rvz}]$, where $[\cdot,\cdot]$ concatenates the sequences.
The objective is to minimize the cross-entropy between the predicted and the ground-truth token at each masked position:
% \vspace{-2mm}
\begin{equation}
\Ls_{\text{mask}}(\rvz; \theta) = \mathop{\E}\limits_{\rvm \sim p_{\gU}} \Big[\sum_{\overline{\ervz}_i = \mask} - \log p_\theta (\ervz_i \mid [\rvc, \overline{\rvz}]) \Big].
\label{eq:loss_mtm}
% \vspace{-2mm}
\end{equation}
During training, MaskGIT randomly samples $\rvm$ 
from a prior distribution $p_{\gU}$ 
where the mask ratio follows a cosine scheduling function $\gamma(\cdot)$~\cite{chang2022maskgit}.
Specifically, it first uniformly samples a per-token mask score $\ervs_i \sim \gU(0, 1)$ to form a sequence denoted as $\rvs$.
Then it samples $r \sim \gU(0, 1)$ and computes a cut-off threshold $s^*$ as the $\lceil \gamma(r)N \rceil$-\textit{th} smallest element in $\rvs$. 
Finally, a mask $\rvm$ is created such that $\ervm_i(x) = \mask$ if $\ervs_i \le s^*$ and $\ervm_i(x) = x$ otherwise.
% , $\forall i\in [1, N]$.

%$\rvs = [\ervs_1, \cdots, \ervs_N] \in \gU(0, 1)^N$.
For inference,
%existing methods typically use the 
the non-autoregressive decoding method~\cite{ghazvininejad2019mask,gu2021fully,kong2021blt} is used to synthesize an image~\cite{chang2022maskgit,lezama2022improved,zhang2021m6}. 
For example, MaskGIT
%transformer $\theta$
generates an image in $K\!\!=\!\!12$ steps~\cite{chang2022maskgit} from a blank canvas with all visual tokens masked out.
At each step, it predicts all tokens in parallel while retaining tokens with the highest prediction scores. 
The remaining tokens are masked and predicted in the next iteration until all tokens are generated. 
Similar to the training stage, the mask ratio is computed by the schedule function $\gamma$, but with a deterministic input as $\gamma(\frac{t}{K})$, where $t$ is the current step.

\begin{sidewaysfigure}[tp]
\centering
\includegraphics[width=\linewidth,trim={0.2cm 0 0 0},clip]{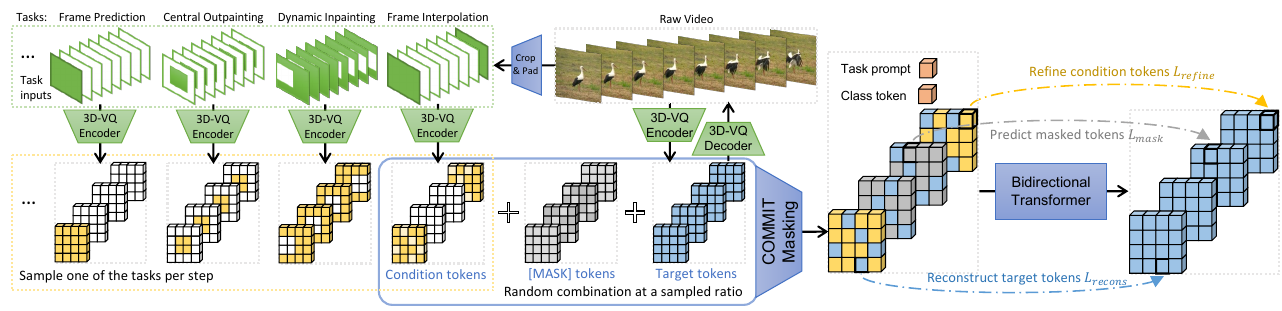}
\caption[\modelname{} pipeline overview]{\textbf{\modelname{} pipeline overview}. The 3D-VQ encoder quantizes a video into discrete tokens, while the 3D-VQ decoder maps them back to the pixel space. 
We sample one of the tasks at each training step and build its condition inputs by cropping and padding the raw video, where \textcolor{ForestGreen}{green} denotes valid pixels and white is padding.
We quantize the condition inputs with the 3D-VQ encoder and select the non-padding part as condition tokens. 
The masked token sequence combines \textcolor{YellowOrange}{condition tokens}, \textcolor{gray}{\mask tokens}, and the \textcolor{RoyalBlue}{target tokens}, with a task prompt and a class token as the prefix. 
The bidirectional transformer learns to predict the target tokens through three objectives: \textcolor{YellowOrange}{refining condition tokens}, \textcolor{gray}{predicting masked tokens}, and \textcolor{RoyalBlue}{reconstructing target tokens}.
% Visual tokens, condition tokens, and \mask are in \textcolor{RoyalBlue}{blue}, \textcolor{YellowOrange}{yellow}, and \textcolor{gray}{gray}, respectively.
}
\label{fig:pipeline}
\end{sidewaysfigure}

\section{MAGVIT: Masked Generative Video Transformer}
% We design 
%
Our goal is to design a multi-task video generation model with high quality and inference efficiency.
We propose MAsked Generative VIdeo Transformer (\modelname{}), a vision transformer framework that leverages masked token modeling and multi-task learning.
% to synthesize video via multi-task masked token modeling.
%
% \modelname{} performs multi-task masked token modeling on quantized video tokens.
\modelname{} generates a video from task-specific condition inputs, such as a frame, a partially-observed video volume, or a class identifier.

The framework consists of two stages.
%Fig.~\ref{fig:pipeline} shows the main steps in the two-stage \modelname{} model.
%MH: no need to say this too many times (give false impression of "follow-up" work with lower novelty)
%following the prior works on image synthesis~\cite{ramesh2021zero,esser2021taming,chang2022maskgit}. 
First, we learn a 3D vector-quantized (VQ) autoencoder to quantize a video into discrete tokens, as introduced in \cref{chap:3dvq}.
In the second stage, we learn a video transformer by multi-task masked token modeling. 

Fig.~\ref{fig:pipeline} illustrates the training in the second stage.
At each training step, we sample one of the tasks with its prompt token,
% a reserved token prompt indicating the task, 
obtain a task-specific conditional mask, and optimize the transformer to predict all target tokens given masked inputs.
% \lu{lijun, please see Yong's comment and add prompt to Fig.2's caption}
%
During inference, we adapt the non-autoregressive decoding method to generate tokens conditionally on the task-specific inputs, which will be detailed in \cref{alg:decoding}.
%
% Finally, the generated tokens are mapped back to the video pixel space by the VQ decoder.

%MH: filler (can remove as we will run out of space)
%The rest of this section details two key modules, \ie an improved VQ video autoencoder (\cref{sec:mth_tok}) and a new way of embedding conditions into the mask (\cref{sec:mth_cmm}).

\subsection{Multi-Task Masked Token Modeling}
\label{sec:mth_cmm}
% In this stage, we learn a bi-directional transformer on the flattened token sequence by MTM. In \cref{sec:mth_mtm} we briefly review the existing MTM with prefix condition. \cref{sec:mth_icmtm} details the proposed MTM method for multi-task training with different condition regions.

% \subsubsection{Background: MTM with Prefix Condition}
% \label{sec:mth_mtm}
% \subsubsection{Interior Conditional MTM}
% \label{sec:mth_icmtm}

%This subsection details a new method for multi-task masked modeling.
%For all video generation tasks, the condition appears as a portion of the target video.
% We consider various video masks
%ing schemes 
%on images 
% as the conditions to facilitate \modelname{} learning for video generation tasks. 
In \modelname{}, we adopt various masking schemes to facilitate learning for video generation tasks with different conditions.
The conditions can be a spatial region for inpainting/outpainting or a few frames for frame prediction/interpolation. 
%
%MH: interior? it is fine but we may find a better term...
We refer to these partially-observed video conditions as \emph{interior conditions}.

We argue that it is suboptimal to directly unmask the tokens corresponding to the region of the interior condition~\cite{chang2022maskgit}. 
%
%This is because the tokens are extracted while observing all spatial-temporal regions in video. Due to the non-local receptive field of convolutions in our 3D tokenizer, the cropped tokens may leak the ground-truth information, leading to a problematic non-causal masking that fails generalization at test time. 
%
As discussed in \cref{sec:mth_tok}, the non-local receptive field of the tokenizer can leak the ground-truth information into the unmasked tokens,
% the cropped tokens can leak the ground-truth information due to the non-local receptive field,
% This is because the tokens are extracted while observing all spatial-temporal regions in the video due to the non-local receptive field in convolutions. 
% Consequently, the cropped tokens can leak the ground-truth information, 
leading to problematic non-causal masking and poor generalization.

We propose a method, COnditional Masked Modeling by Interior Tokens (or \emph{\methodname{}} for short), to embed interior conditions inside the corrupted visual tokens.
%
%MH: filler, remove them as we do not have space
%We will discuss how to use \methodname{} during training in \cref{sec:method_training} and inference in \cref{sec:mth_dec}.
%\methodname{} generalizes the binary mask in MTM~\cite{devlin2019bert} to take multivariate tokens, and transforms the denoising decoding in MTM to a conditional-transition process guided by the multivariate mask.

%by itself is a part of the target video to generate. 
% Prior works use prefix condition which models input conditions as tokens prepended to the visual token sequence.
% However, this may be intractable for modeling the \emph{interior} condition which by itself is a part of the target video to generate. 
% A technical challenge facing \modelname{} to achieve multi-task training is how to model conditions in different tasks. We find that the prefix condition, \ie, prepending condition tokens to the video token sequence, is intractable for modeling . 
% To tackle this issue, \modelname{} proposes a new method that embeds interior conditions into the mask, rendering the non-autoregressive decoding from a denoising process to a new conditional-mask guided generation process.
%This paper uses interior conditions to refer to the condition information that is a part of the video $\rmV$. 
% For interior conditions, prepending the cropped out visual tokens corresponding to the condition region is intractable. 

%\subsubsection{Training}
%\label{sec:method_training}

% \vspace{-4mm}
\paragraph{Training.}
Each training example includes a video $\rmV$ and the optional class annotation $\rvc$.
%the ground-truth annotation for each task and its interior condition, 
The target visual tokens come from the 3D-VQ as \textcolor{RoyalBlue}{$\rvz = f_\gT(\rmV)$}.
% $\rvz \in \sZ^N$ is the flattened visual tokens sequence from $f_\gT(\rmV)$.
At each step, we sample a task prompt $\rho$, obtain the task-specific interior condition pixels, pad it into $\tilde{\rmV}$ with the same shape as $\rmV$, and get the condition tokens \textcolor{YellowOrange}{$\tilde{\rvz} = f_\gT(\tilde{\rmV})$}.
%
% \cref{app:task} 
\cref{sec:vid_task}
lists the padding functions for each task.

At a sampled mark ratio, we randomly replace target tokens $\textcolor{RoyalBlue}{\ervz_i}$, with either
1) the condition token $\textcolor{YellowOrange}{\tilde{\ervz}_i}$, if the corresponding supervoxel of $\textcolor{RoyalBlue}{\ervz_i}$ contains condition pixels; or
2) the special \mask token, otherwise. 
Formally, we compute the \emph{multivariate} conditional mask 
% $\rvm(\widehat{\rmV}; \rvs, s^*) = [\ervm_{i}]_{i=1}^{N}$ 
$\rvm(\cdot \mid \tilde{\rvz})$
as 
% \vspace{-2mm}
% \lu{the below eq appears to be problematic, the definition of m should not depends on $\widehat{\ermZ}_{i, j, k}$. We can a) change it back b) add $\widehat{\ermZ}$ to an input as m.}
\begin{equation}
\ervm(\ervz_i \mid \tilde{\ervz}_i) = 
% x \rightarrow
\begin{cases}
\textcolor{YellowOrange}{\tilde{\ervz}_{i}} & \text{if } \ervs_{i} \le s^* \land \neg \allpadded(\tilde{\ervz}_{i}) \\
\textcolor{gray}{\mask} & \text{if } \ervs_{i} \le s^* \land \allpadded(\tilde{\ervz}_{i}) \\
\textcolor{RoyalBlue}{\ervz_i} &\text{if } \ervs_{i} > s^*
\end{cases},
\label{eq:icmask}
% \vspace{-2mm}
\end{equation}
where $\ervs_{i}$ and $s^*$ are the per-token mask score and the cut-off score introduced in \cref{sec:magvit_background}.
% In addition, $f_\gT$ is the 3D-VQ encoder. 
$\allpadded(\tilde{\ervz}_{i})$ returns whether the corresponding supervoxel of $\tilde{\ervz}_{i}$ in $\tilde{\rmV}$ only contains padding.
% as shown in \cref{fig:pipeline}. 
% The color coding of the three components corresponds to the tokens in \cref{fig:pipeline} and \cref{fig:sample}.

\cref{eq:icmask} indicates that 
\methodname{} embeds interior conditions as corrupted visual tokens into the multivariate mask 
% Therefore, one can no longer sample $\rvm$ from the prior $p_{\gU}$ but from a new distribution $p_\gM$,
$\rvm$, which follows a new distribution $p_\gM$ instead of the prior $p_{\gU}$ for binary masks.
%We sample  $\rvm(\widehat{\rmV}; \rvs, s^*)$ (or $\rvm$ for short)  $\rvm$ from prior $p_\gM$ 
With the corrupted token sequence $\overline{\rvz}=\rvm(\rvz \mid \tilde{\rvz})$ as input,
% \lu{we can skip the flatten part as it is the same as in the discussed MaskGIT.}
% following \cref{eq:apply_mask}.
% For simplification, we use $\rvm$ to denote $\rvm (\widehat{\rmV})$ and sample $\rvm$ from $p_\gV(\widehat{\rmV})$.
%Notice that the mask is integer-valued instead of binary-valued as in the prior work.
%
%
the \emph{multi-task} training objective is
% \vspace{-2mm}
\begin{equation}
\Ls(\rmV; \theta) = \mathop{\E}\limits_{\rho, \hat{\rmV}} \mathop{\E}\limits_{\rvm \sim p_\gM} \Big[\sum_{i} -\log p_\theta (\ervz_i \mid [\rho, \rvc, \overline{\rvz}]) \Big].
\label{eq:mt_loss}
% \vspace{-2mm}
\end{equation}
% where $\rho$ and $\rvc$ are the task prompt and class condition.
%
%both are modeled as prefix conditional tokens. 
We can decompose the loss in \cref{eq:mt_loss} into three parts according to \cref{eq:icmask}: $\Ls_\text{refine}$ refines the task-specific condition tokens, $\Ls_\text{mask}$ predicts masked tokens , and $\Ls_\text{recons}$ reconstructs target tokens.
Let $\overline{\rvc} = [\rho, \rvc, \overline{\rvz}]$ for simplicity,
%\lu{$p_\theta (z_i)$, $z_i$ should be a random variable not a value.}
% \vspace{-4mm}
\begin{equation}
\begin{aligned}
\sum_{i=1}^N - \log p_\theta (\ervz_i \mid [\rho, \rvc,\overline{\rvz}]) =& \underbrace{\sum_{\overline{\ervz}_i = \tilde{\ervz}_i} - \log p_\theta (\ervz_i\mid\overline{\rvc})}_{\text{\textcolor{YellowOrange}{Refine condition tokens $\Ls_\text{refine}$}}} +\underbrace{\sum_{\overline{\ervz}_i = \mask} - \log p_\theta (\ervz_i\mid\overline{\rvc})}_{\text{\textcolor{gray}{Predict masked tokens $\Ls_\text{mask}$}}} \\
&+\underbrace{\sum_{\overline{\ervz}_i = \ervz_i} - \log p_\theta (\ervz_i \mid\overline{\rvc})}_{\text{\textcolor{RoyalBlue}{Reconstruct target tokens $\Ls_\text{recons}$}}} 
% \nonumber
\end{aligned}.
\label{eq:mt_loss_decomposed}
% \vspace{-2mm}
\end{equation}
%\fi
%
% The proposed loss can be decomposed into three parts, \ie, the loss to (1) refine the task-specific condition tokens $\Ls_\text{refine}$, (2) predict masked tokens $\Ls_\text{mask}$, and (3) reconstruct input tokens $\Ls_\text{recons}$. 
%
While $\Ls_\text{mask}$ is the same as the MTM loss in \cref{eq:loss_mtm} and $\Ls_\text{recons}$ sometimes is used as a regularizer (\eg, in NLP tasks), $\Ls_\text{refine}$ is a new component introduced by \methodname{}. 

%We find the \methodname{} masking showing three advantages. 
The  \methodname{} method facilitates multi-task video generation in three aspects. 
First, it provides a correct causal masking for all interior conditions.
Second, it produces a fixed-length sequence for different conditions of arbitrary regional volume, improving training and memory efficiency since no padding tokens are needed.
%
%MH: it is not clear to me what you mean... see the revised sentence 
%Lastly, it obtains better a local-optimal in multi-task video generation. See \cref{sec:exp_abl}.
Third, it achieves state-of-the-art multi-task video generation results (see \cref{tab:cond}).

\begin{algorithm}[tp]
\caption{Non-autoregressive Decoding by \methodname{}}
\label{alg:decoding}
\textbf{Input}: prefix $\rho$ and $\rvc$, condition $\tilde{\rvz}$, steps $K$, temperature $T$ \\
\textbf{Output}: predicted visual tokens $\hat{\rvz}$
\begin{algorithmic}[1]
{\small
% \Require{}
% \Ensure{$\hat{\rvz}$}
\State $\rvs = \vzero$, $s^*=1$, $\hat{\rvz} = \vzero^N$
% \State $\ervm_i \gets f_\gT(\hat{\rmV})_i$ where $\neg \allpadded(\hat{\rmV}, i)$, $\forall i \in [1, N]$
\For {$t \gets 0, 1, \cdots, K - 1$}
\State $\overline{\rvz} \gets \rvm(\hat{\rvz} \mid \tilde{\rvz}; \rvs, s^*)$
\State $\hat{\ervz}_{i} \sim p_{\theta}(\ervz_i \mid [\rho, \rvc, \overline{\rvz}])$, $\forall i$ where $\ervs_i \le s^*$
\State $\ervs_i \gets p_\theta(\hat{\ervz}_i \mid [\rho, \rvc, \overline{\rvz}])$, $\forall i$ where $\ervs_i \le s^*$
\State $\ervs_i \gets \ervs_i + T (1 - \frac{t+1}{K})\gumbel(0, 1)$, $\forall i$ where $\ervs_i < 1$
\State $s^* \gets$ The $\lceil \gamma(\frac{t+1}{K})N \rceil$-\textit{th} smallest value of $\rvs$
\State $\ervs_i \gets 1$, $\forall i$ where $\ervs_i > s^*$
\EndFor \\
\Return $\hat{\rvz} = [\hat{\ervz}_1, \hat{\ervz}_2, \cdots, \hat{\ervz}_N]$
}
\end{algorithmic}
\end{algorithm}

\begin{figure}[tp]
\centering
% \vspace{-1mm}
\begin{subfigure}{0.48\linewidth}
\includegraphics[width=\linewidth,trim={0.25cm 0.2cm 0.1cm 0.2cm},clip]{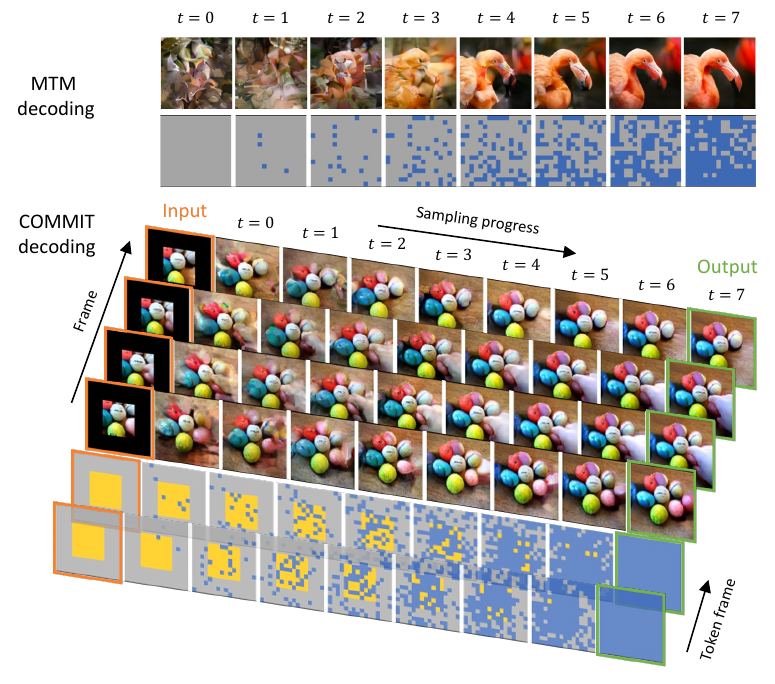}
\end{subfigure}
\begin{subfigure}[b]{0.48\linewidth}
\includegraphics[width=\linewidth,trim={0 0.2cm 0.1cm 0.2cm},clip]{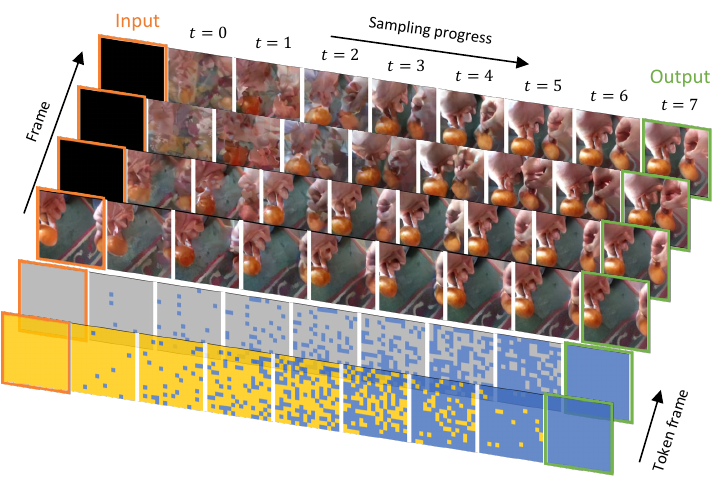}
\end{subfigure}
\caption[Comparison between MTM decoding for image and \methodname{} decoding for video]{\textbf{Comparison between MTM decoding for image~\cite{chang2022maskgit} and \methodname{} decoding for video}. We show the output tokens and image/video at each decoding step $t$, with a central outpainting example for \methodname{} on the left and a frame prediction example on the right.
Unlike the MTM denoising decoding from all \textcolor{gray}{\mask}, \methodname{} performs a conditional generation process toward the \textcolor{RoyalBlue}{output tokens} while gradually replacing the \textcolor{YellowOrange}{interior condition tokens}.
%decoding starts from a multivariate mask that embeds .
% The bottom row shows the input tokens to the transformer at each sampling step, with \textcolor{yellow}{yellow} for condition tokens, \textcolor{gray}{gray} for \mask tokens, and \textcolor{blue}{blue} for decoded tokens.
% The bottom row shows the initial input video and the generated videos at each sampling step. With the task-specific condition inputs, the initial input tokens are built with condition tokens and \mask tokens at $t=0$. At each sampling step, the transformer predicts all tokens in parallel but only the tokens with the highest confidence are preserved into the next step. 
% Visual tokens, condition tokens, and \mask are in \textcolor{SkyBlue}{blue}, \textcolor{YellowOrange}{yellow}, and \textcolor{gray}{gray}, respectively.
Videos and tokens are temporally down-sampled and stacked for visualization.
% See \cref{fig:sample2} for another frame prediction example with COMMIT.
}
\label{fig:sample}
% \vspace{-6mm}
\end{figure}
%
%\subsection{Inference}
%\subsection{Inference: Non-autoregressive Decoding with Interior Condition}
%\label{sec:mth_dec}
% \vspace{-4mm}
\paragraph{Inference.}
%Similar to~\cite{chang2022maskgit}, w
We use a non-autoregressive decoding method to generate video tokens from input conditions in $K$ steps (\eg, $12$).
Each decoding step follows the \methodname{} masking in \cref{eq:icmask} with a gradually reduced mask ratio.
\cref{alg:decoding} outlines the inference procedure.

%We adapt non-autoregressive (NAR) modeling over autoregressive (AR) for its advantages in speed and flexibility in multi-task generation.
% NAR transformers~\cite{chang2022maskgit,lezama2022improved} show faster inference than their AR counterparts (\eg, \cite{esser2021taming}) while offering on-par or superior fidelity and diversity. 

% The traditional NAR decoding starts from a blank canvas with all tokens masked out and generate the full sequence in $K$ steps . 
% In each step, it predicts all tokens in parallel while only retaining the ones with the highest confidence scores. 
% The remaining tokens are masked out and will be predicted in the next iteration until all tokens are generated.

\cref{fig:sample} compares the non-autoregressive image decoding~\cite{chang2022maskgit} and our video decoding procedure.
Different from the MTM decoding in~\cite{chang2022maskgit} which performs denoising from all \mask, \methodname{} decoding starts from a \emph{multivariate} mask that embeds the \textcolor{YellowOrange}{interior conditions}.
Guided by this mask, \cref{alg:decoding} performs a conditional transition process toward the output tokens by replacing a portion of newly generated tokens at each step.
In the end, all tokens are predicted where the interior condition tokens get refined. 
%to improve the consistency with other tokens in the sequence.
% When masking the low-confidence tokens from the previous step, corresponding interior condition tokens are put back.

% \lu{explain why it transforms the denoising decoding in MTM to a conditional-transition process guided by the multivariate mask. Refer to the figure.}

% For example, MaskGIT's parallel decoding algorithm with the MTM-trained transformer executes as:
% \begin{algorithmic}[1]
% {\small
% \Require{$T$, $\rvc$, $\rvs = \vzero$, $\rvm = \mask^N$, $\hat{\rvz} = \varnothing^N$}
% \For {$t \gets 1, 2, \cdots, T$}
% \State $\overline{\rvz} \gets \text{mask}_\text{MTM}(\hat{\rvz}, \rvm)$
% \State $\hat{\ervz}_{i} \sim p_{\theta}(\ervz_i \mid [\rvc, \overline{\rvz}])$, $\forall i$ where $\ervm_i= \mask$
% \State $\ervs_i \gets p_\theta(\hat{\ervz}_i \mid [\rvc, \overline{\rvz}])$, $\forall i$ where $\ervm_i= \mask$
% \State $s^* \gets$ The $\lceil \gamma(\frac{t}{T})N \rceil$th smallest value of $\rvs$
% \State $\ervm_i \gets \varnothing$, $\ervs_i \gets 1$, $\forall i$ where $\ervs_i > s^*$
% \EndFor
% }
% \end{algorithmic}
\subsection{Video Generation Tasks} 
\label{sec:vid_task}
% We consider \emph{ten} tasks for multi-task video generation where each task has a different interior condition and mask:
% Frame Prediction (FP), 
% Frame Interpolation (FI),
% Central Outpainting (OPC),
% Vertical Outpainting (OPV),
% Horizontal Outpainting (OPH),
% Dynamic Outpainting (OPD),
% Central Inpainting (IPC),
% and Dynamic Inpainting (IPD),
% Class-conditional Generation (CG), 
% Class-conditional Frame Prediction (CFP). 
% %MH: I do not think we have space to use bullets. 
% %\begin{itemize}[nosep]
% %\item Class-conditional Generation (CG)
% %\item Frame Prediction (FP)
% %\item Class-conditional Frame Prediction (CFP)
% %\item Frame Interpolation (FI)
% %\item Central Outpainting (OPC)
% %\item Vertical Outpainting (OPV)
% %\item Horizontal Outpainting (OPH)
% %\item Dynamic Outpainting (OPD)
% %\item Central Inpainting (IPC)
% %\item Dynamic Inpainting (IPD)
% %\end{itemize}
% % We will discuss them in \cref{tab:task}
% We provide the detailed definitions in Appendix B.1.
% % \cref{app:task}.
We employ a total of ten tasks for multi-task video generation. Each task is characterized by a few adjustable settings such as interior condition shape, padding function, and optionally prefix condition. \cref{fig:task} illustrates the interior condition regions for each task under the above setup.
Given a video of shape $T\times H \times W$, we define the tasks as following:

\begin{figure}[tp]
\centering
\begin{subfigure}[t]{0.19\linewidth}
\includegraphics[width=\linewidth]{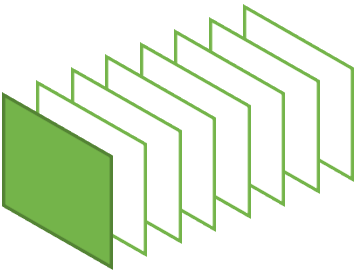}
\caption{FP}
\end{subfigure}
\begin{subfigure}[t]{0.19\linewidth}
\includegraphics[width=\linewidth]{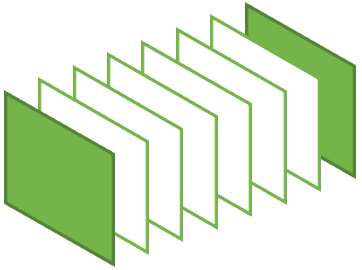}
\caption{FI}
\end{subfigure}
\begin{subfigure}[t]{0.19\linewidth}
\includegraphics[width=\linewidth]{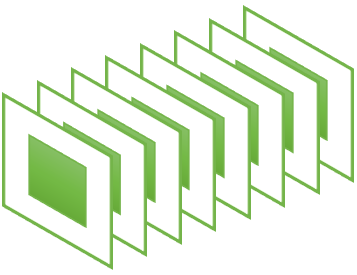}
\caption{OPC}
\end{subfigure}
\begin{subfigure}[t]{0.19\linewidth}
\includegraphics[width=\linewidth]{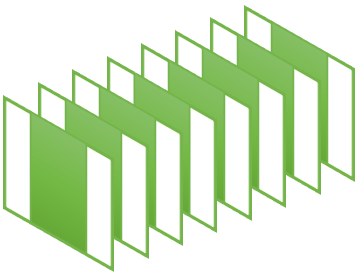}
\caption{OPV}
\end{subfigure}
\begin{subfigure}[t]{0.19\linewidth}
\includegraphics[width=\linewidth]{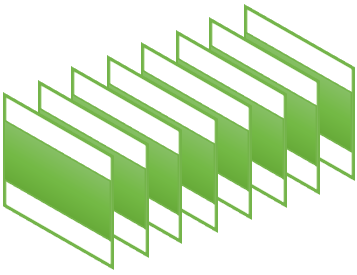}
\caption{OPH}
\end{subfigure}
\begin{subfigure}[t]{0.19\linewidth}
\includegraphics[width=\linewidth]{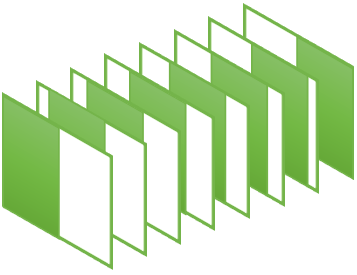}
\caption{OPD}
\end{subfigure}
\begin{subfigure}[t]{0.19\linewidth}
\includegraphics[width=\linewidth]{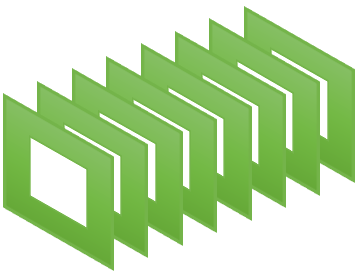}
\caption{IPC}
\end{subfigure}
\begin{subfigure}[t]{0.19\linewidth}
\includegraphics[width=\linewidth]{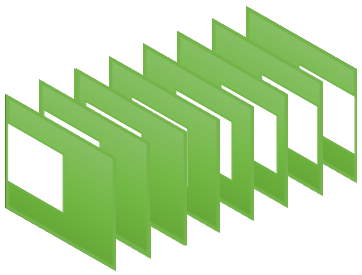}
\caption{IPD}
\end{subfigure}
\begin{subfigure}[t]{0.19\linewidth}
\includegraphics[width=\linewidth]{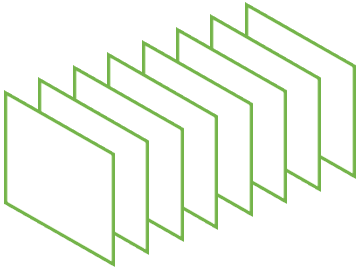}
\caption{CG}
\end{subfigure}
\begin{subfigure}[t]{0.19\linewidth}
\includegraphics[width=\linewidth]{fp_vid}
\caption{CFP}
\end{subfigure}
\caption[Interior condition regions for each task]{\textbf{Interior condition regions for each task}, where \textcolor{ForestGreen}{green} denotes valid pixels and white pixels denote the task-specific paddings discussed in \cref{sec:vid_task}. The tasks are Frame Prediction (FP), Frame Interpolation (FI), Central Outpainting (OPC), Vertical Outpainting (OPV), Horizontal Outpainting (OPH), Dynamic Outpainting (OPD), Central Inpainting (IPC), Dynamic Inpainting (IPD), Class-conditional Generation (CG), and Class-conditional Frame Prediction (CFP).}
\label{fig:task}
\end{figure}

\begin{itemize}
\item Frame Prediction (FP)
\begin{itemize}[nosep]
    \item Interior condition: $t$ frames at the beginning; $t=1$.
    \item Padding: replicate the last given frame.
\end{itemize}
\item Frame Interpolation (FI)
\begin{itemize}[nosep]
    \item Interior condition: $t_1$ frames at the beginning and $t_2$ frames at the end; $t_1=1$, $t_2=1$.
    \item Padding: linear interpolate between the last given frame at the beginning and the first given frame at the end.
\end{itemize}
\item Central Outpainting (OPC)
\begin{itemize}[nosep]
    \item Interior condition: a rectangle at the center with height $h$ and width $w$; $h=0.5H$, $w=0.5W$.
    \item Padding: pad the nearest pixel for each location (\texttt{edge} padding).
\end{itemize}
\item Vertical Outpainting (OPV)
\begin{itemize}[nosep]
    \item Interior condition: a centered vertical strip with width $w$; $w=0.5W$.
    \item Padding: \texttt{edge} padding.
\end{itemize}
\item Horizontal Outpainting (OPH)
\begin{itemize}[nosep]
    \item Interior condition: a centered horizontal strip with height $h$; $h=0.5H$.
    \item Padding: \texttt{edge} padding.
\end{itemize}
\item Dynamic Outpainting (OPD)
\begin{itemize}[nosep]
    \item Interior condition: a moving vertical strip with width $w$; $w=0.5W$.
    \item Direction of movement: left to right.
    \item Padding: zero padding.
\end{itemize}
\item Central Inpainting (IPC)
\begin{itemize}[nosep]
    \item Interior condition: everything but a rectangle at the center with height $h$ and width $w$; $h=0.5H$, $w=0.5W$.
    \item Padding: zero padding.
\end{itemize}
\item Dynamic Inpainting (IPD)
\begin{itemize}[nosep]
    \item Interior condition: everything but a vertically centered moving rectangle with height $h$ and width $w$; $h=0.5H$, $w=0.5W$.
    \item Direction of movement: left to right.
    \item Padding: zero padding.
\end{itemize}
\item Class-conditional Generation (CG)
\begin{itemize}[nosep]
    \item Prefix condition: class label.
\end{itemize}
\item Class-conditional Frame Prediction (CFP)
\begin{itemize}[nosep]
    \item Prefix condition: class label.
    \item Interior condition:  $t$ frames at the beginning; $t=1$.
    \item Padding: replicate the last given frame.
\end{itemize}
\end{itemize}

\section{Experimental Results}

We conduct extensive experiments to demonstrate the video generation quality (\cref{sec:exp_quality}), efficiency (\cref{sec:exp_efficiency}), and flexibility for multi-task generation (\cref{sec:exp_flexibility}).
% In \cref{sec:exp_quality}, we evaluate the generation fidelity on two tasks and three public benchmarks and report notable improvements (up to 77\%) over the previous best results.
% We compare the inference efficiency with existing models in \cref{sec:exp_efficiency}, where our method has an orders-of-magnitude faster generation speed than autoregressive transformers and diffusion models.
% In \cref{sec:exp_flexibility}, we demonstrate the multi-tasking flexibility on two datasets with our proposed set of tasks.
% A single \modelname{} model is capable of all the tasks at good quality.
We show a few generation results here, and refer to the
web page \url{https://magvit.cs.cmu.edu}
for more examples. 
% The source code and trained models will be released.

\subsection{Experimental Setups}

\begin{table*}[tp]
\centering
\begin{tabular}{ccccccc}
\toprule
Model     & Param. & \# heads & \# layers & Hidden size & MLP dim \\
\midrule
\modelname-\textbf{B} & 87 M          & 12     & 12     & 768         & 3072    \\
\modelname-\textbf{L} & 305 M         & 16     & 24     & 1024        & 4096    \\
\modelname-\textbf{H} & 634 M         & 16     & 32     & 1280        & 5120    \\
% \modelname-\textbf{g} & 1.01B         & 16     & 40     & 1408        & 6144  \\
% \modelname-\textbf{G} & 1.85B         & 16     & 48     & 1664        & 8192  \\
% \modelname-\textbf{e} & 2.5B          & 16     & 56     & 1792        & 15360  \\
% \midrule
% VideoULM-\textbf{1B} &1B&WIP&&\\
% VideoULM-\textbf{3B} &3B&WIP&&\\
% VideoULM-\textbf{64B} &64B&WIP&&\\
\bottomrule
\end{tabular}
% \vspace{-2mm}
\caption[Transformer architecture configurations used in \modelname{}]{\textbf{Transformer architecture configurations used in \modelname{}}.}
\label{tab:trans_arch}
% \vspace{-3mm}
\end{table*}

\paragraph{Datasets.}
We evaluate the single-task video generation performance of \modelname{} on three standard benchmarks, \ie, class-conditional generation on UCF-101~\cite{soomro2012ucf101} and frame prediction on BAIR Robot Pushing~\cite{ebert2017self, unterthiner2018towards} (1-frame condition) and Kinetics-600~\cite{carreira2018short} (5-frame condition).
For multi-task video generation, we quantitatively evaluate \modelname{} on BAIR and SSv2~\cite{goyal2017something} on 8-10 tasks.
Furthermore, to evaluate model generalizability, we train models with the same learning recipe on three additional video datasets: nuScenes~\cite{caesar2020nuscenes}, Objectron~\cite{ahmadyan2021objectron}, and 12M Web videos. 
% We show their generated videos in the main paper and quantitative performance in Appendix C.
% \cref{app:adata}. 
% \lu{remember to show their quantitative in Appendix.}

%UCF-101~\cite{soomro2012ucf101} dataset to evaluate the CG performance.

% \vspace{-4mm}
\paragraph{Evaluation metrics.}
The FVD~\cite{unterthiner2018towards} is used as the primary evaluation metric. 
We follow the official implementation in extracting video features with an I3D model trained on Kinetics-400~\cite{carreira2017quo} (\url{https://github.com/google-research/google-research/tree/master/frechet_video_distance}).
We report Inception Score (IS)~\cite{saito2020train} on the UCF-101 dataset which is calculated with a C3D~\cite{tran2015learning} model trained on UCF-101 (\url{https://github.com/pfnet-research/tgan2}). 
We further include image quality metrics: PSNR, SSIM~\cite{wang2004image} and LPIPS~\cite{zhang2018unreasonable} (computed by the VGG features) on the BAIR dataset.
% We use FVD~\cite{unterthiner2018towards} as our primary evaluation metric. 
% %
% Similar to~\cite{ge2022long, ho2022video}, FVD features are extracted with an I3D model trained on Kinetics-400~\cite{carreira2017quo}. 
% %We sample videos from the dataset without replacement to form the real distribution in FVD and use them as the condition inputs for the model. In addition, 
% We also report the Inception Score (IS)~\cite{saito2020train} 
% % on the UCF-101 dataset 
% calculated with a C3D~\cite{tran2015learning} model on UCF-101, and PSNR, SSIM~\cite{wang2004image}, and LPIPS~\cite{zhang2018unreasonable} on BAIR. 
% %
% We report the mean and standard deviation for each metric calculated over four runs. 
%We further includes image quality metrics: PSNR, SSIM~\cite{wang2004image} and LPIPS~\cite{zhang2018unreasonable} on the BAIR dataset.

% We use Fr\'echet Video Distance (FVD)~\cite{unterthiner2018towards} as the major evaluation metric as it captures motion feature and has a high relevance with human evaluation.
% The feature for FVD calculation is extracted with an I3D model trained on Kinetics-400~\cite{carreira2017quo}.

% The ablation study also reports image quality metrics including Peak Signal-to-noise Ratio (PSNR), Structural Similarity Index Measure (SSIM)~\cite{wang2004image} and Learned Perceptual Image Patch Similarity (LPIPS)~\cite{zhang2018unreasonable}.

% \vspace{-4mm}
\paragraph{Model configurations.}
% \modelname{} uses the BERT transformer architecture~\cite{devlin2019bert} .
% in base (B) and large (L) configurations, totaling 87M and 306M parameters, respectively.
%without segment embeddings. 
% The base model \modelname{}-B is a 12-layer transformer with 12 heads and 768 hidden size, totaling 78M parameters. The large model \modelname{}-L has 300M parameters. See \cref{app:train} for details and hyperparameters.
% The input sequence 

We train \modelname{} to generate 16-frame videos at 128$\times$128 resolution, except for BAIR at 64$\times$64.
The proposed 3D-VQ model quantizes a video into 4$\times$16$\times$16 visual tokens,
%into $4\times16\times16$ tokens, which by default downsamples 
% by a factor of 4 temporally and by 8$\times$8 spatially.
where the visual codebook size is 1024.
%
% \lu{modify this part. include a short version here and a full version in the appendix}
%
% \lu{talk about padding functions}
% % \vspace{-4mm}
% \paragraph{Models.}
%MH: try not to say "improved" as it minimizes the novelty and significant of your work. Fix this in the entire text.
%We train the improved 3D-VQGAN model discussed in \cref{sec:mth_tok}
% We train the proposed 3D-VQ model from \cref{sec:mth_tok} as the video tokenizer.
%
% Unless specified otherwise, we compress a video by a factor of 4 temporally and by 8$\times$8 spatially, \eg, a 16-frame video of 128$\times$128 resolution is quantized to 4$\times$16$\times$16 tokens. 
%
% More model architectures are detailed in \cref{app:arch}.
%
We use the BERT transformer~\cite{devlin2019bert} adapted from the
Flaxformer implementation (\url{https://github.com/google/flaxformer}) to model the token sequence, which 
% in base (B) and large (L) configurations, totaling 87M and 306M parameters, respectively.
%without segment embeddings. 
% The base model \modelname{}-B is a 12-layer transformer with 12 heads and 768 hidden size, totaling 78M parameters. The large model \modelname{}-L has 300M parameters. See \cref{app:train} for details and hyperparameters.
% The input sequence 
includes 1 task prompt, 1 class token, and 1024 visual tokens.
%
% Two variants of \modelname{}, \ie, base (B) with 128M parameters and large (L) with 464M, are evaluated. 
Following the transformer configurations in ViT ~\cite{dosovitskiy2020image}, we use two variants of transformers, \ie, base (B) with 87M parameters and large (L) with 306M in all our experiments. \cref{tab:trans_arch} lists the detailed configurations for each variant.
A huge (H) transformer is only used to train on the large Web video dataset and generate demo videos.
We train both stages with the Adam optimizer~\cite{kingma2014adam} in JAX/Flax~\cite{jax2018github,flax2020github} on TPUs.
% \cref{app:arch,app:train}
\cref{sec:magvit_impl} details training configurations.

% Training hyperparameters are provided in \cref{app:train}.
% TPUv2s for the base model and TPUv4s for the large model.
% See \cref{app:bert_arch} for more about transformer configurations.
% Baseline models with prefix condition also have condition tokens which results in a longer sequence.

% \paragraph{Scale}
% In the following experiments, we use two variants of \modelname{}, namely base (B) and large (L).
% Both the tokenizer and the transformer are scaled accordingly.

\subsection{Single-Task Video Generation}
\label{sec:exp_quality}
%We train \modelname{} with a single task to compare with prior works on standard benchmarks.
%% \vspace{-4mm}
\begin{table}[tp]
\centering
% \scriptsize
\begin{tabular}{@{}lcccc@{}} %{@{}l@{\hspace{4pt}}c@{\hspace{4pt}}c@{\hspace{4pt}}c@{\hspace{5pt}}c@{}}
\toprule
Method         & \makecell{Extra\\Video}  & Class      & FVD$\downarrow$          & IS$\uparrow$                \\ \midrule
VGAN\cite{vondrick2016generating} & & \checkmark & -                        & 8.31\mytiny{$\pm$0.09}  \\
TGAN\cite{saito2017temporal}& &            & -                        & 11.85\mytiny{$\pm$0.07} \\
MoCoGAN$^*$\cite{tulyakov2018mocogan}& & \checkmark & -                        & 12.42\mytiny{$\pm$0.07} \\
ProgressiveVGAN\cite{acharya2018towards}&& \checkmark & -                        & 14.56\mytiny{$\pm$0.05} \\
TGAN\cite{saito2017temporal}& & \checkmark & -                        & 15.83\mytiny{$\pm$0.18} \\
RaMViD\cite{hoppe2022diffusion}   & &    &   -  &  21.71\mytiny{$\pm$0.21}  \\
LDVD-GAN\cite{kahembwe2020lower} &&            & -                        & 22.91\mytiny{$\pm$0.19} \\
StyleGAN-V$^{*\#}$\cite{skorokhodov2022stylegan}& &  & - & 23.94\mytiny{$\pm$0.73} \\
VideoGPT\cite{yan2021videogpt}& &            & -                        & 24.69\mytiny{$\pm$0.30} \\
TGANv2\cite{saito2020train}& & \checkmark & 1209\mytiny{$\pm$28} & 28.87\mytiny{$\pm$0.67} \\
MoCoGAN-HD$^\#$\cite{tian2021good}& &            & 838                      & 32.36                       \\
DIGAN\cite{yu2022generating} &&            & 655\mytiny{$\pm$22}  & 29.71\mytiny{$\pm$0.53} \\

DIGAN$^\#$\cite{yu2022generating}& &            & 577\mytiny{$\pm$21}  & 32.70\mytiny{$\pm$0.35} \\
DVD-GAN$^\#$\cite{clark2019adversarial}& & \checkmark & -                        & 32.97\mytiny{$\pm$1.70}  \\
Video Diffusion$^{*\#}$\cite{ho2022video}&&            & -                        & 57.00\mytiny{$\pm$0.62}    \\
TATS\cite{ge2022long} & &            & 420\mytiny{$\pm$18}  & 57.63\mytiny{$\pm$0.24} \\
CCVS+StyleGAN$^\#$\cite{le2021ccvs} &&            & 386\mytiny{$\pm$15}  & 24.47\mytiny{$\pm$0.13} \\
Make-A-Video$^*$\cite{singer2022make} & & \checkmark & 367 & 33.00 \\
TATS\cite{ge2022long} && \checkmark & 332\mytiny{$\pm$18}  & 79.28\mytiny{$\pm$0.38} \\ \hdashline
\color{gray}CogVideo$^*$\cite{hong2022cogvideo} & \color{gray}\checkmark  & \color{gray}\checkmark & \color{gray}626                      & \color{gray}50.46                       \\
\color{gray}Make-A-Video$^*$\cite{singer2022make} & \color{gray}\checkmark & \color{gray}\checkmark & \color{gray}81 & \color{gray}82.55 \\ \midrule
\emph{\modelname}-B-CG (ours) & & \checkmark & \underline{159\mytiny{$\pm$2}} & \underline{83.55\mytiny{$\pm$0.14}}     \\
\emph{\modelname}-L-CG (ours) & & \checkmark & \textbf{76\mytiny{$\pm$2}}     & \textbf{89.27\mytiny{$\pm$0.15}}     \\ \bottomrule
\end{tabular}
\caption[Generation performance on the UCF-101 dataset]{\textbf{Generation performance on the UCF-101 dataset}. 
Methods in \textcolor{gray}{gray} are pretrained on additional large video data.
Methods with $\checkmark$ in the Class column are class-conditional, while the others are unconditional. Methods marked with $^*$ use custom resolutions, while the others are at 128$\times$128.
Methods marked with $^\#$ additionally used the test set in training. }
\label{tab:ucf_full}
\end{table}
\paragraph{Class-conditional generation.}

\begin{figure*}[tp]
\centering
\begin{subfigure}[t]{\linewidth}
\centering
\includegraphics[width=0.75\linewidth]{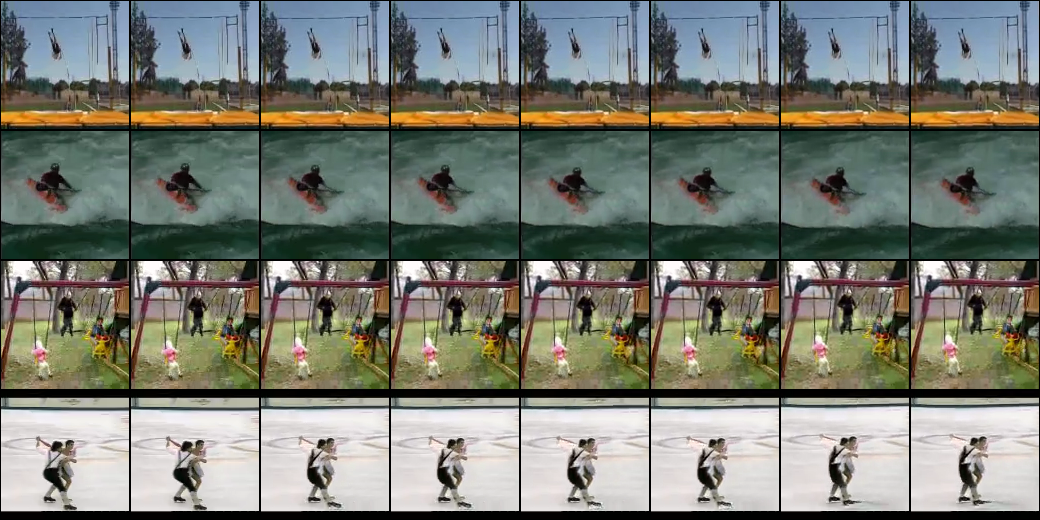}
\caption{CCVS+StyleGAN~\cite{le2021ccvs}
}
\end{subfigure}
\hfill
\begin{subfigure}[t]{\linewidth}
\centering
\includegraphics[width=0.75\linewidth]{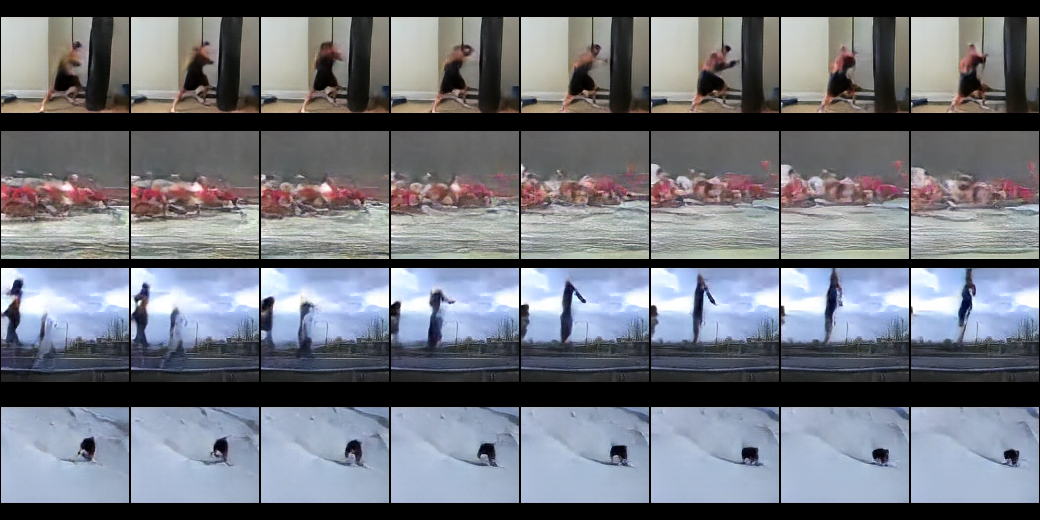}
\caption{TATS~\cite{ge2022long}
}
\end{subfigure}
\hfill
\begin{subfigure}[t]{\linewidth}
\centering
\includegraphics[width=0.75\linewidth]{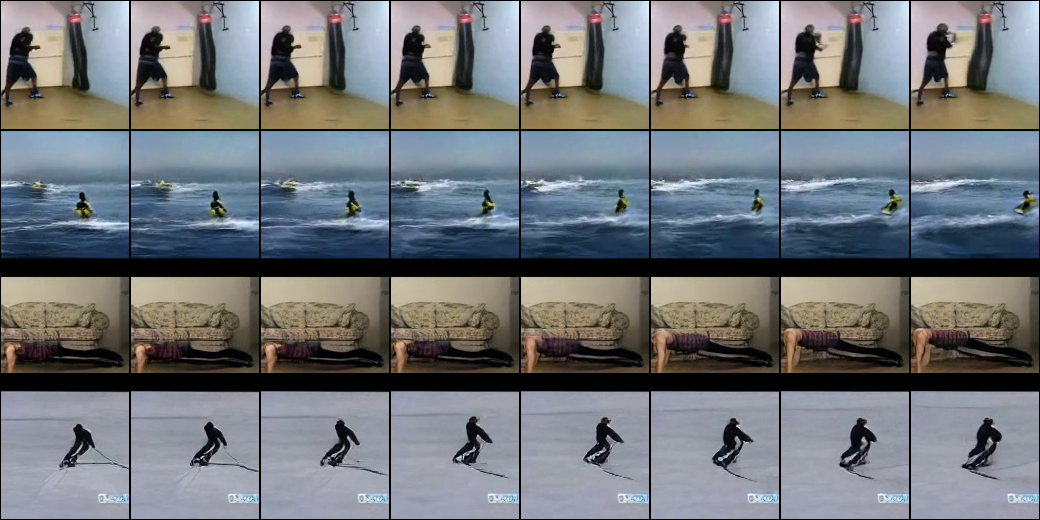}
\caption{\modelname{}-L-CG (ours)
}
\end{subfigure}
\caption[Comparison of class-conditional generation samples on UCF-101]{\textbf{Comparison of class-conditional generation samples on UCF-101}. 16-frame videos are generated at 128$\times$128 resolution 25 fps and shown at 12.5 fps. Samples for \cite{le2021ccvs,ge2022long} are obtained from their official release (\url{https://songweige.github.io/projects/tats/}).
% CCVS+StyleGAN gets a decent single-frame quality attributing to the pretrained StyleGAN, 
% but yields little or no motion.
% TATS generates some motion but with clear artifacts.
% In contrast, our model produces higher-quality frames with substantial motion.
}
\label{fig:ucf_sample}
\end{figure*}

The model is given a class identifier in this task to generate the full video.
% We adopt the broadly used UCF-101~\cite{soomro2012ucf101} dataset to evaluate the CG performance.
% UCF-101 consists of 101 classes and 9.5k videos for training and 3.8k videos for evaluation.
% The benchmark evaluates $16 \times 128 \times 128$ videos and calculates FVD with 10k samples against the training set.
% \cref{tab:ucf_full} shows that \modelname{} surpasses the previous best-published FVD and IS scores. Notably, it outperforms Make-A-Video~\cite{singer2022make} which is pretrained on additional 10M videos with a text-image prior. In contrast, \modelname{} is just trained on the 9.5K training videos of UCF-101.
\cref{tab:ucf_full} shows a detailed comparison with the previously published results on the UCF-101~\cite{soomro2012ucf101} class-conditional video generation benchmark, where the numbers are quoted from the cited papers. 
Note that CogVideo~\cite{hong2022cogvideo} and Make-A-Video~\cite{singer2022make} are pretrained on additional 5-10M videos before finetuning on UCF-101, where Make-A-Video further uses a text-image prior trained on a billion text-image pairs.
The remaining models, including \modelname{}, are only trained on 9.5K training videos of UCF-101, or 13.3K training and testing videos of UCF-101 for those marked with $^\#$. 
% \cref{fig:ucf_sample} provides a visual comparison to the baseline methods.
%even with this base setup. 
%Since no interior condition is given, our training objective is the same as the conventional MTM~\cite{chang2022maskgit} and the result verifies the quality of our 3D tokenizer.
% Our model improves the FVD metric by 77\%, setting a new state-of-the-art at 76 from previous best 332.
% The results validate our framework and verify the quality of our 3D tokenizer.
\cref{fig:ucf_sample} compares the generated videos to baseline models.
%Generated examples by \modelname{} and leading prior models are shown in \cref{fig:ucf}.
We can see that 
CCVS+StyleGAN~\cite{le2021ccvs} gets a decent single-frame quality,
%attributing to the pretrained StyleGAN, 
but yields little or no motion.
TATS~\cite{ge2022long} generates some motion but with artifacts.
In contrast, our model produces higher-quality frames with substantial motion.
% It is worth noting that the quality of these samples is highly limited by the small data size of UCF-101.
%We also increase IS by 13\% to 89.27, which gets very close to the IS of real data 91.36.

% \vspace{-4mm}
\paragraph{Frame prediction.}
\begin{table}[tp]
\centering
% \scriptsize
\begin{tabular}{@{}lcc@{}}
\toprule
Method                   & K600 FVD$\downarrow$   & BAIR FVD$\downarrow$ \\ \midrule
LVT\cite{rakhimov2021latent} & 224.7                      & 126\mytiny{$\pm$3} \\
Video Transformer\cite{weissenborn2019scaling} & 170.0\mytiny{$\pm$5.0} & 94\mytiny{$\pm$2}  \\
CogVideo$^*$\cite{hong2022cogvideo}  & 109.2                      & -                      \\
DVD-GAN-FP\cite{clark2019adversarial} & 69.1\mytiny{$\pm$1.2}  & 110                    \\
CCVS\cite{le2021ccvs}    & 55.0\mytiny{$\pm$1.0}  & 99\mytiny{$\pm$2}  \\
Phenaki\cite{villegas2022phenaki} & 36.4\mytiny{$\pm$0.2} & 97 \\
VideoGPT\cite{yan2021videogpt}  & -                          & 103                    \\
TrIVD-GAN-FP\cite{luc2020transformation} & 25.7\mytiny{$\pm$0.7}  & 103                    \\
Transframer\cite{nash2022transframer}  & 25.4                       & 100                    \\
MaskViT\cite{gupta2022maskvit} & -                          & 94                     \\
FitVid\cite{babaeizadeh2021fitvid} & -                          & 94                     \\
MCVD\cite{voleti2022masked} & - & 90 \\
N\"UWA\cite{wu2021n}    & -                          & 87                     \\
RaMViD\cite{hoppe2022diffusion}        &  16.5     &   84 \\
Video Diffusion\cite{ho2022video}  & \underline{16.2\mytiny{$\pm$0.3}}  & -                     \\ \midrule
\emph{\modelname}-B-FP (ours)          & 24.5\mytiny{$\pm$0.9}     & \underline{76{\mytiny$\pm$0.1}} (47{\mytiny$\pm$0.1})   \\
\emph{\modelname}-L-FP (ours)         & \textbf{9.9\mytiny{$\pm$0.3}}     & \textbf{62{\mytiny$\pm$0.1}}  (31{\mytiny$\pm$0.2})   \\ \bottomrule
\end{tabular}
\caption[Frame prediction performance on the BAIR and Kinetics-600 datasets]{\textbf{Frame prediction performance on the BAIR and Kinetics-600 datasets}.
%Values marked with $^*$ use the new evaluation protocol on BAIR with uniformly sampled test clips. 
- marks that the value is unavailable in their paper or incomparable to others.
The FVD in parentheses uses a debiased evaluation protocol on BAIR detailed in \cref{sec:magvit_impl}.
Methods marked with $^*$ is pretrained on additional large video data.
% that uniformly samples test clips.
%The best values for each metric are in bold, while the second best are in italic.
}
\label{tab:bairk600_full}
\end{table}

\begin{table}[tp]
\centering

%\vspace{-3mm}
% \scriptsize
\begin{tabular}{@{}lcccccc@{}}
\toprule
Method     & FVD$\downarrow$  & PSNR$\uparrow$ & SSIM$\uparrow$  & LPIPS$\downarrow$ \\ \midrule
CCVS\cite{le2021ccvs} & 99  & - & 0.729 & -  \\
MCVD\cite{voleti2022masked} & 90 & 16.9 & 0.780 & -       \\ \midrule
% \emph{\modelname}-B-FP (ours) &   &      &  /   \\
\emph{\modelname}-L-FP (ours) & \textbf{62} & \textbf{19.3}  &   \textbf{0.787}   & 0.123  \\ \bottomrule
\end{tabular}
% \vspace{-1mm}
\caption[Image quality metrics on BAIR frame prediction]{\textbf{Image quality metrics on BAIR frame prediction}. 
% \lu{is SSIM number correct?}
% \lu{remove the star and put the detail into the Appendix}$^*$ generates at $256 \times 256$ while the others are at $64 \times 64$.
}
\label{tab:im_quality}
% \vspace{-5mm}
\end{table}

The model is given a single or a few frames to generate future frames.
% \lu{the replication should be covered in the padding right?}
% The last condition frame is replicated to form the condition video.
%
In \cref{tab:bairk600_full}, we compare \modelname{} against highly-competitive baselines. 
\modelname{} surpasses the previous state-of-the-art FVD on BAIR by a large margin ($84 \xrightarrow[]{} 62$). Inspired by~\cite{unterthiner2018towards}, a ``debiased'' FVD is also reported in the parentheses to overcome the small validation set. See more discussion in \cref{sec:magvit_impl}.
% \cref{app:bair}.
%we also report FVD under a different evaluation protocol aiming at debiasing the small validation set. See more discussions in \cref{app:bair}.
In \cref{tab:im_quality}, it demonstrates better image quality.
%than previous methods.

\begin{figure*}[tp]
\centering
\begin{subfigure}[t]{0.8\linewidth}
\includegraphics[width=\linewidth]{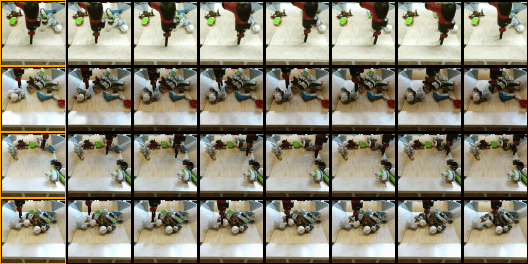}
\caption{RaMViD~\cite{hoppe2022diffusion}}
\end{subfigure}
\begin{subfigure}[t]{0.8\linewidth}
\includegraphics[width=\linewidth]{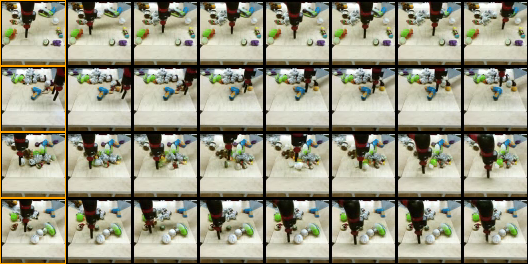}
\caption{\modelname-L-FP (ours)}
\end{subfigure}
\caption[Comparison of frame prediction samples on BAIR unseen evaluation set]{\textbf{Comparison of frame prediction samples on BAIR unseen evaluation set}. 16-frame videos are generated at 64$\times$64 resolution 10 fps given the first frame as condition and shown at 5 fps where condition frames are marked in \textcolor{orange}{orange}. Samples for \cite{hoppe2022diffusion} are obtained from their official release (\url{https://sites.google.com/view/video-diffusion-prediction}).
As shown, the clips produced by \modelname{} maintaining a better visual consistency and spatial-temporal dynamics.}
\label{fig:bair}
\end{figure*}

\begin{figure*}[tp]
\centering
\begin{subfigure}[t]{0.83\linewidth}
\centering
\includegraphics[width=\linewidth]{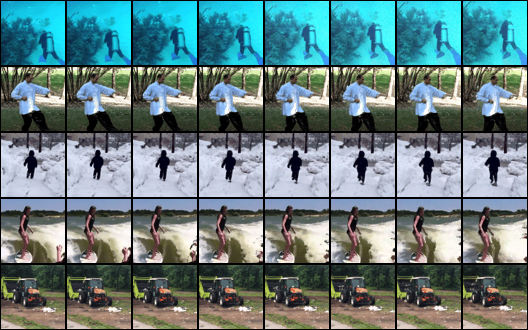}
\caption{RaMViD~\cite{hoppe2022diffusion} at 64$\times$64 resolution, condition information is unavailable.}
\end{subfigure}
\begin{subfigure}[t]{0.83\linewidth}
\centering
\includegraphics[width=\linewidth]{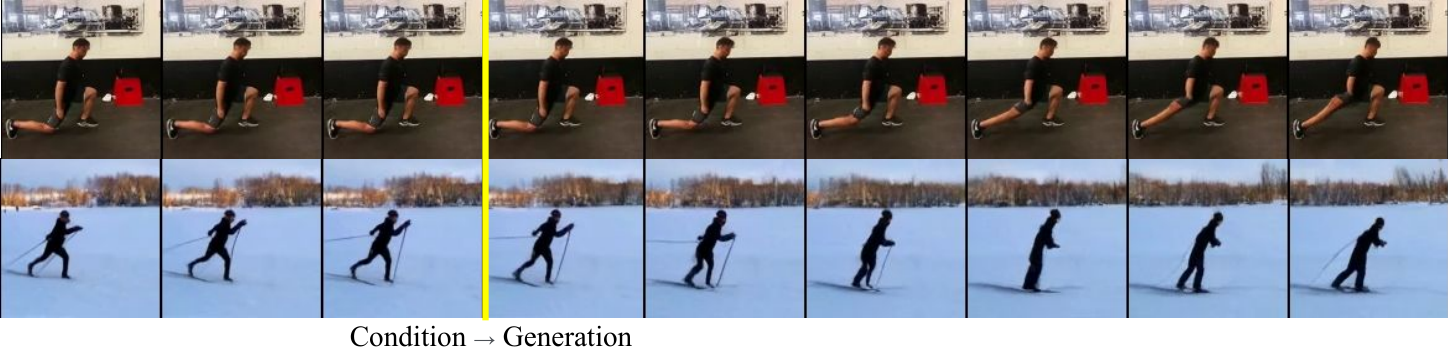}
\caption{\modelname-L-FP (ours) at 128$\times$128 resolution, condition frames are marked in \textcolor{orange}{orange}.}
\end{subfigure}
\caption[Comparison of frame prediction samples on Kinetics-600 unseen evaluation set]{\textbf{Comparison of frame prediction samples on Kinetics-600 unseen evaluation set}. 16-frame videos are generated at 25 fps given 5-frame condition. Samples for \cite{hoppe2022diffusion} are obtained from their official release (\url{https://sites.google.com/view/video-diffusion-prediction}).
As shown, given the conditioned frames, \modelname{} generates plausible actions with greater details.}
\label{fig:k600}
\end{figure*}

On the large dataset of Kinetics-600, it establishes a new state-of-the-art result, improving the previous best FVD in~\cite{ho2022video} from $16.2$ to $9.9$ by a relative $39\%$ improvement. The above results verify \modelname{}'s compelling generation quality, including on the large Kinetics dataset.
\cref{fig:bair} and \cref{fig:k600} below provide visual comparisons to the baseline methods on BAIR and Kinetics-600, respectively.
%on both FP benchmarks, with a +39\% improvement on Kinetics-600 to FVD 9.9 and a 18\% improvement on BAIR to FVD 69.

\subsection{Inference-Time Generation Efficiency}
\label{sec:exp_efficiency}
\begin{figure}[tp]
\centering
\includegraphics[width=\linewidth,trim={0cm 0.25cm 0cm 0.2cm},clip]{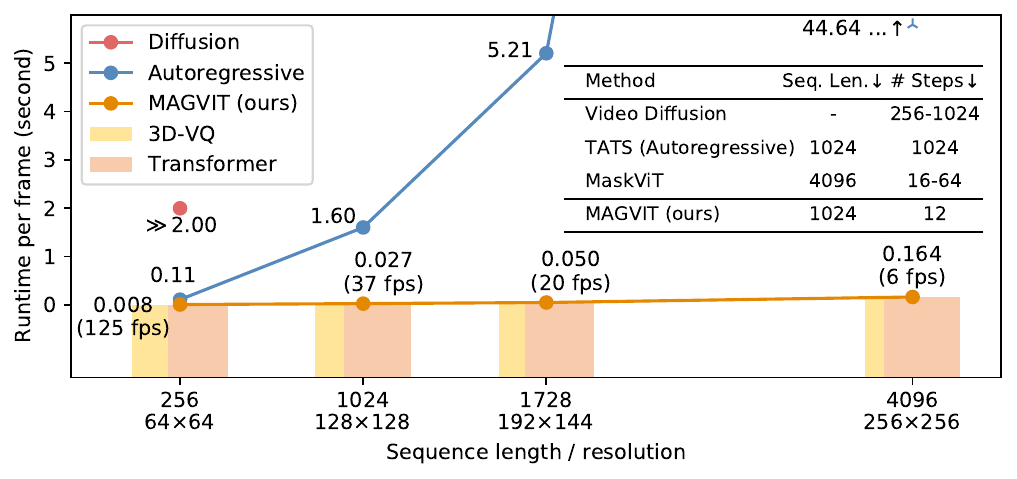}
% \vspace{-5mm}
\caption[Inference-time generation efficiency comparison]{\textbf{Inference-time generation efficiency comparison}. 
The average runtime for generating one frame is measured at different resolutions. 
The colored bars show the time breakdown between the 3D-VQ and the transformer.
The embedded table compares the critical factors of inference efficiency for different methods at 16-frame 128$\times$128, except for Video Diffusion~\cite{ho2022video} at 64$\times$64.
% for a 16-frame video at 128$\times$128 resolution.
% The vertical axis is in log scale for better comparison.
}
\label{fig:efficiency}
% \vspace{-5mm}
\end{figure}

Video generation efficiency is an important metric in many applications. We conduct experiments to validate that \modelname{} offers top speed in video generation.
\cref{fig:efficiency} shows the processing time for each frame on a single V100 GPU at different resolutions.
We compare \modelname-B with an autoregressive transformer of the same size and a diffusion-based model~\cite{ho2022video}.
At 128$\times$128 resolution, \modelname-B runs at $37$ frames-per-second (fps).
% which enters the real-time realm with a relatively old consumer hardware.
When running on a single TPUv4i~\cite{jouppi2021ten},
\modelname{}-B runs at $190$ fps and \modelname{}-L runs at $65$ fps.
%, the larger \modelname{}-L can achieve over $35$ fps.
%\modelname{} opens the gate for real-time neural video synthesis with state-of-the-art generation quality.
% \lu{need to revise the figure.}
% We report the number of decoding steps and total decoding FLOPs of recent video generation models, as listed in \cref{tab:efficiency}.
% The measured computation cost are based on the generation of a 16$\times$128$\times$128 video.
% We also include the number of tokens for transformer-based models.
% Note that
% \input{tables/efficiency}

\cref{fig:efficiency} compares the sequence lengths and inference steps of these models.
Diffusion models~\cite{ho2022video} typically require 256-1000 diffusion steps with a 3D U-Net~\cite{cciccek20163d}.
Autoregressive models, such as TATS~\cite{ge2022long}, decode visual tokens sequentially, which runs $60$ times slower than \modelname{} at 128$\times$128. 
%and $>$250 times slower when the sequence gets longer at $256\times256$.
%Non-autoregressive MaskViT~\cite{gupta2022maskvit} has an improved speed but still falls short compared with our %\modelname{} due to its long sequence from the 2D-VQ tokenizer and relatively more decoding steps.
Compared to the recent non-autoregressive model MaskViT~\cite{gupta2022maskvit}, \modelname{} is $4$ to $16$ times faster due to more efficient decoding on shorter sequences.
%Our 3D-VQ tokenizer can compress both spatially and temporally, while 2D-VQ cannot do temporal compression, resulting in many redundant tokens.
% \footnotetext{\url{https://songweige.github.io/projects/tats/}}

\subsection{Multi-task Video Generation}

\begin{table*}[tp]
\centering
% \scriptsize
\begin{tabular}{@{}lc|ccccccccc@{}}
\toprule
Method &Task &\textbf{Avg}$\downarrow$ & FP & FI & OPC & OPV & OPH & OPD & IPC & IPD  \\ \midrule
\modelname-B-UNC &Single& 150.6 & {\color{mygray}74.0} & {\color{mygray}71.4} & {\color{mygray}119.0} & {\color{mygray}46.7} & {\color{mygray}55.9} & {\color{mygray}389.3} & {\color{mygray}145.0} & {\color{mygray}303.2}  \\
\modelname-B-FP  & Single &201.1 & 47.7 & {\color{mygray}56.2}& {\color{mygray}247.1} & {\color{mygray}118.5} & {\color{mygray}142.7}  & {\color{mygray}366.3} & {\color{mygray}357.3} & {\color{mygray}272.7} \\ \hdashline
\modelname-B-\emph{MT} & Multi& 32.8 & 47.2 & 36.0 & 28.1 & 29.0 & 27.8 & 32.1 & 31.1 & 31.0 \\
\modelname-L-\emph{MT} & Multi& \textbf{22.8} & 31.4 & 26.4 & 21.3 & 21.2 & 19.5 & 20.9 & 21.3 & 20.3  \\\bottomrule
\end{tabular}
% \vspace{-1mm}
\caption[Multi-task generation performance on BAIR]{\textbf{Multi-task generation performance on BAIR evaluated by FVD}. \textcolor{mygray}{Gray} values denote unseen tasks during training. }
\label{tab:mt_bair}
% \vspace{-2mm}
\end{table*}

\begin{sidewaystable}[tp]
\centering
\begin{tabular}{@{}ccccccccccccc@{}}
\toprule
Method & Task & SSv2-\textbf{Avg}$\downarrow$ & FP & FI & OPC & OPV & OPH & OPD  & IPC  & IPD & CG & CFP    \\ \midrule
\modelname-B-UNC & Single & 258.8   & {\color{gray}278.8}  & {\color{gray}91.0} & {\color{gray}67.5} & {\color{gray}27.3}  & {\color{gray}36.2}& {\color{gray}711.5} & {\color{gray}319.3}   & {\color{gray}669.8}  & {\color{gray}107.7}& {\color{gray}279.0} \\
\modelname-B-FP & Single & 402.9   & 59.3 & {\color{gray}76.2} & {\color{gray}213.2} & {\color{gray}81.2} & {\color{gray}86.3}& {\color{gray}632.7}    & {\color{gray}343.1}  & {\color{gray}697.9}  & {\color{gray}1780.0}& 59.3  \\ \hdashline
% \modelname-B-\emph{MT10} & \checkmark & 42.8 & 111.6 & 62.6 & 52.3 & 35.5 & 38.8 & 31.6 & 25.5& 23.6  & 24.5 & 22.4 \\
\modelname-B-\emph{MT} & Multi & 43.4 & 71.5 & 38.0 & 38.8 & 23.3 & 26.1 & 33.4  & 23.3  & 25.3 & 94.7 & 59.3   \\
% \modelname-L-\emph{MT10} & \checkmark & \textbf{29.6} & 99.8 & 35.2 & 30.1 & 25.1 & 21.3 & 23.2 & 17.0& 14.5  & 16.4  & 13.5  \\
\modelname-L-\emph{MT} & Multi & \textbf{27.3} & 33.8 & 25.0 & 21.1 & 16.8 & 17.0& 23.5  & 13.5  & 15.0  & 79.1 & 28.5   \\ \midrule
Masked pixel & - & - & 94\% & 87\% & 75\% & 50\% & 50\% & 50\% & 25\% & 25\% & 100\% & 94\% \\
Masked token & - & - & 75\% & 50\% & 75\% & 50\% & 50\% & 50\% & 25\% & 25\% & 100\% & 75\% \\
% Condition proportion & & & 0\% & 6\% &6\% & 13\% & 25\%& 50\% & 50\%  & 50\% & 75\% & 75\% \\
\bottomrule
\end{tabular}
\caption[Multi-task generation performance on Something-Something-V2]{\textbf{Multi-task generation performance on Something-Something-V2 evaluated by FVD}. \textcolor{gray}{Gray} values denote unseen tasks during training. The bottom two rows list the proportions of masked pixels and tokens for each task.}
\label{tab:ssv2}
\ \\ \ \\
\begin{tabular}{@{}c|c|c|ccccccccccc@{}}
\toprule
Method & \makecell{nuScenes-FP} & \makecell{Objectron-FI} & \makecell{Web12M-\textbf{MT8}} & FP & FI & OPC & OPV & OPH & OPD  & IPC & IPD \\ \midrule
% \modelname-L-\emph{MT10} & \checkmark & \textbf{29.6} & 99.8 & 35.2 & 30.1 & 25.1 & 21.3 & 23.2 & 17.0& 14.5  & 16.4  & 13.5  \\
\modelname-B & 29.3 & - & 33.0 & 84.9 & 33.9 & 34.4 & 21.5 & 22.1& 26.0  & 20.7  & 20.4    \\
\modelname-L & 20.6 & 26.7 & \textbf{21.6} & 45.5 & 30.9 & 19.9 & 15.3 & 14.5 & 20.2  & 12.0  & 14.7    \\
% Condition proportion & & & 0\% & 6\% &6\% & 13\% & 25\%& 50\% & 50\%  & 50\% & 75\% & 75\% \\
\bottomrule
\end{tabular}
\caption[Multi-task generation performance on NuScenes, Objectron, and Web videos]{\textbf{Multi-task generation performance on NuScenes, Objectron, and Web videos evaluated by FVD}.}
\label{tab:nuscene}
\end{sidewaystable}

% \begin{sidewaystable}[tp]
% \centering

% \caption[Generation performance on NuScenes, Objectron, and Web videos]{\textbf{Generation performance on NuScenes, Objectron, and Web videos evaluated by FVD}.}
% \label{tab:other_data}
% \end{sidewaystable}

% \begin{figure*}[tp]
% \centering
% \includegraphics[width=\linewidth,trim={0.8cm 0 0 0},clip]{demo}
% % \vspace{-3mm}
% \caption{\textbf{Multi-task generation samples on four datasets: SSv2~\cite{goyal2017something}, nuScenes~\cite{caesar2020nuscenes}, Objectron~\cite{ahmadyan2021objectron}, and Web videos.} The left column is from a single ten-task model on SSv2, while the top eight rows on the right are from a single eight-task model on Web data.}
% \label{fig:mt_result}
% % \vspace{-3mm}
% \end{figure*}

\label{sec:exp_flexibility}
To demonstrate the flexibility in multi-task video synthesis, we train a single \modelname{} model to perform eight tasks on BAIR or ten tasks on SSv2. 
We do not intend to compare with dedicated models trained on these tasks but to demonstrate a generic model for video synthesis.

%BAIR is two tasks short as it has no labels to support the CG and CFP tasks.
% \vspace{-4mm}
\paragraph{Eight tasks on BAIR.}
We perform a multi-task evaluation on BAIR with eight self-supervised tasks.
\cref{tab:mt_bair} lists the ``debiased'' FVD for each task, where the third column computes the average. 
We compare the multi-task models (MT) with two single-task baselines trained on unconditional generation (UNC) and frame prediction (FP).
% All models are evaluated on all tasks and grey numbers mark the unseen tasks during training.
% Since multi-task models takes longer to train, we train them 3$\times$ longer than the single-task models.

\begin{figure*}[tp]
\centering
\includegraphics[width=\linewidth,trim={0 0 0 0},clip]{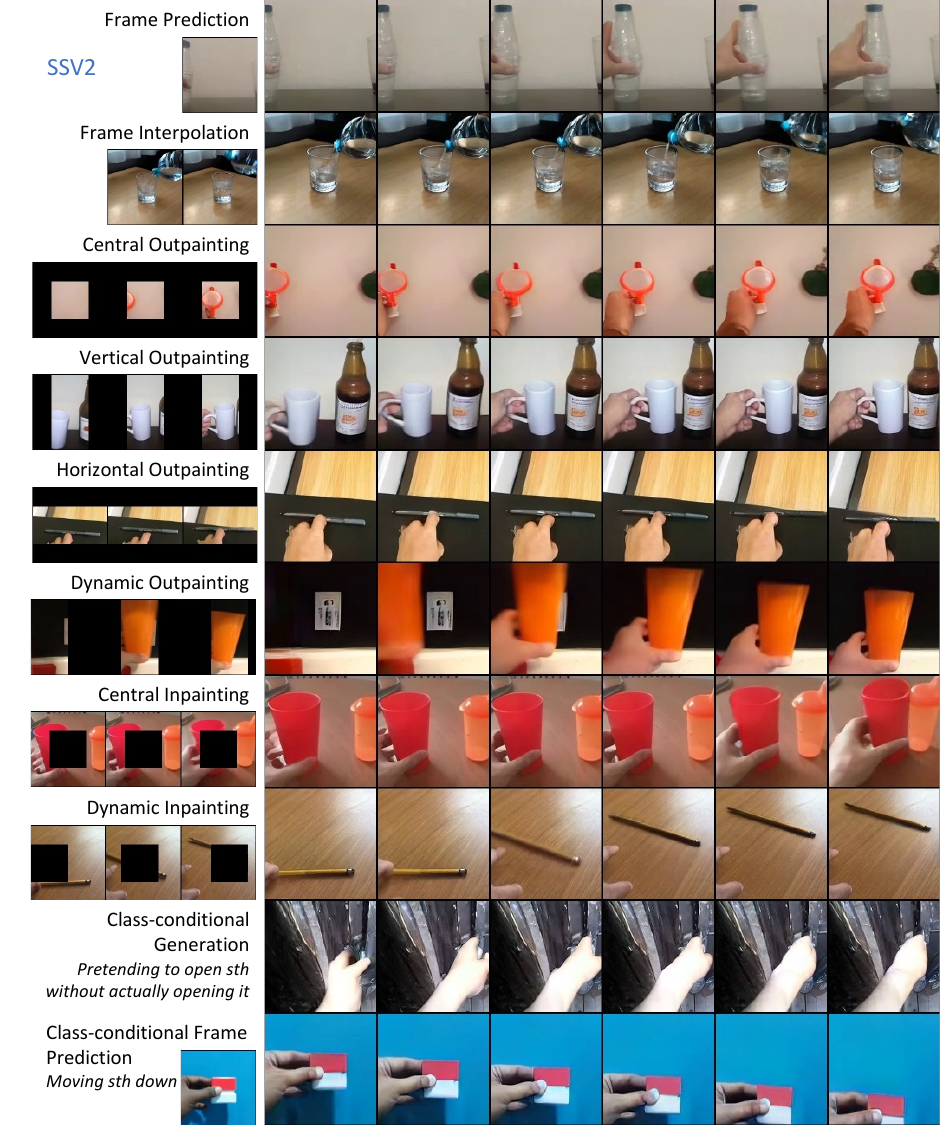}
\caption[Multi-task generation results]{\textbf{Multi-task generation results} for the model only trained on the Something-Something-V2 dataset~\cite{goyal2017something}. The conditions used to generate the shown videos are taken from the Something-Something-V2 evaluation videos.}
\label{fig:mt_result_1}
\end{figure*}

%We use two single-task models as baselines, one for unconditional generation (UNC) and one for frame prediction (FP).
As shown in \cref{tab:mt_bair}, the multi-task models achieve better fidelity across all tasks. Single-task models perform considerably worse on the tasks unseen in training (\textcolor{mygray}{gray} values in \cref{tab:mt_bair}), especially on the tasks that differ more from the training task. 
Compared to the single-task models in their training task, MT performs better with a small gain on FP with the same model size. 
\paragraph{Ten tasks on SSv2.}
We evaluate on the large-scale SSv2 dataset, where \modelname{} needs to synthesize 174 basic actions with everyday objects. We evaluate a total of ten tasks, with two of them using class labels (CG and CFP), as shown in \cref{tab:ssv2}. 
%The average FVD is reported in the last column of \cref{tab:ssv2}
%
% \lu{we should save some space for ablation which is important}
We observe a pattern consistent with BAIR: multi-task models achieve better average FVD across all tasks.
\cref{fig:mt_result_1} shows examples of generated videos for each task.
%SSv2 10-task results show a similar pattern as BAIR 8-task.
% For the CG model, OPV and OPH remain the easiest while OPD and IPD appear to be the hardest.
% The baseline average score is 258.8, with a CG FVD of 107.7.
% The 10-task model achieves an average score of 45.6, with huge improvements on all 9 new tasks compared with the CG model.
% The FVD distribution among the 10 tasks is reversely relevant to the proportion of condition pixels, with CG (0\%) being the most difficult and inpainting (75\%) the easiest. Scaling to the large model can further improve the average score to $29.9$.
The above results substantiate model generalization trained with the proposed multi-task objective.

\begin{figure*}[tp]
\centering
\includegraphics[width=\linewidth,trim={0 0 0 0},clip]{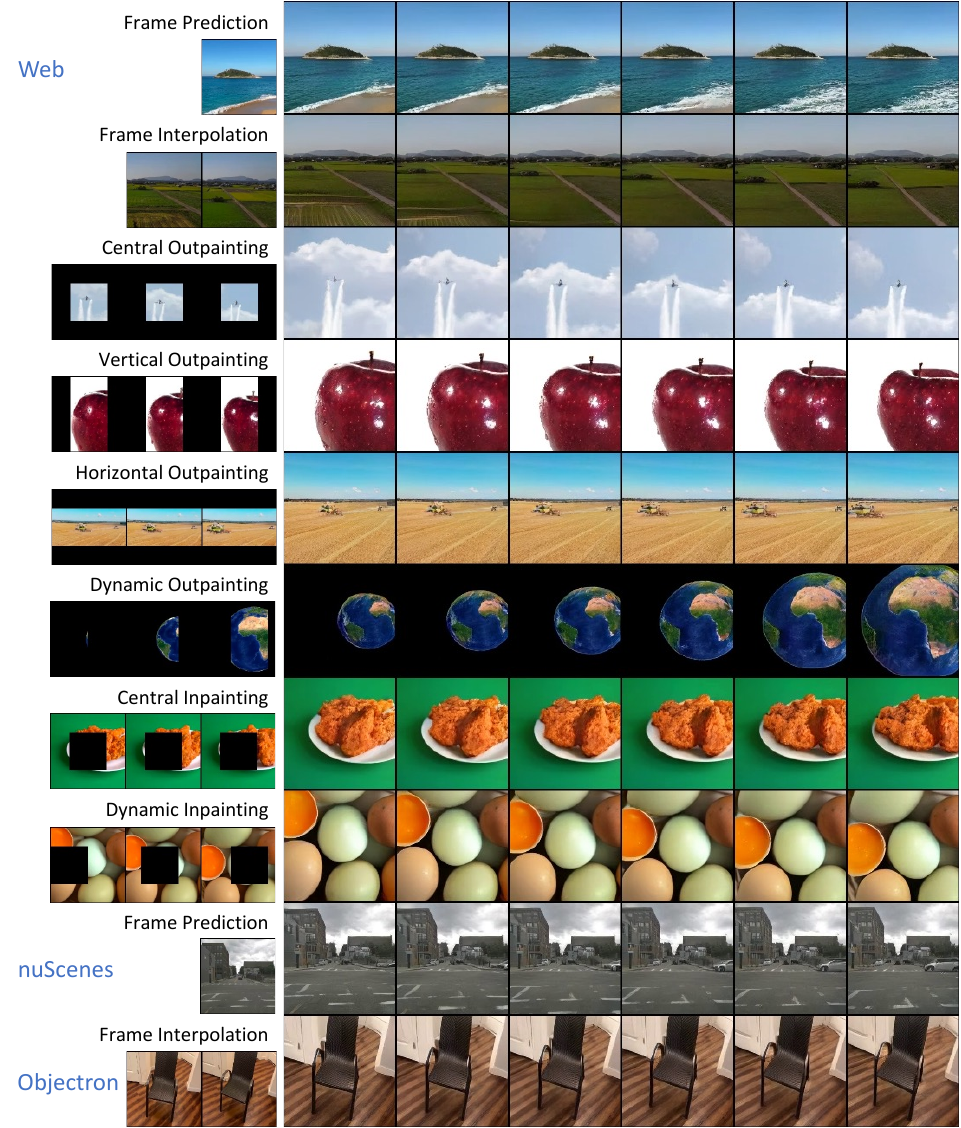}
\caption[Multi-task generation results]{\textbf{Multi-task generation results} for three models trained on nuScenes~\cite{caesar2020nuscenes}, Objectron~\cite{ahmadyan2021objectron}, and 12M Web videos, respectively. The conditions used to generate the shown videos are taken from the evaluation set.}
\label{fig:mt_result_2}
\end{figure*}

\paragraph{Results on nuScenes, Objectron, and 12M Web Videos.}
\cref{tab:nuscene} shows the generation performance on three additional datasets, \ie, nuScenes~\cite{caesar2020nuscenes}, Objectron~\cite{ahmadyan2021objectron}, and Web12M which contains 12 million videos we collected from the web.
We evaluate our model on the frame prediction task on nuScenes, the frame interpolation task on Objectron, and the 8-task suite on the Web videos. 
\cref{fig:mt_result_2} shows examples of generated videos for each task. The results substantiate the generalization performance of \modelname{} on videos from distinct visual domains and the multi-task learning recipe on large-scale data.

%for the evaluation of multi-task video synthesis.
% SSv2 contains 169k videos for training and 25k videos for evaluation in 174 basic actions with everyday objects.
%Its videos are about human performing basic actions with everyday objects.
% We calculate the FVD metric with 50K samples against the training set for the CG task and against the validation task for the other tasks.
%We use a single task model trained with CG as the baseline.

\begin{table}[tp]
\centering
% \scriptsize
\begin{tabular}{@{}llccccccc@{}}%{@{}l@{\hspace{5pt}}l@{\hspace{5pt}}c@{\hspace{6pt}}ccc@{}}
\toprule
\multicolumn{2}{l}{Method} & \makecell{Seq. Length} & \makecell{FP FVD$\downarrow$} & \makecell{MT8 FVD$\downarrow$} \\ \midrule
%C.1 \makecell[l]{Fixed cropped\\(from full videos~\cite{chang2022maskgit})} & 1024  & 946  & 1827 \\
% \multicolumn{2}{l}{Image MTM~\cite{chang2022maskgit}} & 1024  & 946  & 1827 \\
% \multicolumn{2}{l}{MTM with injected condition}  & 1024   & 508 & 76 \\
\multicolumn{2}{l}{Latent masking in MaskGIT~\cite{chang2022maskgit}} & 1024 & 74 & 151 \\
\multicolumn{2}{l}{Prefix condition} & 1024-1792   & 55 & - \\ \midrule
% & $\Ls_\text{mask}$ & $\Ls_\text{recons}$ & $\Ls_\text{refine}$ \\
\multicolumn{1}{l}{\multirow{3}{*}{\makecell{\emph{\methodname}\\(ours)}}} & $\Ls_\text{mask}$ & \multirow{3}{*}{1024} & 388 & 143 \\
& $\Ls_\text{mask}+\Ls_\text{recons}$ & & 51 & 53 \\
% & \checkmark & & \checkmark & & 50 & TBD \\
&$\Ls_\text{mask}+\Ls_\text{recons}+\Ls_\text{refine}$ & & \textbf{48} & \textbf{33} \\ \bottomrule
\end{tabular}
% \vspace{-1mm}
\caption[Comparison of conditional masked token modeling]{\textbf{Comparison of conditional masked token modeling} on BAIR frame prediction (FP) and eight-task (MT8) benchmarks. - indicates we were not able to train to convergence. 
% \lu{the MaskGVT FVD 946 looks too bad. Rerun? Also it is different from the FVD in Table 6.} 
% B.1 obtains condition tokens from the full videos, whereas the others obtain from the condition videos.
}
\label{tab:cond}
% \vspace{-2mm}
\end{table}

\subsection{Ablation Study}
\label{sec:exp_abl}

% % \vspace{-4mm}
\paragraph{Conditional MTM.}
We demonstrate the efficacy of \methodname{} by comparing it with conventional MTM methods, including the
latent masking in MaskGIT for image synthesis~\cite{chang2022maskgit} 
and the commonly-used prefix condition that prepends cropped condition tokens to the input sequence.
%masked token modeling methods on BAIR frame prediction and 8-task benchmarks in \cref{tab:cond}.

\cref{tab:cond} compares these methods on the BAIR dataset where the same 3D-VQ tokenizer is used in all approaches. 
As discussed in \cref{sec:mth_cmm}, latent masking in~\cite{chang2022maskgit}, which directly unmasks tokens of the condition region at inference time, leads to poor generalization, especially for the multi-task setup.
%due to the leaked ground-truth during training, 
% Using condition tokens from the condition videos (C.2) aligns with inference-time input, but introduces a gap with the true output sequence due to the frozen condition tokens.
% Injecting the condition tokens into the sequence before latent masking achieves a better causal
%The gap appears severe in the single-task setup but multi-task training is able to bridge it to some extent.
Prefix condition produces a long sequence of variable length, making it less tractable for multi-task learning.
In contrast, \methodname{} yields a fixed-length sequence and better generalizability for both single- and multi-task setups.
% \lijun{to add qualitative samples}

% \vspace{-4mm}
\paragraph{Training losses.} The bottom section of \cref{tab:cond} shows the contribution of the training loss components in Eq.~\eqref{eq:mt_loss_decomposed}. 
% The result substantiates the proposed refinement loss $\Ls_\text{refine}$. 

% \vspace{-4mm}
\paragraph{Decoding methods.}
% \begin{table}[!ht]
% \centering
% \caption{Comparison between different model architectures on UCF-101 class-conditional generation and BAIR frame prediction benchmarks. We compare between 2D and 3D VQ tokenizers as well as bidirectional (BT) and autoregressive (AT) transformers. \lijun{data to be updated}}
% \label{tab:arch}
% \begin{tabular}{@{}ccccc@{}}
% \toprule
% \multirow{2}{*}{Architecture} & \multirow{2}{*}{\makecell{\# Tokens /\\\# Steps}} & UCF CG & \multirow{2}{*}{\makecell{BAIR\\FP FVD$\downarrow$}}   \\ 
%  & & FVD$\downarrow$ / IS$\uparrow$ & \\ \midrule
% \makecell[l]{\textbf{2D}} \\
% MaskGIT~\cite{chang2022maskgit} & 4096 / 18 & 430 / 72.19 & 206      \\
% MaskViT~\cite{gupta2022maskvit} & 4096 / 18 & - & 94* \\ \midrule
% %  & 8000 & & 160 & 76.76 \\ 
% TATS~\cite{ge2022long} & 1024 / 1024 & 332 / 79.28 & - \\\midrule
% \cite{chang2022maskgit}+3D-VQ & 1024 / 12 & - & 946 \\
% Autoregressive & 1024 / 1024 & 537 / 32.07 & - \\ \midrule
% \modelname & \textbf{1024} / \textbf{12} & \textbf{159} / \textbf{83.55}  & \textbf{47}   \\ \bottomrule
% \end{tabular}
% \end{table}

\begin{table}[tp]
\centering
% \scriptsize
\begin{tabular}{@{}lccccccc@{}} %{@{}l@{\hspace{3pt}}c@{\hspace{5pt}}c@{\hspace{5pt}}c@{\hspace{3pt}}c@{\hspace{3pt}}c@{\hspace{3pt}}c@{\hspace{5pt}}c@{}}
\toprule
Decoding Method & Tokenizer &Type & Param. & Seq. Len.$\downarrow$ & \# Steps$\downarrow$ & \makecell{FVD$\downarrow$} \\ \midrule
%\textbf{2D-VQ} \\
\multirow{2}{*}{MaskGIT~\cite{chang2022maskgit}} & 2D-VQ  &NAR& 53M+87M & 4096 & 12 & 222 (177) \\
 & 3D-VQ &NAR& 41M+87M & 1024 & 12 & 122 (74) \\
MaskViT~\cite{gupta2022maskvit} & 2D-VQ  &NAR& 53M+189M & 4096 & 18 & 94$^*$ \\ \hdashline % 189M
%\textbf{3D-VQ} \\
AR & 3D-VQ &AR&41M+87M & 1024 & 1024 & 91 (56)  \\ \midrule
\emph{\modelname} (ours) & 3D-VQ &NAR&41M+87M & 1024 & 12 & \textbf{76} (\textbf{48})\\ \bottomrule
\end{tabular}
% \vspace{-1mm}
\caption[Comparison of decoding methods]{\textbf{Comparison of decoding methods} on BAIR frame prediction benchmark. 
The number of parameters is broken down as VQ + Transformer.
NAR is non-autoregressive and AR is autoregressive. FVD and debiased FVD (in parentheses) are reported.
$^*$ marks the quoted number from their paper.
% \lu{This 336 is too different from 94. We can either rerun or change our configuration like decoding steps to explain the big gap.}
% \lijun{AR is incorrect, updating}
% \lu{add mt FVD?}
}
\label{tab:arch}
% \vspace{-2mm}
\end{table}

\cref{tab:arch} compares \cref{alg:decoding} with existing autoregressive (AR) and non-autoregressive (NAR) decoding methods. 
We consider two NAR baselines, \ie, MaskGIT~\cite{chang2022maskgit} for image and MaskViT~\cite{gupta2022maskvit} for video synthesis.
%the the MaskGIT~\cite{chang2022maskgit} and MaskViT's NAR methods with both 2D and 3D VQ tokenizers and 
%
% image decoding in MaskGIT~\cite{chang2022maskgit} both 2D and 3D VQ tokenizers and the autoregressive decoding method~\cite{}.
%different architecture choices for both stages on the 
We use the same 3D-VQ tokenizer for MaskGIT, AR, and \modelname{}. 
% For reference, it also includes MaskGIT with a 2D-VQ and the related work MaskViT~\cite{gupta2022maskvit}. 
As shown, the proposed decoding algorithm produces the best quality with the 3D-VQ and has a $4\times$ shorter sequence than the 2D-VQ.
While the AR transformer obtains a reasonable FVD, it takes 
% slower to train and 
over $85\times$ more steps at inference time.

%3D-VQ yields $4\times$ sequence length but 
%UCF-101 class-conditional generation task in \cref{tab:arch}.
% Our \modelname{} uses a 3D tokenizer and a bidirectional transformer.
% Changing to a 2D tokenizer (2D+BT) results in a $4\times$ longer sequence and around $16\times$ computation cost, but worse generation quality.
% Changing to an autoregressive transformer (3D+AT) makes it over $85\times$ slower at inference time, while it takes $4\times$ training cost to reach a comparable performance.
% MaskViT(189M).

\subsection{Implementation Details}\label{sec:magvit_impl}
\paragraph{Training.}
\modelname{} is trained in two stages where we first train the 3D-VQ tokenizer as detailed in \cref{chap:3dvq} and then train the transformer with a frozen tokenizer.
We follow the same learning recipe across all datasets, with the only variation in the number of training epochs.
Here are the training details for the second stage:
\begin{multicols}{2}
\begin{itemize}
    \item Sequence length: 1026.
    \item Hidden dropout rate: $0.1$.
    \item Attention dropout rate: $0.1$.
    \item Mask rate schedule: cosine.
    \item Peak learning rate: $10^{-4}$.
    \item Learning rate schedule: linear warm up and cosine decay.
    \item Optimizer: Adam with $\beta_1=0.9$ and $\beta_2=0.96$.
    \item Weight decay $0.045$.
    \item Label smoothing: $10^{-4}$.
    \item Max gradient norm: $1$.
    \item Batch size: 256.
    \item Speed: \\1.24 steps/sec on 16 TPU-v2 chips for B, 2.70 steps/sec on 32 TPU-v4 chips for L.
\end{itemize}
\end{multicols}

Using more hardware resources can speed up the training.
We train \modelname{} models for each dataset separately.
The training epochs for each dataset are listed in \cref{tab:train_step2}.

\begin{table}[tp]
\centering
\begin{tabular}{@{}lcccccc@{}}
\toprule
Dataset & B & L \\ \midrule
UCF-101 & 2000 & 2000\\
BAIR  & 400 & 800 \\ 
BAIR-MT  & 1200 & 1600 \\
Kinetics-600 & 180 & 360 \\
SSv2 & 720 & 1440 \\
nuScenes  & 2560 & 10240 \\
Objectron  & 1000 & 2000 \\
Web12M & 10 & 20 \\ \bottomrule
\end{tabular}
\caption[Training epochs of \modelname{} transformer for each dataset]{\textbf{Training epochs of \modelname{} transformer for each dataset}.}
\label{tab:train_step2}
\end{table}

\paragraph{Sampling protocols.}
We follow the sampling protocols from previous works~\cite{ge2022long,clark2019adversarial} when eveluating on the standard benchmarks, \ie UCF-101, BAIR, and Kinetics-600.
We sample 16-frame clips from each dataset without replacement to form the real distribution in FVD and extract condition inputs from them to feed to the model.
We continuously run through all the samples required (\eg, 40,000 for UCF-101) with a single data loader and compute the mean and standard deviation for 4 folds.
When evaluating on other datasets, due to the lack of prior works, we adapt the above protocol based on the dataset size to ensure sample diversity.

For our \modelname{} model, we use the following COMMIT decoding hyperparameters by default: cosine schedule, $12$ steps, temperature $4.5$.
Below are detailed setups for each dataset:
% Note that the evaluation resolution may be different from the generation resolution, where the generated samples are bilinear resized to the target resolution.

\begin{itemize}[nosep, leftmargin=*]
    \item UCF-101: 
    \begin{itemize}[nosep, leftmargin=*]
        \item Dataset: 9.5K videos for training, 101 classes.
        \item Number of samples: 10,000$\times$4.
        \item Resolution: 128$\times$128.
        \item Real distribution: random clips from the training videos.
    \end{itemize}
    \item BAIR:
    \begin{itemize}[nosep, leftmargin=*]
        \item Dataset: 43K videos for training and 256 videos for evaluation.
        \item Number of samples: 25,600$\times$4.
        \item Resolution: 64$\times$64.
        \item Real distribution: the first 16-frame clip from each evaluation video.
        \item COMMIT decoding: exponential schedule, temperature 400.
    \end{itemize}
    \item Kinetics-600:
    \begin{itemize}[nosep, leftmargin=*]
        \item Dataset: 384K videos for training and 29K videos for evaluation.
        \item Number of samples: 50,000$\times$4.
        \item Generation resolution: 128$\times$128.
        \item Evaluation resolution: 64$\times$64, via central crop and bilinear resize.
        \item Real distribution: 6 sampled clips (2 temporal windows and 3 spatial crops) from each evaluation video.
        \item COMMIT decoding: uniform schedule, temperature 7.5.
    \end{itemize}
    \item SSv2:
    \begin{itemize}[nosep, leftmargin=*]
        \item Dataset: 169K videos for training and 24K videos for evaluation, 174 classes.
        \item Number of samples: 50,000$\times$4.
        \item Resolution: 128$\times$128.
        \item Real distribution for the CG task: random clips from the training videos.
        \item Real distribution for the other tasks: 2 sampled clips (2 temporal windows and central crop) from each evaluation video.
    \end{itemize}
    \item nuScenes:
    \begin{itemize}[nosep, leftmargin=*]
        \item Dataset: 5.4K videos for training and 0.6K videos for evaluation, front camera only, 32 frames per video.
        \item Number of samples: 50,000$\times$4.
        \item Resolution: 128$\times$128.
        \item Real distribution: 48 sampled clips (16 temporal windows and 3 spatial crops) from each evaluation video.
    \end{itemize}
    \item Objectron:
    \begin{itemize}[nosep, leftmargin=*]
        \item Dataset: 14.4K videos for training and 3.6K videos for evaluation.
        \item Number of samples: 50,000$\times$4.
        \item Resolution: 128$\times$128.
        \item Real distribution: 5 sampled clips (5 temporal windows and central crop) from each evaluation video.
    \end{itemize}
    \item Web12M:
    \begin{itemize}[nosep, leftmargin=*]
        \item Dataset: $\sim$12M videos for training and 26K videos for evaluation.
        \item Number of samples: 50,000$\times$4.
        \item Resolution: 128$\times$128.
        \item Real distribution: randomly sampled clips from evaluation videos.
    \end{itemize}
\end{itemize}

\section{Summary}
In this chapter, we propose \modelname{}, a generic and efficient mask-based video generation model.
We use the high-quality 3D-VQ tokenizer from \cref{chap:3dvq} to quantize a video and
%into visual tokens and train a transformer model on masked token sequences.
design \emph{\methodname{}} for \emph{multi-task} conditional masked token modeling.
%which embeds conditions into a generalized multivariate mask and transforms the denoising decoding into a conditional-transition process.
% The inference of \modelname{} is orders-of-magnitude faster than existing methods, which paves the road for real-time neural video synthesis.
% We demonstrate the flexibility with a single model capable of 10 different generation tasks, as well as datasets from various domains.
We conduct extensive experiments to demonstrate the video generation quality, efficiency, and flexibility for multi-task generation.
Notably, \modelname{} establishes a new state-of-the-art quality for class conditional generation on UCF-101 and frame prediction on BAIR Robot Pushing and Kinetics-600 datasets.

% In comparison, \modelname{} can be trained on multiple self-supervised tasks using only unlabeled data.

\glsresetall
\cleardoublepage
\chapter{Generative Modality Infusion Into Frozen Large Language Models} \label{chap:llm}
\graphicspath{{papers/spae/}}

\renewcommand{\modelname}{SPAE} % Semantical Pyramid AutoEncoder

\paragraph{Overview.}
\Glspl{LLM} have made significant advances in solving a wide range of \acrshort{NLP} tasks, while expanding their capabilities beyond language into other modalities.
However, little success has been made in the generation of another modality with an \Gls{LLM} itself.
In this chapter, we adopt the \acrshort{SPAE} representation proposed in \cref{chap:spae} for enabling frozen \Glspl{LLM} to perform \emph{multi-modal multi-task} understanding and generation involving non-linguistic modalities such as images or videos. 
Our approach is validated through in-context learning experiments with frozen \acrshort{PaLM} 2 and \acrshort{GPT} 3.5 on a diverse set of image understanding and generation tasks.
Our method marks the first successful attempt to enable a frozen \Gls{LLM} to generate image content while surpassing state-of-the-art performance in image understanding tasks, under the same setting, by over 25\%. 

\clearpage

\section{Motivation}

%To facilitate LLMs for zero-shot and few-shot image-text tasks, we propose a method that allows a frozen LLM to perform both understanding and generation tasks involving vision modalities, such as images or videos. We have developed a novel vector quantization (VQ) approach for bidirectional conversion between raw pixels and language words. In contrast to prior works, our approach quantizes the images into readable multilingual words that contain both rich semantic meanings and fine-grained details that are essential for reconstructing the input image.

% Large Language Models (LLMs) have made significant advances in solving a wide range of NLP tasks, while expanding their capabilities beyond language into other modalities.
% However, little success has been made in the generation of another modality with an LLM itself.

%\kevinn{I modified the intro}
%Transformer~\cite{vaswani2017attention} models have achieved state-of-the-art results on natural language processing (NLP), vision, and audio tasks~\cite{brown2020language,chowdhery2022palm,openai2023gpt,yu2022scaling,borsos2022audiolm,googlepalm2}. 
%
Large language models (LLMs) empowered by Transformers~\cite{vaswani2017attention} have achieved remarkable progress in addressing a broad spectrum of Natural Language Processing (NLP) tasks~\cite{brown2020language,chowdhery2022palm,openai2023gpt,googlepalm2}. With the continuous increases in model size and training data,
LLMs are gradually becoming more versatile and agnostic to specific tasks, unlocking new capabilities in solving complex AI tasks~\cite{wei2022emergent}, like question answering, code generation, reasoning, mathematics problem-solving, and understanding humor, among various other applications~\cite{googlepalm2,openai2023gpt}.

% Along with the growth of the model size and training data scale, LLMs are gradually becoming more versatile and agnostic to specific tasks, unlocking a new capabilities in solving complex AI tasks such as question answering, code generation, reasoning, solving mathematics, interrelating humors among other applications~\cite{googlepalm2,openai2023gpt}

%Recently, LLMs are also used for zero-shot and few-shot image-text tasks  \cite{koh2023grounding}. 
%
LLMs capture rich conceptual knowledge about the world in their lexical embeddings. This raises a question:
if provided with the appropriate visual representations as input, \emph{are frozen LLMs capable of solving tasks in visual modalities?} 
Very recently, there have been notable advancements in extending the capabilities of frozen LLMs to tackle image understanding and retrieval tasks~\cite{koh2023grounding,liu2023language}. However, generating a different modality using a frozen LLM that has not been explicitly trained on that modality has proven to be challenging and has had little success.

To facilitate LLMs for such cross-modal tasks,
we propose to learn a vector quantizer to map an image, or some other non-linguistic (``foreign'') modality, 
to the token space of a frozen LLM.
This effectively translates the image into a language that the LLM can comprehend, enabling us to leverage the generative abilities of the LLM to perform image understanding and generation tasks without having to train on image-text pairs.
Specifically, our new approach is that, given an image prompt, 
convert it to a token space with our learned encoder,
use the LLM to generate suitable lexical tokens,
and convert back to pixel space with our learned decoder.

% We introduce a novel Semantic Pyramid AutoEncoder (\modelname{}) that produces a lexical word sequence that (1) carries rich semantics, and (2) retains fine details for signal reconstruction. 
% In contrast to the majority of VQ-VAE approaches~\cite{van2017neural}, our encoder maps to an interpretable discrete latent space, \ie, words. As depicted in \cref{fig:framework}, \modelname{} tokens have a multi-scale representation arranged in a pyramid structure.
% The upper layers of the pyramid comprise semantic-central concepts, while the lower layers prioritize appearance representations that captures the fine details for image reconstruction.
% This design enables us to dynamically adjust the token length to accommodate various tasks, such as using fewer tokens for understanding tasks and more tokens for generation tasks.

%
% Compared to a contemporary work~\cite{liu2023language}, \modelname{} tokens are of rich semantic meaning and suprior reconstrution , enabling us to perform 

% we can retain spatial details by using 

% In addition our is multi-scale representation, produced and superior reconstruction quality.
% %
% In addition, unlike unconstrained VQ-VAE methods,
% our encoder maps to an interpretable discrete latent space.
% %

We verify the plausibility of our approach in an extreme setting of in-context learning~\cite{brown2020language}, without any parameter updates to the LLM.
Our SPAE model is trained standalone, without backpropagating through any language model.
We evaluate our approach on image understanding tasks including image classification, image captioning, and visual question answering.
% We showcase the image generation capabilities of LLMs by leveraging in-context denoising techniques.
We showcase a promising direction to image generation with LLMs by utilizing in-context denoising techniques.
Our method is LLM-agnostic and has been tested with PaLM 2~\cite{googlepalm2} and GPT-3.5~\cite{openai2023gpt}, suggesting compatibility with arbitrary LLMs.
% Code and models are made available at \url{https://github.com/google-research/magvit/projects/spae} for the purpose of reproducible research.

% \mhy{After Kevin's revision, Section is rather short (a bit like abstract). While it is fine, you may consider to take some paragraphs from older versions into this.}
 
%\irfan{I like Kevin's phrasing.  Also, I am getting a bit confused with use of terms like images, videos etc and how they are switched around.}
%
%vivek: added bit of context - feel free to edit out/remove

%Our proposed tokenizer, \emph{Semantic Pyramid AutoEncoder} (\modelname{}), converts raw signals to lexical word sequences, effectively translating the image into a foreign language that the LLM can comprehend. This enables LLMs to perform multimodal tasks such as image/video classification and generation which expands their capabilities beyond traditional language-based tasks. This paper verifies the plausibility of this approach through an extreme scenario, where a frozen LLM can perform multimodal tasks through in-context learning. Our method is LLM-agnostic and has been demonstrated with PaLM-2/Bard and GPT-3.5, suggesting compatibility with arbitrary LLM.

% We propose an autoregressive in-context generation process to facilitate long-term sampling from a foreign modality token distribution.

%and thereby enable LLMs to perform tasks through in-context learning such as image classification, caption, and generation.

%introduce the detailed technique.

%We adopt in-context learning with a black-box frozen large language model to do conditional generation.

%\kevinn{I suggest cutting this to save space}
The main contributions of this work are summarized as follows:
%\irfan{Good list, needs to be a bit concise maybe!}
% % \vspace{-2mm}
\begin{itemize}
\item This is the first successful method,
to the best of our knowledge,
that uses a frozen language model, trained solely on language tokens, to directly generate image content through in-context learning.
% \item We introduce a new SPAE tokenizer producing
% A new visual tokenizer producing language tokens in a pyramid-hierarchical structure with progressively rich semantic meanings and better reconstruction quality
% \item Interpretable 
% interpretable representations of semantic concepts and fine-grained details in the form of multilingual linguistic tokens with adjustable lengths. 
\item We propose a new progressive prompting method that facilitates in-context generation of long cross-modal sequences.
%with a frozen large language model.
\item We evaluate our method on visual understanding and generation tasks, and notably, our approach outperforms the best-published few-shot image classification accuracy~\cite{liu2023language} by an absolute 25\% under the same in-context setting.
% \mhy{missing references}

%Multi-task image generation, combinable; zero-shot image classification improves by around 20\% absolute accuracy.
\end{itemize}

\section{Prior Work}
\paragraph{Multimodal generation with LLMs.}
Advances have been made to expand the capabilities of LLMs beyond language.
For example, Visual ChatGPT~\cite{wu2023visual} uses ChatGPT to generate prompts and executes multimodal tasks through another model, \eg, generating image from text prompts by Stable Diffusion~\cite{rombach2022high}.
FROMAGe~\cite{koh2023grounding} feeds CLIP~\cite{ramesh2021zero} embeddings to OPT~\cite{zhang2022opt} for image understanding and retrieval.
However, it requires backpropagation through the LLM and does not support image synthesis.
This work enables a standalone frozen LLM to understand and generate other modalities which are unseen in training.

\paragraph{Few-shot learning with LLMs.}
%In few-shot learning, a model learns a new task from only a few examples.
In-context learning~\cite{brown2020language,chowdhery2022palm,googlepalm2} facilitates LLMs for few-shot learning via the text interface without parameter updates.
% was popularized along with the GPT-3 model , as a way for language models to learn new tasks from only a few examples. 
% In-context learning can be achieved with zero parameter updates, and 
%
%It has yielded competitive results to their fine-tuned counterparts 
This approach is commonly employed to assess the performance of LLMs on numerous NLP benchmarks, \eg,  classification and question answering~\cite{wang2019superglue}, mathematical reasoning~\cite{lewkowycz2022solving}, and code generation~\cite{yin2022natural}, which yields competitive results to their fine-tuned counterparts.
However, existing few-shot vision-language understanding and generation frameworks~\cite{alayrac2022flamingo,koh2023grounding} still require LLM parameter updates. 
In contrast, our work inherits the in-context learning ability from frozen LLMs.

\section{Progressive In-Context Decoding with LLMs}
% \subsection{}
% \label{sec:prompting}
% \vspace{-2mm}
While our method is more effective when backpropagating through LLMs by prompt~\cite{lester2021power} or adapter tuning~\cite{houlsby2019parameter,hu2021lora}, this chapter focuses on verifying the plausibility in an extreme setting of in-context learning~\cite{brown2020language}.
We demonstrate that LLMs are capable of performing new tasks in foreign modalities without any parameter updates.
%While our method is best with updating the parameters LLMs through back propagation such as finetuning or prompt tuning, LoRA, 
% We adopt the in-context learning~\cite{brown2020language} framework for multimodal understanding and generation tasks with LLMs without parameter updates.
Specifically, a set of $K$ examples $\{(\rvu^i, \rvv^i)\}_{i=1}^K$ are fed to the LLM to learn a new task and answer a query $\hat{\rvu}$ with
% \vspace{-1mm}
\begin{equation}
    \hat{\rvv} \sim \pllm(\cdot \mid \hat{\rvu}; \{(\rvu^i, \rvv^i)\}_{i=1}^K). \label{eq:icl}
\end{equation}
Sampling $\hat{\rvv}$ by a single-pass autoregressive decoding is suboptimal due to the distributional shift in the representation and the presence of exceptionally long sequences, \eg, an image is quantized into over 500 tokens. 
% It is straightforward to generate textual $\hat{\rvv}$ in a single pass of autoregressive decoding, where the LLM has a strong prior about the output distribution.
% However, token sequences from foreign modalities have much smaller likelihoods in the LLM due to the distribution difference and the greater lengths.
To this end, we use a progressive decoding method.
%to effectively generate such sequences.

% \vspace{-4mm}
\subsection{Method Formulation}
\label{sec:prompting}

\paragraph{Progressive generation.}
We generalize \cref{eq:icl} into a multi-pass process, where the LLM learns to generate one segment of the target sequence at a time.
The segment generated from the $t$-th pass is
% Let $s_t$ denote the number of tokens generated in $t$ steps.
% At each step, we sample a segment of the full sequence
\begin{equation}
    \hat{\rvv}_t \sim \pllm(\cdot \mid [\hat{\rvu}, \hat{\rvv}_{<t'}]; \{([\rvu^i, \rvv^i_{<t'}], \rvv^i_t)\}_{i=1}^K),
\end{equation}
where $[\cdot, \cdot]$ indicates concatenation.
$t'$ controls the length of previous segments to condition on, with two common cases:
%\kevin{This is very unclear}
(1) a progressive autoregressive (PAR) process with $t'=t$, where each decoded segment conditions on all previously decoded ones;
(2) a progressive non-autoregressive (PNAR) process with $t'=0$ to sample each segment independently, which greatly reduces the sequence length requirement for the LLM.
% Note that the generation within each segment is always an autoregressive decoding procedure of the LLM.
%
In practice, we use PAR to generate the first few token layers given task-specific conditions, followed by PNAR to generate the remaining token layers conditioned on the previous layers in an unconditional latent refinement process.

\begin{figure}[t]
\centering
\includegraphics[width=\columnwidth]{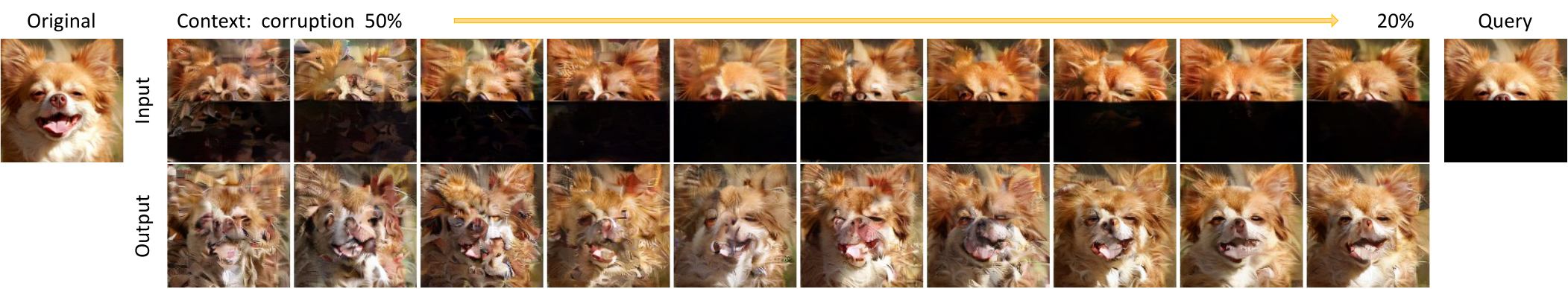}
\caption[An example of in-context denoising]{\textbf{An example of in-context denoising}. The context comprises images randomly corrupted in the token space, gradually ranging from 50\% to 20\%.}
\label{fig:incontext_denoising}
% \vspace{-5mm}
\end{figure}

% \vspace{-4mm}
\paragraph{In-context denoising.}
The learning capacity of an in-context setup is far from sufficient for the entirety of a foreign modality $\gM$. So far, there have been no successful attempts in the literature demonstrating that a frozen LLM can generate image content.
For low-resolution images, LLMs can produce images directly using in-context learning, as will be demonstrated with 32$\times$32 MNIST images~\cite{deng2012mnist}. For higher resolutions, the context length restricts the number of examples. For instance, a context window of 8k tokens can only hold less than a dozen 128$\times$128 images.
Therefore, we operate in a denoising subspace to synthesis beyond 32$\times$32 resolution.
% Therefore, we operate in a denoising subspace to achieve generation.
%
%\kevin{This is best explained by referring to an example image}
Take the image-to-image task in \cref{fig:incontext_denoising} as an example.
The provided context are images randomly corrupted in the token space by $\epsilon(\cdot; r)$, where the corruption ratio $r$ follows a cosine schedule~\cite{chang2022maskgit}.
\begin{equation}
    % \rvv^i \sim \epsilon\big(\mathit{\modelname{}}(\rmI); r_i\big)
    % \rvu^i \sim 
    (\rvu^i, \rvv^i) \sim \Big(\epsilon\big(\gT(\mathtt{mask}(\rmI)); r_i\big), \epsilon\big(\gT(\rmI); r_i\big)\Big), \rmI \in \gM
\end{equation}
where $\gT(\cdot)$ represents the \modelname{} tokenizer and $\gM$ is a small set of raw images.
%\kevin{This is very hard to read; define dummy variables}\lu{fixed}
$\mathtt{mask}(\cdot)$ zeros out pixels of the real image to create the condition image, such as masking out the bottom half for out-painting. The query $\hat{\rvu}$ is always sampled from $\gM$ without noise $\epsilon$.
To ensure the generation is not simply copying the context, we enforce a minimal corruption rate of 20\% such that no identical image from the context matches the real target image.
\subsection{LLM Prompting}

To generate prompts, we utilize \modelname{} from \cref{chap:spae} to quantize an image, or another non-linguistic modality, into a pyramid of lexical tokens. Subsequently, we flatten the tokens by concatenating them layer-by-layer, following a raster scan, and resulting in a 1-D string. This string, representing the image, is referred to as the \textbf{\emph{SPAE string}} in the following prompts.

We use task-specific prompt templates to facilitate answer generation with LLMs.
The LLM output is always parsed by removing leading and trailing whitespace or newline characters.

%Our goal is to model an image, or some other non-linguistic modality (\eg, video or audio)
% are flattened into an \textbf{\emph{SPAE string}} involves:
% \begin{enumerate}[nosep]
%     \item Going from layer 1 to layer $D$.
%     \item Raster scan within each layer.
%     \item Text concatenation without any separators.
% \end{enumerate}

\paragraph{Image classification with GPT 3.5.}
We use the same prompt template as LQAE~\cite{liu2023language} to interact with GPT 3.5.
For a 2-way 1-shot classification between class \emph{lion} and \emph{vase}, the prompt is
\begin{small}
\begin{Verbatim}
For each of the following input output pairs, output is one of 
[`lion', `vase']
###
Input: <SPAE string from a lion image>
Output: lion
###
Input: <SPAE string from a vase image>
Output: vase
###
Input: <SPAE string from the query image>
Output:
\end{Verbatim}
\end{small}
We use greedy decoding to get a maximum of 7 tokens from GPT 3.5.

\paragraph{Image classification with PaLM 2.}
We use the original miniImageNet~\cite{vinyals2016matching} format with PaLM 2.
The prompt looks like
\begin{small}
\begin{Verbatim}[commandchars=\\\{\}]
Answer with "lion" or "vase".

<SPAE string from a lion image>
This is a lion


<SPAE string from a vase image>
This is a vase


<SPAE string from the query image>
\textcolor{gray}{What is this?  # Only used in 5-way 3/5-shot setups}
This is a
\end{Verbatim}
\end{small}
We use greedy decoding to get a maximum of 4 tokens from PaLM 2.

\paragraph{Image captioning.}
We use greedy decoding to get a maximum of 20 tokens before the first newline character with the following prompt:
\begin{small}
\begin{verbatim}
Generate a caption sentence based on words describing an image.

Q: <SPAE string from image 1>
A: <Caption for image 1>

Q: <SPAE string from image 2>
A: <Caption for image 2>

Q: <SPAE string from the query image>
A: 
\end{verbatim}
\end{small}

\paragraph{Visual question answering.}
We use greedy decoding to get a maximum of 4 tokens before the first newline character with the prompt template as
\begin{small}
\begin{Verbatim}
Answer with a single word.

C: <SPAE string from image 1>
Q: <Question for image 1>
A: <Answer for image 1>

C: <SPAE string from image 2>
Q: <Question for image 2>
A: <Answer for image 2>

C: <SPAE string from the query image>
Q: <Question for the query image>
A: 
\end{Verbatim}
\end{small}

\paragraph{Image/video generation with AR decoding.}
For image or video generation tasks, the condition can be a text string or an SPAE string of a condition image.
Suppose we use AR decoding with a stride of 4 tokens.
At the 4th step, the prompt looks like
\begin{small}
\begin{Verbatim}
Learn a new language and predict the 4 tokens following the examples.

C:<condition for image 1>
Q:<SPAE string (token 1-12) for image 1>
A:<SPAE string (token 13-16) for image 1>

C:<condition for image 2>
Q:<SPAE string (token 1-12) for image 2>
A:<SPAE string (token 13-16) for image 2>

C:<condition for the query>
Q:<SPAE string (token 1-12) for the generated image from previous steps>
A:
\end{Verbatim}
\end{small}
We use PaLM 2 to generate 8 predicted sequences for the next 4 tokens, starting with a temperature $T_0=0$.
We use the sentence piece~\cite{kudo2018sentencepiece} tokenizer to tokenize the output string.
If all predictions are shorter than 4 tokens, we retry the LLM prediction with a higher temperature.
At the $i$-th retry, the temperature is given by
\begin{equation}
    T_i = \psi \sum_{j=1}^i 2^j
\end{equation}
where $\psi=0.01$ is used.

\paragraph{Image/video generation with NAR decoding.}
We use NAR decoding to generate \modelname{} layer 6 conditioned on layer 1-5.
With a stride of 16, the prompt at the 3rd step looks like
\begin{small}
\begin{Verbatim}
Predict the outputs following the examples.

Q:<SPAE string from layer 1-5 for image 1>
A:<SPAE string from layer 6 (token 33-48) for image 1>

Q:<SPAE string from layer 1-5 for image 2>
A:<SPAE string from layer 6 (token 33-48) for image 2>

Q:<SPAE string from layer 1-5 for the generated image from AR decoding>
A:
\end{Verbatim}
\end{small}
We use PaLM 2 to generate 8 predicted sequences for the next 16 tokens.
If the sentence piece parsing fails, we retry with the same temperature schedule as in AR decoding.

\section{Experimental Results}
\paragraph{LLM setups.}
The PaLM 2-L API~\cite{googlepalm2} is used for in-context learning with \modelname{}$_\text{PaLM}$.
For a fair comparison with prior works~\cite{liu2023language}, we use \modelname{}$_\text{GPT}$ with the GPT 3.5 \texttt{text-davinci-003} API
%\footnote{
(\url{ https://platform.openai.com/docs/models/gpt-3-5}).

\paragraph{MNIST \modelname{}.}
In addition to the models introduced in \cref{chap:spae}, we train another \modelname{} on the MNIST~\cite{deng2012mnist} dataset with the same architecture setup.
We pad the handwritten digit images from 28$\times$28 to 32$\times$32 pixels, which are then encoded into 4$\times$4 embeddings.
Each image is represented by 37 tokens organized in four layers, with sizes of 1$\times$1, 2$\times$2, 4$\times$4, and 4$\times$4.
We replace the CLIP image embedding with the CLIP text embedding of the label for the semantic loss.
The model is trained for 10k steps with a batch size of 256.
For in-context generation, AR decoding with a stride of 4 is used to produce all 37 tokens.

\subsection{Main Evaluation}
\label{sec:eval}

\begin{table}[p]
    % \vspace{-4mm}
    \centering
    % \small
    \resizebox{1.0\textwidth}{!}{%
    \begin{tabular}{@{}lllcccccccc@{}}
    \toprule
    \multicolumn{3}{r}{Task Induction} &    & \checkmark & \checkmark& \checkmark& \checkmark& \checkmark& \checkmark & \multirow{3}{*}{Avg} \\
    Method &\# Layers & \multicolumn{1}{r}{Inner Shots}   & 1    & 1    & 3    & 5    & 1    & 1    & 1    &                      \\
    & :\ \# Tokens & \multicolumn{1}{r}{Repeats}         & 0    & 0    & 0    & 0    & 1    & 3    & 5    &                      \\
    \midrule
    \multicolumn{5}{l}{\textit{2-Way Classification}} \\
    Frozen~\cite{tsimpoukelli2021multimodal} &- &   & 1.7  & 33.7 & 66   & 66   & 63   & 65   & 63.7 & 51.3                 \\
    LQAE~\cite{liu2023language} & 1: 256 & GPT 3.5 & 1.5  & 35.2 & 68.2 & 69.8 & 68.5 & 68.7 & 65.9 & 53.97                \\
    $\emph{\modelname{}}_\text{GPT}$ (ours) &2: 5& GPT 3.5 & \textit{5.3} &  \textit{77.2}    &   \textit{84.4}   &    \textit{86.0}  & \textit{79.4} & \textit{77.2}  & \textit{77.1}    &    \textit{69.51}                  \\  \hdashline
    \emph{\modelname{}}$_\text{PaLM}$ (ours) & 1: 1 & PaLM 2  & \textbf{34.8} & 77.2 & 81.2 & 80.3 & 74.0 & 73.2 & 71.5 & 70.31                \\
    \emph{\modelname{}}$_\text{PaLM}$ (ours) & 2: 5 & PaLM 2  & 32.2 & 84.0 & 88.5 & 88.4 & \textbf{85.1} & 83.6 & 82.4 & 77.74                \\
    \emph{\modelname{}}$_\text{PaLM}$ (ours) & 3: 21 & PaLM 2  & 27.9 & \textbf{84.8} & \textbf{92.5} & \textbf{92.6} & 84.8 & \textbf{85.2} & \textbf{85.4} & \textbf{79.03}                \\ 
    \emph{\modelname{}}$_\text{PaLM}$ (ours) & 4: 85 & PaLM 2  & 22.8 & 81.1 & 91.4 & 90.4 & 82.6 & 84.3 & 84.7 & 76.76               \\
    \emph{\modelname{}}$_\text{PaLM}$ (ours) & 5: 341 & PaLM 2  & 21.2 & 77.4 & 88.0 & 79.1 & 84.8 & 74.0 & 76.1 & 71.51               \\
    \emph{\modelname{}}$_\text{PaLM}$ (ours) & 6: 597 & PaLM 2  & 21.8 & 73.8 & 70.8 & 62.4 & 64.8 & 62.1 & 58.6 & 59.19               \\ \hdashline
    \emph{\modelname{}}$_\text{PaLM}^\text{disjoint}$ & 2: 5 & PaLM 2 & 24.8 & 79.8 & 84.5 & 83.7 & 80.8 & 78.5 & 78.4 & 72.93 \\
    \emph{\modelname{}}$_\text{PaLM}^\text{disjoint}$ & 3: 21 & PaLM 2  & 21.4 & 81.4 & 89.2 & 87.9 & 82.6 & 81.7 & 80.6 & 74.98                \\ \midrule
    
    \multicolumn{5}{l}{\textit{5-Way Classification}} \\
    % LQAE~\cite{liu2023language} & 1: 256 & GPT 3.5 & 1            & 15.7          & 35.9          & 36.5          & 31.9          & 36.4          & 45.9          & 29.04                \\ \hdashline
    % \emph{\modelname (Ours)} & GPT-3.5 &              &               &               &               &               &               &               &                      \\
    Frozen~\cite{tsimpoukelli2021multimodal} &- &     & 0.9          & 14.5          & 34.7          & 33.8          & 33.8          & 33.3          & 32.8          & 26.26                \\
    LQAE~\cite{liu2023language} & 1: 256 & GPT 3.5 & 1            & 15.7          & 35.9          & 36.5          & 31.9          & 36.4          & 45.9          & 29.04                \\
    $\emph{\modelname{}}_\text{GPT}$ (ours) &2: 5& GPT 3.5 & \textit{4.3} &  \textit{63.0}    &   \textit{63.4}   &    \textit{60.6}  & \textit{61.9} & \textit{62.1}  & \textit{62.1}    &    \textit{53.91}                  \\ \hdashline
    \emph{\modelname{}}$_\text{PaLM}$ (ours) & 1: 1  & PaLM 2     & \textbf{26.8} & 52.0 & 50.9 & 49.9 & 51.9 & 48.4 & 47.9 & 46.83       \\ 
    \emph{\modelname{}}$_\text{PaLM}$ (ours)  & 2: 5  & PaLM 2     & 23.6 & 64.2 & 68.0 & 69.9 & 63.4 & 62.0 & 60.2 & 58.76       \\ 
    \emph{\modelname{}}$_\text{PaLM}$ (ours)  & 3: 21 & PaLM 2     & 20.2 & \textbf{65.1} & \textbf{73.7} & \textbf{74.3} & \textbf{66.4} & \textbf{67.0} & 66.3 & \textbf{61.86}       \\ 
    \emph{\modelname{}}$_\text{PaLM}$ (ours)  & 4: 85  & PaLM 2     & 16.1 & 58.5 & 67.2 & 69.1 & 64.0 & 66.4 & \textbf{67.4} & 58.39       \\ 
    \emph{\modelname{}}$_\text{PaLM}$ (ours)  & 5: 341  & PaLM 2     & 12.1 & 46.3 & 55.9 & 67.2 & 43.3 & 46.3 & - & -      \\ 
    \emph{\modelname{}}$_\text{PaLM}$ (ours)  & 6: 597  & PaLM 2     & 12.1 & 35.7 & - & - & - & - & - & -      \\ 
    % $\emph{\modelname{}}_\text{PaLM}$ (Ours) & PaLM v2 & \textbf{34.8} & \textbf{84.8} & \textbf{92.5} & \textbf{92.6} & \textbf{85.1} & \textbf{85.2} & \textbf{85.4} & \textbf{80.06}  \\   
    % & \makecell(r){$d=$} & $1$ & $3$ & $3$ & $3$ & $2$ & $3$ & $3$ \\ 
    \bottomrule
    \end{tabular}
    }%
    \caption[Few-shot classification accuracy on the mini-ImageNet benchmarks]{\textbf{Few-shot classification accuracy} on the mini-ImageNet benchmarks.
    \modelname{}$_\text{GPT}$ and \modelname{}$_\text{PaLM}$ are trained using different vocabularies and embedding sources, with different prompt templates for in-context learning.
    They show the broad compatibility of \modelname{} but are not for a comparison between the LLMs.
    The best performance with GPT is in italics while the overall best is in bold. - means value unavailable due to an infeasible sequence length.}
    \label{tab:imagenet_2way_full}
    % \vspace{-4mm}
    \end{table}
    
\begin{table}[p]
\centering
\begin{tabular}{@{}lcccc@{}}
\toprule
Method & Inner Shots & 1 & 3 & 5 \\ \midrule
Frozen~\cite{tsimpoukelli2021multimodal} & & 7.8 & 10.1 & 10.5 \\
$\emph{\modelname{}}_\text{PaLM}$ (ours) & PaLM 2 & \textbf{14.3} & \textbf{15.9} & \textbf{15.1} \\
\bottomrule
\end{tabular}
\caption[Few-shot VQA performance on Real-Fast-VQA]{\textbf{Few-shot VQA performance on Real-Fast-VQA}.}
\label{tab:vqa}
\end{table}

\begin{figure}[tp]
% \vspace{-2mm}
\centering
\includegraphics[width=\columnwidth,trim={0.6cm 0 1.2cm 0.2cm},clip]{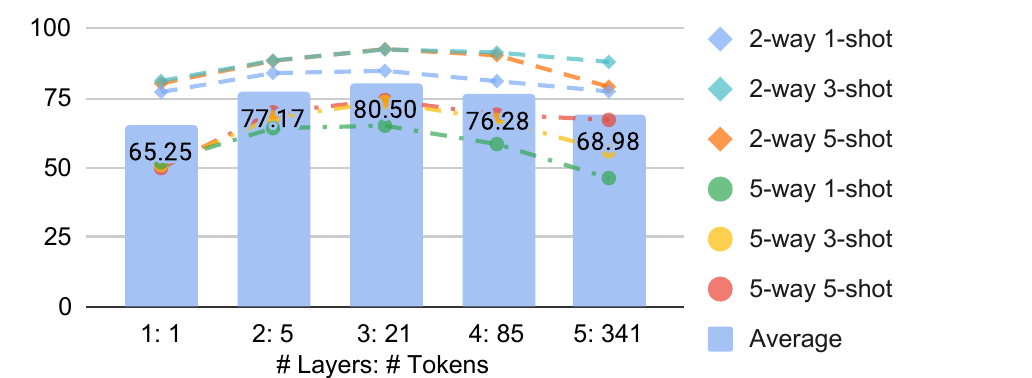}
\caption[Few-shot classification accuracy on mini-ImageNet]{\textbf{Few-shot classification accuracy} on mini-ImageNet using different \modelname{}$_\text{PaLM}$ layers.}
\label{fig:clf}
% \vspace{-3mm}
\end{figure}

% \vspace{-3mm}
\paragraph{Few-shot image classification.}
We evaluate the in-context image understanding capability with a frozen LLM on the mini-ImageNet~\cite{vinyals2016matching} few-shot classification benchmark.
A set of tokenized images and class labels are fed to the language model as context for classification of a new image. 
Following \cite{tsimpoukelli2021multimodal,liu2023language}, we evaluate 14 settings controlled by four factors regarding the content of each test case: 
(1) task induction: whether including a preamble to specify the output space;
%, \eg, \texttt{"Answer with `dog' or `cat'"}; 
(2) number of ways: the number of categories; 
(3) number of inner shots: the number of unique examples for each category; 
(4) number of repeats: the number of times that each unique example is repeated.

%% \vspace{-4mm}
%\paragraph{Comparison with state-of-the-art.}
We compare \modelname{} with the state-of-the-art methods Frozen~\cite{tsimpoukelli2021multimodal} and LQAE~\cite{liu2023language}.
%The training of both baseline methods requires backpropagation through BERT or LLMs to learn the tokenizer.
%while ours solely relies on precomputing similarity scores using CLIP.
%
As shown in \cref{tab:imagenet_2way_full}, \modelname{}$_\text{GPT}$ consistently outperforms LQAE, both using the same GPT 3.5 model and in-context format, while using only 2\% of its tokens.
\cref{fig:clf} shows the performance trend when using different number of \modelname{}$_\text{PaLM}$ layers across six settings with task induction and 0 repeats.
%, \cf more results in Appendix B.
%
%Our model using a single layer already significantly outperforms the LQAE baseline, suggesting the high semantic similarity of \modelname{} tokens to the images.
%
\modelname{}$_\text{PaLM}$ with 3 layers achieves the best performance which balances between sufficient semantics and an image sequence length that is optimal for LLM in-context learning.
Overall, \modelname{}$_\text{PaLM}$ yields $+25\%$ and $+32\%$ average accuracy improvement over the state-of-the-art on the 2-way and 5-way benchmarks in \cref{tab:imagenet_2way_full}.
%
%These substantial improvements demonstrate that \modelname{} learns a more effective semantic mapping from images to natural languages, which empowers LLMs to conduct image classifications out of the box.

% [To shrink]
% Overall, \modelname{} exhibits superior performance compared to the baseline models. 
% The average accuracy gains on the 2-way and 5-way few-shot classification tasks are $+26.06$ and $+32.82$ respectively over the previous best model LQAE. 

% Moreover, we found the advantage is more pronounced in the context of the more challenging 5-way classification task. 
% It is very interesting that the baseline models lose the few-shot capability (almost 0 accuracy) without the task induction but our model performs not bad in this case. 
% The observation that increased inner shots can improve model performance aligns with our intuition that greater exposure to task-specific information helps the model develop a better understanding of the task. 
% We also investigate the effect of the layer from which we extract the image tokens. 
% Our results suggest that the earlier layers are more conducive to good performance in the classification tasks as the low layer probably concentrates on abstract and high-level information. We find the third layer averagely yields the best results in the few-shot classification. More detailed results can be found in Figure~\ref{fig:clf} to corroborate our findings.

\paragraph{Few-shot visual question answering.}

\cref{tab:vqa} provides quantitative results on the visual question answering (VQA) task.
We compare with the baseline Frozen~\cite{tsimpoukelli2021multimodal} method on the Real-Fast-VQA~\cite{tsimpoukelli2021multimodal} benchmark for few-shot learning. 
As shown, \modelname{} consistently outperforms Frozen. 
Unlike Frozen, \modelname{} training does not require backpropagation through the LLM.

\subsection{Qualitative Studies}
% \vspace{-3mm}
This section explores the capability of a frozen PaLM 2, trained solely on language tokens, in performing multimodal tasks using in-context learning.
% We showcase \modelname{} with generative text 
% visual reasoning tasks to demonstrate in-context multimodal generation with \modelname{} and a frozen PaLM 2-L model.
% All samples are generated with a  model.
We adopt a two-stage decoding process for image generation.
In stage one, we use AR decoding to produce the first 5 \modelname{} layers with task-specific conditions.
Stage two is a task-agnostic NAR decoding process for layer 6 conditioned on the first 5 layers.
% followed by NAR decoding with a stride of 16 for layer 6, which takes a total of 102 steps per image. 
%and 358 steps per video.

\begin{sidewaysfigure}
\centering
% % \vspace{-2mm}
\includegraphics[width=\columnwidth]{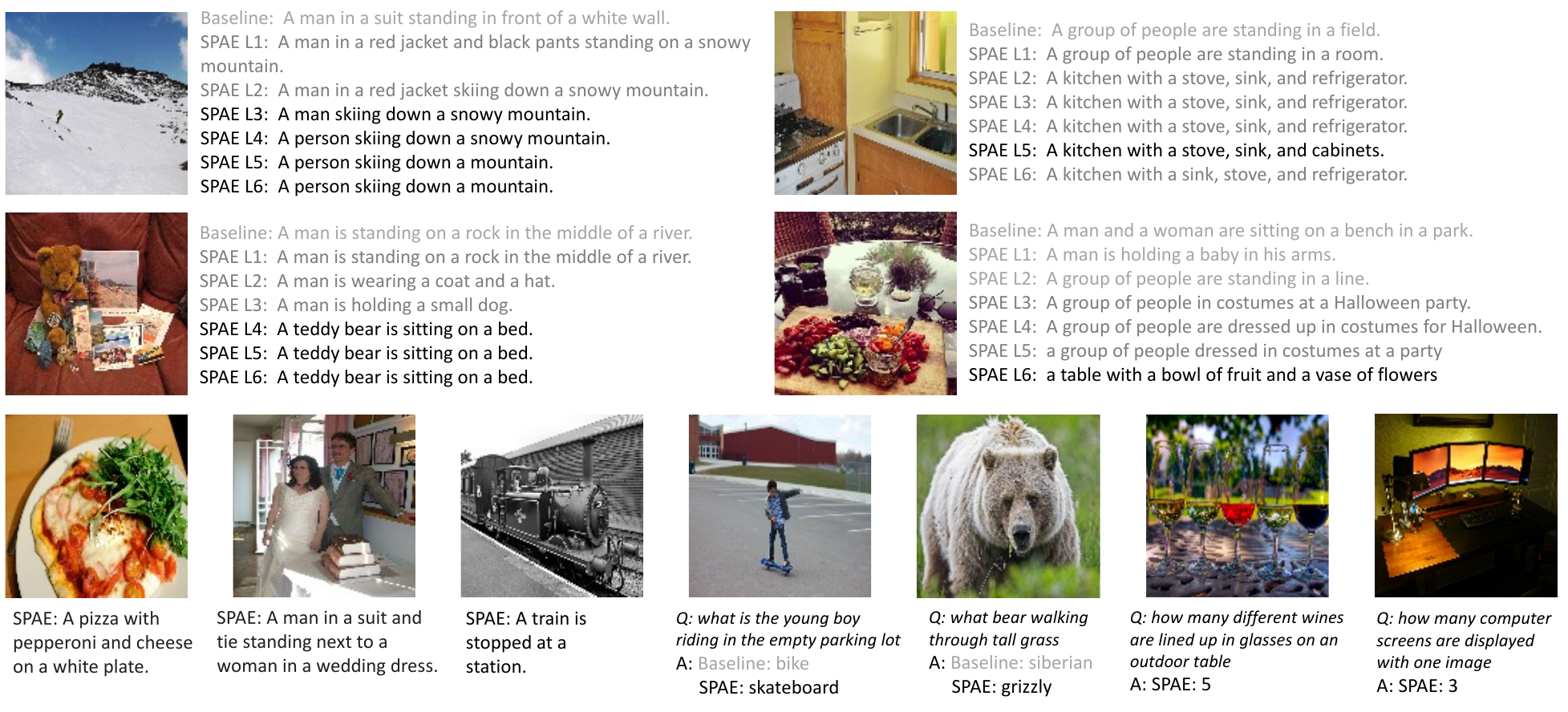}
\caption[Qualitative samples of image-to-text generation]{\textbf{Qualitative samples of image-to-text generation}: image captioning and VQA. We compare between different layers of \modelname{} (L1-L6) and a baseline model without semantic guidance or pyramid SAQ. }
\label{fig:im2tx}
% \vspace{-4mm}
\end{sidewaysfigure}

\begin{figure}[tp]
\centering
\includegraphics[width=\columnwidth,trim={0 0 0 0},clip]{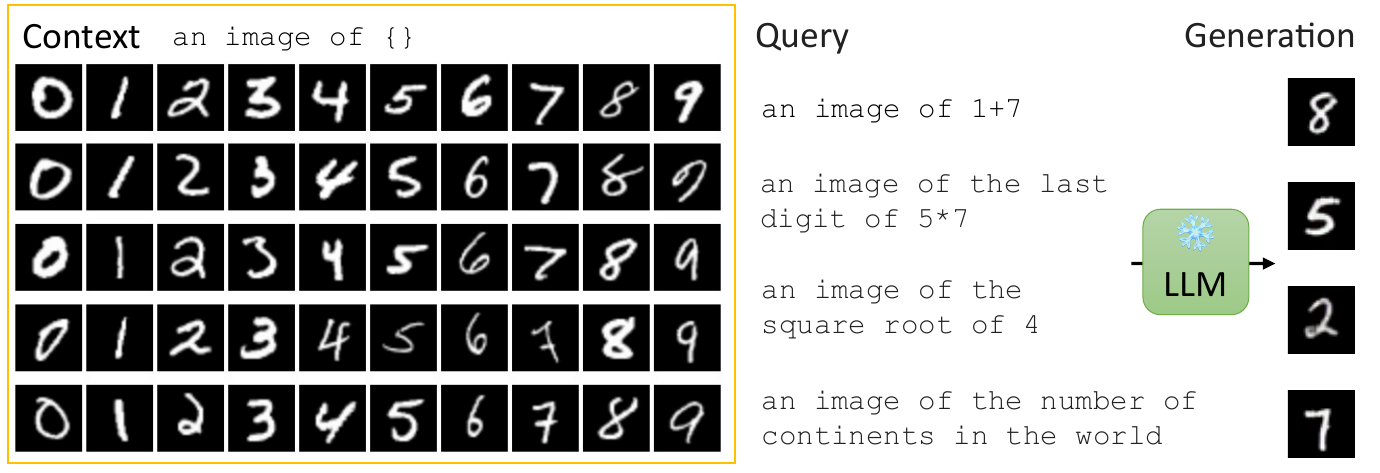}
\caption[Examples of text-to-image generation on MNIST using the frozen PaLM 2 model]{\textbf{Examples of text-to-image generation on MNIST using the frozen PaLM 2 model}. 
%\modelname{} enables PaLM 2 to comprehend handwritten digits and generate images in response to complex queries. 
We use SPAE to tokenize 50 handwritten images as the context and ask PaLM 2, an LLM trained solely on text tokens, to answer complex queries that require generating digit images through SPAE as the output. To achieve this, we use \modelname{} to convert the image into lexical tokens and construct prompts. Then, we ask PaLM 2 to generate suitable answers for the prompts. Finally, we convert the answer tokens back into the pixel space and display them in the figure. Note that the generated digit images do not appear identical to any of the samples provided in the context.}
\label{fig:reason}
% \vspace{-2mm}
\end{figure}

% \vspace{-4mm}
\paragraph{Image to text and VQA.}
% text generation tasks conditioned on an image and optionally other texts: 
We examine two tasks involving visual-text reasoning
(1) image captioning on COCO~\cite{lin2014microsoft} captions; and
(2) visual question answering (VQA) on COCO-QA~\cite{ren2015exploring}.
For both tasks, we provide 10 unique training examples as prompts.
In the case of VQA, 10 different answers are presented to form a 10-way 1-shot setup.

We compare \modelname{} to a baseline model trained with the same frozen language codebook but without the proposed semantic guidance or pyramid SAQ. 
%This baseline is similar to LQAE~\cite{liu2023language} dropping BERT reconstruction
As shown in \cref{fig:im2tx}, when fed with baseline tokens, the LLM randomly hallucinates a caption or guesses an answer simply based on the question.
Similar hallucination can happen if we only use the first two layers of \modelname{} or five words to represent an image, as it provides insufficient context for captioning.
Reasonable captions start to appear with 4 layers or 85 words representing an image, while complex scenes may still need the full 6 layers of 597 words.

% \vspace{-4mm}
\paragraph{LLM generating MNIST images.}
% At the first page of this supplementary document, 
\cref{fig:reason} shows a few image generation examples on the MNIST~\cite{deng2012mnist} dataset.
The frozen LLM learns about handwritten digit images through 50 context samples tokenized by \modelname{} trained on MNIST.
Each sample consists of a preamble \texttt{"an image of $k$"} and the lexical tokens representing an image of digit $k$.
% The LLM learns the \modelname{} strings of handwritten digit images through the 50 context samples.
% Each sample 
Then we can ask the LLM to answer questions with digit images.
Specifically, with a query of \texttt{"an image of 1+7"}, we can use progressive AR decoding with the LLM to produce a token sequence that can be decoded into an image of $8$ by \modelname{}.
We test with complex questions requiring mathematical reasoning or common sense knowledge, and the LLM is able to respond correctly.
In addition, the generated digit images appear different from all context samples.
This demonstrates the cross-modal reasoning capability enabled by \modelname{} and a frozen LLM, with images generated over the text-only interface.

\begin{figure*}[p]
% \vspace{-4mm}
\centering
\includegraphics[width=\linewidth,trim={0.2cm 0 0 0},clip]{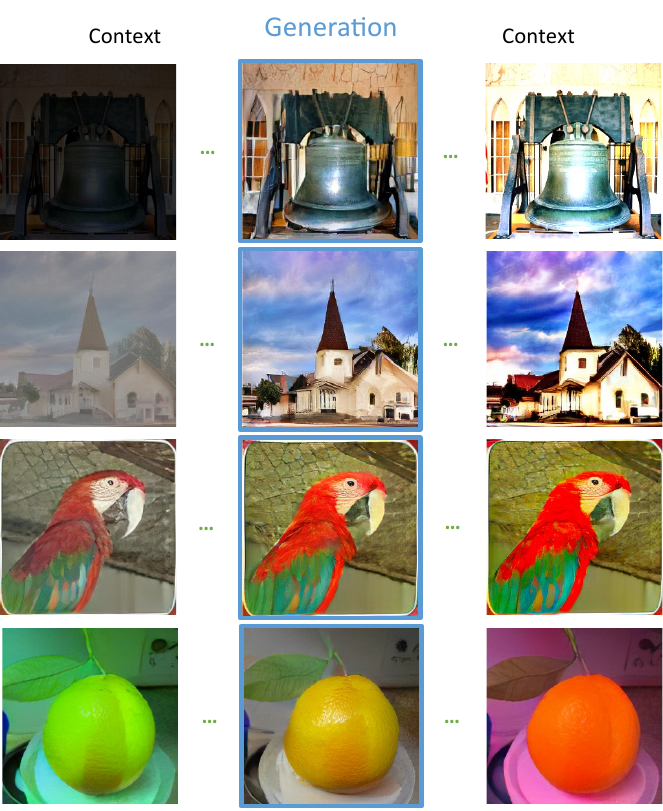}
% \vspace{-5mm}
\caption[Examples of conditional image interpolation at 256$\times$256 resolution]{\textbf{Examples of conditional image interpolation} at 256x256 resolution. The LLM is provided with eight condition images for the interpolation following the setup in \cref{fig:tx2im}. }
\label{fig:gen256}
% \vspace{-4mm}
\end{figure*}

\begin{figure}[tp]
% \vspace{-4mm}
\centering
\includegraphics[width=\columnwidth,trim={0.3cm 0 0 0},clip]{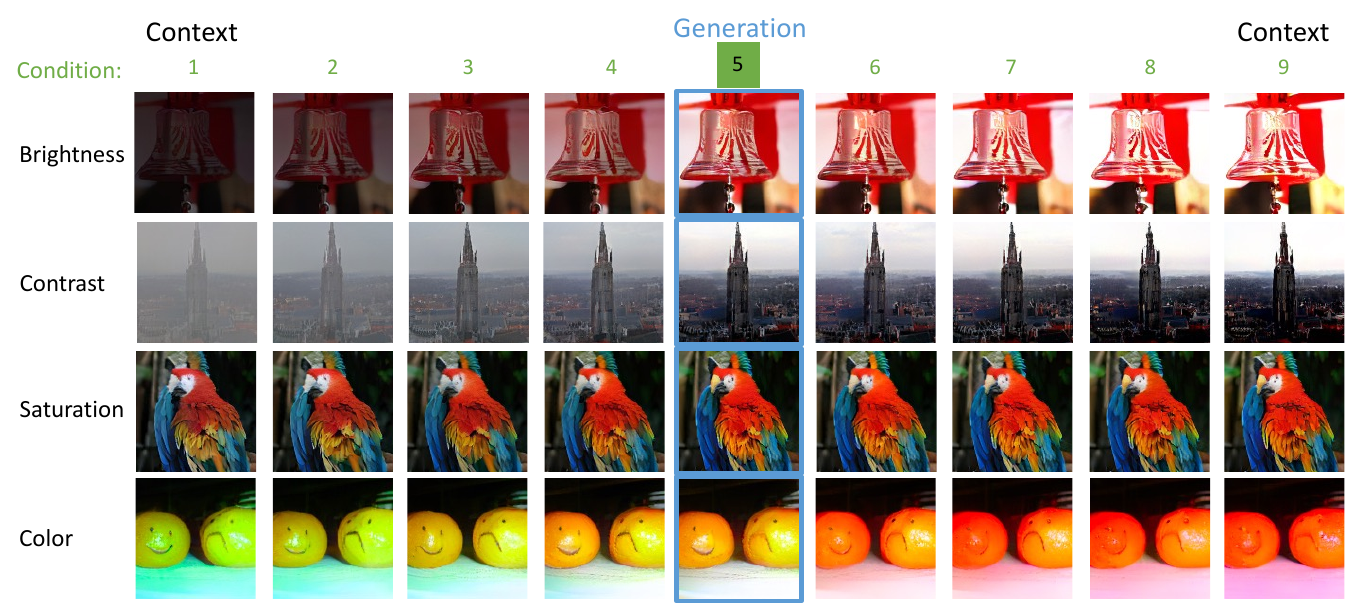}
\caption[Examples of conditional image interpolation]{\textbf{Examples of conditional image interpolation} of different image transformations.}
\label{fig:tx2im}
% \vspace{-1mm}
\end{figure}

\paragraph{Conditional image interpolation.}
To the best of our knowledge, there have been no successful attempts that demonstrate generic image generation capability using a frozen LLM. To this end, we define a very simple setup to explore the interpolation capability of LLM, where the conditions are integers from 1 to 9.
The target images are created with different pixel-space transformations:
\begin{itemize}
    \item Brightness: $[\pm0.8, \pm0.6, \pm0.4, \pm0.2]$.
    \item Contrast: $[\pm0.8, \pm0.6, \pm0.4, \pm0.2]$.
    \item Saturation: $[\pm0.4, \pm0.3, \pm0.2, \pm0.1]$.
    \item Color (RGB): $[(0.6, 1.4, 1), (0.7, 1.3, 1), (0.8, 1.2, 1), (0.9, 1.1, 1),$ \\$(1.1, 0.9, 1), (1.2, 0.8, 1), (1.3, 0.7, 1), (1.4, 0.6, 1)]$
\end{itemize}
Overflow pixels are clipped to $[0, 255]$.
As shown in \cref{fig:tx2im}, images 1-4 and 6-9 are fed as context to produce image 5, where the model interpolates the variable property. 
\cref{fig:gen256} shows generated samples at 256$\times$256 resolution under the same setup.

\paragraph{Conditional image denoising.}
% To the best of our knowledge, there have been no successful attempts that demonstrate generic image generation capability using a frozen LLM.
We also define a denoising setup to explore the capability of LLMs.
% We transition to a more challenging task to generate an image from a condition image.
The conditional image denoising tasks include image outpainting, deblur, inpainting, location translation, rotation, \etc.
%Note the images are generated using the context 
%is a in-context generation using the context . 
Note that, in order to generate images for each task, we utilize 10 pairs of noisy examples with corruption rates ranging from 50\% to 20\%, as discussed in \cref{sec:prompting}. 
We use PAR decoding to produce the first 5 token layers with task-specific conditions, followed by task-agnostic PNAR decoding to fill in layer 6.

\begin{figure}[tp]
\centering
\includegraphics[width=\columnwidth,trim={0 0 0 0},clip]{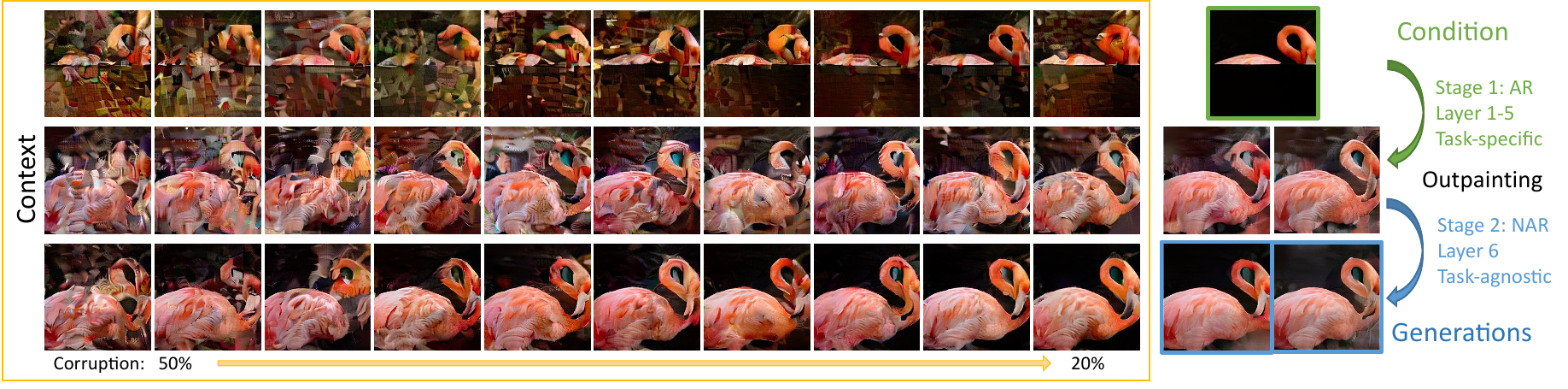}
\includegraphics[width=\columnwidth,trim={0 2.5cm 0 0},clip]{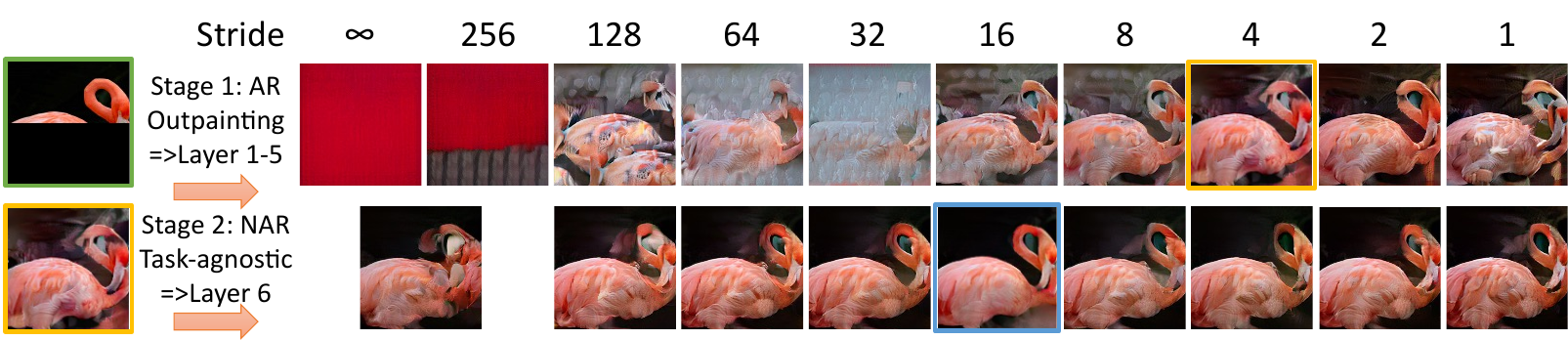}
\caption[Examples of conditional image denoising]{\textbf{Examples of conditional image denoising}. We compare different decoding strides for both stages.
% \textbf{Examples of conditional image denoising}. We compare different decoding strides for both stages. 
Yellow and blue boxes indicate the selected results.
% with an outpainting task.
% Samples from other tasks are shown at the bottom. 
The LLM is provided with ten pairs of noisy examples shown at the top half.
Multiple different outputs can be obtained from the same set of context samples.
% Bottom are samples from other tasks. Each generated image use ten pairs of in-context denoising examples omitted in the figure, which can be found in the Appendix.
}
\label{fig:im2im}
% \vspace{-1mm}
\end{figure}

\begin{figure*}[tp]
    % \vspace{-3mm}
\centering
\includegraphics[width=\linewidth,trim={0.2cm 0 0 0},clip]{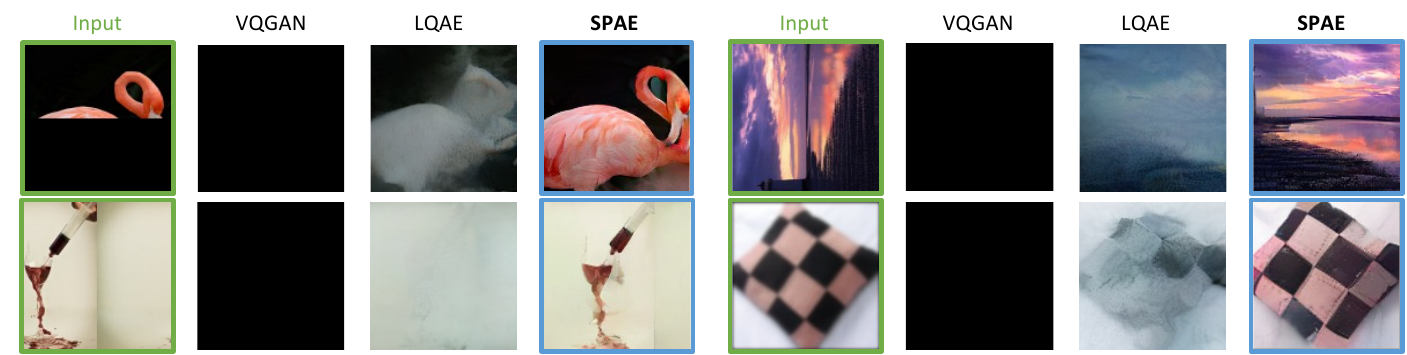}
% \vspace{-5mm}
\caption[Comparison on conditional image denoising with different tokenizers]{\textbf{Comparison on conditional image denoising with different tokenizers}. All models use the same decoding setup with the same ten pairs of prompt images from \cref{fig:im2im,fig:im2im_ctx}.}
\label{fig:imagegen_compare}
% \vspace{-4mm}
\end{figure*}

\begin{figure}[p]
    % \vspace{-5mm}
\centering
% \begin{subfigure}{\textwidth}
% \includegraphics[width=\columnwidth,trim={0 0 0 0},clip]{figures/flamingo_opv.pdf}
% \caption{Outpainting the bottom half. Two generated images are shown.}
% \end{subfigure}
% \begin{subfigure}{\textwidth}
% \includegraphics[width=\columnwidth,trim={0 0 0 0},clip]{figures/flamingo_opv.pdf}
% \caption{Outpainting the bottom half. Two generated images are shown.}
% \end{subfigure}
\begin{subfigure}{\textwidth}
\includegraphics[width=\columnwidth,trim={0 0 0 0},clip]{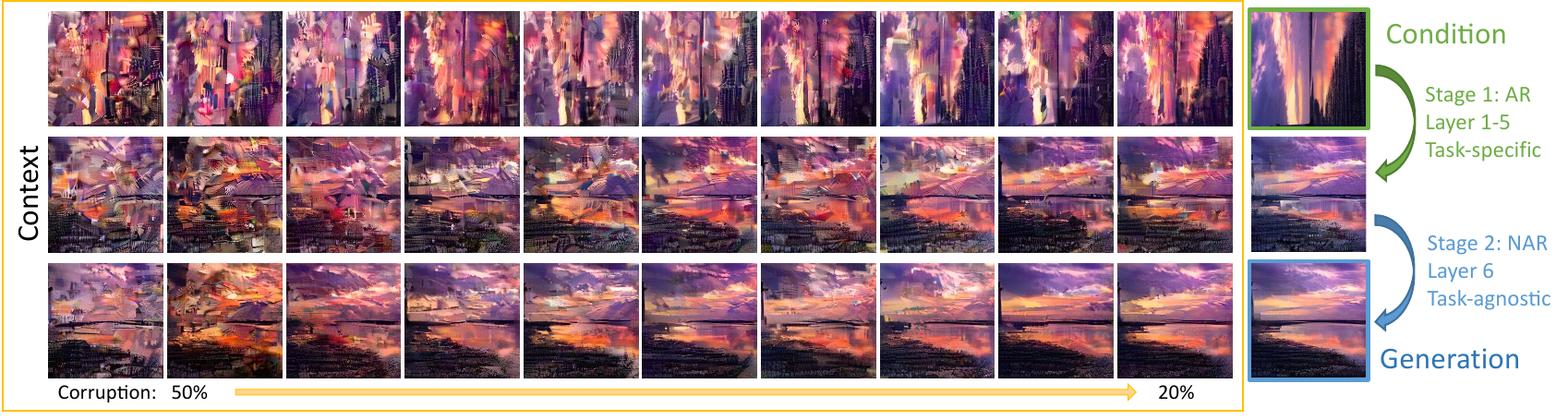}
\caption{Clockwise rotation by 90 degrees.}
\end{subfigure}
\begin{subfigure}{\textwidth}
\includegraphics[width=\columnwidth,trim={0 0 0 0},clip]{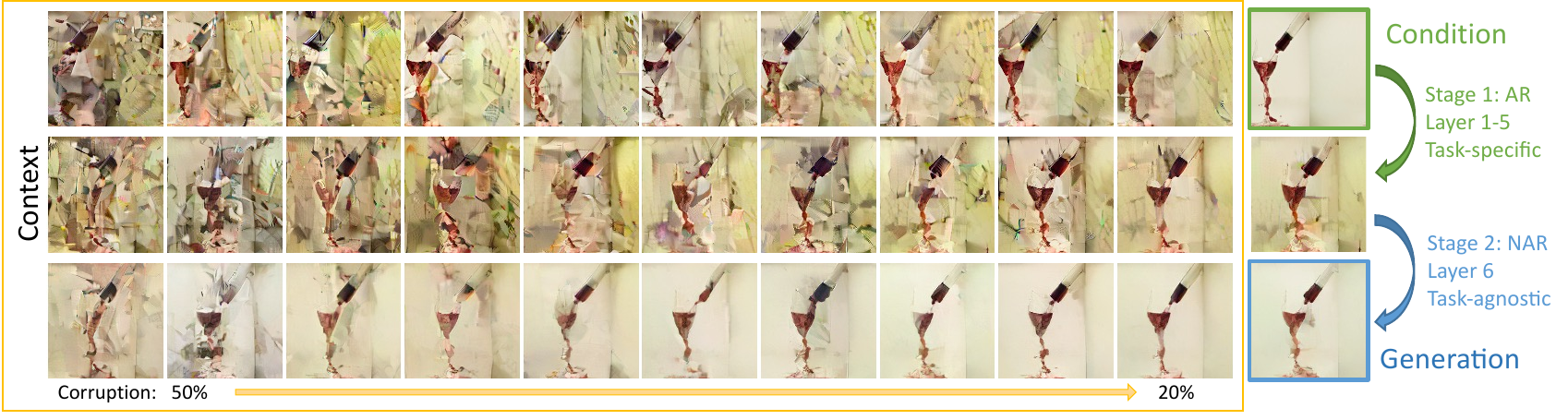}
\caption{Spatial translation to the right.}
\end{subfigure}
\begin{subfigure}{\textwidth}
\includegraphics[width=\columnwidth,trim={0 0 0 0},clip]{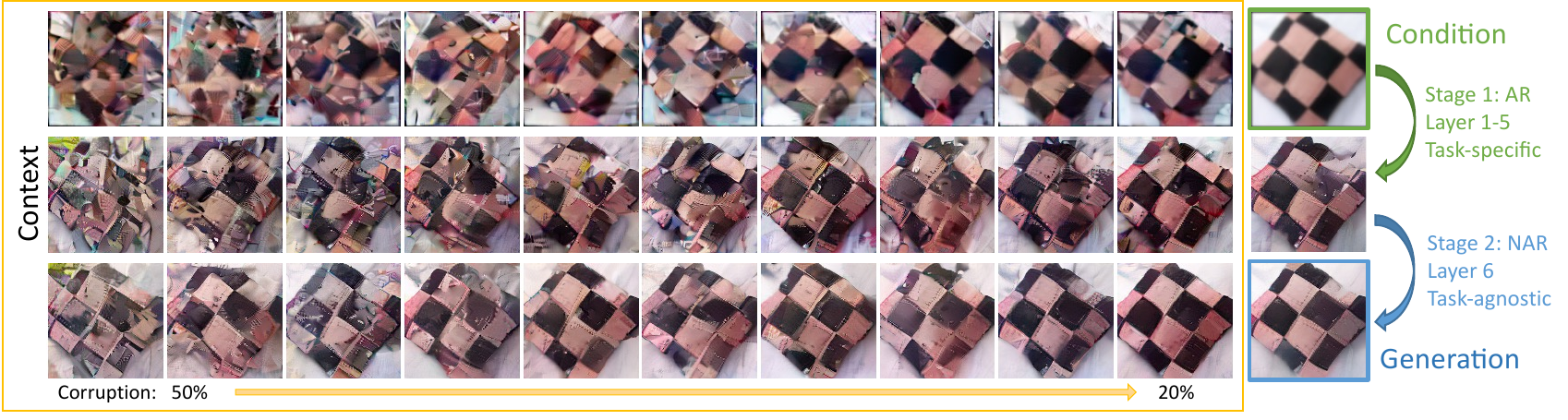}
\caption{Deblur from a gaussian filter.}
\end{subfigure}
% \vspace{-4mm}
\caption[Examples of conditional image denoising]{\textbf{Examples of conditional image denoising}. All input samples for the in-context learning are presented for the examples in \cref{fig:imagegen_compare}.
%  The LLM generates novel images based on the provided context.
}
\label{fig:im2im_ctx}
% \vspace{-4mm}
\end{figure}

\begin{figure}[tp]
% \vspace{-5mm}
\centering
\begin{subfigure}{\textwidth}
\includegraphics[width=\columnwidth,trim={0 0 0 0},clip]{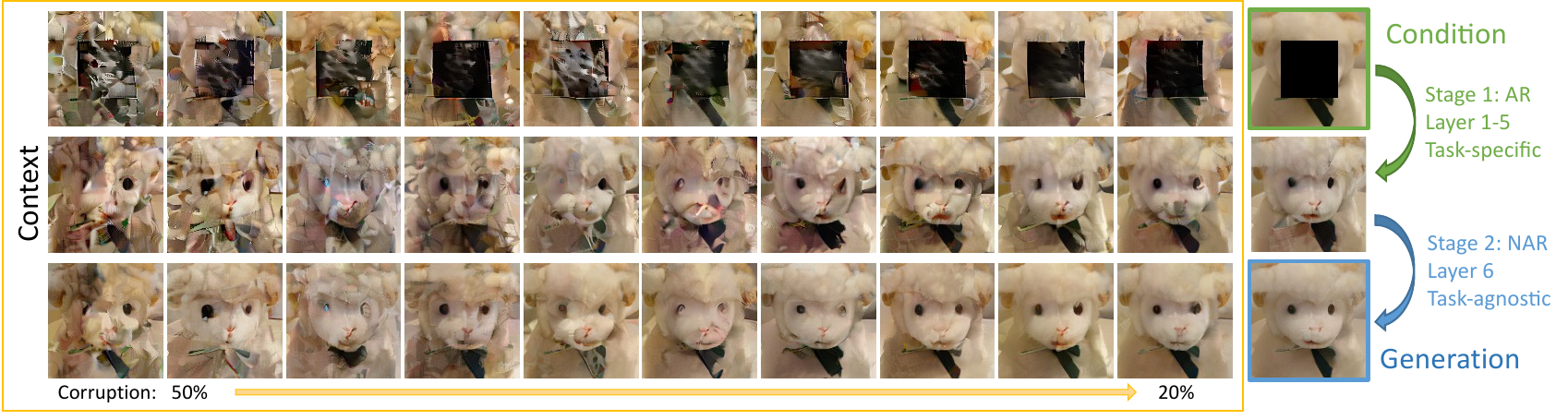}
\caption{Inpainting the center region.}
\end{subfigure}
\begin{subfigure}{\textwidth}
\includegraphics[width=\columnwidth,trim={0 0 0 0},clip]{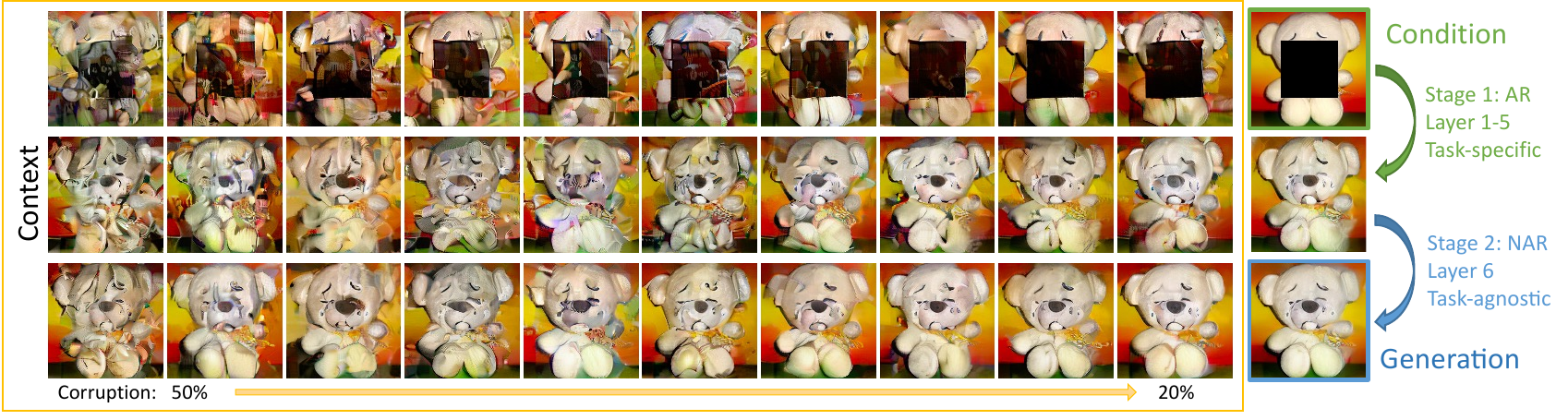}
\caption{Inpainting the center region.}
\end{subfigure}
\begin{subfigure}{\textwidth}
\includegraphics[width=\columnwidth,trim={0 0 0 0},clip]{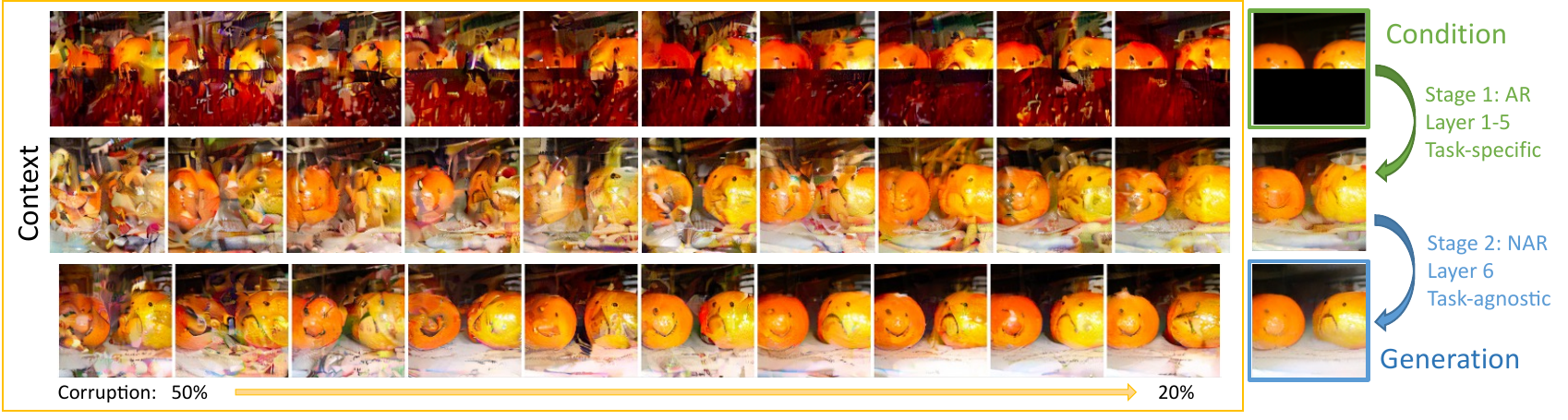}
\caption{Outpainting the bottom half.}
\end{subfigure}
% \begin{subfigure}{\textwidth}
% \includegraphics[width=\columnwidth,trim={0 0 0 0},clip]{figures/sheep_ipc.pdf}
% \caption{Inpainting the center region.}
% \end{subfigure}
% \begin{subfigure}{\textwidth}
% \includegraphics[width=\columnwidth,trim={0 0 0 0},clip]{figures/cup_translation.pdf}
% \caption{Spatial translation to the right.}
% \end{subfigure}
% \begin{subfigure}{\textwidth}
% \includegraphics[width=\columnwidth,trim={0 0 0 0},clip]{figures/scene_rotation.pdf}
% \caption{Clockwise rotation by 90 degrees.}
% \end{subfigure}
% \vspace{-4mm}
\caption[Examples of conditional image denoising]{\textbf{Examples of conditional image denoising}. 
% The LLM generates novel images based on the provided context image pairs.
}
\label{fig:im2im_ctx2}
% \vspace{-4mm}
\end{figure}

The bottom rows of \cref{fig:im2im} compare the output quality from different decoding strides with the same set of context examples.
Single-step decoding with infinity stride fails to produce a reasonable image, which validates the proposed progressive generation approach.
In \cref{fig:imagegen_compare}, we qualitatively compare \modelname{} with baseline methods VQGAN and LQAE using the same in-context denoising procedure. 
As shown, VQGAN fails to produce reasonable images, in part because many words in the LLM output are out of its vocabulary. 
LQAE only produces vague object contours but cannot recover any visual details. On the contrary, SPAE can generate reasonable images. 
\cref{fig:im2im_ctx} visualizes the input pairs for the conditional image denoising samples in \cref{fig:imagegen_compare}, with more examples in \cref{fig:im2im_ctx2}.
% Under the in-context denoising setup, the LLM generates novel images based on the provided context, where multiple different generations can be obtained as shown in \cref{fig:im2im}.

\begin{figure}[tp]
% \vspace{-5mm}
\centering
\includegraphics[width=\columnwidth,trim={0 0 0 0},clip]{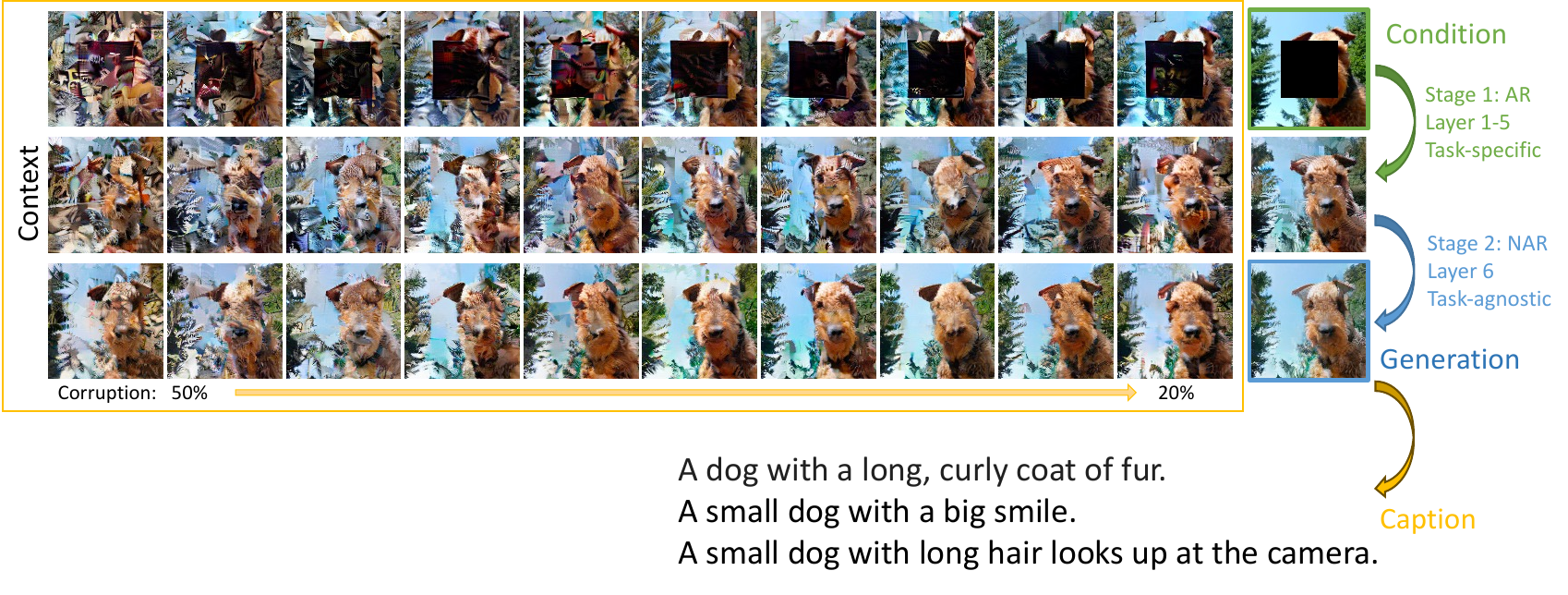}
\caption[Examples of multi-modal outputs]{\textbf{Examples of multi-modal outputs} from the LLM. 
% All input samples for the in-context learning are presented for the example in Fig. 8.
% The LLM generates a novel image with multiple captions based on the provided context.
}
\label{fig:imtx_ctx}
% \vspace{-4mm}
\end{figure}

\newpage
% \vspace{-4mm}
\paragraph{Multimodal outputs.}
\cref{fig:imtx_ctx} shows a task requiring a single LLM to output both image and text, where it first inpaints the center region of an image using in-context denoising and then creates multiple captions for the output image. 
% \paragraph{Generating image and text using a single model.}
% \cref{fig:imtx_ctx} shows a task requiring a single LLM to generate both image and text, where it first inpaints the center region of an image with novel contents using in-context denoising and then creates multiple captions for the generated image. 
% Please refer to the Appendix for the complete context inputs.%, which is omitted in \cref{fig:im2imtx}.

% Lu: showcase a task needs both language and image generation capability.

\begin{sidewaysfigure}[tp]
% \vspace{-5mm}
\centering
\includegraphics[width=\textwidth,trim={0 0 0 0}]{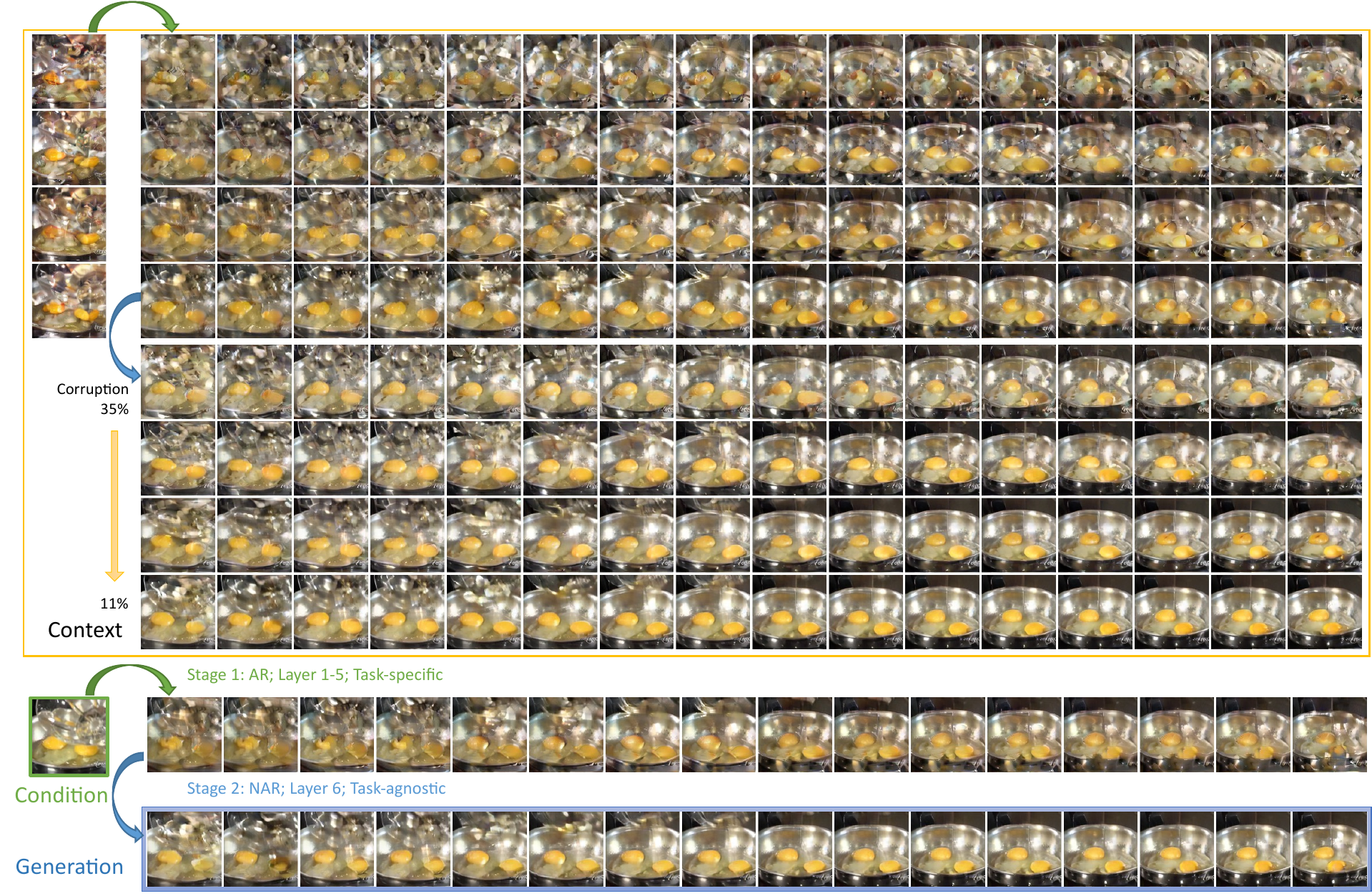}
\caption[Examples of image-to-video denoising]{\textbf{Examples of image-to-video denoising}: frame prediction. We follow the same two-stage generation procedure as in conditional image denoising tasks. Due to the sequence length limit, only four samples can be fit into the context. The generated video clip appear visually different from the context samples, especially around the reflections of the bowl. }
\label{fig:im2vid}
% \vspace{-4mm}
\end{sidewaysfigure}
% \vspace{-4mm}

% \paragraph{Image to video denoising generation.} 
% \cref{fig:im2vid} shows an image-to-video example with the frame prediction task.
% The input is one frame tokenized by the image \modelname{}, while the output is a 16-frame clip tokenized by the video \modelname{}.
% We follow the same two-stage procedure as image-to-image generation, with more steps in each stage to account for the longer sequence.
% Due to the sequence length limit, only four samples can be fit into the context, which limits the generation performance.
% Nevertheless, this demonstrates, for the first time, video generation capabilities of a frozen LLM.
\paragraph{Image-to-video denoising.}
\cref{fig:im2vid} shows an image-to-video example with the frame prediction task using progressive in-context denoising.
The input is one frame tokenized by the image \modelname{}, while the output is a 16-frame clip tokenized by the video \modelname{}.
We follow the same two-stage procedure as conditional image denoising, with more steps in each stage to account for the longer sequence.
Due to the sequence length limit, only four samples can be fit into the context, which limits LLM's performance for this task.
% \mhy{cite Appendix A/B/C (as you did in other paragraphs)? or simply say the supplementary material (and change others to this)}

\section{Summary}
% \section{Highlighted Items for Approval}

% \red{We compare the performance of GPT3.5 and PaLM 2 on the in-context classification tasks.}

% \red{To work with Bard, we used the PaLM 2 vocabulary that is not public accessible. We will not release the vocaulary. But we will need to show an example of 100 words in this vocabulary as shown in \cref{fig:teaser} \cref{fig:pyramid_ex}.}

% \red{We used an internal API to access the Bard/ULM 340B model instead of the public Bard API.}

Our work unveils the untapped potential of frozen Large Language Models (LLMs) in tackling \emph{multi-modal multi-task} understanding and generation involving images and videos, without requiring explicit training on these modalities.
This is achieved with the new method, \modelname{} from \cref{chap:spae}, which converts between visual content and lexical tokens of variable length, imbued with rich semantic meaning.
Our findings show the great potential of harnessing the vast knowledge and reasoning capabilities of LLMs in the field of computer vision, transcending the limitations of language-only tasks.

\paragraph{Limitations.}
The capability of in-context learning is significantly constrained by the acceptable sequence length.
Although our results suggest the plausibility of image generation, the quality and diversity is still far from the recent text-to-image models trained on paired image and text data.

%\emph{Broader impact.}
\paragraph{Broader impact.}
We showcase the untapped potential of frozen Large Language Models (LLMs) in multimodal understanding and generation tasks involving images and videos, without requiring explicit training on these modalities. As an initial research proof-of-concept, we focus on in-context learning, which has limitations in learning context and constrained capabilities. Consequently, there is still a substantial gap to the recent specialized models for text-to-image (\eg, Stable Diffusion) or image-to-text that have been specifically trained using billions of text-image pairs.

The potential impact of our research lies in its influence on future studies, specifically in the area of interacting with pretrained LLMs to enhance their understanding and generation capabilities in the visual modality. For instance, our work can be extended to explore finetuning or adapter tuning of LLMs on large-scale text-image datasets.
Future research in these directions may implicate ethical issues around fairness and transparency, which need to be carefully considered beyond the quality measurements employed here.
We have found that the generated tokens occasionally include slang terms or words that create inappropriate connotations related to the subject depicted in the image or video. Such concerns must be thoroughly considered and effectively addressed prior to deploying this method in real-world applications.

\glsresetall
\cleardoublepage
\chapter{Scalable Generative Multi-Modal Transformer}\label{chap:videopoet}
\graphicspath{{papers/videopoet/}}

\renewcommand{\modelname}{VideoPoet}
\newcommand{\videoframe}[1]{
\includegraphics[width=0.22\textwidth]{#1}}

\paragraph{Overview.}
% \begin{abstract}
Leveraging the representation we have built in \cref{chap:magvitv2}, we present \modelname{}, a model for synthesizing high-quality videos from a large variety of conditioning signals.
\modelname{} employs a decoder-only transformer architecture that enables \emph{multi-task} generation of \emph{multi-modal} outputs -- including videos, images, and audio.
The training protocol follows that of \Glspl{LLM}, consisting of two stages: Pretraining and task-specific adaptation. During pretraining, \modelname{} incorporates a mixture of multi-modal generative objectives within an autoregressive Transformer framework.
The pretrained \Gls{LLM} serves as a foundation that is adapted to a range of video generation tasks.
We present results demonstrating the model's state-of-the-art capabilities in zero-shot video generation, specifically highlighting the generation of high-fidelity motions. 
Project page: \url{http://sites.research.google/videopoet/}.
% \end{abstract}

\clearpage

\section{Motivation}

\begin{sidewaysfigure}
\centering
\includegraphics[width=\textwidth,trim={0 0.6cm 0 0},clip]{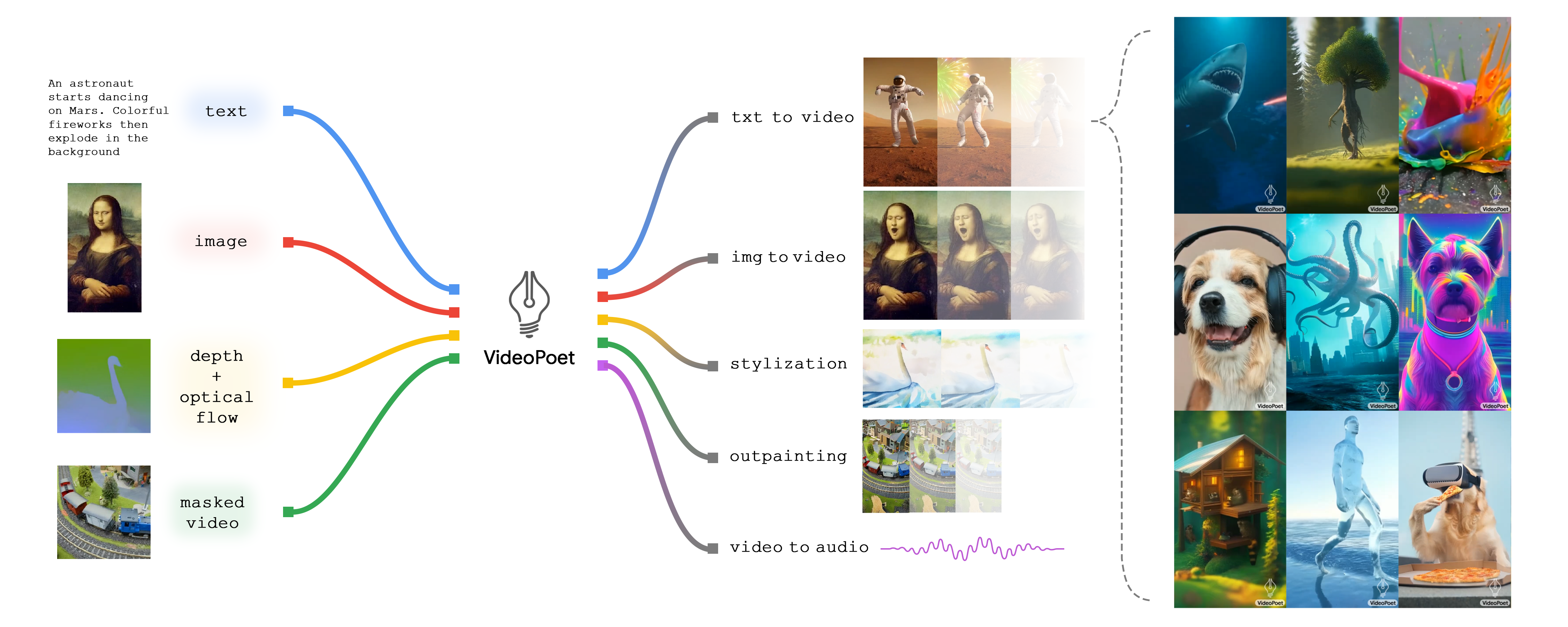}
% \vspace{-2mm}
\caption[\modelname{} overview]{\textbf{\modelname{} overview}: a versatile video generator that conditions on multiple types of inputs and performs a variety of video generation tasks.}
\label{fig:info_graphic}
% \vspace{-4mm}
\end{sidewaysfigure}

% \vspace{-8mm}
% \section{Introduction} 
% \vspace{-1mm}
% \label{sec:intro}

Recently, there has been a surge of generative video models capable of a variety of video creation tasks.
These include text-to-video~\cite{zhang2023show,singer2022make}, image-to-video~\cite{yu2023video}, video-to-video stylization~\cite{chen2023control,chai2023stablevideo,voleti2022masked}, and video editing~\cite{ceylan2023pix2video,wang2023zero,geyer2023tokenflow} among other video applications.
Most existing models employ diffusion-based methods for video generation. 
These video models typically start with a pretrained image model, such as Stable Diffusion~\cite{rombach2022high,podell2023sdxl}, that produces high-fidelity images for individual frames, and then fine-tune the model to improve temporal consistency across video frames. 

While Large Language Models (LLMs) are commonly used as foundation models across various modalities including language~\cite{brown2020language}, code~\cite{li2023starcoder,openai2023gpt}, audio~\cite{rubenstein2023audiopalm}, speech~\cite{agostinelli2023musiclm}, and robotics~\cite{driess2023palm,brohan2023rt}, the diffusion model remains the predominant approach for video generation.
Although early research has demonstrated the effectiveness
of LLMs in text-to-image generation, \eg, DALL-E~\cite{ramesh2022hierarchical}, Parti~\cite{yu2022scaling} and CogView~\cite{ding2021cogview}, and text-to-video, \eg, CogVideo~\cite{hong2022cogvideo}, language models have not reached a level of quality on par with video diffusion models in tasks like text-to-video generation as shown in previous studies~\cite{nash2022transframer,villegas2022phenaki}.
In contrast to training exclusively for text-to-video tasks, the generative model of LLMs in the language domain emphasizes a large pretraining stage to learn a foundation~\cite{bommasani2021opportunities} by examining pretraining tasks that extend beyond text-to-video generation.

A notable advantage of employing LLMs in video generation  
lies in the ease of integrating existing LLM frameworks. 
This integration allows for reusing LLM infrastructure and leverages the optimizations our community has developed over many years for LLMs, including optimizations in learning recipes for model scaling~\cite{brown2020language,chowdhery2022palm}, training and inference infrastructure~\cite{du2022glam}, hardware, among other advancements.
This couples with their flexibility in encoding many diverse tasks in the same model~\cite{raffel2020exploring},
%~\cite{raffel2020exploring,brown2020language,hoffmann2022training,chowdhery2022palm,nash2022transframer}, 
which stands in contrast to most diffusion models where architectural changes and adapter modules are the dominant approach used to adapt the model to more diverse tasks~\cite{zhang2023adding}.
%~\cite{zhang2023adding,chen2023control,esser2023structure,liew2023magicedit}.

In this chapter, we exploit language models for video generation, following the canonical training protocols of LLMs in the language domain.
We introduce \emph{\modelname{}}, a language model for video generation. \modelname{} employs a decoder-only LLM architecture~\cite{googlepalm2,openai2023gpt} that admits image, video, and audio modalities as discrete tokens, each produced by their respective tokenizer.

The training process of \modelname{} consists of two stages: (1) pretraining and (2) task-adaptation.
During pretraining, \modelname{} incorporates a mixture of multimodal pretraining objectives within an autoregressive transformer framework.
After pretraining, the model functions as a versatile multi-task video generation model such as text-to-video, image-to-video, video editing and video-to-video stylization.
These capabilities are inherently integrated into a single LLM, rather than relying on a separate generative model controlled by text prompts~\cite{tang2023any}.
During subsequent task-adaptation, the pretrained model can be further fine-tuned either to enhance its generation quality on the training tasks or to perform new tasks.

%\lu{add a disclaimer about the claim not claiming SOTA on the audio, video-to-video, and others; feasibility exploration of many taks in the same model.}
Experiments show \modelname{}'s state-of-the-art capabilities in generating videos with large and high-fidelity motions. 
Through the powerful capabilities of the transformer architecture, \modelname{} can be straightforwardly trained on a multi-task, multimodal generative objective, allowing for generating consistent and realistic motion driven by text or other prompts. 
Furthermore, \modelname{} can synthesize coherent long videos of up to 10 seconds by autoregressively extending the content, conditioned on the last second of the generated video. 
%
%More results and videos are availalbe in the supplementary material. 
%MH: move this part to experiment al results.
%\textcolor{blue}{See the supplementary materials for the generated videos.}

We also demonstrate that \modelname{} is capable of zero-shot video generation. We use the term ``zero-shot video generation'' as \modelname{} processes new text, image, or video inputs that diverge from the training data distribution.
Furthermore, \modelname{} handles new tasks not included in its training. For example, \modelname{} is able to perform new editing tasks by sequentially chaining training tasks together. %\Cref{sec:capabilities_whole}.
The main contributions of this work are:
\begin{itemize}%[nosep, leftmargin=*]
    \item A method for training a Large Language Model (LLM) specifically for video generation tasks, utilizing tokenized video data that incorporates both text-paired and unpaired video data.
    % \item An approach to video super-resolution that increases spatial resolution within the latent token space using a bidirectional transformer with efficient windowed local attention.
    \item Evaluations and demonstrations to highlight  \modelname{}'s competitive and state-of-the-art performance, especially in generating realistic and interesting videos with motion.
\end{itemize}

\section{Prior Work}

\begin{sidewaysfigure}
\centering
\includegraphics[width=\textwidth]{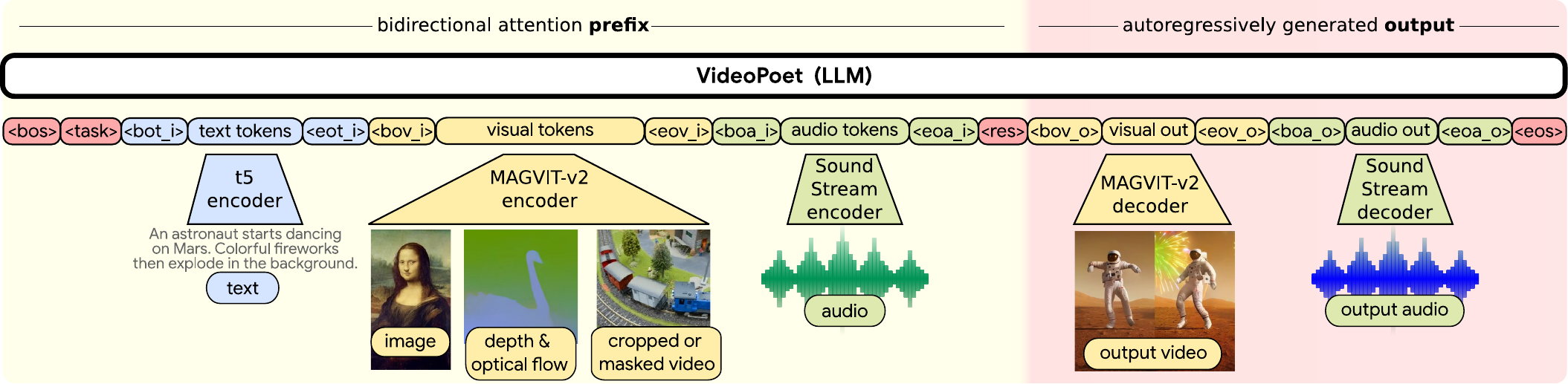}
\caption[Sequence layout for \modelname{}]{
    \textbf{Sequence layout for \modelname{}}. We encode all modalities into the discrete token space, so that we can directly use large language model architectures for video generation.
    We denote special tokens in $<>$ (see Table~\ref{tab:special_tokens} for definitions).
    The modality agnostic tokens are in darker red; the text related components are in blue; the vision related components are in yellow; the audio related components are in green.
    The left portion of the layout on light yellow represents the bidirectional prefix inputs.
    The right portion on darker red represents the autoregressively generated outputs with causal attention.
}
\label{fig:vffm_schematic}
% \vspace-4mm}
\end{sidewaysfigure}

% \vspace-2mm}
% \section{Related Work} 
% \label{sec:related_work}
% % \vspace-1mm}

\paragraph{Video diffusion models.}
Recently, numerous video generation methods use diffusion-based methods for text-to-video~\cite{ho2022imagen,blattmann2023align,zhang2023show,blattmann2023stable,he2023latent,zhou2022magicvideo,wang2023modelscope,ge2023preserve,wang2023internvid,wang2023videofactory,singer2022make,zhang2023show,zeng2023make} and video-to-video editing~\cite{liew2023magicedit,feng2023ccedit,esser2023structure,chen2023control}. 
As video diffusion models are usually derived from text-to-image diffusion models~\cite{ramesh2021zero,saharia2022photorealistic}, additional tasks and modalities are added via inference tricks~\cite{meng2021sdedit}, architectural changes~\cite{esser2023structure,liew2023magicedit} and adapter layers~\cite{zhang2023adding,guo2023animatediff}. 
Although these models are composable after training, they are not trained end-to-end in a unified framework.
Our multitask pretraining strategy in a single model improves performance and provides zero-shot video generation capabilities.

% \vspace-4mm}
\paragraph{Language models for video and image generation.}
Video language models are typically 
 derived from the general family of transformer-based language models~\cite{vaswani2017attention,raffel2020exploring} that easily combine multiple tasks in pretraining and demonstrate powerful zero-shot capabilities.
Image generation language models can generate images autoregressively~\cite{yu2022scaling} or via masked prediction~\cite{chang2022maskgit,chang2023muse}.
Both families have been extended to text-to-video~\cite{hong2022cogvideo,villegas2022phenaki,hu2023gaia,yan2021videogpt} using paired data.
While other text-to-video work with transformers only leverages video-text pairs for training, we also leverage unpaired videos (without text) and the same video for different tasks.
Since video language models can flexibly incorporate numerous tasks~\cite{yu2022magvit,nash2022transframer}, including video-to-video, we extend this family of work to text- and multimodal-conditioned tasks in this work with a synergistic pretraining strategy across various tasks.

% %Parti~\cite{yu2022scaling} is a language model that autoregressively predicts image tokens.
% In contrast to this approach, however, we find that analogously pretraining on video-text pairs aloneise not sufficient forhigh-qualityy synthesis, sowe  devise a multitask pretraining strategy to improve results.
%we alsoe show that these tasks provide powerful zero-shot video editing capabilities.

% Studies of non-autoregressive language models called masked LM include MaskGIT~\cite{chang2022maskgit}-style training for image~\cite{chang2023muse} and video generation~\cite{villegas2022phenaki}.

% % "Generating Images with Multimodal Language Models"
% Gaia-1 model by Hu et al\cite{hu2023gaia} take a language model approach to video generation but is constrained to autonomous driving scenarios.

%  ~\cite{esser2020taming},

% \paragraph{Video Generation} %Text-to-Video}
% The rise of text-to-image models~\cite{ramesh2021zero,saharia2022photorealistic,ramesh2022hierarchical,yu2022scaling} enabled text-to-video models. 

% There are two distinct families of models: diffusion models~\cite{ho2022imagen,blattmann2023align,zhang2023show,tang2023any} and language models~\cite{hong2022cogvideo,villegas2022phenaki}.

% codi-gen~\cite{tang2023any}

%-------------------------------------------------------------------------

% \vspace-4mm}
\paragraph{Pretraining task design in LLMs.}

As language models can easily incorporate multiple training tasks, task selection is an important area of research. 
GPT-3~\cite{brown2020language} and PaLM~\cite{chowdhery2022palm} demonstrate that training LLMs on diverse tasks leads to positive scaling effects on zero- and few-shot tasks.
Other approaches show that masking approaches are a valuable learning target~\cite{hoffmann2022training,yu2022magvit,yu2023language}. 
As the model size grows, training data must grow as well~\cite{hoffmann2022training} to maintain similar performance. 
Our pretraining strategy enables using the same video for multiple training tasks even without paired text. 
This design facilitates training on a large quantity of video-only examples, thereby decreasing the demand for video-text pairs.

\section{VideoPoet Model Design}
% % \vspace{-4mm}
% \section{Model Overview} \label{sec:method}
% % \vspace{-1mm}
We propose an effective method for video generation and related tasks from different input signals by leveraging large language models.
Our model consists of modality-specific tokenizers and a language model backbone~(\cref{fig:vffm_schematic}).
The tokenizers map input data -- \ie image pixels, video frames, and audio waveforms -- into discrete tokens in a \emph{unified} vocabulary.
The visual and audio tokens are flattened into a sequence of integers.
Next, the LLM accepts these tokens as input along with text embeddings, and is responsible for generative multi-task and multimodal modeling. 
As illustrated in \cref{fig:vffm_schematic}, \modelname{} conditions on text embeddings, visual tokens, and audio tokens, and autoregressively predicts visual and audio tokens.
%
% Subsequently, the super-resolution module increases the resolution of the video outputs while refining visual details for higher quality.
% In the following, we discuss design specifics
% which allow our  LLM to generate across video and audio modalities.

%-------------------------------------------------------------------------
% \vspace{-3mm}
\subsection{Tokenization}
\label{sec:tokenization}
% \vspace{-2mm}

We employ the MAGVIT-v2~\cite{yu2023language} tokenizer introduced in \cref{chap:magvitv2} for joint image and video tokenization, and the  SoundStream~\cite{zeghidour2021soundstream} tokenizer for audio. 
Visual and audio vocabularies are concatenated into a unified vocabulary. 
%The unified vocabulary is constructed as follow: the initial 256 codes are reserved for special tokens and task prompts. Subsequently, the next 262,144 codes are allocated for image and video tokenization. This is followed by 4,096 audio codes.
The text modality is represented by embeddings. 
%for its better performance compared with training with text tokens from scratch. 

% \vspace{-4mm}
\paragraph{Image and video tokenizer.}
Visual tokenizer is key to generating high-quality video content, often determining the upper limit of achievable video generation quality~\cite{yu2023language}.
After analyzing existing tokenizers~\cite{esser2021taming,villegas2022phenaki,yu2022magvit,yu2023spae}, we choose the MAGVIT-v2~\cite{yu2023language} tokenizer due to its performance in visual quality and high compression capabilities, which effectively reduce the sequence length required by the LLM, thereby facilitating more efficient and effective learning.
%
%In particular, it represents both images and videos as a sequence of discrete tokens in a unified large vocabulary.
%
Specifically, a video clip is encoded and quantized into a sequence of integers, with a decoder mapping them back to the pixel space. MAGVIT-v2 tokenizes 17-frame 2.125-second 128$\times$128 resolution videos sampled at 8 fps to produce a latent shape of $(5, 16, 16)$, which is then flattened into 1280 tokens, with a vocabulary size of $2^{18}$. % = 262,144
%To facilitate the generation of short-form content for mobile, 
We also tokenize videos into a portrait aspect ratio at 128$\times$224 resolution, producing a latent shape of $(5, 28, 16)$, or 2240 tokens.

% \vspace{-1mm}
We enforce causal temporal dependency, which facilitates the generation of longer videos. 
To jointly represent images and videos, we encode the initial frame of a video or a static image into tokens with a consistent shape of $(1, 16, 16)$.
%which can be used to represent a static image. 
%And then, every 4-frame chunks are encoded into (1,16,16) tokens. These tokens are concatenated on the first (temporal) dimension.
We use the COMMIT~\cite{yu2022magvit} encoding scheme to tokenize the inpainting and outpainting tasks. 
%In simple terms, COMMIT encoding processes the input condition video and the target video differently to avoid information leakage during tokenization. The former involves tokenization of the conditional video with pixel masks applied, while the latter uses tokenization on the entire unmasked video.

% \vspace{-4mm}
\paragraph{Audio tokenizer.} We tokenize audio clips with a pretrained SoundStream~\cite{zeghidour2021soundstream} tokenizer.
We embed 2.125 seconds of audio to produce 106 latent frames with a residual vector quantizer (RVQ) of four levels.
To improve audio generation performance, we transpose the clip before flattening so that the model predicts the full audio clip at each RVQ granularity level before moving on to the finer grained levels.
Finally, each RVQ level has a disjoint vocabulary with each level containing 1,024 codes. 
This results in a combined audio vocabulary size of 4,096 codes.

% \vspace{-4mm}
\paragraph{Text embedding as input.}
%A strong text encoding is important for text-to-video generation. 
Pretrained text representations, in general, outperform training our model by learning text tokens from scratch.
We use pretrained language embeddings from a frozen T5 XL encoder~\cite{raffel2020exploring}. 
For tasks with text guidance, such as text-to-video, T5 XL embeddings are projected into the transformer's embedding space with a linear layer.

% at our model scale so the model can allocate more capacity to generating and understanding vision and audio modalities.

% Therefore, instead of inputting text tokens into the model directly, we first input the tokens into a frozen pretrained T5 XL~\cite{raffel2020exploring} encoder to produce a sequence of text embeddings.

%-------------------------------------------------------------------------
% \vspace{-3mm}
\subsection{Language Model Backbone}
% \vspace{-2mm}
After converting the image, video, and audio modalities into discrete tokens within a shared vocabulary, we can directly leverage a language model to generate videos and audios in the token space.
We use a prefix language model with a decoder-only architecture as the backbone.
By constructing different patterns of input tokens to output tokens during training, we can control the tasks the model is able to perform as explained in \cref{sec:task_prompt_design}.
\subsection{Task Prompt Design}
\label{sec:task_prompt_design}
% \vspace{-2mm}

We design a pretraining task mixture, each with a defined prefix input and output. The model conditions on the prefix, applying the loss solely to the output.
\cref{fig:vffm_schematic} shows a typical input-output sequence layout.
For each task, the input sequence may include three types of values: text embeddings (T5), visual tokens (MAGVIT-v2), and audio tokens (SoundStream). 
The model outputs two types of tokens: visual and audio tokens. To facilitate training, \modelname{} employs \emph{special tokens}, as listed in \cref{tab:special_tokens}. 
In the following, we describe key designs for the task prompts.

% \vspace{-4mm}
\paragraph{Pretraining tasks.}
We consider the following tasks:

\begin{itemize}
    \item {Unconditioned video generation}: Generate video frames without conditioning on an input.
    \item {Text-to-video (T2V)}: Generate video from a text prompt. 
    \item {Video future prediction (FP)}: Given an input video of variable length, predict future frames. 
    \item {Image-to-video (I2V)}: Given the first frame of a video as an input image, predict the future frames. 
    \item {Video inpainting/outpainting (Painting)}: Given a masked video, predict the video with the masked contents filled in. 
    \item {Video stylization}: Given text, optical flow, and depth, predict the video frames (\cref{sec:task_prompt_design}). 
    \item {Audio-to-video}: Given an input audio waveform, predict the corresponding video. 
    \item {Video-to-audio}: Given an input video, predict the corresponding audio waveform.
    \item {Audio-video continuation} (AVCont): given an input frame and its audio, predict the rest of the video and audio.
\end{itemize}
% , with or without the first frame from a video

To indicate the type of task, we condition on the \texttt{<task>} token, which has a unique value for each unique output.
We note that not all input variations need a new \texttt{<task>}; the model adapts to different context signals for identical outputs. 
For instance, text-to-video, image-to-video, and unconditioned video generation share the same \texttt{<task>}.
If a modality is absent in a task, related input/output tokens and special tokens are excluded, shortening the sequence.

% \vspace{-3mm}
\paragraph{Representing an image as a video.}
% In the input sequence, we leave out the end-of-sequence token (\texttt{<eos>}) in the text-to-image pretraining task so the model keeps generating more video tokens at inference time.
% This setting causes the model not to distinguish between video or image data, further facilitating information sharing between the two modalities.
% This results in higher quality initial frames being predicted, which reduces the errors and artifacts in future frames.
In text-to-image pretraining, we omit the \texttt{<eos>} and \texttt{<eov\_o>} tokens from the input sequence, enabling continuous token generation for inference of longer videos.
This approach blurs the boundary between video and image generation tasks, enhancing cross-modality information sharing. 
This design leads to the prediction of higher-quality initial frames and reduces errors and artifacts in subsequent frames.

% \vspace{-3mm}
\paragraph{Video token format.} 
We generate video tokens at two resolutions, 128$\times$128 and 128$\times$224, each available in two lengths: 17 frames and 41 frames, both encoded at 8 frames per second. 
Special conditioning tokens are used to signal the desired resolutions and durations for video generation.
%We combine the two resolutions and two video lengths, which leads to a total of 4 combinations.
Images are a special case of a 1-frame video, which we tokenize at 128$\times$128 resolution.
\subsection{Training Strategy} 
\label{sec:training}
% \vspace{-2mm}
% We train on image-text pairs and video with or without text or audio.
% Both text and sound are noisy and may not match the visual content.
% The model is trained on approximately 2 trillion tokens across all modalities.

% For multi-task training, we use the Alternating Gradient Descent (AGD) method~\cite{akbari2023alternating} to train videos of varying lengths. 
% %
% %MH: not sure what this means. need to revise 
% We design the tasks in the AGD format resulting in a near 0\% padding ratio, lower than that of the packing approach~\cite{raffel2020exploring}. 
% %
% This is accomplished by grouping tasks by sequence length and alternately sampling one group at each iteration.
% %
% %MH: not sure what this menas
% Since sequence lengths are fixed and vary significantly across tasks, \eg, first frame and long video generation, we achieve efficient training with minimal padding.

We find that sampling from image and video datasets uniformly across time can lead to suboptimal results, as training on images can enhance the model's understanding of objects but does not capture any motions that are represented in video data.
Thus, we devise a two-stage pretraining strategy, where we augment our sampling weights to sample  image data 90\% of the time and video data 10\% of the time for the first 25\% iterations of training.
We then switch to training on video 90\% and image 10\% for the remaining iterations. 

We fine-tune our pretrained model for enhanced performance on specific tasks or for new task adaptation, such as text-to-video and image-to-video tasks, using a high-quality data subset. This results in improved generation quality, consistent with \cite{zhou2023lima}, and addresses decoding collapse issues, characterized by repetitive token predictions. 
Such fine-tuning not only diversifies outputs but also allows for a higher classifier-free guidance scale~\cite{ho2021classifier}, boosting overall quality. 
%Additionally, we adapt the model for video-to-audio generation.
% After pretraining, we can fine-tune the pretrained model to either improve its performance on specific tasks or to enable it to undertake new tasks, \ie, task adaption. For example, we finetune the model on both text-to-video and image-to-video tasks using a high-quality data subset. We observe improved generation quality which aligns with findings from \cite{zhou2023lima}. Furthermore, we note that the fine-tuned model mitigates the issue of decoding collapse, characterized by the degradation of predictions into repetitive tokens. This improvement not only enhances the model's output diversity but also enables increasing the Classifier-Free Guidance~\cite{ho2022classifier} scale, leading to an overall enhancement in quality. In addition, we also finetune the pretrained model to perform video-to-audio generation.
 \label{sec:videopoet_method}

\section{Experimental Results}

%-------------------------------------------------------------------------

% \vspace{-3mm}
% \section{Experiments} \label{sec:experiments}
% % \vspace{-1mm}

%-------------------------------------------------------------------------
% \vspace{-1mm}
\subsection{Experimental Setup}
% % \vspace{-1mm}

%% \vspace{-5mm}
\paragraph{Training tasks.}
We train the model on a mixture of pretraining tasks as detailed in~\cref{sec:task_prompt_design}.
We finetune a model on a high-quality training subset for text-to-video evaluations, as discussed in \cref{sec:training}.
Unless explicitly stated, we do not finetune on specific tasks for evaluations.

% \vspace{-4mm}
\paragraph{Datasets.} 
\label{sec:exp_data}
We train on a total of 1B image-text pairs and $\sim$270M videos ($\sim$100M with paired text, of which $\sim$50M are used for high-quality finetuning,  and $\sim$170M with paired audio) from the public internet and other sources, \ie around 2 trillion tokens across all modalities. 
%
% Both text and sound are rather noisy and may not match the visual content.
%
The data has been filtered to remove egregious content and sampled to improve contextual and demographic diversity.

% \vspace{-4mm}
\paragraph{Evaluation protocol.}
We employ a zero-shot generation evaluation protocol, 
as the model has not been trained on the training data of target benchmarks.
Specifically, the evaluation benchmark includes two text-to-video generation datasets, MSR-VTT~\cite{xu2016msr} and UCF-101~\cite{soomro2012ucf101}, as well as the frame prediction task on Kinetics 600 (K600)~\cite{carreira2018short}, in which the first 5 frames are provided as the condition to predict the next 11 frames. We also include inpainting and outpainting tasks~\cite{yu2022magvit} on Something-Something V2 (SSv2)~\cite{goyal2017something}. 
% Additionally, we assess stylization tasks using a subset of the DAVIS dataset\footnote{\url{https://davischallenge.org/}}~\cite{Perazzi2016davis}, as further detailed below.

We adopt widely used metrics such as Fréchet Video Distance (FVD)~\cite{unterthiner2018towards}, CLIP similarity score~\cite{wu2021godiva}, and Inception Score (IS)~\cite{saito2020train} for evaluation. 
Note that the specific metrics and evaluation methods vary across different datasets. Detailed information on these variations can be found in \cref{sec:add_details}. 
% \textcolor{black}{
% We include examples of the generated videos in the supplementary materials.}
%More results and videos are availalbe in the supplementary material. 

% \vspace{-2mm}
\begin{sidewaystable}[tp]
\centering
%{-2mm}
% \scriptsize
\begin{tabular}{lcccccc|ccccc}
\toprule
\multirow{4}{*}{Method} & \multicolumn{6}{c|}{Pretraining Tasks}                                                                                                             & \multicolumn{5}{c}{Zero-shot Evaluation Benchmark}                                                                                                                                 \\ \cline{2-12} 
 & \multirow{3}{*}{T2I} & \multirow{3}{*}{T2V} & \multirow{3}{*}{Uncond} & \multirow{3}{*}{FP} & \multirow{3}{*}{Painting} & \multirow{3}{*}{AVCont} & \multicolumn{2}{c}{T2V}                                 & \multicolumn{1}{c}{FP}   & \multicolumn{1}{c}{Inpainting} & \multicolumn{1}{c}{Outpainting}  \\ 
    &                      &                      &                         &                     &                           &                         & \multicolumn{1}{c}{MSR-VTT} & \multicolumn{1}{c}{UCF101} & \multicolumn{1}{c}{K600} & \multicolumn{1}{c}{SSv2}       & \multicolumn{1}{c}{SSv2}     \\
    &                      &                      &                         &                     &                           &    & \multicolumn{1}{c}{CLIPSIM $\uparrow$}   & \multicolumn{1}{c}{FVD $\downarrow$}    & \multicolumn{1}{c}{FVD $\downarrow$}  & \multicolumn{1}{c}{FVD $\downarrow$}        & \multicolumn{1}{c}{FVD $\downarrow$}           \\
\midrule
T2V  && \checkmark&  &  &  & & 0.244&822&759&2,333&2,310                     \\
T2V+I & \checkmark & \checkmark &  &&&& 0.247&1,025&794&2,118&1,916                        \\
SSL & & & \checkmark & \checkmark & \checkmark & \checkmark & 0.226&1,742&700&1,093&1,500 \\
NO T2I &  & \checkmark & \checkmark & \checkmark & \checkmark & \checkmark & 0.235&1,008&755&95&389 \\
ALL & \checkmark & \checkmark& \checkmark & \checkmark & \checkmark & \checkmark& 0.240&1,085&729&127&636 \\ 
\hdashline
ALL (8B) & \checkmark & \checkmark& \checkmark & \checkmark & \checkmark & \checkmark & 0.305	& 355	& 687	& 4.7 &	13.76  \\
\bottomrule
\end{tabular}
\caption[Pretraining task analysis on 300M models]{\textbf{Pretraining task analysis on 300M models.} The top rows list models with 300M parameters, trained on a subset of the data, and are comparable to each other. The last row shows an 8B model trained on the entire dataset. \textbf{T2I} (text-to-image), \textbf{T2V} (text-to-video), \textbf{FP} (frame prediction), \textbf{Painting} (inpainting/outpainting), \textbf{Uncond} (unconditional generation), \textbf{AVCont} (audio-video continuation), and \textbf{SSL} (self-supervised learning).
}
\label{tab:pretraining}
%{-3mm}
\end{sidewaystable}
\subsection{Pretraining Task Analysis}
% \vspace{-2mm}
We investigate the learning capabilities of different combinations of pretraining tasks using a model with 300 million parameters. 
All task combinations are trained using a learning rate of $10^{-3}$ for the same number of steps (300k) with a batch size of 1024.

For the analysis of pretraining tasks, we consider text-to-video (T2V), text-to-image (T2I), and four self-supervised learning (SSL) tasks: frame prediction (FP), central inpainting and central outpainting (Painting)~\cite{yu2022magvit} and audio-video continuation (AVCont) where the model is provided with the first frame and its corresponding audio to predict the subsequent 16 frames and matching audio. 
% %
% We use uniform sampling among the selected tasks,
% %
% When fewer tasks are selected, they are trained more extensively. 
% %
For each video task, we uniformly select 20\% of training samples from a random subset of 50 million videos.
For the text-to-image task, we randomly sample 50 million text-image pairs from our training dataset.
For tasks involving audio, our sampling is exclusive to videos that contain an audio track.

The evaluation results are presented in \cref{tab:pretraining}.
We assess a model across the four tasks within the zero-shot evaluation benchmark: the T2V task on MSR-VTT~\cite{xu2016msr} and UCF 101~\cite{soomro2012ucf101}, FP on K600~\cite{carreira2018short}, and the Painting tasks on SSv2~\cite{goyal2017something}.
In these experiments, we employ a single model to perform all the tasks. 
%The evaluation on K600 and SSv2 are is detailed in~\cref{sec:ssl_task}. And the evaluation on the text-to-video task will be discussed in \cref{sec:exp_t2v}.
The model is not trained on the training data of these evaluation datasets, and thus it is a zero-shot evaluation.

The top rows of~\cref{tab:pretraining} depict each pretraining task configuration of the 300 million parameter model, which are comparable in their setups.
Our evaluation benchmarks span diverse visual domains, posing a challenge to achieving consistent improvement across all of them.
Nevertheless, incorporating all pretraining tasks results in the best overall performance, on average, across all evaluated tasks. 
Additionally, the significant disparity observed in the ``SSL'' row suggests the limitations of self-supervised training and underscores the necessity for text-paired data during training. 
The last row, ``ALL (8B)'', is the model with 8 billion parameters, trained on the pretraining tasks as discussed in \cref{sec:videopoet_method} and utilized significantly more compute. 

\subsection{Model Scaling and Performance}
% %{-2mm}
To analyze model performance versus model scale, we use a subset of the training set without text-paired data and a slightly different task prompt design. 
We evaluate the video generation quality using FVD~\cite{unterthiner2018towards} and audio generation quality using the Fréchet Audio Distance (FAD), which uses the VGGish model as the embedding function~\cite{hershey2017cnn}. 
Both FVD and FAD metrics are calculated using a held-out subset of 25 thousand videos.

\begin{figure*}[h]
\centering
\begin{subfigure}{0.49\textwidth}  
\includegraphics[width=\linewidth]{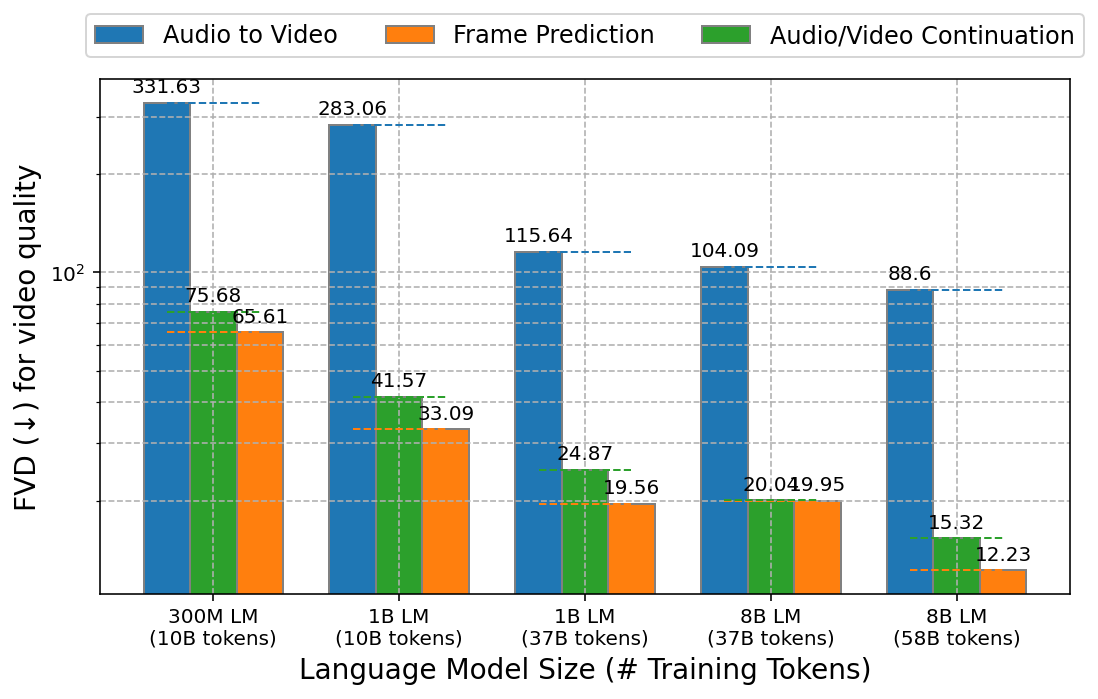}
\subcaption{Video generation quality in FVD ($\downarrow$).}
\end{subfigure}
\begin{subfigure}{0.49\textwidth}  
\includegraphics[width=\linewidth]{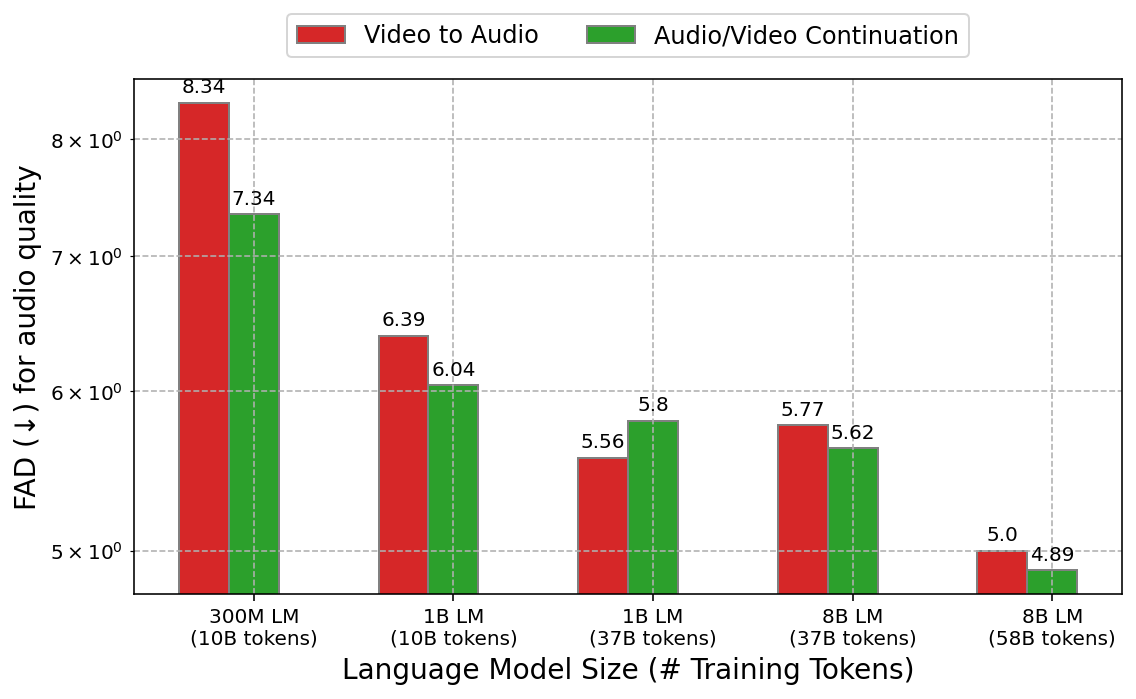}
\subcaption{Audio generation quality in FAD  ($\downarrow$).}
\end{subfigure}
\caption[Effects of model and data scale on video and audio generation quality]{\textbf{Effects of model and data scale on video and audio generation quality}. The performance, depicted on a log-log scale, improves significantly when we scale up the model and training data. 
Language models with 300 million, 1 billion, and 8 billion parameters are trained on datasets comprising 10, 37, and 58 billion visual and audio tokens, respectively.
} 
\label{fig:model_scale}
% %{-6mm}
\end{figure*}

\cref{fig:model_scale} shows that as the model size grows and the amount of training data increases, performance improves across visual and audiovisual tasks.
After obtaining the above results, we retrain our 1B and 8B models using the task design and text-paired training data discussed in \cref{sec:videopoet_method}.
%
% \cref{sec:appendix-sxs} 

\paragraph{Qualitative comparison of 1B and 8B models.} %\label{sec:appendix-sxs}

\begin{figure*}[tp]
% \addtocounter{figure}{-1}

\begin{subfigure}{1.0\linewidth}
\centering
\includegraphics[width=0.11\textwidth]{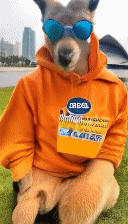}
\includegraphics[width=0.11\textwidth]{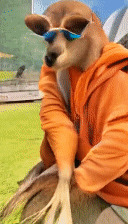}
\includegraphics[width=0.11\textwidth]{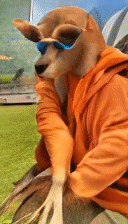}
\includegraphics[width=0.11\textwidth]{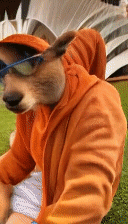}
\hspace{0.03\textwidth}
\includegraphics[width=0.11\textwidth]{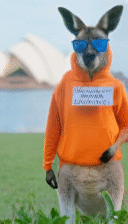}
\includegraphics[width=0.11\textwidth]{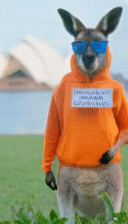}
\includegraphics[width=0.11\textwidth]{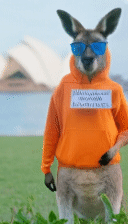}
\includegraphics[width=0.11\textwidth]{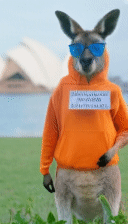}
~
{\raggedright \textbf{prompt:} A portrait photo of a kangaroo wearing an orange hoodie and blue sunglasses standing on the grass
in front of the Sydney Opera House holding a sign on the chest that says Welcome Friends!
\ \\
\ \\}

\end{subfigure}

\vspace{-2mm}

\begin{subfigure}{1.0\linewidth}
\centering
\includegraphics[width=0.11\textwidth]{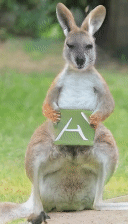}
\includegraphics[width=0.11\textwidth]{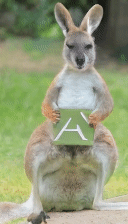}
\includegraphics[width=0.11\textwidth]{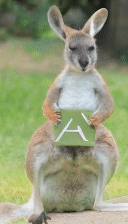}
\includegraphics[width=0.11\textwidth]{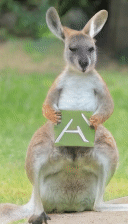}
\hspace{0.03\textwidth}
\includegraphics[width=0.11\textwidth]{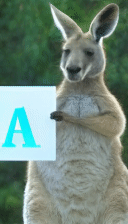}
\includegraphics[width=0.11\textwidth]{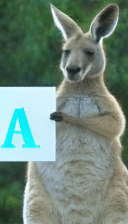}
\includegraphics[width=0.11\textwidth]{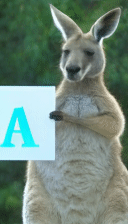}
\includegraphics[width=0.11\textwidth]{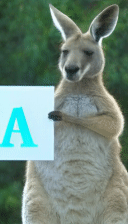}
{\raggedright \textbf{prompt:} A kangaroo holding a sign with the letter A on it
\ \\
\ \\}

\end{subfigure}

\vspace{-2mm}

\begin{subfigure}{1.0\linewidth}
\centering
\includegraphics[width=0.11\textwidth]{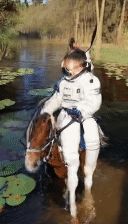}
\includegraphics[width=0.11\textwidth]{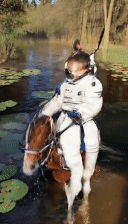}
\includegraphics[width=0.11\textwidth]{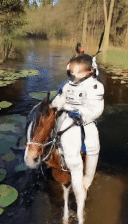}
\includegraphics[width=0.11\textwidth]{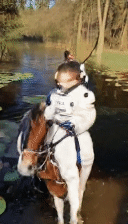}
\hspace{0.03\textwidth}
\includegraphics[width=0.11\textwidth]{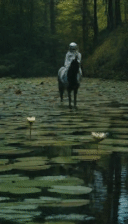}
\includegraphics[width=0.11\textwidth]{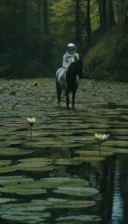}
\includegraphics[width=0.11\textwidth]{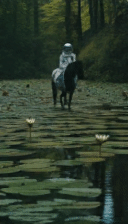}
\includegraphics[width=0.11\textwidth]{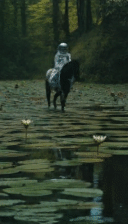}
{\raggedright \textbf{prompt:} A photo of an astronaut riding a horse in the forest. There is a river in front of them with water lilies
\ \\
\ \\}

\end{subfigure}

\vspace{-2mm}

\begin{subfigure}{1.0\linewidth}
\centering
\includegraphics[width=0.11\textwidth]{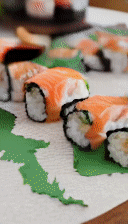}
\includegraphics[width=0.11\textwidth]{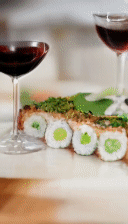}
\includegraphics[width=0.11\textwidth]{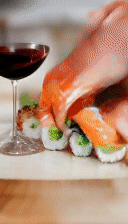}
\includegraphics[width=0.11\textwidth]{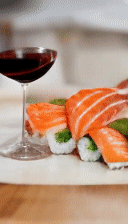}
\hspace{0.03\textwidth}
\includegraphics[width=0.11\textwidth]{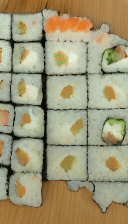}
\includegraphics[width=0.11\textwidth]{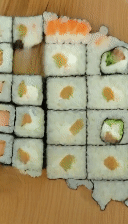}
\includegraphics[width=0.11\textwidth]{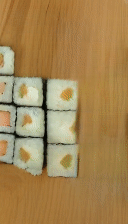}
\includegraphics[width=0.11\textwidth]{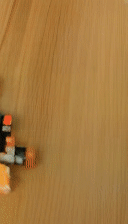}
{\raggedright \textbf{prompt:} A zoomed out map of the United States made out of sushi. It is on a table next to a glass of red wine. Pieces of sushi disappear one by one
\ \\
\ \\}

\end{subfigure}

\vspace{-2mm}

\begin{subfigure}{1.0\linewidth}
\centering
\includegraphics[width=0.11\textwidth]{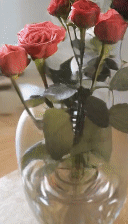}
\includegraphics[width=0.11\textwidth]{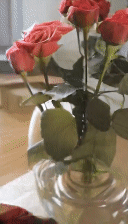}
\includegraphics[width=0.11\textwidth]{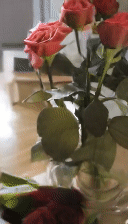}
\includegraphics[width=0.11\textwidth]{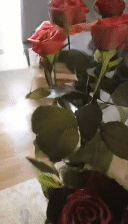}
\hspace{0.03\textwidth}
\includegraphics[width=0.11\textwidth]{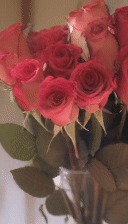}
\includegraphics[width=0.11\textwidth]{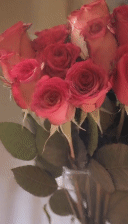}
\includegraphics[width=0.11\textwidth]{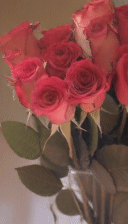}
\includegraphics[width=0.11\textwidth]{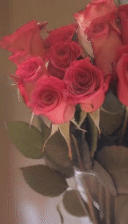}
{\raggedright \textbf{prompt:} Rotating around a vase holding a dozen roses
\ \\
\ \\}

\end{subfigure}

\vspace{-5mm}

\caption[A comparison between 1B and 8B parameter models]{\textbf{A comparison between 1B (left) and 8B (right) parameter models} on the same prompt and settings.}

\label{fig:sxs_1b_8b}
\end{figure*}

We show a qualitative comparison of the 1B and 8B pretrained models. 
% Increasing the model size improved temporal consistency, prompt fidelity, and motion dynamics while adding capabilities for limited text rendering, spatial understanding, and counting.
In Figure~\ref{fig:sxs_1b_8b}, we show outputs of 1B and 8B parameter models on the same prompts.
Four frames from the best video output of each model in a batch of four text-to-video samples were selected to represent the model.
In the first row, the 1B model is unstable with large changes to the subject over time and misses elements from the complex prompt.
This prompt was originally used for scaling comparisons in~\cite{yu2022scaling}, and compared to a dedicated image-only model, our model does not preserve text as well given the training data used. In the second row, we use a simpler text task and show that the 8B model can represent a single letter clearly, but the 1B model still produces artifacts.
In the third row, we show that the 8B model learns spatial positioning such as the river being in front of the astronaut and horse. In the fourth row, we show that the 8B parameter model learned a stop motion style to have items disappear ``one by one" and can follow a complicated layout from a long prompt.
In contrast, the 1B model includes all of the nouns, but is unstable over time and does not follow the layout indicated in the prompt. In the bottom row, we show that the 8B model understands counts of objects in that it displays a full bouquet (though 12 roses are not explicitly in frame) and smooth consistent motion as opposed to the 5 roses and distorting objects produced by the 1B model. 
Overall, scaling the model improves temporal consistency, prompt fidelity, and motion dynamics while adding capabilities for limited text rendering, spatial understanding, and counting.

% -----------------------------

% \subsection{Additional Generated Examples} 
% We include most generated videos in the supplementary materials for an enhanced visualization of motion and visual quality, in addition to \cref{fig:outpaint_stylize} and \cref{fig:camera_movement}.

% \input{figures/fig_chain_mona_yawn}

% \vspace{-2mm}
\subsection{Comparison with the State-of-the-Art}
% % \vspace{-2mm}
\begin{table}[tp]
\centering
% \scriptsize
%{-2mm}
% \resizebox{\linewidth}{!}{%
\begin{tabular}{lcccc}
\toprule
      Model & \multicolumn{2}{c}{MSR-VTT} & \multicolumn{2}{c}{UCF-101} \\
      & \multicolumn{1}{c}{CLIPSIM} & \multicolumn{1}{c}{FVD} & \multicolumn{1}{c}{FVD} & \multicolumn{1}{c}{IS} \\
\midrule
CogVideo (EN)~\cite{hong2022cogvideo} & \multicolumn{1}{c}{0.2631} & 1294 & 702 & 25.27 \\
MagicVideo~\cite{zhou2022magicvideo} & \multicolumn{1}{c}{-} & 998 & 655 & - \\
Video LDM~\cite{blattmann2023align} & \multicolumn{1}{c}{0.2929} & - &  551 & 33.45 \\
ModelScopeT2V~\cite{wang2023modelscope} & \multicolumn{1}{c}{0.2930} & 550 & - & - \\
InternVid~\cite{wang2023internvid} & \multicolumn{1}{c}{0.2951} & - & 617 & 21.04 \\
VideoFactory~\cite{wang2023videofactory} & \multicolumn{1}{c}{0.3005} & - & 410 & - \\
Make-A-Video~\cite{singer2022make} & \multicolumn{1}{c}{0.3049} & - & 367 & 33.00 \\
Show-1~\cite{zhang2023show} & \multicolumn{1}{c}{0.3072} & 538 & 394 & 35.42 \\ \midrule
\bf \modelname{} (Pretrain) & \multicolumn{1}{c}{0.3049}  & \bf 213 & \bf 355 & \bf 38.44\\
\bf \modelname{} (Task adapt) & \multicolumn{1}{c}{\bf 0.3123} & - & - & - \\
\bottomrule
\end{tabular}
% }
\caption[Comparison on zero-shot text-to-video benchmarks]{\textbf{Comparison on zero-shot text-to-video benchmarks.}
%  See Appendix~\ref{sec:appendix-evaluation} for evaluation details.
}
\label{tab:t2v_sota}
%{-2mm}
\end{table}
\paragraph{Text-to-Video (T2V).}
\label{sec:exp_t2v}
\cref{tab:t2v_sota} shows zero-shot text-to-video evaluation results on the common MSR-VTT~\cite{xu2016msr} and UCF-101~\cite{soomro2012ucf101} datasets. 
Our model performs favorably in terms of CLIP similarity and FVD scores on MSR-VTT and UCF-101. 
The pretrained foundation model already achieves competitive performance on all metrics.
After finetuned on high-quality subset of text-video pairs, \modelname{} achieves even better CLIPSIM on MSR-VTT.
More details on the evaluation settings can be found in~\cref{sec:add_details}.

\begin{figure}[tp] %p]
\centering

\begin{subfigure}{\linewidth} 
    \centering 
\includegraphics[width=0.67\linewidth]{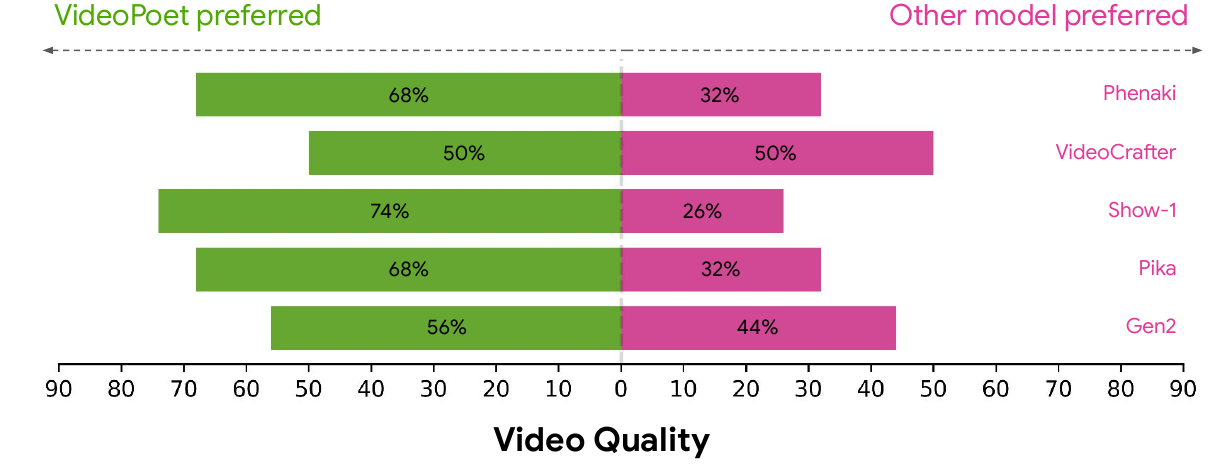}
\end{subfigure}
\begin{subfigure}{\linewidth} 
    \centering 
\includegraphics[width=0.67\linewidth]{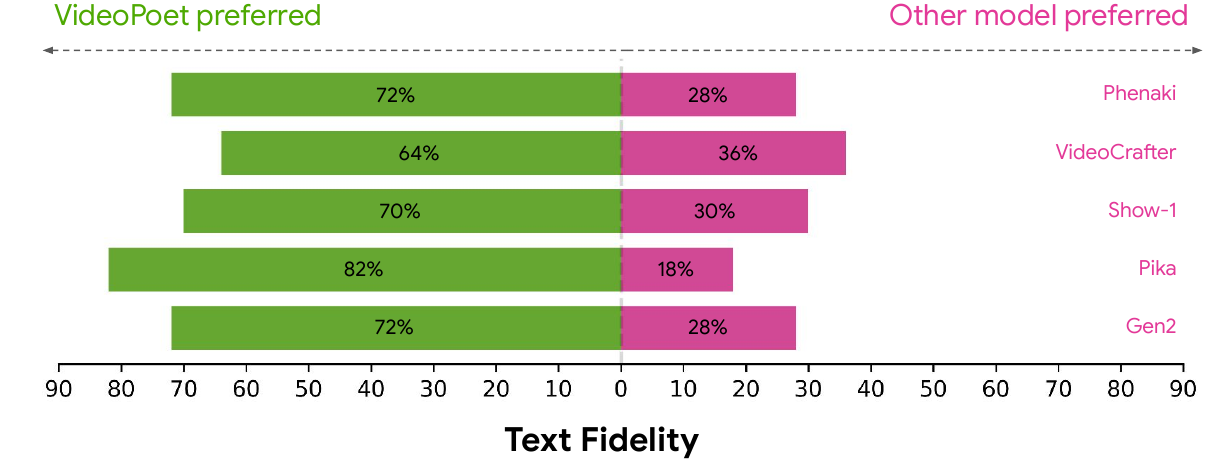}
\end{subfigure}
\begin{subfigure}{\linewidth} 
    \centering 
\includegraphics[width=0.67\linewidth]{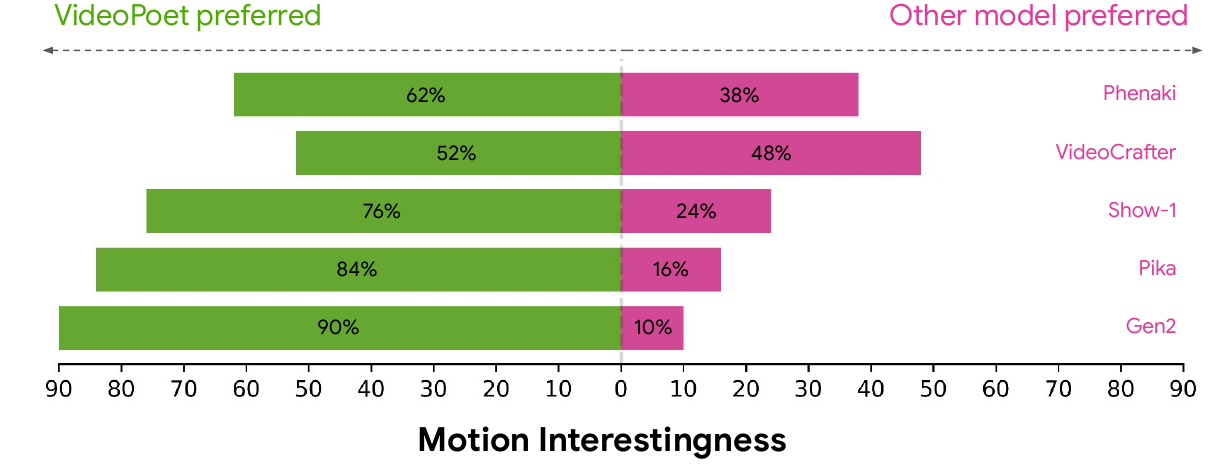}
\end{subfigure}
\begin{subfigure}{\linewidth}  
    \centering
\includegraphics[width=0.67\linewidth]{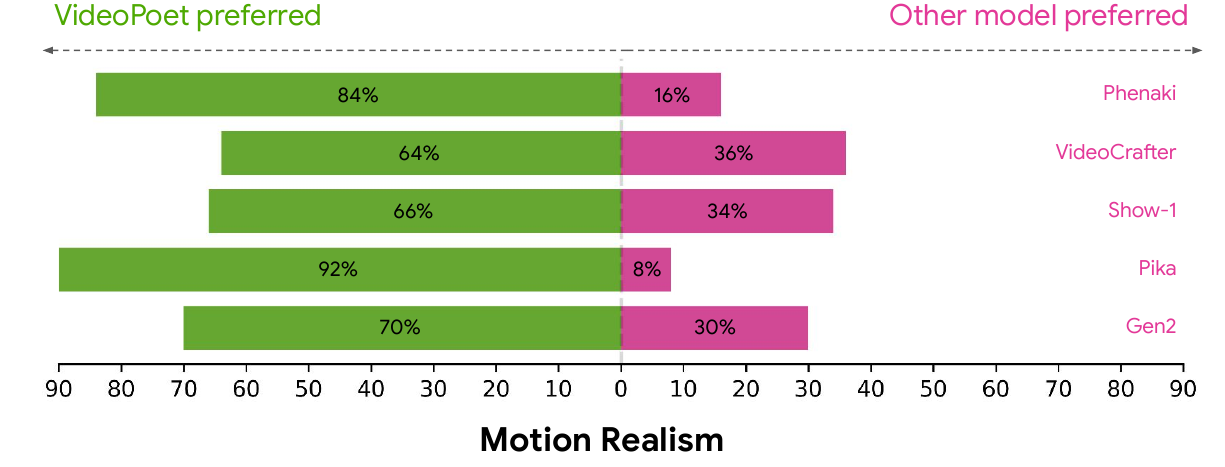}
\end{subfigure}
\begin{subfigure}{\linewidth}  
    \centering
\includegraphics[width=0.67\linewidth]{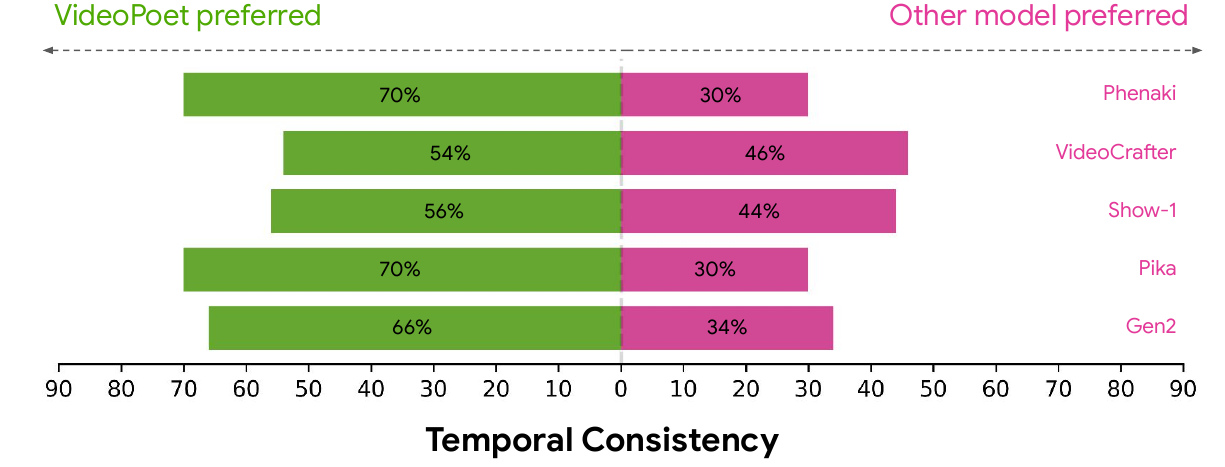}
\end{subfigure}
% \includegraphics[width=0.48\linewidth]{figures/Text_Fidelity_v3.pdf}
% \includegraphics[width=0.48\linewidth]{figures/Video_Quality_v3.pdf}%%{4mm}
%%{4mm}

%\begin{subfigure}{0.24\textwidth}  
%\includegraphics[width=0.4\linewidth]{figures/Video_Quality_v2.pdf}%%{4mm}
%\end{subfigure}

%\begin{subfigure}{0.42\textwidth}  
% \includegraphics[width=0.44\linewidth]{figures/Motion_Interestingness_v3.pdf}
% \includegraphics[width=0.48\linewidth]{figures/Motion_Realism_v3.pdf}
% \includegraphics[width=0.48\linewidth]{figures/Temporal_Consistency_v3.pdf}
%%{4mm}
%\end{subfigure}

%\begin{subfigure}{0.42\textwidth}  
%\includegraphics[width=0.99\linewidth]{figures/Motion_Realism_v2.pdf}
%%{4mm}
%\end{subfigure}

%\begin{subfigure}{0.42\textwidth}  

%\end{subfigure}
%{-4mm}
\caption[Human evaluation results on text-to-video (T2V) generation]{
% \dan{TODO: Jon to update these with finalized plots} \textbf{User evaluations of \modelname{} against  
% leading text-to-video models.}
\textbf{Human evaluation results on text-to-video (T2V) generation.}
Green and pink bars represent the proportion of trials where \modelname{} was \textit{preferred over} or \textit{less preferred} to an alternative, respectively.
% We evaluate against Phenaki~\cite{villegas2022phenaki}, VideoCrafter~\cite{chen2023videocrafter1}, Show-1~\cite{zhang2023show}, and Pika~\cite{pika2023}, examples generated as of January 2024.
}
\label{fig:human_t2v_evals}
%{-4mm}
\end{figure}
% \vspace{-4mm}
\paragraph{Human Evaluations with Text-to-Video (T2V).}
% \lu{need to replace it with the latest user study. Please reduce the length and put the rest to Appendix.}
We analyze \modelname{} using human raters and compare with other recent
models: Show-1~\cite{zhang2023show},
VideoCrafter~\cite{chen2023videocrafter1}, Phenaki~\cite{villegas2022phenaki},  Pika~\cite{pika2023}, 
and Gen2~\cite{runway2023}.
Show-1, VideoCrafter, Pika, and Runway are video diffusion models
while Phenaki is a token-based model using  masked token modeling~\cite{chang2022maskgit}.
We ran the most up-to-date model versions as of January 2024.

We first develop a unified evaluation prompt bank consisting of prompts from a variety of categories and styles.
Our prompts are
sourced from published prompt sets 
(\eg, Show-1, Video LDM~\cite{blattmann2023align}).
We select the prompts prior to generating videos
and fix these choices after initial selection. 
We also select preferentially for prompts that contain an explicit mention of motion so that the evaluation would not be biased for models that generate high quality videos that are almost still (e.g., ``person jumping off of a chair'' over ``person standing on a chair''). 

\begin{figure}[tp]
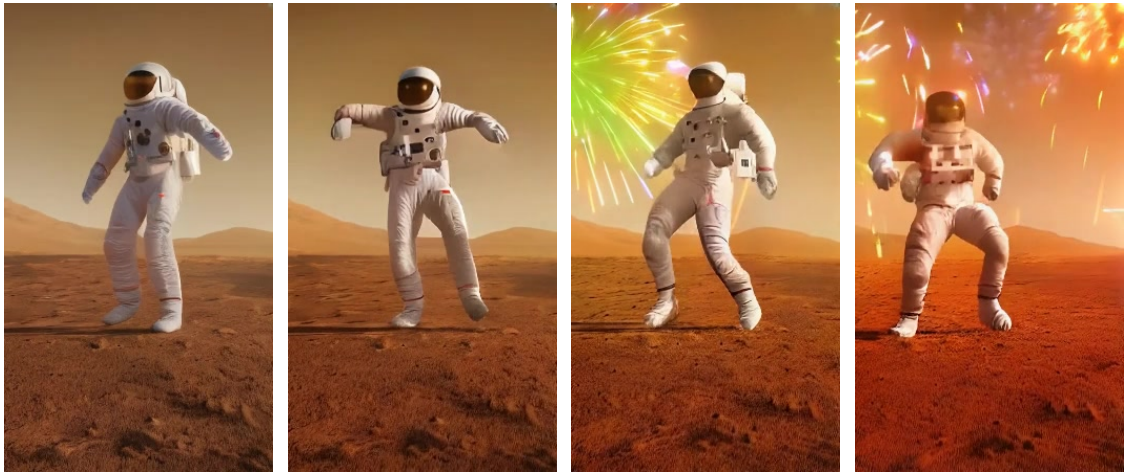

\addtocounter{figure}{-1}

\begin{subfigure}{1.0\linewidth}
\centering
\videoframe{figures/video_frames/astronaut_00001.jpg}
\videoframe{figures/video_frames/astronaut_00029.jpg}
\videoframe{figures/video_frames/astronaut_00060.jpg}
\videoframe{figures/video_frames/astronaut_00081.jpg}

\end{subfigure}

\caption[10-Second long video generation example]{\textbf{10-Second long video generation example.} By predicting 1-second video segments from an initial 1-second clip, \modelname{} can iteratively generate videos of extended lengths.}
\label{fig:long_video}
% \vspace{-4mm}
\end{figure}

For this user study we use the fine-tuned version of \modelname{} as discussed in \cref{sec:training}
% The finetuned model discussed in \cref{sec:training} is used for the user study.
% We then compare each model against \modelname{}
and compare against alternative models
in a side-by-side fashion for each prompt. Raters are shown videos generated by two models at a time (in randomized order so as to not bias raters). 
Not all methods generate videos at the same size or aspect ratio, and we resize each video to a fixed area while maintaining its original aspect ratio. 
Raters are then asked to compare the videos along 5 dimensions and
for each dimension to report which video is better. 
The 5 dimensions are: (1) text fidelity (which video follows the text prompt most faithfully), (2) video quality, (3) motion ``interestingness'', (4) motion realism, and (5) temporal consistency.
Raters are required to go through a collection of training examples for each of these 5 dimensions before they begin. 

Our findings are summarized in \cref{fig:human_t2v_evals}, where
green and pink bars  represent the  proportion  of trials  where \modelname{} was \textit{preferred} 
% an alternative,
% \textit{similar to}, 
or \textit{less preferred} over an alternative, respectively.
%
% Experiments where the green segment is larger than the pink
% segment means that \modelname{} is preferred over the alternative on average.
%
In summary, \modelname{} outperforms all baseline models
along almost all of the dimensions 
%(text fidelity, quality, motion interestingness and realism) 
and achieves its most significant wins along the motion categories.

% On temporal consistency, \modelname{} shows performance on-par with Phenaki and VideoCrafter but slightly underperforms the Show-1 model.
% %
% This is due to an inherent trade-off with motion interestingness, \ie, a static scene is more temporally consistent but is less interesting.
% %
% More interesting larger motions necessitate more possibilities of producing noticable artifacts vs. safer small motions.
% %

% \vspace{-2mm}
\subsection{LLM's Diverse Capabilities in Video Generation}
\label{sec:capabilities_whole}
% % \vspace{-2mm}

This subsection presents several capabilities we discover from the pretrained \modelname{}, shedding light on the LLM's promising potential in video generation. By combining the flexibility of our autoregressive language model to perform diverse tasks such as extending video in time, inpainting, outpainting, and stylization, \modelname{} accomplishes multiple tasks in a unified model.

\begin{figure}[tp]
% \addtocounter{figure}{-1}

% \begin{subfigure}{1.0\linewidth}
% \centering
% \videoframe{figures/video_frames/milk_drop_00001.jpg}
% \videoframe{figures/video_frames/milk_drop_00004.jpg}
% \videoframe{figures/video_frames/milk_drop_00008.jpg}
% \videoframe{figures/video_frames/milk_drop_00014.jpg}
% {\textbf{Animated from Photograph}
% \ \\
% \ \\}
% \end{subfigure}

\begin{subfigure}{1.0\linewidth}
\centering
\videoframe{figures/video_frames/flag_00001.jpg}
\videoframe{figures/video_frames/flag_00006.jpg}
\videoframe{figures/video_frames/flag_00011.jpg}
\videoframe{figures/video_frames/flag_00017.jpg}
{\textbf{Animated from historical photo}
\ \\
\ \\}
\end{subfigure}

% \vspace{-3mm}
\begin{subfigure}{1.0\linewidth}
\centering
\videoframe{figures/video_frames/ship_00001.jpg}
\videoframe{figures/video_frames/ship_00003.jpg}
\videoframe{figures/video_frames/ship_00009.jpg}
\videoframe{figures/video_frames/ship_00015.jpg}
{\textbf{Animated from painting}
\ \\}
\end{subfigure}
% \vspace{-3mm}
\caption[Examples of videos animated from still images]{\textbf{Examples of videos animated from still images} plus text prompts tailored to each initial image.}
\label{fig:animated_images}
\end{figure}
% \vspace{-4mm}
\paragraph{Coherent long video generation and image-to-video.}

A benefit of an decoder-based language model is that it pairs well with autoregressively extending generation in time. 
We present two different variants: Generating longer videos~(\cref{fig:long_video}) and converting images to videos. 
Encoding the first frame independently allows us to convert any image into the initial frame of a video without padding. Subsequent frames are generated by predicting remaining tokens, transforming the image into a video as shown in \cref{fig:animated_images}\footnote{For image-to-video examples we source images from Wikimedia Commons: https://commons.wikimedia.org/wiki/Main\_Page}.

This results in the capability to generate videos longer than 10 seconds or to allow users to iteratively extend video clips based on previously generated video, and produces temporally consistent videos without significant distortion.
Such capabilities are rarely observed in contemporary diffusion models.

% \vspace{-4mm}
\paragraph{Zero-shot video editing and task chaining.} \label{sec:capabilities}
\begin{figure}[tp]
% \addtocounter{figure}{-1}
\centering
\begin{subfigure}{1.0\linewidth}
\centering
\videoframe{figures/video_frames/mona_yawn_lores_00001.jpg}
\videoframe{figures/video_frames/mona_yawn_lores_00006.jpg}
\videoframe{figures/video_frames/mona_yawn_lores_00011.jpg}
\videoframe{figures/video_frames/mona_yawn_lores_00017.jpg}
{\textbf{Animated from still image}
\ \\
\ \\}
\end{subfigure}
% \vspace{-4mm}

\begin{subfigure}{1.0\linewidth}
\centering
\videoframe{figures/video_frames/mona_yawn_snow_00001.jpg}
\videoframe{figures/video_frames/mona_yawn_snow_00006.jpg}
\videoframe{figures/video_frames/mona_yawn_snow_00011.jpg}
\videoframe{figures/video_frames/mona_yawn_snow_00017.jpg}
{\textbf{Stylized video} \\
\textbf{Prompt:} An oil painting of a snowman with a red hat opening their mouth to yawn
\ \\}
\end{subfigure}
\caption[Example of zero-shot video editing via task chaining]{\textbf{Example of zero-shot video editing via task chaining} (text conditioned image-to-video and stylization) 
%-- the original painting is first animated via a text prompt and then stylized via another text prompt.
}
\label{fig:animate_stylize}
% \vspace{-2mm}
\end{figure}

With the multi-task pretraining, \modelname{} exhibits task generalization that can be chained together to perform novel tasks.
We show the model can apply image-to-video animation followed by video-to-video stylization in \cref{fig:animate_stylize}.
\cref{fig:outpaint_stylize} shows another example applying video-to-video outpainting, followed by editing them with additional video-to-video effects.
% In the supplementary materials, we include more examples.
%
At each stage, the quality of the output is sufficient to remain in-distribution (\ie teacher forcing) for the next stage without noticeable artifacts.
These capabilities can be attributed to our multimodal task design within an LLM transformer framework that allows for modeling multimodal content using a single transformer architecture over a unified vocabulary. 
%Our approach contrasts with others, such as diffusion models, which typically solve these tasks by adopting multiple individually tuned adapter models to control the diffusion process~\cite{esser2023structure,liew2023magicedit,guo2023animatediff}.

% \vspace{-4mm}
% \paragraph{Zero-shot video stylization.}
% Stylization results are presented in \cref{sec:stylization} where the structure and text are used as prefixes to guide the language model. Unlike other stylization methods that employ adapter modules such as cross-attention networks~\cite{zhang2023adding} or latent blending~\cite{meng2021sdedit}, our approach stylizes videos within an LLM as one of several generative tasks.
% %We observe temporally coherent generations of objects in a video scene with dynamic, and meaningful motion (see \cref{fig:long_video}).
% % To predict the future frames, despite the model only being able to view up to a short temporal context, such as the first frame or the first second of video, the model is able to keep the motion, style, and identity of objects consistent across more than a total of 8 seconds of video output.

% \vspace{-4mm}
\paragraph{3D structure, camera motion, visual styles.}
% Because our training spans videos, images, and text, we can prompt our model to demonstrate many aspects of understanding about the world including 3D structures, camera motions, and visual styles learned from these different sources.
Even though we do not add specific training data or losses to encourage 3D consistency, our model
can rotate around objects and predict reasonable visualizations of the backside of objects. Additionally, with only a small proportion of input videos and texts describing camera motion, our model can use short text prompts to apply a range of camera motions to image-to-video and text-to-video generations (see \cref{fig:camera_movement}).

%\begin{figure*}[h]
\begin{figure}[tp]
% \addtocounter{figure}{-1} %Prevent subfigures increasing main figure count  % for some reason, this needed to be commented out to get this figure to have a different number than the preceeding figure.
\centering
% Arc shot
\begin{subfigure}[b]{0.5\linewidth}
%{0.5\textwidth}
\centering
\videoframe{figures/video_frames/train_original_00001.jpg}
\videoframe{figures/video_frames/train_original_00006.jpg}
\videoframe{figures/video_frames/train_original_00011.jpg}
\videoframe{figures/video_frames/train_original_00017.jpg}
%\captionof{figure}{Example of directed camera movement }
%\end{figure}
{\textbf{Original Video}
\ \\
\ \\}
\end{subfigure}

\begin{subfigure}[b]{0.5\linewidth}
\centering
\videoframe{figures/video_frames/train_outpaint_00001.jpg}
\videoframe{figures/video_frames/train_outpaint_00006.jpg}
\videoframe{figures/video_frames/train_outpaint_00011.jpg}
\videoframe{figures/video_frames/train_outpaint_00017.jpg}
{\textbf{Outpainted Video}
\ \\
\ \\}
\end{subfigure}

\begin{subfigure}[b]{0.5\linewidth}
\centering
\videoframe{figures/video_frames/train_food_00001.jpg}
\videoframe{figures/video_frames/train_food_00006.jpg}
\videoframe{figures/video_frames/train_food_00011.jpg}
\videoframe{figures/video_frames/train_food_00017.jpg}
{\textbf{Stylized Video} \\
\textbf{Prompt:} A gingerbread and candy train on a track
\ \\}
\end{subfigure}

\caption[Example of zero-shot video editing via task chaining]{\textbf{Example of zero-shot video editing via task chaining} (outpainting and stylization) -- the original video is first outpainted and then stylized via a text prompt.}
\label{fig:outpaint_stylize}
\end{figure}

\begin{figure}[tp]
% \addtocounter{figure}{-1}
\centering
\begin{subfigure}{0.5\linewidth}

\centering
\videoframe{figures/video_frames/adventure_arc_shot_00001.jpg}
\videoframe{figures/video_frames/adventure_arc_shot_00006.jpg}
\videoframe{figures/video_frames/adventure_arc_shot_00011.jpg}
\videoframe{figures/video_frames/adventure_arc_shot_00017.jpg}
{\textbf{ Camera Motion}: Arc shot
\ \\
\ \\}
\end{subfigure}

\begin{subfigure}{0.5\linewidth}
\centering
\videoframe{figures/video_frames/adventure_fpv_00001.jpg}
\videoframe{figures/video_frames/adventure_fpv_00006.jpg}
\videoframe{figures/video_frames/adventure_fpv_00011.jpg}
\videoframe{figures/video_frames/adventure_fpv_00017.jpg}
{\textbf{ Camera Motion}: FPV drone shot
\ \\}
\end{subfigure}

\caption[Examples of directed camera movement]{\textbf{Examples of directed camera movement} from the same initial frame.}
\label{fig:camera_movement}
\end{figure}

\subsection{Additional Implementation Details}  \label{sec:add_details}

%---------------------------------------------------------------------------------------
% Moved from the main text

% \subsubsection{Additional Implementation Details}
\paragraph{Training settings.}
The unified vocabulary is constructed as follows: the initial 256 codes are reserved for special tokens and task prompts. \Cref{tab:special_tokens} lists some examples of special tokens. Subsequently, the next 262,144 codes are allocated for image and video tokenization. This is followed by 4,096 audio codes. We also include a small text vocabulary of English words. Overall, this produces a total vocabulary size of approximately 300,000.

Since the first frame is tokenized separately, MAGVIT-v2 allows images to be represented in the same vocabulary as video.
In addition to being more compact, images provide many learnable characteristics that are not typically represented in videos, such as strong visual styles (\eg, art paintings), objects which are infrequently seen in video, rich captions, and significantly more text-image paired training data.
When training on images, we resize the images to 128$\times$128 which are then tokenized to a latent shape of $(1, 16, 16)$, or 256 tokens.
We scale the MAGVIT-v2 model's size and train it on the datasets discussed in \Cref{sec:exp_data}. The training follows two steps: image training, inflation~\cite{yu2023language} and video training. Due to images requiring fewer tokens, we can include roughly 5$\times$ more images per batch than videos, \ie 256 image tokens vs. 1280 video tokens. We use up to a maximum of 64 text tokens for all of our experiments. For the \texttt{<res>} token, the resolution is only specified for 128$\times$224 output, 128$\times$128 resolution is assumed otherwise.

The video-to-video tasks use the COMMIT encoding~\cite{yu2022magvit} to obtain the tokens for the tasks such as inpainting and outpainting.
Text is encoded as T5 XL embeddings~\cite{raffel2020exploring} and are inserted into reserved sequence positions right after the \texttt{<bot\_i>} token as shown in \Cref{fig:vffm_schematic}.
 
% Additionally, we assess stylization tasks using a subset of the DAVIS dataset\footnote{\url{https://davischallenge.org/}}~\cite{Perazzi2016davis}, as further detailed below.

% \begin{table*}
% \centering
% \begin{tabular}{ll}
% \toprule
%                                             Prefix &                                            Target \\
% \midrule
% \bos full\_generation (\textemd text\_embedding(64) \bovin \eovin &             \bov video\_tokens256 \\
% \bottomrule
% \end{tabular}
% \end{table*}

\begin{table}[tp]
\centering
\begin{tabular}{cl}
\toprule
Special Token                          & \multicolumn{1}{c}{Usage} \\
\midrule
\texttt{<bos>} & Beginning of sequence     \\
\texttt{<task>} & Task to perform for this sequence \\
\texttt{<bot\_i>}& Beginning of the text input. \\
\texttt{<eot\_i>}& End of the text input. \\
\texttt{<bov\_i>}& Beginning of the visual input. \\
\texttt{<eov\_i>}& End of the video input. \\
\texttt{<boa\_i>}& Beginning of the audio input. \\
\texttt{<eoa\_i>}& End of the audio input. \\
\texttt{<source>}& The source of the video to generate. \\
\texttt{<res>} & Output resolution for the video. \\
\texttt{<bov\_o>}& Beginning of the video output. \\
\texttt{<eov\_o>}& End of the video output. \\
\texttt{<boa\_o>}& Beginning of the audio output. \\
\texttt{<eoa\_o>}& End of the audio output. \\
\texttt{<eos>}& End of the entire sequence. \\
\bottomrule
\end{tabular}
\caption[List of representative special tokens]{\textbf{List of representative special tokens} used in training and inference. }
\label{tab:special_tokens}
\end{table}

\paragraph{Zero-shot text-to-video evaluation settings.}
% \label{sec:appendix-evaluation}
We report the details of our zero-shot text-to-video settings here.
We note that some details are missing in previous papers and different papers use different settings. Hence, we provide all the details and hope this evaluation setting can serve as a standard text-to-video generation benchmark.
Our results are reported on the 8B model and we adopt classifier-free guidance~\cite{ho2021classifier}.
%{-2mm}
All metrics are evaluated on generated videos containing 16 frames with a resolution of 256$\times$256.
We first generate videos of 128$\times$128 resolution and then resize to 256$\times$256 via bicubic upsampling.

\emph{Zero-shot MSR-VTT}.
For CLIP score, we used all 59,794 captions from the MSR-VTT test set. We use CLIP ViT-B/16 model following Phenaki~\cite{villegas2022phenaki}.
We note that some papers use other CLIP models, \eg, VideoLDM~\cite{blattmann2023align} uses ViT-B/32.
Our CLIP score evaluated on the ViT-B/32 backbone for MSR-VTT is 30.01.
For the FVD metric, to evaluate on a wide range of captions as well as to be comparable with previous papers that evaluate on 2,048 videos, we evaluate on the first 40,960 captions in the MSR-VTT test set. More specifically, we report the FVD metrics on 2048 videos with 20 repeats.
The FVD real features are extracted from 2,048 videos sampled from the MSR-VTT test set.
We sample the central 16 frames of each real video, without any temporal downsampling, \ie, we use the original fps in the MSR-VTT dataset (30 fps as reported in \cite{xu2016msr}). The FVD is evaluated with an I3D model trained on Kinetics-400.

\emph{Zero-shot UCF-101}.
Following VDM~\cite{ho2022video}, we sample 10,000 videos from the UCF-101 test set and use their categories as the text prompts to generate 10,000 videos.
We use the class text prompts provided in PYoCo~\cite{ge2023preserve} to represent the 101 categories.
To compute the FVD real features, we sample 10K videos from the training set, following TGAN2~\cite{saito2020train}.
We sample the central 16 frames for each real video , without any temporal downsampling, \ie, we use the original fps in the UCF-101 dataset (25 fps as reported in \cite{soomro2012ucf101}).
The FVD metric is evaluated with an I3D model trained on Kinetics-400 and the IS metric is evaluated with a C3D model trained on UCF-101.

\paragraph{Self-supervised tasks evaluation settings.}
% \label{sec:ssl_task}
Self-supervised learning tasks include frame prediction on K600 with 5 frames as condition, as well as inpainting and outpainting on SSv2.
FVD~\cite{unterthiner2018towards} is used as the primary metric, calculated with 16 frames at 128$\times$128 resolution.
We follow MAGVIT~\cite{yu2022magvit} in evaluating these tasks against the respective real distribution, using 50000$\times$4 samples for K600 and 50000 samples for SSv2.

\section{Summary}
% \vspace{-2mm}
% \section{Conclusion}
% \vspace{-1mm}

\modelname{} demonstrates the potential of a large language model that is trained on discrete visual, text and audio tokens, in generating videos of compelling state-of-the-art quality.
A particular strength of our model lies in its ability to
generate high-fidelity, large, and complex motions.
Our large language model formulation benefits from training
across a variety of \emph{multi-modal} tasks with a unified architecture and vocabulary.
Consequently, the pretrained model is adept at \emph{multi-task} video creation, and serves as a foundation for a diverse variety of video generation related capabilities, including multiple forms of editing.

\paragraph{Limitations.}

Despite VideoPoet demonstrating highly competitive performance of LLMs relative to state-of-the-art models, certain limitations are still observed.
For example, the RGB frame reconstruction from compressed and quantized tokens place an upper bound on the generative model's visual fidelity. Second, the per-frame aesthetic biases in static scenes does not match the best baseline.
This difference is largely due to a choice of training data, where we focus our training on more natural aesthetics % and excluded some other sources common in other work.
and excluded some sources containing copyrighted images, such as LAION~\cite{schuhmann2022laion}, which is commonly used in other work.
% This shortfall is primarily due to the limited image training data, which is constrained to images for which we have rights or legal clearance to use, thereby excluding copyrighted images such as those found in the LAION dataset~\cite{schuhmann2022laion}.
Finally, small objects and fine-grained details, especially when coupled with significant motions, remains difficult within the token-based modeling.
%and so put an upper bound on the per-frame quality for a fixed amount of tokens.
% The video tokenizer is also trained on 8 fps video, which can lead to distored artifacts when reconstructing videos with fast motion.

% Per-frame quality is low due to the bad image\lu{complete this}.
%, which has been noted to be difficult for many existing video generation models~\cite{liu2023evalcrafter}. In addition, these controls can be added on top of a wide range of styles,  such as watercolor or oil paintings. 

% These stylization training sources are primarily observed in the text-image training data.
% The ability to generalize across and combine these different types of styles to produce large motions following text prompts underscores the strength of our model's understanding of objects in a temporal context.

\glsresetall
\cleardoublepage
\part*{Conclusion}
\addcontentsline{toc}{part}{Conclusion}
\cleardoublepage
\chapter*{Conclusion}

\section*{Summary}
\addcontentsline{toc}{section}{Summary}

This thesis has explored the development of multi-task models for generating videos
and other modalities under diverse conditions, as well as for understanding and
compression applications.
Throughout this research, we have demonstrated the effectiveness of integrating multiple tasks, crafting high-fidelity latent representations, and generating multiple modalities.

Initial work on separate multi-task and multi-modal setups with pixel-space models has revealed potential, but also demonstrated the constraints imposed by fixed label spaces and specialized architectures. This has underscored the importance of pursuing more flexible and adaptable designs.

Seeking compact and informative representations of high-dimensional visual data, we have developed spatial-temporal video tokenizers that have maintained remarkable fidelity. We have introduced a novel method for bidirectional mapping between visual observations and interpretable linguistic concepts.  Additionally, we have created a scalable visual token representation that has proven beneficial across generation, compression, and understanding tasks. These advancements have marked early instances of language models outperforming diffusion models in visual synthesis and a video tokenizer exceeding the performance of industry-standard codecs.

Building upon these multi-modal latent spaces, we have investigated the design of multi-task generative models.  Our masked multi-task transformer has demonstrated favorable performance in video generation across quality, efficiency, and flexibility. We have successfully enabled a language model, trained exclusively on text, to generate images and videos.  Finally, we have constructed a scalable multi-modal generative transformer trained from scratch, capable of generating videos and audio under various conditions.  This model has notably generated high-fidelity videos with corresponding audio in a zero-shot setting.

\section*{Contributions}
\addcontentsline{toc}{section}{Contributions}

\paragraph{}
We briefly summarize our contributions organized by these included publications:

\begin{itemize}
    \item \cref{chap:argus}: \cite{yu2022argus++} \textbf{Lijun Yu}, Yijun Qian, Wenhe Liu, and Alexander G. Hauptmann. \\
    \emph{Argus++: Robust Real-Time Activity Detection for Unconstrained Video Streams with Overlapping Cube Proposals.} \\
    In WACV 2022 HADCV Workshop.
    \begin{itemize}
    \item We propose Argus++, a real-time activity detection system for unconstrained video streams, which is robust across different scenarios.
    \item We introduce overlapping spatial-temporal cubes as the core concept of activity proposals to ensure coverage and completeness of activity detection through over-sampling.
    \item The proposed system has achieved outstanding performance in a large series of activity detection benchmarks, including CVPR ActivityNet ActEV 2021, NIST ActEV SDL UF/KF, TRECVID ActEV 2020/2021, and ICCV ROAD 2021.
    \end{itemize}

    \item \cref{chap:doc}: \cite{yu2023document} \textbf{Lijun Yu}, Jin Miao, Xiaoyu Sun, Jiayi Chen, Alexander G. Hauptmann, Hanjun Dai, and Wei Wei. \\
    \emph{DocumentNet: Bridging the Data Gap in Document Pre-Training.} \\
    In EMNLP 2023 Industry Track.
    \begin{itemize}
    % \item We introduce a massive-scale document dataset DocumentNet collected  over the Internet with a proposed ontology covering about 400 keywords.
    \item We propose a pretraining-finetuning paradigm with lightweight UniFormer models and three objectives for unified token representation.
    \item Extensive experiments on VDER tasks demonstrate the favorable performance of  UniFormer pretrained on DocumentNet compared to the commonly used IIT-CDIP.
    \end{itemize}

    \item \cref{chap:3dvq,chap:magvit}: \cite{yu2022magvit} \textbf{Lijun Yu}, Yong Cheng, Kihyuk Sohn, José Lezama, Han Zhang, Huiwen Chang, Alexander G. Hauptmann, Ming-Hsuan Yang, Yuan Hao, Irfan Essa and Lu Jiang. \\
    \emph{MAGVIT: Masked Generative Video Transformer.} \\
    In CVPR 2023 as a Highlight paper. 
    \begin{itemize}
    \item To the best of our knowledge, we present the first masked multi-task transformer for efficient video generation and manipulation. We show that a trained model can perform ten different tasks at inference time.
    \item We propose an effective embedding method with diverse masks for numerous video generation tasks. 
    \item We show that MAGVIT achieves the best-published fidelity on three widely-used benchmarks, including  UCF-101, BAIR Robot Pushing, and Kinetics-600 datasets.
    \end{itemize}

    \item \cref{chap:spae,chap:llm}: \cite{yu2023spae} \textbf{Lijun Yu}, Yong Cheng, Zhiruo Wang, Vivek Kumar, Wolfgang Macherey, Yanping Huang, David A. Ross, Irfan Essa, Yonatan Bisk, Ming-Hsuan Yang, Kevin Murphy, Alexander G. Hauptmann, Lu Jiang. \\
    \emph{SPAE: Semantic Pyramid AutoEncoder for Multimodal Generation with Frozen LLMs.} \\
    In NeurIPS 2023 as a Spotlight paper. 
    \begin{itemize}
    \item This is the first successful method,
    to the best of our knowledge,
    that uses a frozen language model, trained solely on language tokens, to directly generate image content through in-context learning.
    \item We propose a new progressive prompting method that facilitates in-context generation of long cross-modal sequences.
    %with a frozen large language model.
    \item We evaluate our method on visual understanding and generation tasks, and notably, our approach outperforms the best-published few-shot image classification accuracy~\cite{liu2023language} by an absolute 25\% under the same in-context setting.
    \end{itemize}

    \item \cref{chap:magvitv2}: \cite{yu2023language} \textbf{Lijun Yu}, Jos\'e Lezama, Nitesh B. Gundavarapu, Luca Versari, Kihyuk Sohn, David Minnen, Yong Cheng, Agrim Gupta, Xiuye Gu, Alexander G. Hauptmann, Boqing Gong, Ming-Hsuan Yang, Irfan Essa, David A. Ross, and Lu Jiang. \\
    \emph{Language Model Beats Diffusion -- Tokenizer is Key to Visual Generation.} \\
    To appear in ICLR 2024.
    \begin{itemize}
    \item A new video tokenizer that outperforms the previously best-performing video tokenizer in three areas: visual generation, video compression, and action recognition.
    \item A novel lookup-free quantization approach that enables improving the visual generation quality of language models by learning a large vocabulary.
    \item To the best of our knowledge, the first evidence suggesting that a language model can outperform diffusion models on ImageNet when provided with the same training data, an equivalent model size, and a similar training budget.
    \item A video compressor with better quality than HEVC and VVC, at similar bit rates, according to user studies. To our knowledge, this is the first successful attempt of a visual tokenizer designed for video generation to achieve comparable results to standard codecs.
    \end{itemize}

    \item \cref{chap:videopoet}: \cite{kondratyuk2023videopoet} Dan Kondratyuk$^*$, \textbf{Lijun Yu}$^*$, Xiuye Gu$^*$, Jos\'e Lezama$^*$, Jonathan Huang$^*$, Grant Schindler, Rachel Hornung, Vighnesh Birodkar, Jimmy Yan, Ming-Chang Chiu, Krishna Somandepalli, Hassan Akbari, Yair Alon, Yong Cheng, Josh Dillon, Agrim Gupta, Meera Hahn, Anja Hauth, David Hendon, Alonso Martinez, David Minnen, Mikhail Sirotenko, Kihyuk Sohn, Xuan Yang, Hartwig Adam, Ming-Hsuan Yang, Irfan Essa, Huisheng Wang, David A. Ross, Bryan Seybold$^*$, and Lu Jiang$^*$. $^*$ indicates equal contribution.\\
    \emph{VideoPoet: A Large Language Model for Zero-Shot Video Generation}. \\
    To appear in ICML 2024. \\
    In this large collaborative effort, my contribution primarily includes
    \begin{itemize}
    \item A method for training a Large Language Model (LLM) specifically for video generation tasks utilizing tokenized video frames and corresponding audio.
    \item A preliminary study on the scaling behavior of VideoPoet on audio-visual generation tasks.
    % \item Co-designed a method for training a Large Language Model (LLM) specifically for video generation tasks, utilizing tokenized video data that incorporates both text-paired and unpaired video data.
    % \item An approach to video super-resolution that increases spatial resolution within the latent token space using a bidirectional transformer with efficient windowed local attention.
    % \item Contributed to evaluations and demonstrations to highlight VideoPoet's competitive and state-of-the-art performance, especially in generating realistic and interesting videos with motion.
    \end{itemize}
\end{itemize}

\paragraph{}
In addition, we have the following publications referenced:

\begin{itemize}
    \item \cite{yu2018traffic} \textbf{Lijun Yu}, Dawei Zhang, Xiangqun Chen, and Alexander G Hauptmann. \\ 
    \emph{Traffic Danger Recognition With Surveillance Cameras Without Training Data.}\\
    In AVSS 2018.
    \item \cite{yu2019training} \textbf{Lijun Yu}, Peng Chen, Wenhe Liu, Guoliang Kang, and Alexander G Hauptmann. \\
    \emph{Training-Free Monocular 3D Event Detection System for Traffic Surveillance.} \\
    In Big Data 2019.
    \item \cite{yu2020cmu} \textbf{Lijun Yu}, Yijun Qian, Wenhe Liu, and Alexander G Hauptmann. \\
    \emph{CMU Informedia at TRECVID 2020: Activity Detection with Dense Spatio-Temporal Proposals.} \\
    In TRECVID 2020.
    \item \cite{qian2020electricity} Yijun Qian$^*$, \textbf{Lijun Yu$^*$}, Wenhe Liu, and Alexander G Hauptmann. $^*$ indicates equal contribution.\\
    \emph{ELECTRICITY: An Efficient Multi-Camera Vehicle Tracking System for Intelligent City.} \\
    In CVPRW 2020. 
    \item \cite{yu2020zero} \textbf{Lijun Yu}, Qianyu Feng, Yijun Qian, Wenhe Liu, and Alexander G Hauptmann. \\
    \emph{Zero-VIRUS: Zero-shot Vehicle Route Understanding System for Intelligent Transportation.} \\
    In CVPRW 2020.
    \item \cite{yu_cmu_2021} \textbf{Lijun Yu}, Yijun Qian, Wenhe Liu, and Alexander G Hauptmann. \\
    \emph{CMU Informedia at TRECVID 2021: Activity Detection with Argus++.} \\
    In TRECVID 2021.
    \item \cite{qian2022rethinking} Yijun Qian, \textbf{Lijun Yu}, Wenhe Liu, and Alexander G Hauptmann. \\
    \emph{Rethinking Zero-Shot Action Recognition: Learning from Latent Atomic Actions.} \\
    In ECCV 2022.
    \item \cite{sun2022score} Haoran Sun, \textbf{Lijun Yu}, Bo Dai, Dale Schuurmans, and Hanjun Dai. \\
    \emph{Score-based Continuous-time Discrete Diffusion Models.} \\
    In ICLR 2022.
    \item \cite{michaels2024audio} Jackson Michaels, Juncheng Li, Laura Yao, \textbf{Lijun Yu}, Zach Wood-Doughty, and Florian Metze. \\
    \emph{Audio-Journey: Open Domain Latent Diffusion Based Text-to-Audio Generation.} \\
    In ICASSP 2024.
    % \item \cite{zhao2024improving} Lingxiao Zhao, Xueying Ding, \textbf{Lijun Yu} \etal. Improving and Unifying Discrete \& Continuous-time Discrete Denoising Diffusion. In submission to ICML 2024.
    \item \cite{gupta2023photorealistic} Agrim Gupta, \textbf{Lijun Yu}, Kihyuk Sohn, Xiuye Gu, Meera Hahn, Li Fei-Fei, Irfan Essa, Lu Jiang, and Jos\'e Lezama. \\
    \emph{Photorealistic Video Generation with Diffusion Models.} \\
    In submission to ECCV 2024.
    % \item \cite{li2023audio} Juncheng B Li, Jackson Sam Michaels, Laura Yao, \textbf{Lijun Yu} \etal. Audio-Journey: Efficient Visual+LLM-aided Audio Encodec Diffusion. In ICML ES-FoMO 2023.
\end{itemize}

\paragraph{}
During the course, we have given these invited talks:

\begin{itemize}
    \item Traffic Danger Recognition With Surveillance Cameras Without Training Data.
    \begin{itemize}
        \item ICCV Demo, Oct 2019.
        \item TRECVID, Nov 2019.
    \end{itemize}
    \item Argus++: Real-Time Activity Detection in Unknown Facilities with Dense Spatio-Temporal Proposals.
    \begin{itemize}
        \item WACV HADCV, Jan 2021.
        \item CVPR ActivityNet, Jun 2021.
    \end{itemize}
    \item ArgusRoad: Road Activity Detection with Connectionist Spatio-Temporal Proposals.
    \begin{itemize}
        \item ICCV ROAD, Oct 2021.
    \end{itemize}
    \item Language Model Beats Diffusion: Tokenizer is Key to Visual Generation.
    \begin{itemize}
        \item Hong Kong University of Science and Technology, Jan 2024.
        \item Institute of Computing Technology, Chinese Academy of Sciences, Jan 2024.
        \item Baidu, Jan 2024.
        \item New York University, Feb 2024.
        \item ByteDance, Feb 2024.
        \item Peking University Alumni Association of Nothern California, Mar 2024.
        \item Kunlun Tech, Mar 2024.
        \item California Institute of Technology, Mar 2024.
        \item Adobe, Apr 2024.
        \item Hong Kong Shanghai AI Forum, Apr 2024.        
    \end{itemize}    
\end{itemize}

\paragraph{}
We have won the first place at the following international challenges:

\begin{itemize}
    \item TRECVID ActEV, 2019 \& 2020.
    \item CVPR AI City: City-Scale Multi-Camera Vehicle Tracking, 2020.
    \item MediaEval Sports Video: Stroke Classification, 2021.
    \item WACV ActEV SDL Unknown Facility, 2021.
    \item CVPR ActivityNet: Kinetics-700, 2021.
    \item ICCV ROAD: Action Detection, 2021.
    \item CVPR ActivityNet: ActEV, 2021 \& 2022.
\end{itemize}

\section*{Applications}
\addcontentsline{toc}{section}{Applications}

The technologies developed within this thesis hold the potential of revolutionizing multiple sectors, offering applications that stretch from creative content creation to enhancing communication and predictive simulations.

One of the most immediate applications lies in the realm of digital content creation. For movie makers and vloggers, the ability to generate high-fidelity videos with corresponding audio on demand could streamline the production process, reducing the need for extensive filming schedules and potentially lowering production costs. Filmmakers could leverage these models to create realistic scenes or backgrounds without the logistical challenges of on-location shoots, while vloggers might use them to enhance their content with high-quality visuals created from textual descriptions, making their production process more efficient and creative.

In the sphere of social media and digital platforms, these technologies could pioneer a new era of content discovery and engagement. By replacing traditional recommendation systems with automated content generation based on user preferences and interaction history, platforms akin to Instagram could offer an endless stream of personalized, generated content. This could not only enhance user engagement through highly customized experiences but also address some of the challenges associated with content moderation by generating safer, tailored content that adheres to platform guidelines.

Another impactful application is in communication, specifically in translating sign language into text or speech and vice versa in real time. This could dramatically improve the accessibility of digital content for the deaf and hard-of-hearing community, making information more inclusive and fostering a more connected world. The ability of these models to understand and generate video content opens the door to real-time sign language interpretation services, which could be integrated into video conferencing platforms, educational content, and public broadcasting.

Furthermore, the implications for scientific research and predictive modeling are profound. In weather forecasting, these technologies could enhance the accuracy of predictions by generating and analyzing complex weather patterns from vast datasets, potentially offering more reliable forecasts that could aid in disaster preparedness and climate research. In the realm of physics, the ability to simulate intricate systems through generated visual content could advance our understanding of phenomena that are challenging to observe directly, such as subatomic particle interactions or astronomical events.

These applications merely scratch the surface of what is possible with the advancements presented in this thesis. As these technologies continue to evolve, they will likely uncover new opportunities and challenges, pushing the boundaries of creativity, communication, and scientific exploration.

\section*{Limitations and Future Work}
\addcontentsline{toc}{section}{Limitations and Future Work}

\paragraph{Limitations.}
While the advancements discussed in this thesis represent significant progress in the field of artificial intelligence, particularly in video generation, they are not without limitations. 
\begin{itemize}
    \item \emph{Imperfect realism}: Models may struggle to perfectly reproduce subtle details and emotional nuances, hindering use in applications demanding absolute realism (e.g., film, VR).
    \item \emph{Controllability}: Generated output may not always perfectly match complex or specific conditions in the input text. This limits use in fields requiring precise visual representation.
    \item \emph{Resolution and length}: Current technology could struggle with high-resolution or long-duration videos, impeding use in industries reliant on detailed, lengthy visualizations.
    \item \emph{Efficiency and cost}: The amount of compute required by current method remains infeasible to generate large-scale media content, \eg, the daily uploaded amount to YouTube.
    \item \emph{Data bias}: Biases in training data can perpetuate harmful stereotypes or lack of diversity within the generated content.
\end{itemize}

\paragraph{Disccusion.}
Beyond the approaches presented in this thesis, the relevant literature involves latent diffusion transformer~\cite{peebles2022scalable}, which has been promoted by the very recent release of Sora~\cite{videoworldsimulators2024} for minute-long full-HD high-quality video generation.
% Very recently, the video generation field has witnessed the demo release of Sora~\cite{videoworldsimulators2024}, which is capable of generating minute-long full-HD videos in high quality.
% While many design details are uncertain from its vague technical report, it is using a different technique from those presented in this thesis, namely latent diffusion transformer~\cite{peebles2022scalable}.
Concurrently, we have also worked on diffusion transformers in W.A.L.T~\cite{gupta2023photorealistic} through a collaborative effort, utilizing the MAGVIT-v2 tokenizer.
It remains challenging to compare our models with the proprietary Sora side-by-side especially due to its undisclosed training compute and dataset, which are estimated to be orders of magnitude larger.

Through the MAGVIT series, VideoPoet, and W.A.L.T, we have studied video generation with masked, autoregressive, and diffusion transformers.
While it is beyond the scope of this thesis, a conceptual comparison between them appears valuable for further study.
\begin{itemize}
    \item \emph{Latent model}: All of these transformer models operate in a compressed latent space, where the encoder compresses and the decoder decompresses. This is partly due to the high-dimensional nature of videos and the compute cost of transformers. As we have demonstrated in this thesis, crafting high-quality latent representation is crucial for video generation. Our MAGVIT-v2 tokenizer is widely applicable for these generative transformers.
    \item \emph{Architecture}: All of these transformers share a similar architecture for sequence modeling, with different input and output layers to adapt for discrete or continuous latents, according to the chosen objective. The overall training is agnostic to architecture details, where efficiency optimizations such as windowed attention or even state-space models could be adopted.
    \item \emph{Continuous \vs discrete space}: Autoregressive transformers and masked transformers predict discrete variables, whereas diffusion models predict continuous variables. Equivalent information can be stored in either discrete or continuous representations, where preliminary MAGVIT-v2 experiments show 4 LFQ bits are roughly equal to one VAE channel. Discrete tokenizers are usually more challenging to train than continuous versions due to the non-differentiable quantization layer and the strict bandwidth. Comparing to the regression of a continuous variable with Gaussian prior, or even simpler regression of the added noise, categorical prediction of a uniform discrete variable is more challenging for the model, but may also lead to better scalability.
    \item \emph{Decoding}: To produce an output of $\rvx \in \sR^{n}$, diffusion decoding is factorized along an extra axis with $t$ levels of noise, where conditional independence between the elements of $\rvx$ can be assumed given a previous prediction. On the other hand, autoregressive decoding is factorized along the axis of $n$, where the joint distribution is enforced by sampling one position at a time. Masked decoding can be viewed as a discrete diffusion process with absorbing noise~\cite{austin2021structured}, where independent samples are drawn for the positions decoded at the same step. An $n$-step decoding is the price that autoregressive models pay to sample from the exact joint distribution. Diffusion models approximate the joint distribution through the denoising of independent Gaussian noises, where the approximation gets more precise as $t \rightarrow n$. One nice property of diffusion decoding is that it can correct or refine the previous decoded sample, while autoregressive cannot.
    \item \emph{Inference efficiency}: Despite taking $n$ steps, the theoretical compute cost of autoregressive decoding is equivalent to just one decoding step of diffusion or masked models. However, token-by-token autoregressive decoding is usually memory-bound by the KV-cache on modern hardware, decaying the actual performance. On the other hand, diffusion decoding is compute-bound since it runs multiple full forward passes without caching.
    \item \emph{Training efficiency}: Autoregressive transformer has the best training efficiency among these models, as one forward pass can cover the entire decoding trajectory thanks to the causal attention mask. Diffusion and masked transformers usually sample a random noise ratio for each forward pass, which corresponds to one decoding step.
    \item \emph{Causality}: Diffusion and masked transformers adopt bidirectional attention for full dependency, whereas autoregressive models only allow the prediction to depend on the past. While enforcing a raster-scan order for spatial autoregressive modeling may not be the optimal solution, temporal causality remains critical for long generation and streamed generation in interactive setups.
    \item \emph{Unification}: Autoregressive video generation models such as VideoPoet can be easily unified with existing LLM approaches and benefit from the relevant optimizations, which is one of the top advantages of this approach. It remains challenging to unify diffusion and LLM into one model.
\end{itemize}

\paragraph{Suggestions.}
As technologies evolve, future approaches are likely to appear differently from what we have presented in this thesis.
However, a few suggestions from our experiences might remain helpful for a longer period of time than the methodologies themselves.
\begin{itemize}
    \item \emph{Transformers} are never designed to produce a large amount of data. Even with an optimistic estimation of 10M sequence length with $2^{20}$ vocabulary size, the model can only produce 25MB data.
    This is still far from an hour-long full-HD video which can be several gigabytes. Therefore, either a new architecture or a highly-compressed latent is required.
    \item \emph{Hardware-software co-design} could play an important role for new techniques to be deployed at scale. For example, hardware accelerators for video codecs could be a potential solution to improve generation efficiency by 100$\times$. In addition, the coarse-to-fine setup in discrete cosine transformation could be a better order than the raster scan. 
    \item \emph{Modality unification} is worth exploring where one model should be able to jointly learn the generation of multiple modalities. There is a huge space to search for the optimal learning recipe which should take care of the inter-modal interaction and also enable post-training addition of new modalities at minimal costs.
    \item \emph{Realtime interactive} video generation, if made possible, could be applied to a wide array of scenarios, including gaming, story telling, software UI, world simulator, \etc. This could become the next major milestone in the history of computer software.
    \item \emph{Embodied intelligence} could benefit from neural simulators and end-to-end controller with visual world knowledge, both enabled via generative video learning. As LLMs become widely deployed, robotics might be a great direction for next-generation PhD students.
    \item \emph{Foundation models} should learn from raw signals like video and audio rather than the distilled human knowledge from text. The endless sources for these raw signals could enable better scalability than exhausted text data. Models can be pretrained without text and just aligned with text for human interaction. This sets up a better stage to enable the discovery of new rules of the universe without restrictions from human prior.
\end{itemize}

\glsresetall
\cleardoublepage
% \appendix
% \input{chapters/appendix.tex}

\backmatter

%\renewcommand{\baselinestretch}{1.0}\normalsize

% By default \bibsection is \chapter*, but we really want this to show
% up in the table of contents and pdf bookmarks.

%\newcommand{\bibpreamble}{This text goes between the ``Bibliography''
%  header and the actual list of references}
\bibliographystyle{plainnat}
\bibliography{reference} %your bib file

\cleardoublepage

\end{document}